\documentclass{article}

\usepackage[nonatbib,preprint]{neurips_2021}

\usepackage[utf8]{inputenc} %
\usepackage[T1]{fontenc}    %
\usepackage{hyperref}       %
\usepackage{url}            %
\usepackage{booktabs}       %
\usepackage{amsfonts}       %
\usepackage{nicefrac}       %
\usepackage{microtype}      %
\usepackage{xcolor}         %
\usepackage{graphicx}
\usepackage{subfig}
\usepackage[export]{adjustbox}
\extrafloats{1000}

\usepackage{bookmark,hyperref}

\usepackage[
backend=biber,
style=numeric-comp,
sorting=ynt
]{biblatex}
\usepackage{setspace}

\newcommand{\arxiv}[1]{#1}
\newcommand{\neurips}[1]{}
\newcommand{\Action}{\mathcal{A}}
\newcommand{\State}{\mathcal{S}}

\usepackage{amsmath,amsfonts,bm}

\def\eqref#1{equation~\ref{#1}}

\def\1{\bm{1}}

\DeclareMathAlphabet{\mathsfit}{\encodingdefault}{\sfdefault}{m}{sl}
\SetMathAlphabet{\mathsfit}{bold}{\encodingdefault}{\sfdefault}{bx}{n}

\newcommand{\E}{\mathbb{E}}

\newcommand{\R}{\mathbb{R}}

\newcommand{\softplus}{\mathrm{softplus}}
\newcommand{\KL}{D_{\mathrm{KL}}}
\newcommand{\JS}{D_{\mathrm{JS}}}

\newcounter{choice}[section]
\DeclareRobustCommand{\setchoice}[1]{%
   \refstepcounter{choice}%
 \label{choice:#1}}
\newcommand{\choicep}[1]{\texttt{C\ref{choice:#1}}}
\newcommand{\choicet}[1]{\texttt{\csname#1\endcsname{ }(C\ref{choice:#1})}}
\newcommand{\choicetable}[1]{\texttt{C\ref{choice:#1}} & \texttt{\csname#1\endcsname}}

\newcommand{\dchoicet}[1]{\choicet{#1}\setchoice{#1}}

\neurips{\everypar{\looseness=-1}}
\title{What Matters for Adversarial Imitation Learning?}

\author{
  Manu Orsini\thanks{Equal contribution.},
  Anton Raichuk\footnotemark[1],
  L\'eonard Hussenot\footnotemark[1]  \thanks{Univ. de Lille, CNRS, Inria Scool, UMR 9189 CRIStAL.}, \\
  \textbf{Damien Vincent,
  Robert Dadashi, 
  Sertan Girgin,} \\
  \textbf{Matthieu Geist,
  Olivier Bachem,
  Olivier Pietquin,
  Marcin Andrychowicz\thanks{Corresponding author. E-mail: \texttt{marcina@google.com}.}}
  \\ \\
  Google Research, Brain Team
}

\begin{document}

\maketitle
\begin{abstract}
Adversarial imitation learning has become a popular framework for imitation in continuous control. Over the years, several variations of its components were proposed to enhance the performance of the learned policies as well as the sample complexity of the algorithm. In practice, these choices are rarely tested all together in rigorous empirical studies.
It is therefore difficult to discuss and understand what choices, among the high-level algorithmic options as well as  low-level implementation details, matter.
To tackle this issue, we implement more than 50 of these choices in a generic adversarial imitation learning framework
and investigate their impacts in a large-scale study (>500k trained agents) with both synthetic and human-generated demonstrations.
\neurips{We analyze the key results and highlight the most surprising findings.}
\arxiv{While many of our findings confirm common practices, some of them are
surprising or even contradict prior work.
In particular, our results suggest that artificial %
demonstrations are not a good proxy for human data
and that the very common practice of evaluating imitation algorithms only with synthetic demonstrations
may lead to algorithms which perform poorly in the more realistic scenarios with human demonstrations.
}
\end{abstract}
\section{Introduction}\label{introduction}

Reinforcement Learning (RL) has shown its ability to perform complex tasks in contexts where clear reward functions can be set-up (e.g. +1 for winning a chess game)
\cite{silver2016mastering, berner2019dota,andrychowicz2020learning,vinyals2019grandmaster} but for many real-world applications, designing a correct reward function is either tedious or impossible \cite{popov2017data},
while demonstrating a correct behavior is often easy and cheap.
Therefore, imitation learning (IL, \cite{SCHAAL1999233,argall2009survey}) might be the key to unlock the resolution of more complex tasks,
such as autonomous driving, for which reward functions are much harder to design.%

The simplest approach to IL is Behavioral Cloning (BC, \cite{bc}) which uses supervised learning to predict the expert's action for any given state. However, BC is often unreliable as prediction errors compound in the course of an episode. Adversarial Imitation Learning (AIL, \cite{gail}) aims to remedy this using inspiration from Generative Adversarial Networks (GANs, \cite{goodfellow2014generative})
and Inverse RL \cite{russell1998learning,ng2000algorithms,ziebart2008maximum}: the policy is trained to generate trajectories
that are indistinguishable from the expert's ones.
As in GANs, this is formalized as a two-player game where a discriminator is co-trained to distinguish between the policy and expert trajectories (or states). See App.~\ref{app:gail} for a brief introduction to AIL.

\looseness=-1
A myriad of improvements over the original AIL algorithm were proposed over the years \cite{airl,dac,fairl,lipschitzness,pugail}, from changing the discriminator’s loss function
\cite{airl} to switching from on-policy to off-policy agents \cite{dac}. However, their relative performance is rarely studied in a controlled setting, and never these changes were all compared together.
The performance of these high-level choices may also depend on the low-level implementation details,
which might be not even mentioned in the publications
\cite{islam2017reproducibility, henderson2018deep, tucker2018mirage, mujoco123},
as well as the hyperparameters (HPs) used.
Thus, assessing whether the proposed changes are the reason for the presented improvements becomes extremely difficult. This lack of proper comparisons slows down the overall research in imitation learning and the industrial applicability of these methods.

\looseness=-1
We investigate such high- and low-level choices
in depth and study their impact on the algorithm performance.
Hence, as \textbf{our key contributions}, we
\textbf{(1)} implement a highly-configurable generic AIL algorithm, with various axes of variation
(>50 HPs), %
including 4 different RL algorithms and 7 regularization schemes for the discriminator,
\textbf{(2)} conduct a large-scale (>500k trained agents) experimental study %
on 10 continuous-control tasks
\arxiv{\footnote{
A \emph{task} is defined by an environment
and the demonstrator type (either human or RL agent).
}}
and \textbf{(3)} analyze the experimental results to provide practical insights and recommendations
for designing novel and using existing AIL algorithms.

\looseness=-1
\paragraph{Most surprising finding \#1: regularizers.}
While many of our findings confirm common practices in AIL research, some of them are
surprising or even contradict prior work.
In particular, we find that standard regularizers from Supervised Learning --- dropout \cite{dropout} and
weight decay \cite{decay} often perform similarly to the regularizers designed specifically for
adversarial learning like gradient penalty \cite{gp}. Moreover, for easier environments
(which were often the only ones used in prior work), we find that it is possible to
achieve excellent results without using any explicit discriminator regularization.

\looseness=-1
\paragraph{Most surprising finding \#2: human demonstrations.}
Not only does the performance of AIL %
heavily depend on whether the
demonstrations were collected from a human operator or generated by an RL algorithm,
but %
the relative performance of algorithmic choices also depends on the demonstration
source.
Our results suggest that artificial %
demonstrations are not a good proxy for human data
and that the very common practice of evaluating IL algorithms only with synthetic demonstrations
may lead to algorithms which perform poorly in the more realistic scenarios with human demonstrations.

\arxiv{
\paragraph{Paper outline.} In Sec.~\ref{sec:design}, we describe our experimental setting and the performance metrics used.
We then present and analyze the results related to the agent (Sec.~\ref{sec:results-agent})
and discriminator (Sec.~\ref{sec:results-discriminator}) training.
Afterwards, we compare RL-generated and human-collected demonstrations (Sec.~\ref{sec:results-human})
and analyze the choices influencing the computational cost of running the algorithm (Sec.~\ref{sec:results-trade-offs}).
The appendices contain the details of the different algorithmic choices in AIL (App.~\ref{app:choices}) as well as the raw results of the 
experiments
(App.~\ref{exp_wide}--\ref{exp_tradeoffs}).
}

\looseness=-1
\section{Experimental design}\label{sec:design}

\looseness=-1
\paragraph{Environments.}
We focus on continuous-control tasks as robotics appears as one of the main potential applications of IL and a vast majority of the IL
literature thus focuses on it.
In particular, we run experiments with five widely used environments from OpenAI Gym \cite{gym}: \texttt{HalfCheetah-v2}, \texttt{Hopper-v2}, \texttt{Walker2d-v2},
\texttt{Ant-v2}, and \texttt{Humanoid-v2} and three manipulation environments
from Adroit \cite{rajeswaran2017learning}: \texttt{pen-v0}, \texttt{door-v0}, and \texttt{hammer-v0}.
\arxiv{All the environments are shown in Fig.~\ref{fig:envs}.}
The Adroit tasks consist in aligning a pen with a target orientation,
opening a door and hammering a nail with a 5-fingered hand.
\arxiv{
These two benchmarks bring orthogonal contributions.
The former focuses on locomotion but has 5 environments with different state/action dimensionality.
The latter, more varied in term of tasks, has an almost constant state-action space.
}

\arxiv{
\begin{figure}[!htb]
\centering
\begin{minipage}{\linewidth}
  \centering
  \includegraphics[height=3cm]{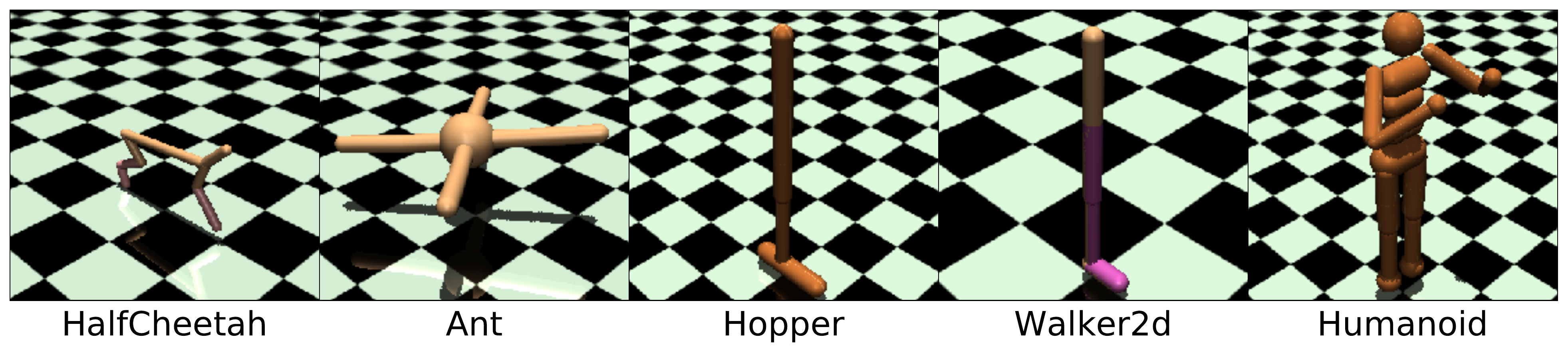}
  \includegraphics[height=3cm]{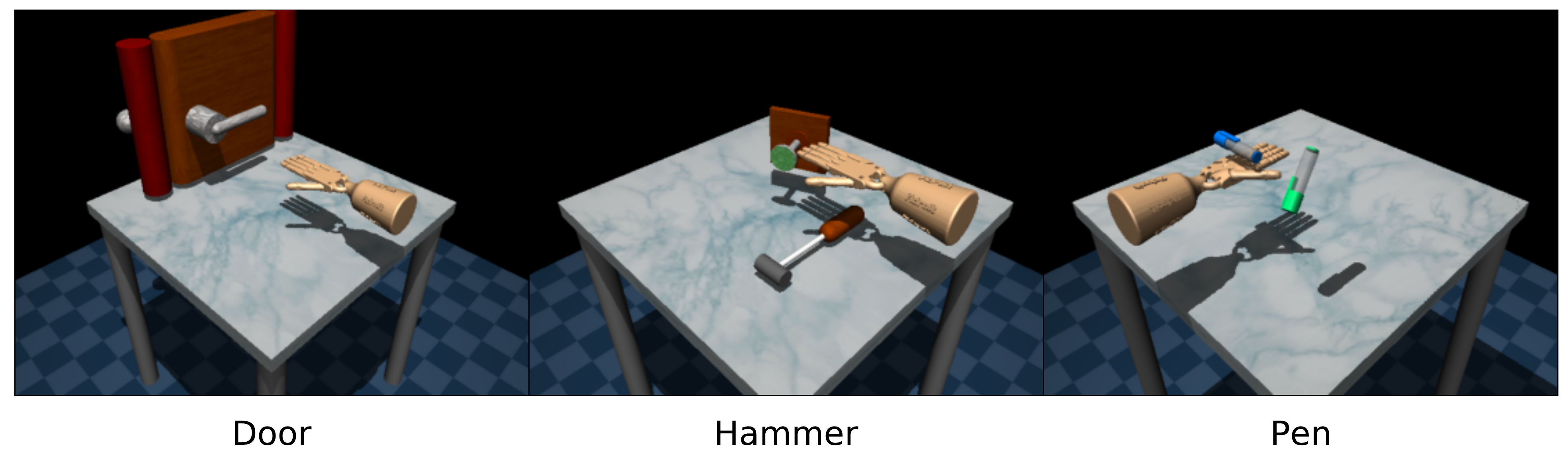}
\end{minipage}
\caption{Environments: OpenAI Gym (top) and Adroit (bottom).}
\label{fig:envs}
\end{figure}
}

\looseness=-1
\paragraph{Demonstrations.}
For the Gym tasks, we generate demonstrations with a SAC \cite{sac}
agent trained on the environment reward.
For the Adroit environments, we use the ``expert'' and ``human'' datasets from D4RL \cite{d4rl}, %
\arxiv{\footnote{
For \texttt{pen}, we only use the ``expert'' dataset,  the ``human'' one
consisting of a single (yet very long) trajectory.
}}
which are, respectively, generated by an RL agent and collected from a human operator.
As far as we know, our work is the first to solve these tasks with human datasets
in the imitation setup
(most of the prior work concentrated on Offline RL).
For all environments, we use 11 demonstration trajectories.
Following prior work \cite{gail, dac, fairl}, we subsample expert demonstrations
by only using every $20^\text{th}$ state-action pair to make the tasks harder.

\looseness=-1
\paragraph{Adversarial Imitation Learning algorithms.}
We researched prior work on AIL algorithms and made a list of commonly used
design decisions like policy objectives or discriminator regularization techniques.
We also included a number of natural options which we have not encountered in literature
(e.g. dropout \cite{dropout} in the discriminator or clipping rewards bigger than a threshold).
All choices are listed and explained in App.~\ref{app:choices}.
Then, we implemented a single highly-configurable AIL agent which exposes
all these choices as configuration options in the Acme framework \cite{acme}
using JAX \cite{jax} for automatic differentiation
and Flax \cite{flax} for neural networks computation.
The configuration space is so wide that it covers the whole family of
AIL algorithms, in particular, it mostly covers the setups
from AIRL%
\arxiv{\footnote{\looseness=-1 Seminal AIRL uses TRPO \cite{trpo} to train the policy, not supported in our implementation (PPO~\cite{ppo} used here).}}
\cite{airl} and DAC \cite{dac}.
We plan to open source the agent implementation.

\looseness=-1
\paragraph{Experimental design.}
We created a large HP sweep (57 HPs swept, >120k agents trained) in which each HP is sampled uniformly at random
from a discrete set and independently from the other HPs.
We manually ensured that the sampling ranges of all HPs are appropriate
and cover the optimal values.
Then, we analyzed the results of this initial experiment
(called \emph{wide}, detailed description and results in App.~\ref{exp_wide}),
removed clearly suboptimal options
and ran another experiment with the pruned sampling ranges
(called \emph{main}, 43 HPs swept, >250k agents trained, detailed description and results in App.~\ref{exp_main}).
The latter experiment serves as the basis for most of the conclusions
drawn in this paper but we also run a few additional experiments to investigate
some additional questions (App.~\ref{exp_tradeoffs} and App.~\ref{app:additional-plots}).

This pruning of the HP space guarantees that we draw conclusions based on training configurations
which are highly competitive (training curves can be found in Fig.~\ref{fig:main_training_curves})
while using a large HP sweep (including, for example, multiple different RL algorithms)
ensures that our conclusions are robust and valid 
not only for a single RL algorithm and specific values of HPs, but are more
generally applicable.
Moreover, many choices may have strong interactions with other related choices, for example
we find a surprisingly strong interaction between the discriminator regularization scheme
and the discriminator learning rate (Sec.~\ref{sec:results-discriminator}).
This means that such choices need to be tuned together (as it is the case in our study)
and experiments where only a single choice is varied but the interacting choices are kept fixed
may lead to  misleading conclusions.

\looseness=-1
\paragraph{Performance measure.}
For each HP configuration and each of the 10 environment-dataset pairs
we train a policy and evaluate it 10 times through the training
by running it for 50 episodes and computing the average undiscounted return
using the environment reward.
We then average these scores to obtain a single performance score %
which approximates the area under the learning curve.
This ensures we assign higher scores to HP configurations that learn quickly.

\looseness=-1
\paragraph{Analysis.} We consider two different analyses for each choice\footnote{
This analysis is based on a similar type of study focused on on-policy
RL algorithms \cite{mujoco123}.}:%

\textbf{\emph{Conditional 95th percentile}}: For each potential value of that choice (e.g., \texttt{RL Algorithm} = \texttt{PPO}), we look at the performance distribution of sampled configurations with that value.
We report the 95th percentile of the performance as well as
error bars based on bootstrapping.\footnote{
We compute each metric $20$ times based on a randomly selected half of all training runs,
and then report the mean of these 20 measurements while the error bars show
mean-std and mean+std.}
This corresponds to an estimate of the performance one can expect if all other choices
were tuned with random search and a limited budget of roughly 13 HP configurations\footnote{The probability that all 13 configurations score worse than the 95th percentile
is equal $0.95^{13} \approx 50\%$.}.
All scores are normalized so that $0$ corresponds to a random policy
and $1$ to the expert performance (expert scores can be found in App.~\ref{app:expert_scores}).

\textbf{\emph{Distribution of choice within top 5\% configurations.}}
We further consider for each choice the distribution of values among the top 5\%
HP configurations.
In particular, we measure the ratio of the frequency of the given value in the top 5\% of
HP configurations with the best performance
to the frequency of this value among all HP configurations.
If certain values are over-represented in the top models (ratio higher than 1),
this indicates that the specific choice is important for %
good performance.

\looseness=-1
\section{What matters for the agent training?}\label{sec:results-agent}

\looseness=-1
\paragraph{Summary of key findings.}
The AIRL reward function perform best for synthetic demonstrations
while $-\ln(1-D)$ is better for human demonstrations.
Using explicit absorbing state is crucial in environments
with variable length episodes.
Observation normalization strongly affects the performance.
Using an off-policy RL algorithm
is necessary for good sample complexity
while replaying expert data and pretraining with BC
improves the performance only slightly.

\looseness=-1
\paragraph{Implicit reward function.}
In this section, we investigate choices related to agent training with AIL,
the most salient of which is probably the choice of the implicit reward function.
Let $D(s,a)$ be the probability of classifying the given state-action pair as \emph{expert} by the discriminator%
\arxiv{\footnote{
Some prior works, including GAIL\cite{gail}, use the opposite notation, with
$D(s,a)$ the \emph{non-expert} probability.}}.
In particular, we run experiments with the following reward functions:
$r(s,a)=-\log (1-D(s,a))$ (used in the original GAIL paper \cite{gail}),
$r(s,a)=\log D(s,a)-\log (1-D(s,a))$ (called the AIRL reward \cite{airl}),
$r(s,a)=\log D(s,a)$ (a natural choice we have not encountered in literature),
and the FAIRL \cite{fairl} reward function $r(s,a)=-h(s,a)\cdot e^{h(s,a)}$,
where $h(s,a)$ is the discriminator logit\arxiv{\footnote{It can also be expressed as $h(s,a)=\log D(s,a)-\log(1-D(s,a))$.}}.
It can be shown that, under the assumption that all episodes
have the same length, maximizing these reward functions corresponds to
the minimization of different divergences between the marginal
state-action distribution of the expert and the policy.
See \cite{fairl} for an in-depth discussion on this topic.
We also consider clipping the rewards with absolute values bigger than a threshold which is a HP.

\begin{figure}[h]
  \centering
    \subfloat[\centering wide HP search]{{\includegraphics[height=4cm]{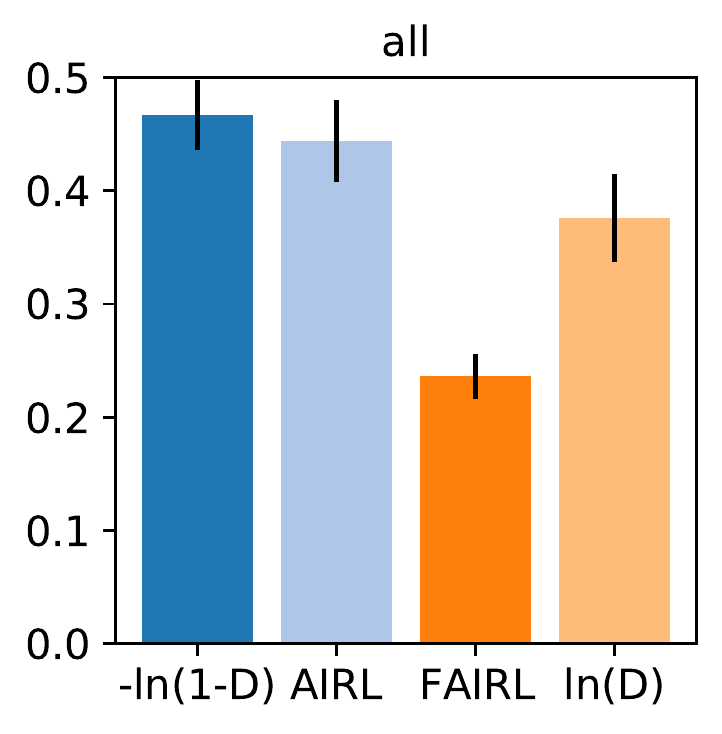} }}%
    \qquad
    \subfloat[\centering no absorbing state]{{\includegraphics[height=4cm]{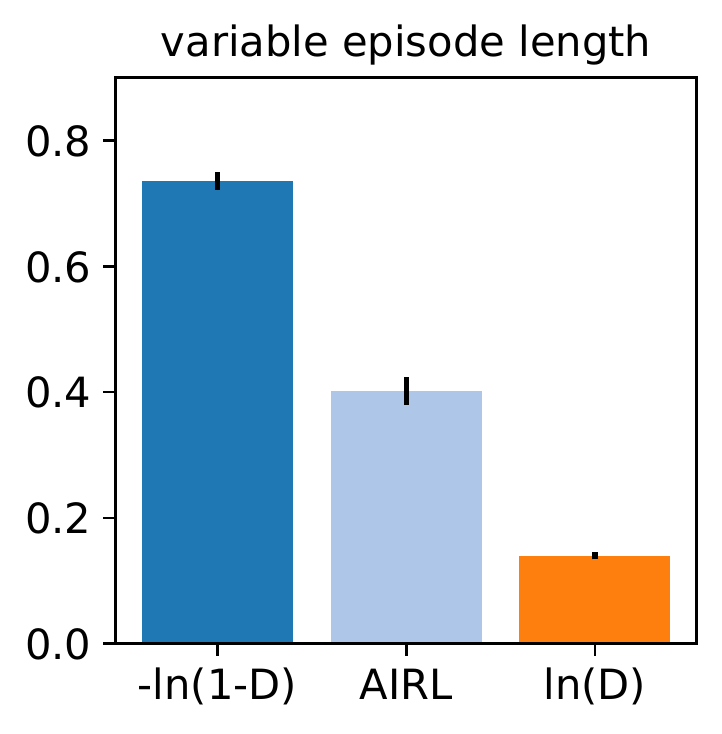} }}%
    \qquad
    \subfloat[\centering absorbing state]{{\includegraphics[height=4cm]{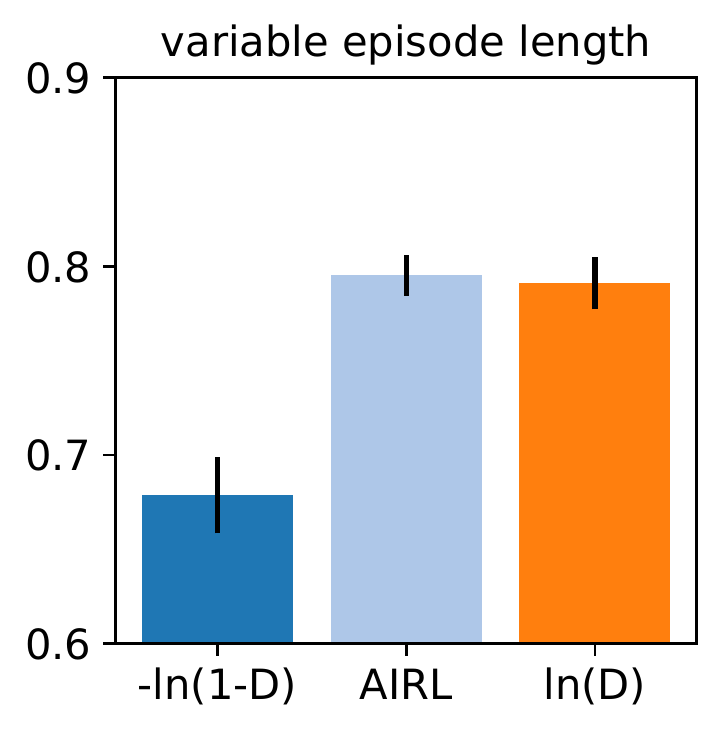} }}%
    \caption{
    Comparison of different reward functions.
    The bars show the 95th percentile across HPs sampling of the \emph{average} policy performance during training.
    Plot (a) shows the results averaged across all 10 tasks.
    Plots (b) and (c) show the performance on the subset of environments
    with variable length episodes when the
    absorbing state is disabled (b) or enabled (c).
    See Fig.~\ref{fig:wide__gin_reward_function__macro_value} and Fig.~\ref{fig:corr_abs_reward} for the individual results in all environments.}%
  \label{fig:results-rewards}
\end{figure}

\looseness=-1
The FAIRL reward performed much worse than all %
others in the initial wide experiment (Fig.~\ref{fig:results-rewards}a)
and therefore was not included in our main experiment.
This is mostly caused by its inferior performance with off-policy RL algorithms
(Fig.~\ref{fig:corr_rl_reward}).
Moreover, reward clipping significantly helps the FAIRL reward (Fig.~\ref{fig:fairl_clipping})
while it does not help the other reward functions
apart from some small gains for $-\ln(1-D)$
(Fig.~\ref{fig:extra_reward_clipping}).
Therefore, we suspect that the poor performance of the FAIRL reward function may be caused by 
its exponential term which may have very high magnitudes.
Moreover, the FAIRL paper \cite{fairl} mentions
that the FAIRL reward is more sensitive to HPs
than other reward functions which could
also explain its poor performance in our experiments.

\looseness=-1
Fig.~\ref{fig:main__gin_reward_function__macro_value} shows that the $\ln(D)$ reward functions
performs a bit worse than the other two reward functions in the main experiment.
Five out of the ten tasks used in our experiments have variable length episodes
with longer episodes correlated with better behaviour\arxiv{\footnote{The episodes are terminated earlier if the simulated robot falls over or if the pen is dropped.}}
(\texttt{Hopper}, \texttt{Walker2d}, \texttt{Ant}, \texttt{Humanoid}, \texttt{pen}) --- on these tasks we can notice
that $r(s,a)=-\ln (1-D(s,a))$ often performs best and $r(s,a)=\ln D(s,a)$ worst.
This can be explained by the fact that $-\ln (1-D(s,a))>0$ and $\ln D(s,a)<0$
which means that the former reward encourages longer episodes and the latter one shorter ones \cite{dac}.
Absorbing state (described in App.~\ref{app:choices-imitation-rl}) is a technique introduced in the DAC paper \cite{dac}
to mitigate the mentioned bias and encourage the policy to generate episodes
of similar length to demonstrations.
In Fig.~\ref{fig:results-rewards}b-c
we show how the performance of different reward functions compares
in the environments with variable length episodes
depending on whether the absorbing state is used.
We can notice that without the absorbing state $r(s,a)=-\ln (1-D(s,a))>0$
performs much better in the environments with variable episode length
which suggests that the learning is driven to a large extent
by the reward bias and not actual imitation of the expert behaviour \cite{dac}.
This effect disappears when the absorbing state is enabled (Fig.~\ref{fig:results-rewards}c).

Fig.~\ref{fig:corr_abs_reward} shows the performance
of different reward functions in all environments
conditioned on whether the absorbing state is used.
If the absorbing state is used,
the AIRL reward function performs best
in all the environments with RL-generated demonstrations,
and $\ln(D)$ performs only marginally worse.
The $-\ln(1-D)$ reward function underperforms on
the \texttt{Humanoid} and \texttt{pen} tasks while performing best with human datasets.
We provide some hypothesis for this behaviour in Sec.~\ref{sec:results-human},
where we discuss human demonstrations in more details.

\looseness=-1
\paragraph{Observation normalization.} We consider observation normalization which is applied to the inputs of all neural networks
involved in AIL (policy, critic and discriminator).
The normalization aims to transform the observations so that that each observation coordinate
has mean $0$ and standard deviation $1$.
In particular, we consider computing the normalization statistics either
using only the expert demonstrations so that the normalization is \emph{fixed} throughout the training,
or using data from the policy being trained (called \emph{online}). See App.~\ref{app:choices-normalization} for more details.
Fig.~\ref{\neurips{fig:main_obs_normalization}\arxiv{fig:results-obs-norm}} shows that input normalization significantly influences the performance
with the effects on performance being often much larger than those of algorithmic
choices like the reward function or RL algorithm used.
Surprisingly, normalizing observations can either significantly improve or diminish performance
and whether the fixed or online normalization performs better is also environment dependent.
\arxiv{
\arxiv{
\begin{figure}[h]
  \centering
  \includegraphics[height=4cm]{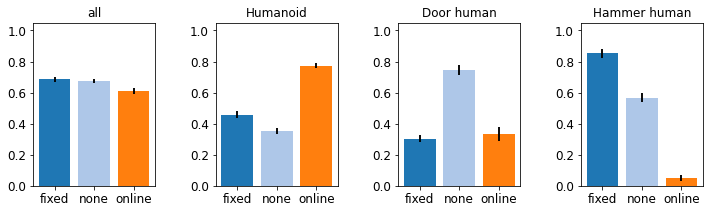}
  \caption{Comparison of observation normalization schemes.
  The bars show the 95th percentile of performance.
  The leftmost plot shows the results averaged across all 10 tasks.
  See Fig.~\ref{fig:main_obs_normalization} for the results on all environments.}
  \label{fig:results-obs-norm}
\end{figure}
}
}

\looseness=-1
\paragraph{Replaying expert data.}
When demonstrations as well as %
external rewards %
are available, it is common for RL algorithms to sample batches for
off-policy updates 
from the demonstrations in addition to the replay buffer
\cite{dqfn, r2d3}.
We varied the ratio of the policy to expert data being replayed
but found only very minor %
gains (Fig.~\ref{fig:expert_replay}).
Moreover, in the cases when we see some benefits,
it is usually best to replay 16--64 times more policy than expert data.
On some tasks (\texttt{Humanoid}) replaying even a single expert transitions every
256 agent ones significantly hurts performance.
We suspect that, in contrast to RL with demonstrations,
we see little benefit from replaying expert data in the setup with learned
rewards because (1) replaying expert data mostly helps when the reward signal is sparse
(not the case for discriminator-based rewards), and
(2) discriminator may overfit to the expert demonstrations
which could result in incorrectly high rewards being assigned
to expert transitions.

\looseness=-1
\paragraph{Pretraining with BC.} We also experiment with pretraining a policy with Behavioral Cloning (BC, \cite{bc})
at the beginning of training. Despite starting from a much better policy than a random one, we usually observe that the policy quality
deteriorates quickly at the beginning of training (see the \texttt{pen} task in Fig.~\ref{fig:results-best})
due to being updated using randomly initialized critic and discriminator networks,
and the overall gain from pretraining is very small
in most environments (Fig.~\ref{fig:main_pretrain_with_bc}).

\paragraph{RL algorithms.}
\looseness=-1
We run experiments with four different RL algorithms,
three of which are off-policy algorithms (SAC \cite{sac}, TD3 \cite{td3} and D4PG \cite{d4pg}),
as well as PPO \cite{ppo} which is nearly on-policy. %
Fig.~\ref{\neurips{fig:wide_direct_rl_algorithm}\arxiv{fig:results-rl}} shows that
the sample complexity of PPO is significantly worse than that of the
off-policy algorithms while all off-policy algorithms
perform overall similarly.

\arxiv{
\begin{figure}[h]
  \centering
  \includegraphics[height=4cm]{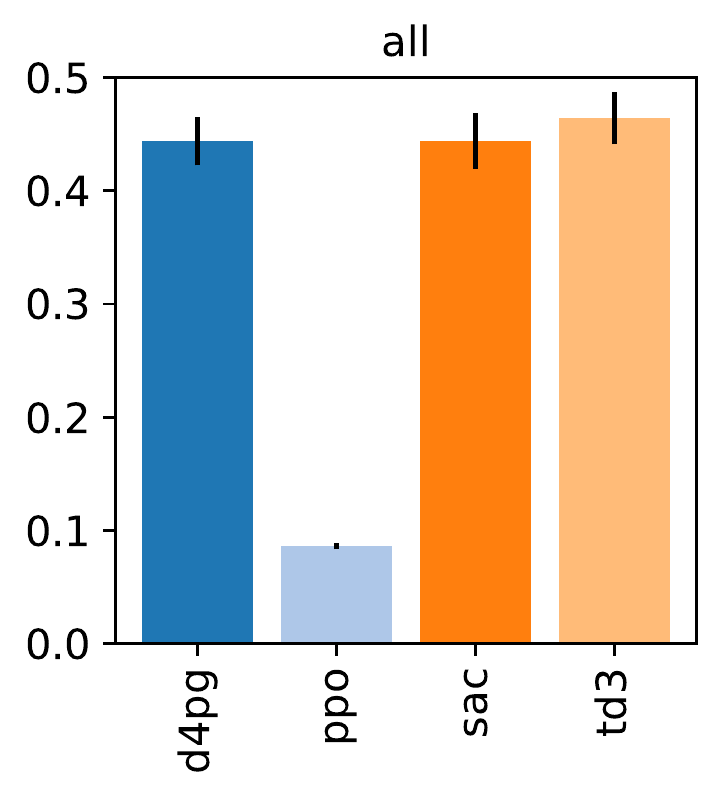}
  \caption{Comparison of RL algorithms (wide HP search).
  See Fig.~\ref{fig:wide_direct_rl_algorithm} for the results on individual environments.}
  \label{fig:results-rl}
\end{figure}
}

\paragraph{RL algorithms HPs.}
Fig.~\ref{fig:wide_discount} shows that the discount factor is one of the most important HPs with the values of $0.97-0.99$ performing well on all tasks.
Fig.~\ref{fig:main_max_replay_size} shows that in most environments it is better not to erase any data from the RL replay buffer
and always sample from all the experience encountered so far.
It is common in RL to use a noise-free version of the policy during evaluation
and we observe that it indeed improves the performance (Fig.~\ref{fig:main_eval_behavior_policy_type}).
The policy MLP size does not matter much (Figs.~\ref{fig:main_num_policy_layers}-\ref{fig:main_policy_layer_size})
while bigger critic networks perform significantly better\arxiv{\footnote{
We thus only include critics with at least two hidden layers with the size at least 128 in the main experiment.}}
(Figs.~\ref{fig:wide_num_critic_layers}-\ref{fig:wide_critic_layer_size}).
Regarding activation functions\arxiv{\footnote{
We use the same activation function in the policy and critic networks.}},
relu performs on par or better than tanh in all
environments apart from \texttt{door} %
in which tanh is significantly better (Fig.~\ref{fig:main_activation}).
Our implementation of TD3 optionally applies gradient clipping\arxiv{\footnote{The reason for that is that the DAC paper \cite{dac} uses TD3 with gradient clipping.}} but it does not affect the performance much (Fig.~\ref{fig:main_td3_gradient_clipping}).
D4PG can use n-step returns%
, this improves the performance on the Adroit tasks
but hurts  on the Gym suite (Fig.~\ref{fig:main_n_step}).

\section{What matters for the discriminator training?}\label{sec:results-discriminator}\looseness=-1

\paragraph{Summary of key findings.}\looseness=-1
MLP discriminators perform on par or better than AIL-specific architectures.
Explicit discriminator regularization is only important in more complicated environments (\texttt{Humanoid} and harder ones).
Spectral norm is overall the best regularizer but 
standard regularizers from supervised learning often perform on par.
Optimal learning rate for the discriminator
may be 2--2.5 orders of magnitude lower than the one for the RL agent.

\paragraph{Discriminator input.} In this section we look at the choices related to the discriminator training.
Fig.~\ref{fig:main__gin_discriminator_input__macro_value} shows how the performance depends on the discriminator input.
We can observe that while it is beneficial to feed actions as well as states to the discriminator,
the state-only demonstrations perform almost as well.
Interestingly, on the \texttt{door} task with human data, it is better to
ignore the expert actions. We explore the results with human demonstrations in more
depth in Sec.~\ref{sec:results-human}.

\paragraph{Discriminator architecture.} Regarding the discriminator network, %
our basic architecture is an MLP
but we also consider two modifications introduced in AIRL \cite{airl}:
a reward shaping term and a $\log \pi(a|s)$ logit shift
which introduces a dependence on the current policy
(only applicable to RL algorithms with stochastic policies, which in our case are PPO and SAC).
See App.~\ref{app:choices-discriminator-param} for a detailed description of these techniques.
Fig.~\ref{fig:wide__gin_make_discriminator_subtract_logpi} shows that the logit shift significantly hurts the performance.
This is mainly due to the fact that it does not work well with SAC
which is  off-policy (Fig.~\ref{fig:corr_rl_logpi}).
Fig.~\ref{fig:main__gin_make_discriminator_discriminator_module} shows that the shaping term does not affect the performance much.
While the modifications from AIRL does not improve the sample complexity in our experiments, it is worth
mentioning that they were introduced for another purpose, namely the recovery of transferable reward functions.

Regarding the size of the discriminator MLP(s),
the best results on all tasks are obtained with a single hidden layer (Fig.~\ref{fig:main__gin_discriminator__MLP_num_layers}),
while the size of the hidden layer is of secondary importance (if it is not very small)
with the exception of the tasks with human data where fewer hidden units perform significantly better
(Fig.~\ref{fig:main__gin_discriminator__MLP_num_units}).
All tested discriminator activation functions perform overall similarly
while sigmoid performs best with human demonstrations (Fig.~\ref{fig:main__gin_discriminator__MLP_activation}).

\paragraph{Discriminator training.} Fig.~\ref{fig:main__gin_GAILBuilder_max_replay_size} shows that it is best to use as large as possible replay buffers for sampling
negative examples (i.e. agent transitions).
As noticed in prior work, the initialization of the last \emph{policy} layer
can significantly influence the performance in RL \cite{mujoco123},
thus we tried initializing the last \emph{discriminator} layer with smaller weights but it
does not make much difference (Fig~\ref{fig:main__gin_discriminator__MLP_last_layer_kernel_init_scale}).

\paragraph{Discriminator regularization.}
\looseness=-1
An overfitting or too accurate discriminator can make agent's training challenging, and therefore
it is common to use additional regularization techniques
when training the AIL discriminator (or GANs in general).
We run experiments with a number of regularizers %
commonly used with AIL, namely
Gradient Penalty \cite{gp} (GP, used e.g. in \cite{dac}),
spectral norm \cite{spectral} (e.g. in \cite{lipschitzness}),
Mixup \cite{mixup} (e.g. in \cite{bee}),
as well as using the PUGAIL loss \cite{pugail}
instead of the standard cross entropy loss to train the discriminator.
Apart from the above regularizers, we also run experiments with regularizers
commonly used in Supervised Learning, namely
dropout \cite{dropout}, the weight decay \cite{decay} variant
from AdamW \cite{adamw} as well as the entropy bonus of the 
discriminator output treated as a Bernoulli distribution.
The detailed description of all these regularization techniques can be found in App.~\ref{app:choices-regularizers}.

\begin{figure}[h]
  \centering
  \includegraphics[width=.8\textwidth]{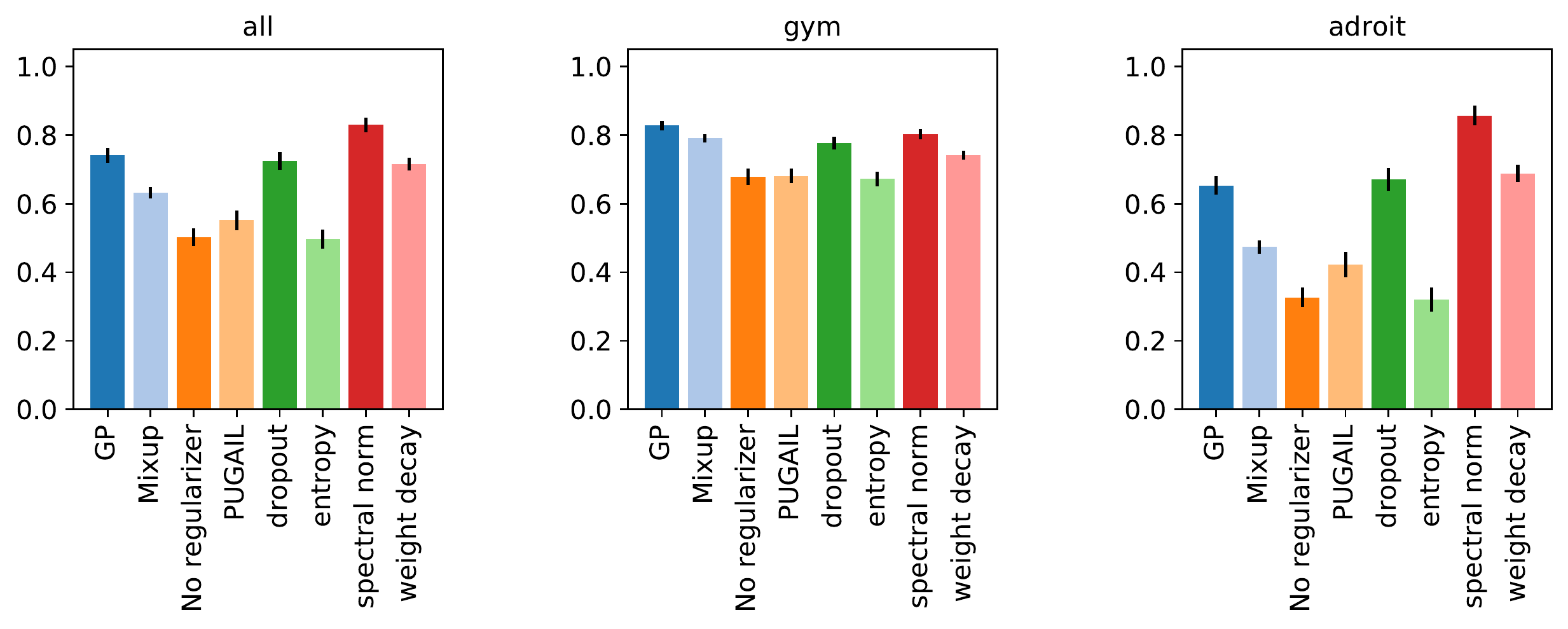}
  \caption{The 95th percentile of performance for different discriminator regularizers.
  The central plot shows the average performance across 5 tasks from OpenAI gym and the right one
  the average performance across 5 tasks from the Adroit suite.
  See Fig.~\ref{fig:main__gin_regularizer__macro_value} for the plots for individual environments.}
  \label{fig:regularizers}
\end{figure}

\looseness=-1
Fig.~\ref{fig:regularizers} shows how the performance depends on the regularizer. %
Spectral normalization performs overall best,  while
GP, dropout and weight decay all perform on par with each other
and only a bit worse than spectral normalization.
We find this conclusion to be quite surprising given that
we have not seen dropout or weight decay being used with AIL in literature.
We  also notice that the regularization is generally more
important on harder tasks like \texttt{Humanoid} or the tasks in the Adroit suite
(Fig.~\ref{fig:main__gin_regularizer__macro_value}).

Most of the regularizers investigated in this section have their own HPs
and therefore the comparison of different regularizers depends on
how these HPs are sampled.
As we randomly sample the
regularizer-specific HPs in this analysis, our approach favours regularizers that
are not too sensitive to their HPs. At the same time, there might be regularizers that are sensitive
to their HPs but for which good settings may be easily found.
Fig.~\ref{fig:main_perf_new_regularizer_0} shows that even
if we condition on choosing the optimal HPs for each regularizer,
the relative ranking of regularizers does not change.

\looseness=-1
Moreover, there might be correlations between the regularizer %
and other HPs, %
therefore their relative performance may depend
on the distribution of all other HPs.
In fact, we have found two such surprising correlations.
Fig.~\ref{fig:corr_reg_lr_main} shows the performance conditioned on the regularizer used \emph{as well}
as the discriminator learning rate.
We %
notice that for PUGAIL, entropy and no regularization,
the performance significantly increases for lower discriminator learning rates
and the best performing discriminator learning rate ($10^{-6}$)
is in fact 2-2.5 orders of magnitude lower than the best learning rate for
the RL algorithm
($0.0001$--$0.0003$, Figs.~\ref{fig:wide_ppo_learning_rate}, \ref{fig:main_td3_policy_learning_rate}, \ref{fig:main_td3_critic_learning_rate}, \ref{fig:main_sac_learning_rate}, \ref{fig:main_d4pg_learning_rate}).\footnote{
The optimal learning rate for those regularizers was the smallest one
included in the main experiment.
We also run an additional sweep with smaller rates but found
that even lower ones do not perform better (Fig.~\ref{fig:corr_reg_lr_low}).
}
On the other hand, the remaining regularizers are not too sensitive to the discriminator learning rate.
This means that the performance gap between PUGAIL, entropy and no regularization
and the other regularizers
is to some degree caused by the fact that the former ones
are more sensitive to the learning rate and 
may be smaller than suggested by Fig.~\ref{fig:regularizers} if we
adjust for the appropriate choice of the discriminator learning rate.
We can notice that PUGAIL and entropy are the only regularizers which only change the discriminator loss
but do not affect the internals of the discriminator neural network.
Given that they are the only two regularizers benefiting from very low discriminator
learning rate, we suspect that it means that a very low learning rate can play a regularizing role
in the absence of an explicit regularization inside the network.

\looseness=-1
Another surprising correlation is that in some environments,
the regularizer interacts strongly with
observation normalization (described App.~\ref{app:choices-normalization})
employed on discriminator inputs (see Fig.~\ref{fig:corr_reg_obs_ant} for an example on \texttt{Ant}).
These two correlations highlight the difficulty of comparing regularizers,
and algorithmic choices more broadly,
as their performance significantly depends on the distribution of other HPs.

We also supplement our analysis by comparing the performance of different
regularizers for the \emph{best} found HPs.
More precisely, we choose the best value for each HP in the main experiment
(listed in App.~\ref{app:best}) and run them with different regularizers.
To account for the mentioned correlations with the discriminator learning rate
and observation normalization, we also include these two choices in the HP sweep
and choose the best performing variant (as measure by the area under the learning curve)
for each regularizer and each environment.

\begin{figure}[h]
  \centering
  \includegraphics[width=\textwidth]{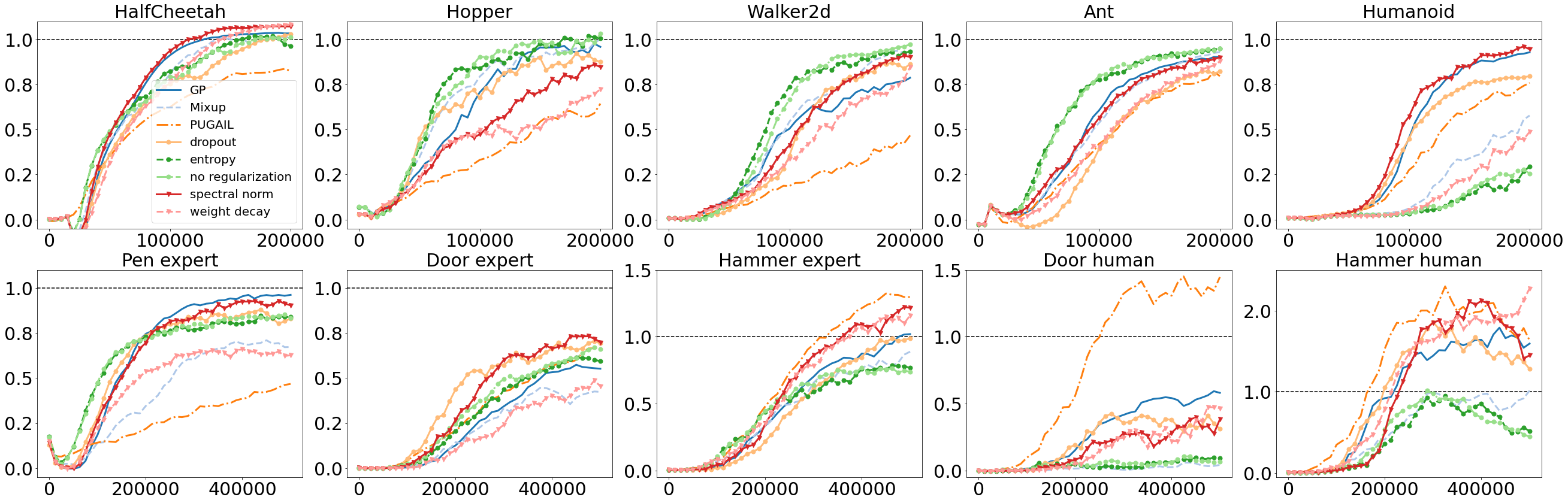}
  \caption{Learning curves for different discriminator regularizers when the other HPs are set to the best performing value across all tasks. The y-axis shows the average policy return normalized
  so that $0$ corresponds to a random policy and $1$ to the expert. See App.~\ref{app:best} for the HPs used. The plots shows the averages across 30 random seeds. Best seen in color.}
  \label{fig:results-best}
\end{figure}

While it is not guaranteed that the performance is going to be good at all
because we greedily choose the best performing value for each HP
and there might be some unaccounted HP correlations,
we find that the performance is very competitive (Fig.~\ref{fig:results-best}).
Notice that we use the same HPs in \emph{all} environments\arxiv{\footnote{
Apart from the discriminator learning rate and observation normalization used.}}
and the performance can be probably improved by varying some HPs between the environments,
or at least between the two environment suites.

\looseness=-1
We notice that on the four easiest tasks (\texttt{HalfCheetah}, \texttt{Hopper}, \texttt{Walker2d}, \texttt{Ant}),
investigated discriminator regularizers  provide no, or only minor performance improvements
and excellent results can be achieved without them.
On the tasks where regularization is beneficial, we usually see that there are multiple
regularizers performing similarly well, with spectral normalization being one of the best regularizers
in all tasks apart from the two tasks with human data where PUGAIL performs better.

\paragraph{Regularizers-specific HPs.}
For GP, the target gradient norm of $1$ is slightly better in most environments but the value of $0$ is significantly better in \texttt{hammer-human} (Fig.~\ref{fig:main__gin_add_gradient_penalty_gradient_penalty_target}),
while the penalty strength of $1$ performs best overall (Fig.~\ref{fig:main__gin_add_gradient_penalty_gradient_penalty_coefficient}).
For dropout, it is important to apply it not only to hidden layers
but also to inputs (Fig.~\ref{fig:main__gin_discriminator__MLP_input_dropout_rate}) and the best results are obtained
for 50\% input dropout and 75\% hidden activations dropout
(Figs.~\ref{fig:main__gin_discriminator__MLP_input_dropout_rate},~\ref{fig:main__gin_discriminator__MLP_hidden_dropout_rate}
and~\ref{fig:main_perf_new_regularizer_0}).
For weight decay, the optimal decay coefficient in the AIL setup is much larger than the values
typically used for Supervised Learning,  the value $\lambda=10$ performs best in our experiments
(Fig.~\ref{fig:main__gin_discriminator__optax_adamw_weight_decay}).
For Mixup, $\alpha=1$ outperforms the other values on almost all tested environments
(Fig.~\ref{fig:main__gin_gail_loss_mixup_alpha}).
For PUGAIL, the unbounded version performs much better on the Adroit suite,
while the bounded version is better on the gym tasks
(Fig.~\ref{fig:main__gin_pugail_loss_pugail_beta}),
and positive class prior of $\eta=0.7$ performs well on most tasks
(Fig.~\ref{fig:main__gin_pugail_loss_positive_class_prior}).
For the discriminator entropy bonus, the values around $0.03$ performed best overall
(Fig.~\ref{fig:main__gin_gail_loss_entropy_coefficient}).
All experiments with spectral normalization enforce the Lipschitz constant of $1$
for each weight matrix.%
\arxiv{\footnote{\looseness=-1 There may be different
Lipschitz constants for different networks depending on those of related activations.}}

\neurips{
\paragraph{How to train efficiently?}
So far we have analysed how HPs affect the
sample complexity of AIL algorithms.
For the analysis of the HPs which influence sample complexity
as well as the computational cost of running an algorithm
see App.~\ref{app:results-trade-offs}.
In particular, we describe there a simple code optimization relying on processing multiple
batches at once
which makes training 2-3x faster in wall clock time without affecting
the sample complexity (Fig.~\ref{fig:steps-per-second}).
}

\section{Are synthetic demonstrations a good proxy for human data?}\label{sec:results-human}

\paragraph{Summary of key findings}
Human demonstrations significantly differ from synthetic ones.
Learning from human demonstrations benefits more from discriminator regularization
and may work better with different discriminator inputs and reward functions
than RL-generated demonstrations.

Using a dataset of human demonstrations comes with a number of additional challenges. Compared to  synthetic demonstrations, the human policy can be multi-modal in that for a given state different decisions might be chosen. A typical example occurs when the human demonstrator remains idle for some time (for example to think about the next action) before taking the actual relevant action: we have two modes in that state, the relevant action has a low probability while the idle action has a very high probability. The human policy might not be exactly markovian either. Those differences are significant enough that the conclusions on synthetic datasets might not hold anymore.

In this section, we focus on the Adroit \texttt{door} and \texttt{hammer} environments for which we run experiments with human as well as synthetic demonstrations.
\footnote{For \texttt{pen}, we only use the ``expert'' dataset,  the ``human'' one
consists of a single (yet very long) trajectory.
} 
Note that on top of the aforementioned challenges, the setup with the Adroit environments using human demonstrations exhibits a few additional specifics. The demonstrations were collected letting the human decide when the task is completed: said in a different way, the demonstrator is offered an additional action to jump directly to a terminal state and this action is not available to the agent imitating the expert. The end result is a dataset of demonstrations of variable length while the agent can only generate episodes consisting of exactly 200 transitions. Note that there was no time limit imposed on the demonstrator and some of the demonstrations have a length greater than 200 transitions. Getting to the exact same state distribution as the human expert may be impossible,
and imitation learning algorithms may have to make some trade-offs.
The additional specificity of that setup is that the reward of the environment is not exactly what the human demonstrator optimized. In the \texttt{door} environment, the reward provided by the environment is the highest when the door is
\emph{fully} opened while the human might abort the task slightly before getting the highest reward. However, overall, we consider the reward provided by the environment as a reasonable metric to assess the quality of the trained policies.
Moreover, in the \texttt{hammer} environment, some demonstrations have a low return and we suspect those are not successful demonstrations.\footnote{D4RL datasets \cite{d4rl} contain only the policy observations
and not the simulator states and therefore it is not straightforward to visualize the demonstrations.}

\looseness=-1
\paragraph{Discriminator regularization.} When comparing the results for RL-generated (\texttt{adroit-expert}\footnote{
We do not include \texttt{pen}  in the \texttt{adroit-expert} plots so that both \texttt{adroit-expert} and \texttt{adroit-human}
show the results averages across the \texttt{door} and \texttt{hammer} tasks and differ only in the demonstrations used.}) and human demonstrations (\texttt{adroit-human})
we can notice differences on a number of HPs related to the discriminator training.
Human demonstrations benefit more from using discriminator regularizers (Fig.~\ref{fig:main__gin_regularizer__macro_value})
and they also work better with smaller discriminator networks (Fig.~\ref{fig:main__gin_discriminator__MLP_num_units})
trained with lower learning rates (Fig.~\ref{fig:main__gin_discriminator__optax_adamw_learning_rate}).
The increased need for regularization suggest that it is easier to overfit to the idiosyncrasies of human demonstrations than to those of RL policies.

\begin{figure}[h]
  \centering
    \subfloat[\centering discriminator input]{{ \includegraphics[width=6.5cm,valign=t]{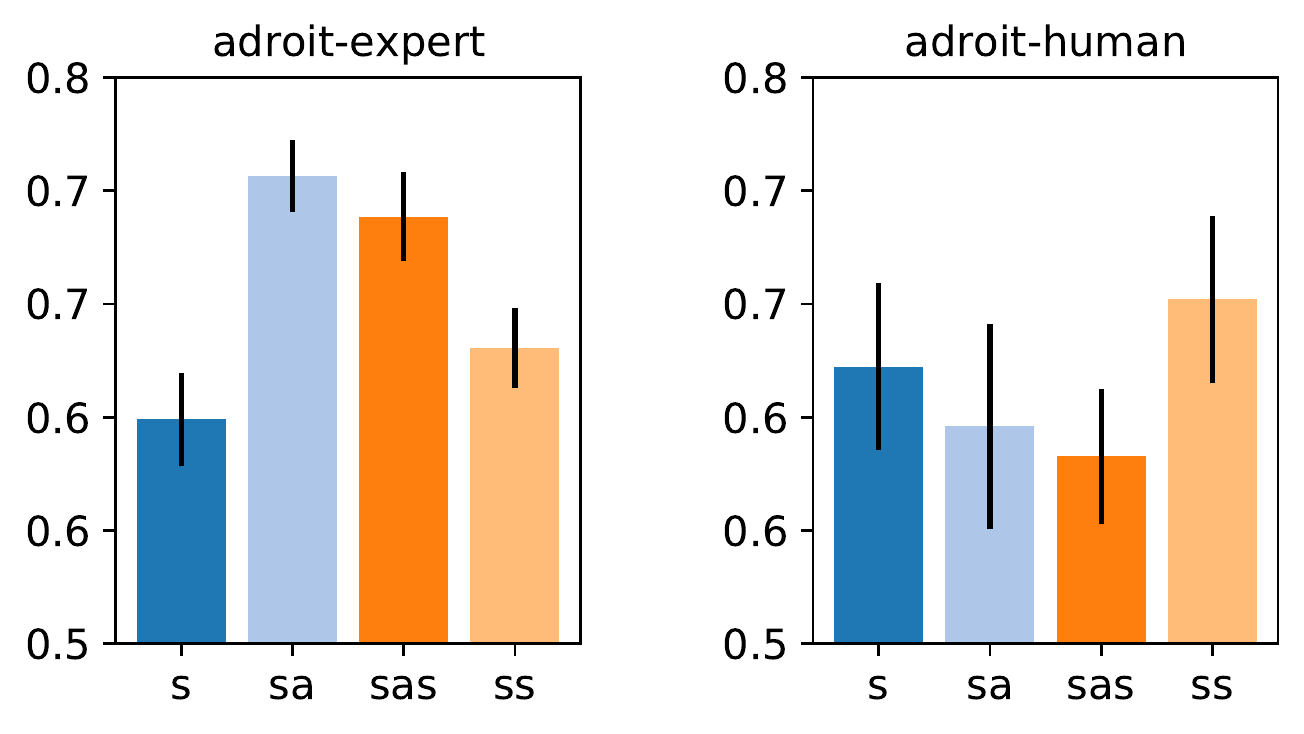}
    \vphantom{\includegraphics[width=6.5cm,valign=t]{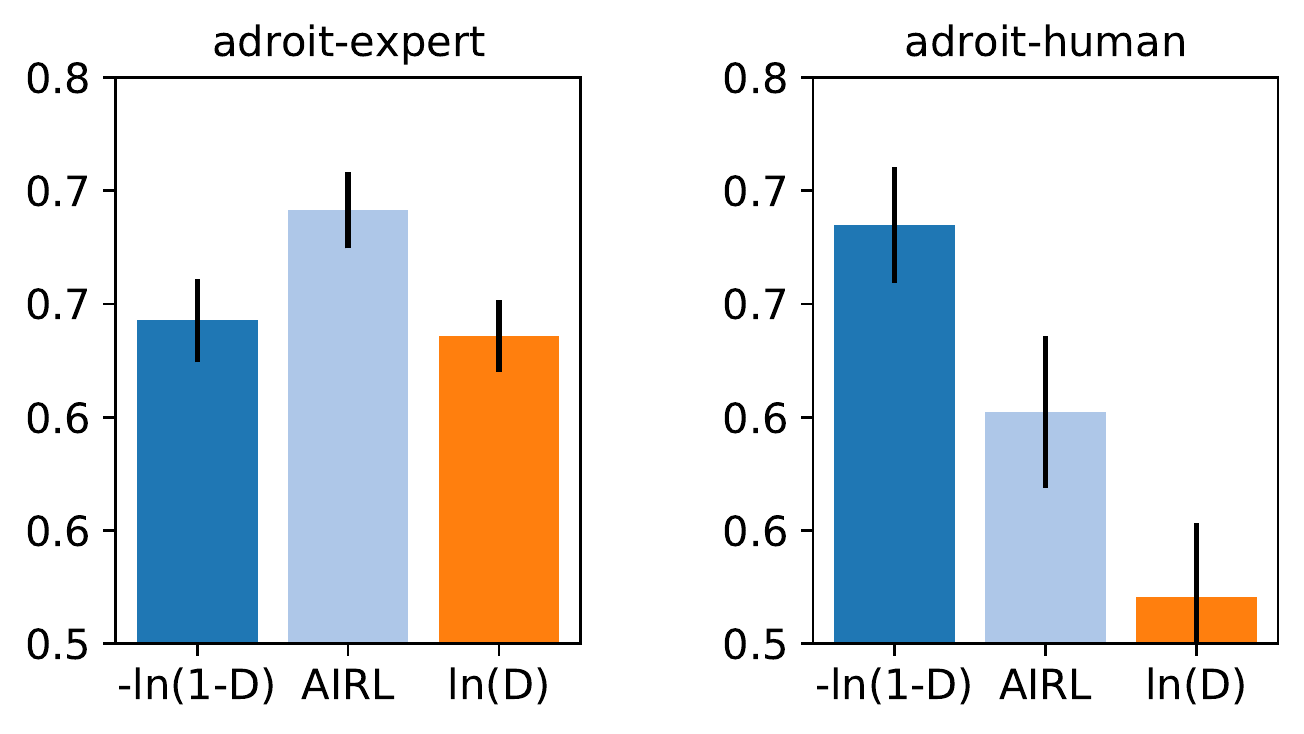}} }}
    \qquad
    \subfloat[\centering reward functions]{{\includegraphics[width=6.5cm,valign=t]{custom_plots/human_perf__gin_reward_function__macro_value_height_3.pdf}}}%
    \caption{Comparison of discriminator inputs (a) and reward functions (b)
    for environments with human demonstrations.
    See Fig.~\ref{fig:main__gin_discriminator_input__macro_value} and
    Fig.~\ref{fig:main__gin_reward_function__macro_value} for the individual results in all environments.}%
  \label{fig:results-human}
\end{figure}

\looseness=-1
\paragraph{Discriminator input.}
Fig.~\ref{fig:results-human}a shows the performance given the
discriminator input depending on the demonstration source.
For most tasks with RL-generated demonstrations, %
feeding actions as well as states improves the performance (Fig.~\ref{fig:main__gin_discriminator_input__macro_value}).
Yet, the opposite holds when human demonstrations are used.
We suspect that it might be caused by the mentioned issue with demonstrations lengths which forces the policy to repeat a similar movement
but with a different speed than the demonstrator.

\paragraph{Reward functions.}
Finally, we look at how the relative performance of different reward functions depends on
the demonstration source.
Fig.~\ref{fig:results-human}b shows that for RL-generated demonstrations the best reward function is AIRL
while $-\ln(1-D)$ performs better with human demonstrations.
Under the assumption that the discriminator is optimal,
these two reward functions correspond to the minimization of different divergences
between the state (or state-action depending on the discriminator input)
occupancy measures of the policy \neurips{(denoted $\pi$)} and the expert \neurips{(denoted $E$)}
\neurips{.}
\arxiv{ --- See
Table~\ref{table:divergences} for the details.

\begin{table}[h]
  \caption{Reward functions and corresponding divergences.
  $\pi$ and $E$ denote the state occupancy measures of, respectively, the policy and the expert.
  The proofs can be found in \cite{gail} and \cite{fairl}.
  The \emph{bounded} and \emph{symmetric} column show whether the given \emph{divergence}
  is bounded or symmetric.
  $D$ denotes the probability of being classified as \emph{expert}.}
  \label{table:divergences}
  \centering
  \begin{tabular}{lllll}
    \toprule
    Paper & Reward     & Divergence     & Bounded & Symmetric \\
    \midrule
    GAIL \cite{gail} & $-\ln(1-D)$ & $\JS(\pi||E)$ & \checkmark & \checkmark \\
    AIRL \cite{airl} & $\ln(D)-\ln(1-D)$ & $\KL(\pi||E)$ & $\times$ & $\times$  \\
    \bottomrule
  \end{tabular}
\end{table}
}

The reward function performing best with human demonstrations ($-\ln(1-D)$)
corresponds to the minimization of the Jensen-Shannon divergence (proof in \cite{gail}).\arxiv{\footnote{
$\JS(P||Q)=\KL(P||M)+KL(Q||M)$, where $M=\frac{P+Q}{2}$.}}
Interestingly, this divergence is symmetric ($\JS(\pi||E)=\JS(E||\pi)$)
and bounded ($0 \le \JS(\pi||E) \le \ln(2)$).
For AIL, the symmetry means that it penalizes the policy for doing things the expert never does
with exactly the same weight as for not doing some of the things the expert does
while the boundedness means that the penalty for not visiting a single state is always finite.
We suspect that this boundedness is beneficial for learning with human demonstrations
because it may not be possible to exactly match the human distribution for the reasons explained earlier.

In contrast to Jensen-Shannon, the $\KL(\pi||E)$ divergence
which is optimized by the AIRL reward (proof in \cite{fairl}) is neither symmetric, nor bounded
--- it penalizes the policy much more heavily for doing the things the expert never does
that for not doing all the things the expert does and the penalty for visiting a single state the expert never visits
is infinite (assuming a perfect discriminator).

While it is hard to draw any general conclusions only from the two investigated environments
for which we had access to human demonstrations,
our analysis shows that the differences between synthetic and human-generated demonstrations
can influence the relative performance of different algorithmic choices.
This suggests that RL-generated data are not a good proxy for human demonstrations
and that the very common practice of evaluating IL %
only with synthetic demonstrations
may lead to algorithms which perform poorly in the more realistic scenarios with human demonstrations.

\arxiv{
\section{How to train efficiently?}\label{sec:results-trade-offs}
So far we have analysed how HPs affect the
performance of AIL algorithms measured after fixed numbers of
environment steps.
Here we look at the HPs which influence sample complexity
as well as the computational cost of running an algorithm.
Raw experiment report can be found in App.~\ref{exp_tradeoffs}.

\paragraph{Batch size and replay ratio.} One of the main factors influencing the throughput of a particular
imitation algorithm is the number of times each transition is
replayed on average and the batch size used.\footnote{We use the same batch size
for the policy and actor networks while the discriminator batch size is effectively two times
larger because its batches contain always \choicet{batchsize} demonstration transitions
and \choicet{batchsize} policy transitions.
The replay ratio is the same for all networks with the exception of the
discriminator which can have its replay ratio doubled
depending on the value of
\choicet{discriminatortorlupdatesratio}.
See App.~\ref{app:choices-d-training} for details.}
See App.~\ref{app:choices-algorithm} for the detailed description of the
HPs involved.
Fig.~\ref{fig:tradeoffs_batch_size} shows that smaller batches perform overall better
(given a fixed replay ratio)
and increasing the replay ratio improves the performance, at least
up to some threshold depending on the environment
(Fig.~\ref{fig:tradeoffs_samples_per_insert}).
There is a very strong correlation between the two HPs
--- Fig.~\ref{fig:corr_batch_replay}
shows that for most batch sizes, the optimal replay ratio is equal to the batch size,
which corresponds to replaying exactly one batch of data per environment step.
If we compare different batch sizes under the ratio of batches to environment steps fixed to one,
the performance is mostly independent of the batch size (Fig.~\ref{fig:corr_batch_replay}).

While in most of our experiment the discriminator and the RL
agent are trained with exactly the same number of batches,
we also tried doubling the number of discriminator batches.
Fig.~\ref{fig:tradeoffs_discriminator_to_rl_updates_ratio}
shows that it improves the performance slightly on the Adroit suite.

\paragraph{Combining multiple batches.}
We also consider processing multiple batches at once for improved accelerator (GPU or TPU) utilization.
In particular, we sample
an $N$-times larger batch from a replay buffer, split it back into $N$ smaller/proper batches on an accelerator, and process them sequentially.
In order to keep the replay ratio unaffected, we decrease the frequency of updates
accordingly, e.g. instead of performing one gradient update for every environment step, we perform
$N$ gradients updates every $N$ environment steps.
We apply this technique to the discriminator as well as the RL agent training.
The effect on the sample complexity of the algorithm can be seen in Fig.~\ref{fig:tradeoffs_grad_updates_per_batch}. There is a small negative effect for values larger or equal to 16. The effect of this parameter on the throughput of our system could be observed in Fig.~\ref{fig:steps-per-second}. The value of 8 provides a good compromise: almost no noticeable
sample complexity regression while decreasing the training time by 2--3 times.

\begin{figure}[h]
  \centering
  \includegraphics[width=\textwidth]{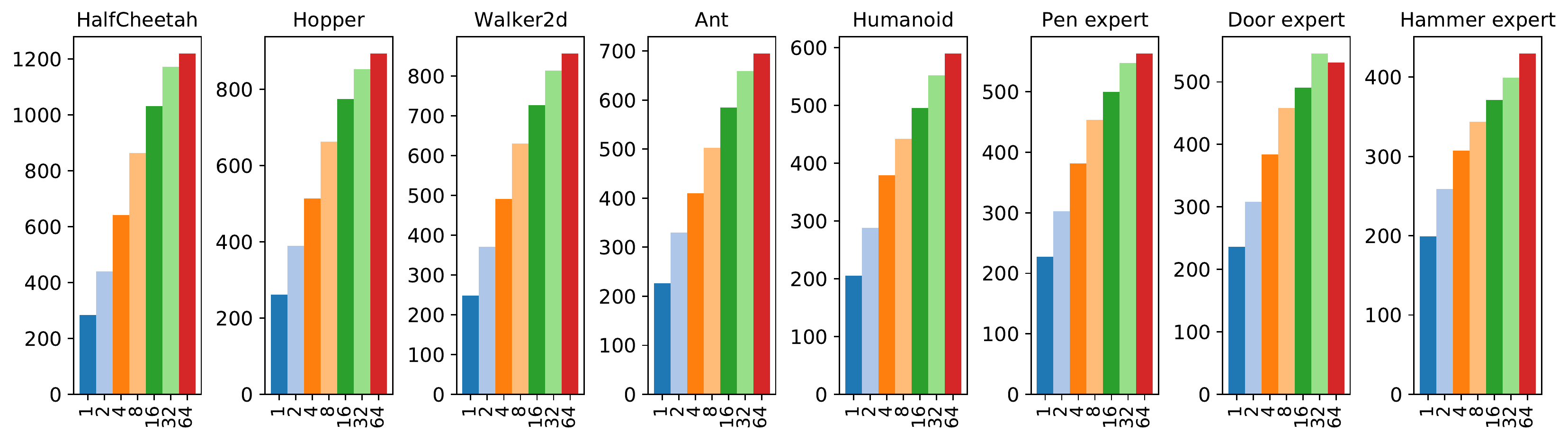}
  \caption{Training speed (in terms of environment steps per second) for
  combining multiple batches.
  The x-axis denotes the number of batches combined.
  Other HPs are set to the best performing value across all tasks (listed in App.~\ref{app:best}).
  The plots shows the averages across 10 random seeds.}
  \label{fig:steps-per-second}
\end{figure}
}

\section{Related work}\label{sec:related}

The most similar work to ours is probably \cite{lipschitzness}
which compares the performance of different discriminator regularizers and
concludes that gradient penalty is necessary for achieving good performance with off-policy
AIL algorithms. In contrast to \cite{lipschitzness}, which uses a single HP configuration,
we run large-scale experiments with very wide HP sweeps which allows us to
reach more robust conclusions.
In particular, we are able to achieve excellent sample complexity
on all the environments used in \cite{lipschitzness}\arxiv{\footnote{
\cite{lipschitzness} evaluated gradient penalty in the off-policy setup in the following environments:
\texttt{HalfCheetah}, \texttt{Hopper}, \texttt{Walker2d} and \texttt{Ant}, as well as \texttt{InvertedPendulum}
which we did not use due to its simplicity.}} without using any explicit
discriminator regularizer (Fig.~\ref{fig:results-best}).

\arxiv{
Another empirical study of IL algorithms
is \cite{hpil}, which investigates the problem of HP selection in IL
under the assumption that the reward function is not available for the HP selection.
}

The methodology of our study is mostly based on \cite{mujoco123}
which analyzed the importance of different choices for
on-policy actor-critic methods.
Our work is also similar to other large-scale studies
done in other fields of Deep Learning, e.g.
model-based RL \cite{langlois2019benchmarking},
GANs \cite{lucic2018gans}, NLP \cite{kaplan2020scaling}, disentangled representations~\cite{locatello2018challenging}
and convolution network architectures \cite{radosavovic2020designing}.

\section{Conclusions}\label{sec:conclusions}
In this empirical study, we investigate in depth many aspects of the AIL framework %
including discriminator architecture, training and regularization %
as well as many choices related to the agent training.
Our key findings can be divided into three categories:
(1)~Corroborating prior work, e.g. for the underlying RL problem, off-policy algorithms are more sample efficient than on-policy ones; %
(2)~Adding nuances to previous studies, e.g. while the regularization schemes
encouraging Lipschitzness improve the performance,
more classical regularizers like dropout or weight decay often perform on par;
(3)~Raising concerns: we observe a high discrepancy between the results for RL-generated and human data.
We hope this study will be helpful to anyone using or designing AIL algorithms.  

\arxiv{
\section*{Acknowledgments}
We thank Kamyar Ghasemipour for the discussions related to the FAIRL reward function
and Lucas Beyer for the feedback on an earlier version of the manuscript.
}

\newpage
{
\printbibliography
}

\appendix

\neurips{\section*{Checklist}

\begin{enumerate}

\item For all authors...
\begin{enumerate}
  \item Do the main claims made in the abstract and introduction accurately reflect the paper's contributions and scope?
    \answerYes
  \item Did you describe the limitations of your work?
    \answerYes
  \item Did you discuss any potential negative societal impacts of your work?
    \answerNA{We study general purpose imitation algorithms which are not related to any particular application.}
  \item Have you read the ethics review guidelines and ensured that your paper conforms to them?
    \answerYes
\end{enumerate}

\item If you are including theoretical results...
\begin{enumerate}
  \item Did you state the full set of assumptions of all theoretical results?
    \answerNA
	\item Did you include complete proofs of all theoretical results?
    \answerNA
\end{enumerate}

\item If you ran experiments...
\begin{enumerate}
  \item Did you include the code, data, and instructions needed to reproduce the main experimental results (either in the supplemental material or as a URL)?
    \answerNo
  \item Did you specify all the training details (e.g., data splits, hyperparameters, how they were chosen)?
    \answerYes
	\item Did you report error bars (e.g., with respect to the random seed after running experiments multiple times)?
    \answerYes
	\item Did you include the total amount of compute and the type of resources used (e.g., type of GPUs, internal cluster, or cloud provider)?
    \answerNo
\end{enumerate}

\item If you are using existing assets (e.g., code, data, models) or curating/releasing new assets...
\begin{enumerate}
  \item If your work uses existing assets, did you cite the creators?
    \answerYes
  \item Did you mention the license of the assets?
    \answerNo{D4RL demonstrations are available under Apache License 2.0}
  \item Did you include any new assets either in the supplemental material or as a URL?
    \answerNo
  \item Did you discuss whether and how consent was obtained from people whose data you're using/curating?
    \answerNA
  \item Did you discuss whether the data you are using/curating contains personally identifiable information or offensive content?
    \answerNA
\end{enumerate}

\item If you used crowdsourcing or conducted research with human subjects...
\begin{enumerate}
  \item Did you include the full text of instructions given to participants and screenshots, if applicable?
    \answerNA
  \item Did you describe any potential participant risks, with links to Institutional Review Board (IRB) approvals, if applicable?
    \answerNA
  \item Did you include the estimated hourly wage paid to participants and the total amount spent on participant compensation?
    \answerNA
\end{enumerate}

\end{enumerate}
}

\clearpage
\begin{spacing}{0.1}
\tableofcontents
\end{spacing}

\neurips{
\section{How to train efficiently?}\label{app:results-trade-offs}

}

\section{Reinforcement Learning Background}\label{app:rl}

We consider the standard reinforcement learning formalism
consisting of an agent interacting with an environment.
To simplify the exposition we assume in this section that the environment is fully observable.
An environment
is described by
a set of states $\State$,
a set of actions $\Action$,
a distribution of initial states $p(s_0)$,
a reward function $r : \State \times \Action \rightarrow \R$,
transition probabilities $p(s_{t+1}|s_t,a_t)$ ($t$ is a timestep index explained later),
termination probabilities $T(s_t,a_t)$
and a discount factor $\gamma \in [0,1]$.

A policy $\pi$ is a mapping from state to a distribution over actions.
Every episode starts by sampling an initial state $s_0$.
At every timestep $t$ the agent produces an action based on the current state:
$a_t \sim \pi(\cdot|s_t)$.
In turn, the agent receives a reward $r_t=r(s_t,a_t)$ and the environment's state is updated.
With probability $T(s_t,a_t)$ the episode is terminated, and otherwise the
new environments state $s_{t+1}$ is sampled from $p(\cdot|s_t,a_t)$.
The discounted sum of future rewards, also referred to as the \emph{return}, is defined as
$R_t=\sum_{i=t}^\infty \gamma^{i-t} r_i$.
The agent's goal is to find the policy $\pi$ which maximizes the expected return $\E_\pi [R_0|s_0]$, where
the expectation is taken over the initial state distribution, the policy, and environment transitions accordingly to the dynamics
specified above.

\section{Adversarial Imitation Learning Background}\label{app:gail}
See App.~\ref{app:rl} for a very brief introduction to RL and the notation used in this section.

Drawing inspiration from Inverse Reinforcement Learning \cite{ng2000algorithms, maxent} and Generative Adversarial Networks (GANs, \cite{goodfellow2014generative}), adversarial imitation learning \cite{gail}
aims at learning a behavior similar to that of the expert given a set of expert demonstrations $\mathcal{D}_{expert}$
and the ability to interact with the environment.

To do so, the agent with policy $\pi$ is initialized randomly and interacts with the environment.
A discriminator network $D$ is trained to distinguish between samples
coming from the agent $(s_t, a_t, s_{t+1}) \sim \mathcal{D}_{\pi}$
and samples coming from the expert dataset $(s_t, a_t, s_{t+1}) \sim \mathcal{D}_{expert}$ with a cross-entropy loss.
A reward function for the policy is then defined based on the discriminator prediction, e.g.
$r(s,a)=-\ln(1-D(s,a))$, where $D(s,a)$ denotes the probability of classifying the
state-action pair as expert by the discriminator.
The agent is then trained with an RL algorithm to maximize this reward and thus fool the discriminator.
As in GANs, the training of the discriminator and that of the agent (here playing the role of the \emph{generator}) are interleaved.
Therefore, at the high level, the algorithm repeats the following steps in a loop:
(1) interact with the environment using the current policy and store the experience in a replay buffer,
(2) update the discriminator,
(3) perform an RL update accordingly to the RL algorithm used.

\section{List of Investigated Choices}\label{app:choices}

In this section we list all algorithmic choices which we consider in our experiments.
See App.~\ref{app:gail} for an introduction to adversarial imitation and the notation used
in this section.
For convenience, we mark each of the choices with a number (e.g., \choicep{directrlalgorithm}) and a fixed name (e.g. \choicet{directrlalgorithm})
that can be easily used to find a description of the choice in this section.

\subsection{Reinforcement Learning algorithms}\label{app:choices-algorithm}

In all experiments we use MLPs for the policy and critic/value networks
and sample the following HPs controlling the networks architectures:
\dchoicet{numpolicylayers} (the number of \emph{hidden} layers),
\dchoicet{policylayersize},
\dchoicet{numcriticlayers},
\dchoicet{criticlayersize},
\dchoicet{activation},
as well as \dchoicet{discount} and \dchoicet{batchsize}.
All networks are optimized with the Adam \cite{adam} 
optimizer.

We sample \dchoicet{directrlalgorithm} from the following options:

\paragraph{Proximal Policy Optimization (PPO, \cite{ppo})}\label{app:choices-ppo}
For PPO, \choicet{batchsize} denotes the number of experience
fragments, each of consisting \dchoicet{ppounrolllength} transitions, collected in each policy update step.
In each policy update step, we perform \dchoicet{pponumepochs}
passes over the gathered data when in each pass the data is split into
\dchoicet{pponumminibatches} minibatches.
We use the PPO loss with the clipping threshold set by
\dchoicet{ppoclippingepsilon} and add an entropy loss with the
coefficient specified by \dchoicet{ppoentropycost}.
We also sample \dchoicet{ppolearningrate}, 
and the GAE \cite{gae} returns mixing coefficient \dchoicet{ppogaelambda}.

\paragraph{Soft Actor Critic (SAC, \cite{sac})}\label{app:choices-sac}
We use a version of SAC with a policy entropy constraint \cite{sac2}.
In particular, we choose \dchoicet{sactargetentropyperdimension}
and that set the entropy constraint so that the policy entropy is not lower than
the number of action dimensions times this value.
We also sweep \dchoicet{saclearningrate} and the target network polyak averaging coefficient
\dchoicet{sactau}
(the target network is updates after each minibatch).

\paragraph{Twin Delayed Deep Deterministic Policy Gradient (TD3, \cite{td3})}\label{app:choices-td3}
For TD3, we sweep
\dchoicet{tdtpolicylearningrate} and
\dchoicet{tdtcriticlearningrate} separately,
as well as sample \dchoicet{rlsigma}.
Following the original publication,
we update the actor only using every other minibatch while the critic networks uses all minibatches.
The target network is updated after every minibatch with the polyak
coefficient fixed to $0.005$.
Following DAC \cite{dac}, we clip actor gradients with magnitudes bigger than
\dchoicet{tdtgradientclipping}.

\paragraph{Distributed Distributional Deterministic Policy Gradients
(D4PG, \cite{d4pg})}\label{app:choices-d4pg}
This algorithm is similar to TD3 but uses a distributional C51-style critic \cite{c51}
outputting distributions over
\dchoicet{numatoms} atoms spaced equally between
-\dchoicet{vmax} and \choicet{vmax}
as well as \dchoicet{nstep} returns.
In contrast to the original D4PG \cite{d4pg},
we use a single actor and do not use prioritized replay.
The target network is fully updated every $100$ training batches.
As usual, we also sweep \dchoicet{dfpglearningrate}.

Moreover, for off-policy algorithm (SAC, TD3 and D4PG) we
sample \dchoicet{samplesperinsert} which denotes the average
number of times each transition is replayed.
This is achieved in the following way ---
if \choicet{samplesperinsert} $\ge$ \choicet{batchsize}
than we replay \choicet{samplesperinsert} $/$ \choicet{batchsize}
batches (each with \choicet{batchsize} transitions)
after every environment step.
If \choicet{batchsize} $>$ \choicet{samplesperinsert},
we replay a single batch every \choicet{batchsize} $/$ \choicet{samplesperinsert}
transitions.
The transitions for replay are sampled uniformly from a FIFO
replay buffer of size \dchoicet{maxreplaysize}
and we start training whenever we have at least 10k transition in the buffer.

For the RL algorithms which train stochastic policies (PPO and SAC)
we use a Gaussian distribution followed by tanh to squash actions into
the $[-1, 1]$ range.\footnote{The action coordinates are scaled to $[-1, 1]$ regardless of the RL algorithm used.}
More precisely, the policy network output is split into two parts ---
$\mu$ and $\rho$, and the action distribution used during training is $\tanh(\mathcal{N}(\mu,\,\softplus(\rho)+0.001))$.
For policy evaluation, we choose \dchoicet{evalbehaviorpolicytype}
from the following options:
\begin{itemize}
    \item \emph{stochastic}: sample from the distribution (same as behavioral policy used during training),
    \item \emph{mode}: use the mode of the Gaussian instead of sampling,
    \item \emph{average}: sample five action from the distribution and take the average of them.
\end{itemize}

\subsection{Imitation-specific changes to RL}\label{app:choices-imitation-rl}

\paragraph{Reward function} Let $D$ denote the probability that a state-action pair $(s,a)$ is classified as \emph{expert} by the discriminator
while $h$ is the discriminator logit, i.e. $D=\sigma(h)$ where $\sigma$ denotes the sigmoid function.
Depending on the value of \dchoicet{gailreward} we use one of the following reward functions
(for completeness we write the formulas as a function of $D$ as well as $h$):
\begin{itemize}
\item $r(s,a)=-\ln (1-D)=\softplus(h)$ (used in the original GAIL paper\footnote{The GAIL paper uses
the inverse convention in which $D$ denotes the probability as being classified as \emph{non-expert}.} \cite{gail}),
\item $r(s,a)=\ln D-\ln (1-D)=h$ (introduced in AIRL \cite{airl}).
\item $r(s,a)=\ln D=-\softplus(-h)$,
\item $r(s,a)=-he^h$ (introduced in FAIRL \cite{fairl}).
\end{itemize}
We also clip rewards with the absolute values higher than \dchoicet{gailmaxrewardmagnitude}.

\paragraph{Absorbing state}
We optionally (if \dchoicet{explicitabsorbingstate}\texttt{=True})
apply the absorbing state technique from DAC \cite{dac}.
This technique encourages the agent to generate episodes of similar length
to the ones of the expert.
In particular, the demonstration and agent episodes are processed in the following way:
for each terminal transition, we replace it with a non-terminal
transition to a special absorbing state\footnote{In practice, this is done by adding a special
bit to every observation which is set to zero for normal observations and one for
the absorbing state. The remaining bits of the absorbing state are all zeros.}
and also add a transition from the absorbing state to itself with a zero action.

\paragraph{Replaying demonstrations}
For off-policy RL algorithms, we optionally (if \dchoicet{expertreplay}$\not = \infty$) sample
batches for RL training not only from the replay buffer,
but also from the demonstrations.
In particular, the ratio of policy to expert data in each minibatch
is equal to \choicet{expertreplay}.

\paragraph{Initialization with behavior cloning}
We optionally (if \dchoicet{pretrainwithbc}\texttt{=True})
pre-train the policy network offline at the beginning of training using Behavior Cloning \cite{bc}.
In particular, we perform 100k gradient steps with Adam on the MSE loss,
using learning rate $10^{-4}$ and batch size $256$.

\subsection{Discriminator parameterization}\label{app:choices-discriminator-param}

Depending on the value of \dchoicet{gailinput}, the discriminator
is fed single states, state-action pairs, state-state pairs or state-action-state tuples.

Our basic discriminator architecture is an MLP with \dchoicet{gailmlpnumlayers}
hidden layers, each of size \dchoicet{gailmlpnumwidth}
with the activation function specified by \dchoicet{gailmlpactivation}.
Its output is interpreted as the logit of the probability of being classified as expert,
i.e. for a state-action-state tuple $(s,a,s')$ we have
$D(s,a,s')=\sigma(f(s,a,s'))$, where $D$ is the probability of classifying the tuple $(s,a,s')$
as expert, $\sigma$ denotes the sigmoid function,
and $f$ is a learnable function represented as an MLP.

We also consider two modifications introduced in the AIRL \cite{airl} paper.
The first one (enabled if \dchoicet{gaildiscriminatormodule}\texttt{=True}) adds a reward shaping
term where the $f$ function is parameterized in the following way:
$f(s,a,s')=g(s,a,s')+\gamma h(s')-h(s)$ where $g$ and $h$ are MLPs
parameterized as described above,
and $\gamma$ is the RL discount factor.\footnote{The inputs fed to $g$ are specified by \choicet{gailinput}.}
The second modification (enabled if \dchoicet{subtractlogp}\texttt{=True}) parameterizes the discriminator as
$D(s,a,s')=\frac{\exp(f(s,a,s'))}{\exp(f(s,a,s'))+\pi(a|s)}$,
where $\pi$ is the current agent policy.
It can be easy shown that it is equivalent to $D(s,a,s')=\sigma((f(s,a,s')-\log \pi(a|s))$
so this just shifts the logits by $\log \pi(a|s)$.

\subsection{Discriminator training}\label{app:choices-d-training}

All discriminator weight matrices use the \texttt{lecun\_uniform} initializer from JAX \cite{jax}.
The last discriminator layer initialization is additionally multiplied by \dchoicet{gailmlplastlayerinitscale}.

The discriminator is trained with the Adam \cite{adam} optimizer,
the learning rate specified by \dchoicet{gaildiscriminatorlearningrate}
and the cross-entropy loss. %
Each data batch contains exactly \choicet{batchsize} expert transitions
and \choicet{batchsize} policy transitions.
The policy transitions are sampled uniformly from a FIFO replay buffer
of size \dchoicet{gailmaxreplaysize}.

We perform
\dchoicet{discriminatortorlupdatesratio}
discriminator gradient steps for each RL gradient step.
More precisely, after each environment step,
we compute the number of RL gradient steps as described in App.~\ref{app:choices-algorithm},
and perform \choicet{discriminatortorlupdatesratio} that many discriminator gradient steps
\emph{before} performing the RL update.

\subsection{Discriminator regularization}\label{app:choices-regularizers}
Depending on the value of \dchoicet{regularizer},
we optionally apply one of the following regularizers to the discriminator:

\paragraph{Gradient Penalty (GP, \cite{gp})}

Gradient penalty is parameterized with \dchoicet{gptarget}
and \dchoicet{gpcoef}.
This regularizer adds an extra term in the discriminator loss that encourages
the discriminator gradient to be close to $k$
on a convex combination of positive (expert) and negative (policy) data.
In particular, for an expert data $x\sim \mathcal{D}_{expert}$ and policy data $\widetilde{x}\sim \mathcal{D}_{\pi}$,
the gradient penalty is defined as
$\lambda (||\nabla_{\hat{x}} D(\hat{x})||_2-k)^2$,
where $\hat{x}$ is a convex combination of $x$ and $\widetilde{x}$, i.e.
$\hat{x}:= \epsilon x +(1-\epsilon)\widetilde{x}$
and $\epsilon$ follows a uniform distribution: $\epsilon \sim U[0,1]$.
In practice,
$k$ is usually chosen to be $0$ (penalty for high gradients) or $1$ (penalty for gradients with norms far from $1$).
Our gradient penalty implementation uses the gradient of the discriminator
logit instead of the classification probability.

\paragraph{Spectral normalization \cite{spectral}} 

Spectral normalization guarantees that the discriminator is 1-Lipschitz: $|D(x_2) - D(x_1)| \leq ||x_2 -x_1||$.
It does so by dividing each dense layer matrix by its highest eigenvalue which can be efficiently computed
with the power iteration method. See \cite{spectral} for details.

\paragraph{Mixup \cite{mixup}}

Mixup is parameterized with \dchoicet{mixupalpha}
and relies on training the discriminator on a convex combination of positive (expert) and negative (policy) data.
With expert data $x\sim \mathcal{D}_{expert}$ and policy data $\widetilde{x}\sim \mathcal{D}_{\pi}$,
let $\epsilon$ follow a Beta distribution: $\epsilon \sim Beta(\alpha, \alpha)$.
Instead of training the discriminator on $x$ and $\widetilde{x}$ separately,
we only train it on the convex combination of them
$\hat{x}:= \epsilon x +(1-\epsilon)\widetilde{x}$
with the label being the convex combinations of the labels,
i.e. expert with probability $\epsilon$ and non-expert
with probability $1-\epsilon$, so that the loss is $-\epsilon \ln D(\hat{x}) -(1-\epsilon) \ln (1-D(\hat{x}))$.

\paragraph{Positive Unlabeled GAIL (PUGAIL, \cite{pugail})}

Normally the discriminator is trained under the assumption that expert trajectories are positive examples and policy trajectories are negative examples. The PUGAIL loss assumes instead that policy trajectories are a mix of positive and negative examples.

With \dchoicet{pugailpositiveclassprior} denoting the assumed proportion of positive samples in the policy data and
\dchoicet{pugailbeta} being a clipping threshold, the discriminator is trained with the following loss:
$$\eta\hat{\E}_{x\sim \mathcal{D}_{expert}}[-\ln(D(x))] + \max \left( -\beta,\, \hat{\E}_{x\sim \mathcal{D}_{\pi}}[-\ln(1-D(x))] - \eta\hat{\E}_{x\sim \mathcal{D}_{expert}}[-\ln(1 -D(x))] \right).$$

\paragraph{Dropout \cite{dropout}}
We apply dropout to the hidden layers (\dchoicet{dropouthiddenrate})
as well as inputs (\dchoicet{dropoutinputrate}).
See \cite{dropout} for the description of dropout.

\paragraph{Weight decay \cite{decay, adamw}}
Weight decay is parameterized with a parameter controlling its strength
\dchoicet{regweightdecay}.
Normally, weight decay is applied by adding a sum of the squares of the
network parameters to the loss.
However, this may interact negatively with an adaptive gradient optimizer like Adam \cite{adam}
unless the optimizer is modified appropriately \cite{adamw}.
In our experiments, we use a version of Adam with weight decay called
AdamW \cite{adamw} from the Optax library \cite{optax}.
See \cite{adamw} for the details.

\paragraph{Entropy bonus}
Similarly to entropy bonus in RL,
we also experiment with adding to the discriminator loss
a term proportional to the entropy of the discriminator output treated as a Bernoulli distribution:
$\lambda \left(D \ln D + (1-D) \ln (1-D) \right)$
where \dchoicet{regentropycoef} is a HP.

\subsection{Observation normalization}\label{app:choices-normalization}
We optionally apply input normalization
(choice \dchoicet{obsnormalization})
which transforms linearly the
observations to all neural networks
(in the RL algorithm
as well as the discriminator)
so that each coordinate has approximately
mean equal zero and standard deviation equal one.
This is done by subtracting from each observation
$\mu$ and dividing by $\max(\rho, 0.001)$,
where $\mu$ and $\rho$ are the empirical mean and standard deviation of either all demonstrations (we call it \emph{fixed} normalization because it does not change during training)
or the empirical mean and standard deviation of all the observations
encountered by the policy being trained so far (called \emph{online}
because it changes during training).

\subsection{Combining multiple batches}
We consider processing multiple batches at once for improved accelerator (GPU or TPU) utilization (choice \dchoicet{gradupdatesperbatch}).
In particular, we sample
an $N$-times larger batch from a replay buffer, split it back into $N$ smaller/proper batches on an accelerator, and process them sequentially.
In order to keep the replay ratio unaffected, we decrease the frequency of updates
accordingly, e.g. instead of performing one gradient update for every environment step, we perform
$N$ gradients updates every $N$ environment steps.
We apply this technique to the discriminator as well as the RL agent training.

\clearpage\section{Best hyperparameter values}\label{app:best}
\label{sec:default-settings}
\label{sec:default_settings}
Table~\ref{table:best} shows the best value found for each HP in the main experiment.
See App.~\ref{exp_main} for the full experimental report.
The sample complexity can be slightly improved
by decreasing \choicet{gradupdatesperbatch}
and increasing \choicet{discriminatortorlupdatesratio}.
We used the suboptimal values from Table~\ref{table:best}
because they give a good trade-off between sample complexity and runtime.
\choicet{gaildiscriminatorlearningrate} equal $10^{-6}$
is better when PUGAIL, entropy or no discriminator regularizer is used,
and $3 \cdot 10^{-5}$ is better otherwise.
The performance of observation normalization schemes
depends heavily on the environment \emph{and}
discriminator regularization used.
For completeness, we present the best HPs
for all discriminator regularizers.

\begin{table}[h]
  \caption{Best hyperparameter configuration.}
  \label{table:best}
  \centering
  \begin{tabular}{clr}
    \toprule
    Choice & Name & Best value \\
    \midrule
\choicetable{numpolicylayers} & $2$ \\
\choicetable{policylayersize} & $256$ \\
\choicetable{numcriticlayers} & $2$\\
\choicetable{criticlayersize} & $256$ \\
\choicetable{activation} & ReLu \\
\choicetable{discount} & $0.97$ \\
\choicetable{batchsize} & $256$ \\
\choicetable{directrlalgorithm} & \texttt{SAC} \\

\choicetable{sactargetentropyperdimension} & $-0.5$ \\
\choicetable{saclearningrate} & $3\cdot10^{-4}$ \\
\choicetable{sactau} & $0.01$ \\

\choicetable{samplesperinsert} & $256$ \\
\choicetable{maxreplaysize} & $3\cdot10^6$ \\
\choicetable{evalbehaviorpolicytype} & \texttt{mode} \\
\choicetable{gailreward} & \texttt{AIRL} \\
\choicetable{gailmaxrewardmagnitude} & $\infty$ \\

\choicetable{explicitabsorbingstate} & \texttt{True} \\
\choicetable{expertreplay} & $\infty$ \\
\choicetable{pretrainwithbc} & \texttt{True} \\

\choicetable{gailinput} & $(s,\,a)$ \\
\choicetable{gailmlpnumlayers} & $1$ \\
\choicetable{gailmlpnumwidth} & $64$ \\
\choicetable{gailmlpactivation} & \texttt{ReLu} \\
\choicetable{gaildiscriminatormodule} & \texttt{False} \\
\choicetable{subtractlogp} & \texttt{False} \\

\choicetable{gailmlplastlayerinitscale} & $1$ \\
\choicetable{gaildiscriminatorlearningrate} & $10^{-6}$ or $3\cdot10^{-5}$\\
\choicetable{gailmaxreplaysize} & $3\cdot10^6$ \\
\choicetable{discriminatortorlupdatesratio} & $1$ \\

\choicetable{regularizer} & \texttt{spectral normalization} \\
\choicetable{gptarget} & $0$\\
\choicetable{gpcoef} & $1$\\
\choicetable{mixupalpha} & $1$\\
\choicetable{pugailpositiveclassprior} & $0.7$ \\
\choicetable{pugailbeta} & $\infty$\\
\choicetable{dropouthiddenrate} & $75\%$ \\
\choicetable{dropoutinputrate} & $50\%$ \\
\choicetable{regweightdecay} & $10$ \\
\choicetable{regentropycoef} & $0.03$ \\

\choicetable{obsnormalization} & depends on the environment \\

\choicetable{gradupdatesperbatch} & $8$ \\

    \bottomrule
  \end{tabular}
\end{table}
\clearpage\section{Expert and random policy scores}\label{app:expert_scores}

\begin{table}[h]
  \caption{Expert and random policy scores used to normalize the performance for all tasks.}
  \label{table:expert_scores}
  \centering
  \begin{tabular}{lll}
    \toprule
    Task & Random policy score & Expert score \\
    \midrule
    \texttt{HalfCheetah-v2} & -282 & 8770 \\
    \texttt{Hopper-v2} & 18 & 2798 \\ 
    \texttt{Walker2d-v2} & 1.6 & 4118 \\
    \texttt{Ant-v2} & -59 & 5637 \\
    \texttt{Humanoid-v2} & 123 & 9115 \\
    \midrule
    \texttt{pen-expert-v0} & 94 &  3078 \\
    \texttt{door-expert-v0} & -56 & 2882 \\
    \texttt{door-human-v0} & -56 & 796 \\
    \texttt{hammer-expert-v0}  & -274 & 12794 \\
    \texttt{hammer-human-v0}  & -274 & 3071 \\
    \bottomrule
  \end{tabular}
\end{table}

\graphicspath{{reports/}}
\clearpage
\section{Experiment wide}
\label{exp_wide}
\subsection{Design}
\label{exp_design_wide}
For each of the 10 tasks, we sampled 12083 choice configurations where we sampled the following choices independently and uniformly from the following ranges:
\begin{itemize}
    \item \choicet{directrlalgorithm}: \{d4pg, ppo, sac, td3\}
    \begin{itemize}
        \item For the case ``\choicet{directrlalgorithm} = sac'', we further sampled the sub-choices:
        \begin{itemize}
            \item \choicet{saclearningrate}: \{0.0001, 0.0003, 0.001\}
            \item \choicet{sactargetentropyperdimension}: \{-2.0, -1.0, -0.5, 0.0\}
            \item \choicet{sactau}: \{0.001, 0.003, 0.01, 0.03\}
            \item \choicet{subtractlogp}: \{False, True\}
            \item \choicet{batchsize}: \{256.0\}
        \end{itemize}
        \item For the case ``\choicet{directrlalgorithm} = d4pg'', we further sampled the sub-choices:
        \begin{itemize}
            \item \choicet{dfpglearningrate}: \{3e-05, 0.0001, 0.0003\}
            \item \choicet{rlsigma}: \{0.1, 0.2, 0.3, 0.5\}
            \item \choicet{vmax}: \{150.0, 750.0, 1500.0\}
            \item \choicet{numatoms}: \{51.0, 101.0, 201.0, 401.0\}
            \item \choicet{nstep}: \{1.0, 3.0, 5.0\}
            \item \choicet{batchsize}: \{256.0\}
        \end{itemize}
        \item For the case ``\choicet{directrlalgorithm} = td3'', we further sampled the sub-choices:
        \begin{itemize}
            \item \choicet{tdtpolicylearningrate}: \{0.0001, 0.0003, 0.001\}
            \item \choicet{tdtcriticlearningrate}: \{0.0001, 0.0003, 0.001\}
            \item \choicet{tdtgradientclipping}: \{40.0, $\infty$\}
            \item \choicet{rlsigma}: \{0.1, 0.2, 0.3, 0.5\}
            \item \choicet{batchsize}: \{256.0\}
        \end{itemize}
        \item For the case ``\choicet{directrlalgorithm} = ppo'', we further sampled the sub-choices:
        \begin{itemize}
            \item \choicet{ppolearningrate}: \{3e-05, 0.0001, 0.0003\}
            \item \choicet{pponumepochs}: \{2.0, 5.0, 10.0, 20.0\}
            \item \choicet{ppoentropycost}: \{0.0, 0.001, 0.003, 0.01, 0.03, 0.1\}
            \item \choicet{pponumminibatches}: \{8.0, 16.0, 32.0, 64.0\}
            \item \choicet{ppounrolllength}: \{4.0, 8.0, 16.0, 32.0\}
            \item \choicet{ppoclippingepsilon}: \{0.1, 0.2, 0.3\}
            \item \choicet{ppogaelambda}: \{0.8, 0.9, 0.95, 0.99\}
            \item \choicet{subtractlogp}: \{False, True\}
            \item \choicet{batchsize}: \{64.0, 128.0, 256.0\}
        \end{itemize}
    \end{itemize}
    \item \choicet{maxreplaysize}: \{300000.0, 1000000.0, 3000000.0\}
    \item \choicet{numpolicylayers}: \{1, 2, 3\}
    \item \choicet{policylayersize}: \{64, 128, 256, 512\}
    \item \choicet{numcriticlayers}: \{1, 2, 3\}
    \item \choicet{criticlayersize}: \{64, 128, 256, 512\}
    \item \choicet{activation}: \{relu, tanh\}
    \item \choicet{discount}: \{0.9, 0.97, 0.99, 0.997\}
    \item \choicet{pretrainwithbc}: \{False, True\}
    \item \choicet{explicitabsorbingstate}: \{False, True\}
    \item \choicet{gailmaxreplaysize}: \{300000, 1000000, 3000000\}
    \item \choicet{gaildiscriminatormodule}: \{False, True\}
    \item \choicet{gailinput}: \{s, sa, sas, ss\}
    \item \choicet{gailmlpnumlayers}: \{1, 2, 3\}
    \item \choicet{gailmlpnumwidth}: \{16, 32, 64, 128, 256, 512\}
    \item \choicet{gailmlpactivation}: \{elu, leaky\_relu, relu, sigmoid, swish, tanh\}
    \item \choicet{gailmlplastlayerinitscale}: \{0.001, 1.0\}
    \item \choicet{regularizer}: \{GP, Mixup, No regularizer, PUGAIL, dropout, entropy, spectral norm, weight decay\}
    \begin{itemize}
        \item For the case ``\choicet{regularizer} = GP'', we further sampled the sub-choices:
        \begin{itemize}
            \item \choicet{gpcoef}: \{0.1, 1.0, 10.0\}
            \item \choicet{gptarget}: \{0.0, 1.0\}
        \end{itemize}
        \item For the case ``\choicet{regularizer} = Mixup'', we further sampled the sub-choices:
        \begin{itemize}
            \item \choicet{mixupalpha}: \{0.1, 0.4, 1.0\}
        \end{itemize}
        \item For the case ``\choicet{regularizer} = PUGAIL'', we further sampled the sub-choices:
        \begin{itemize}
            \item \choicet{pugailpositiveclassprior}: \{0.25, 0.5, 0.7\}
            \item \choicet{pugailbeta}: \{0.0, 0.7, $\infty$\}
        \end{itemize}
        \item For the case ``\choicet{regularizer} = entropy'', we further sampled the sub-choices:
        \begin{itemize}
            \item \choicet{regentropycoef}: \{0.0003, 0.001, 0.003, 0.01, 0.03, 0.1, 0.3\}
        \end{itemize}
        \item For the case ``\choicet{regularizer} = weight decay'', we further sampled the sub-choices:
        \begin{itemize}
            \item \choicet{regweightdecay}: \{0.3, 1.0, 3.0, 10.0, 30.0\}
        \end{itemize}
        \item For the case ``\choicet{regularizer} = dropout'', we further sampled the sub-choices:
        \begin{itemize}
            \item \choicet{dropoutinputrate}: \{0.0, 0.25, 0.5, 0.75\}
            \item \choicet{dropouthiddenrate}: \{0.25, 0.5, 0.75\}
        \end{itemize}
    \end{itemize}
    \item \choicet{obsnormalization}: \{fixed, none\}
    \item \choicet{evalbehaviorpolicytype}: \{average, mode, stochastic\}
    \item \choicet{gaildiscriminatorlearningrate}: \{1e-06, 3e-06, 1e-05, 3e-05, 0.0001, 0.0003\}
    \item \choicet{gailmaxrewardmagnitude}: \{0.5, 1.0, 2.0, 5.0, 10.0, 50.0, $\infty$\}
    \item \choicet{gailreward}: \{-ln(1-D), AIRL, FAIRL, ln(D)\}
    \item \choicet{samplesperinsert}: \{256\}
    \item \choicet{discriminatortorlupdatesratio}: \{1\}
    \item \choicet{gradupdatesperbatch}: \{8\}
\end{itemize}

\subsection{Results}
\label{exp_results_wide}
\begin{figure}[ht]
\begin{center}
\centerline{\includegraphics[width=1\textwidth]{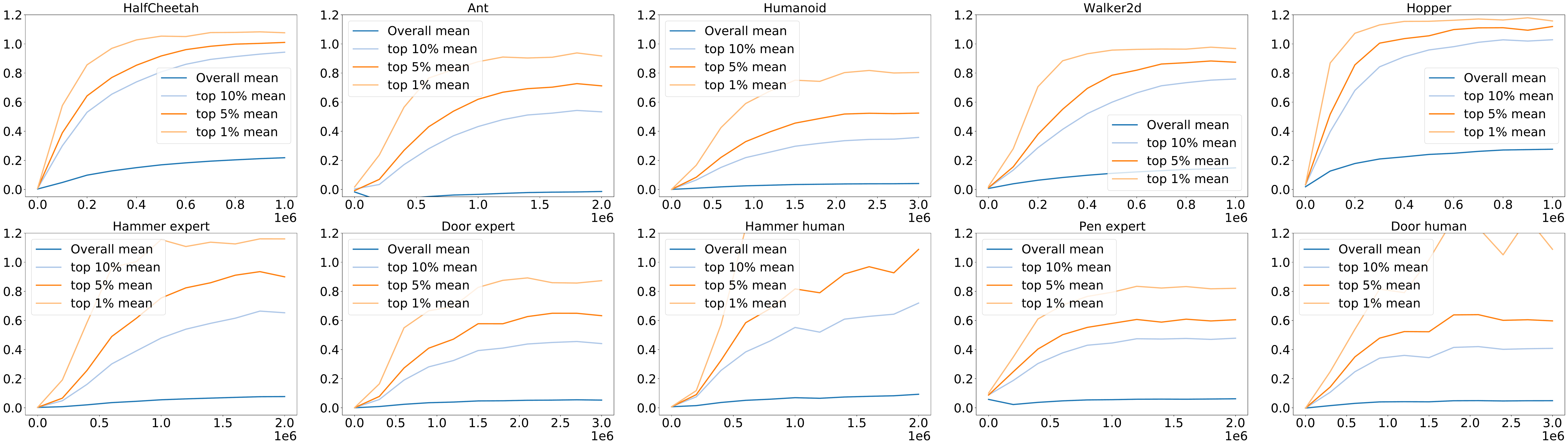}}
\caption{Training curves.}
\label{fig:wide_training_curves}
\end{center}
\end{figure}

\begin{figure}[ht]
\begin{center}
\centerline{\includegraphics[height=4.5cm,width=1\textwidth]{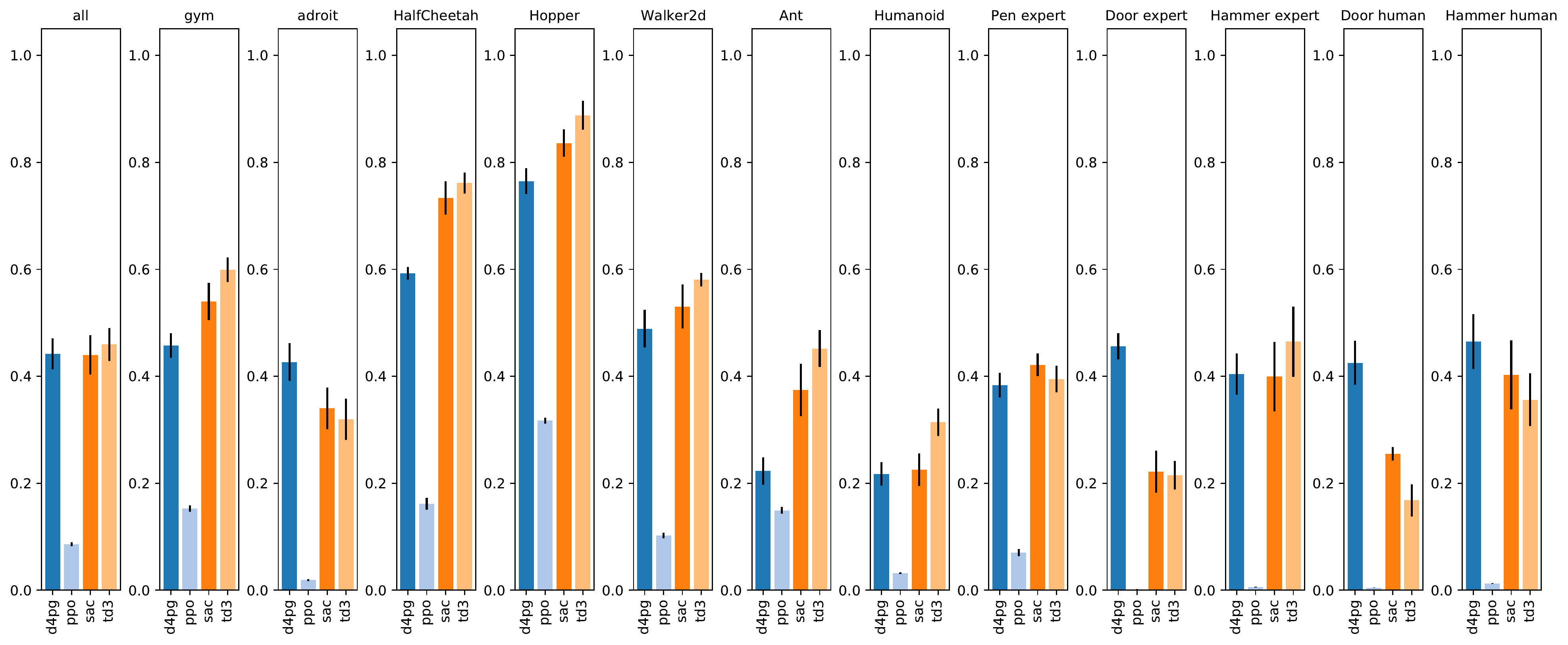}}
\centerline{\includegraphics[height=4.5cm,width=1\textwidth]{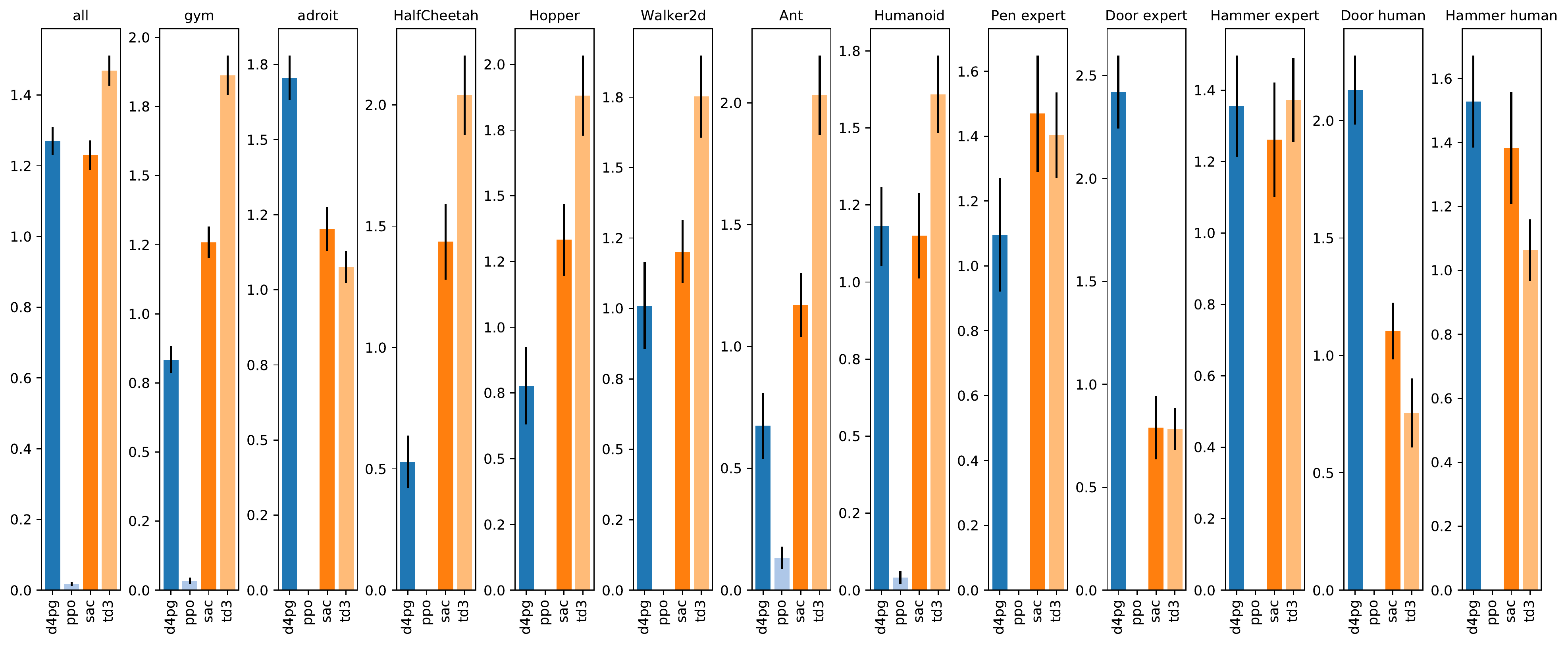}}
\caption{Analysis of choice \choicet{directrlalgorithm}: 95th percentile of performance scores conditioned on choice (top) and distribution of choices in top 5\% of configurations (bottom).}
\label{fig:wide_direct_rl_algorithm}
\end{center}
\end{figure}

\begin{figure}[ht]
\begin{center}
\centerline{\includegraphics[height=4.5cm,width=1\textwidth]{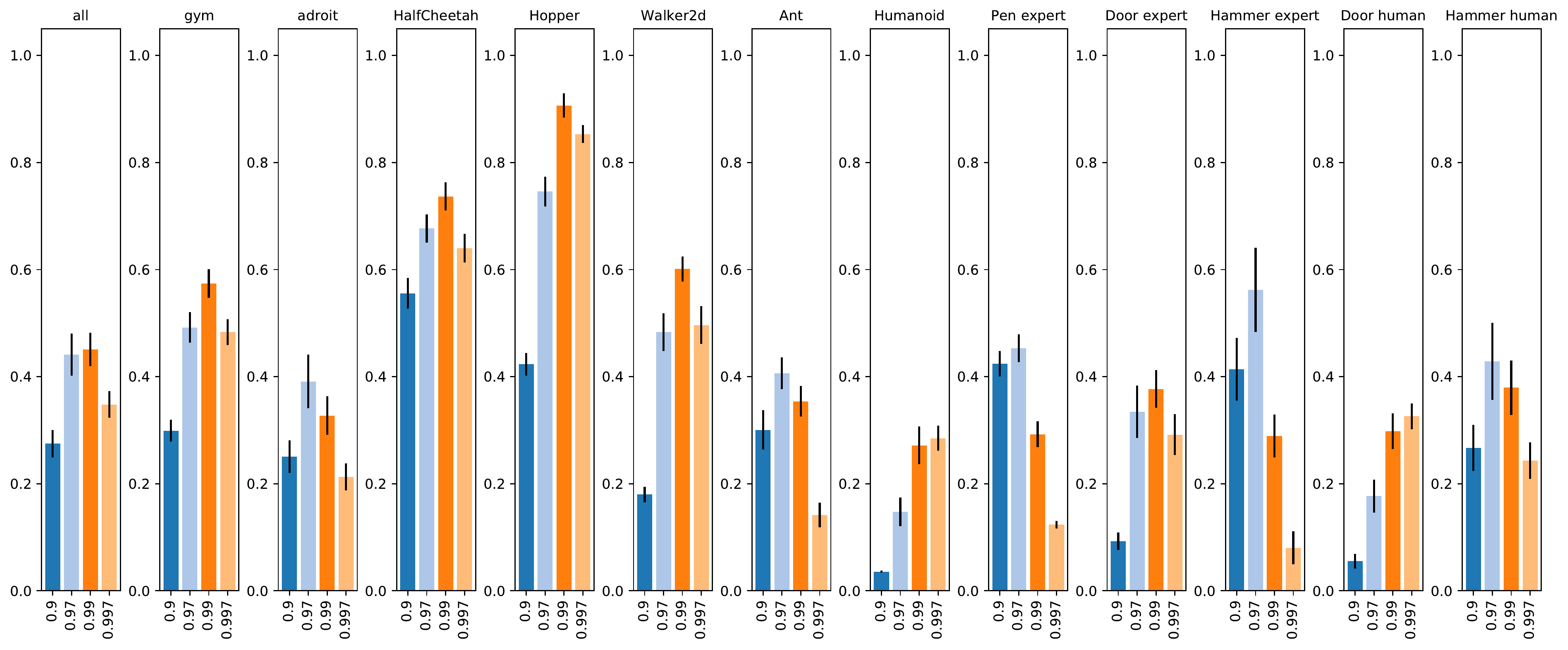}}
\centerline{\includegraphics[height=4.5cm,width=1\textwidth]{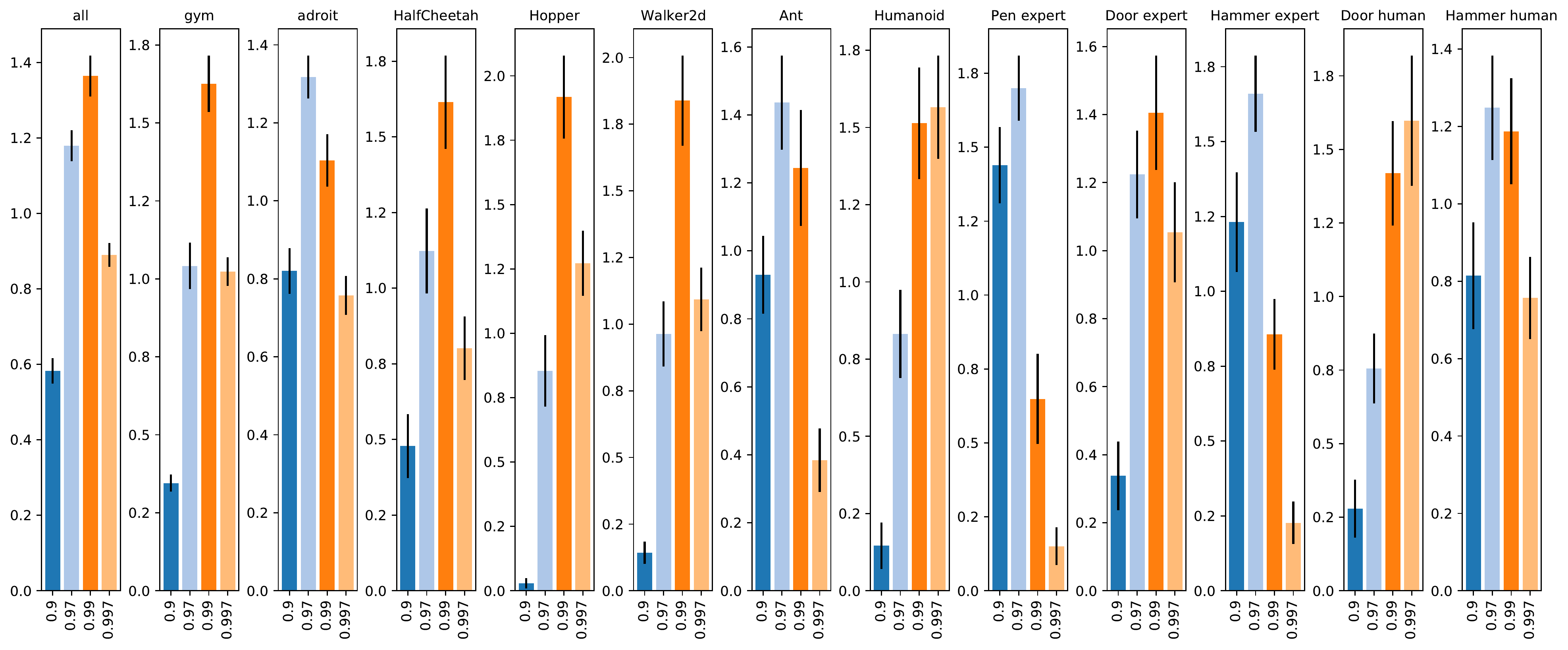}}
\caption{Analysis of choice \choicet{discount}: 95th percentile of performance scores conditioned on choice (top) and distribution of choices in top 5\% of configurations (bottom).}
\label{fig:wide_discount}
\end{center}
\end{figure}

\begin{figure}[ht]
\begin{center}
\centerline{\includegraphics[height=4.5cm,width=1\textwidth]{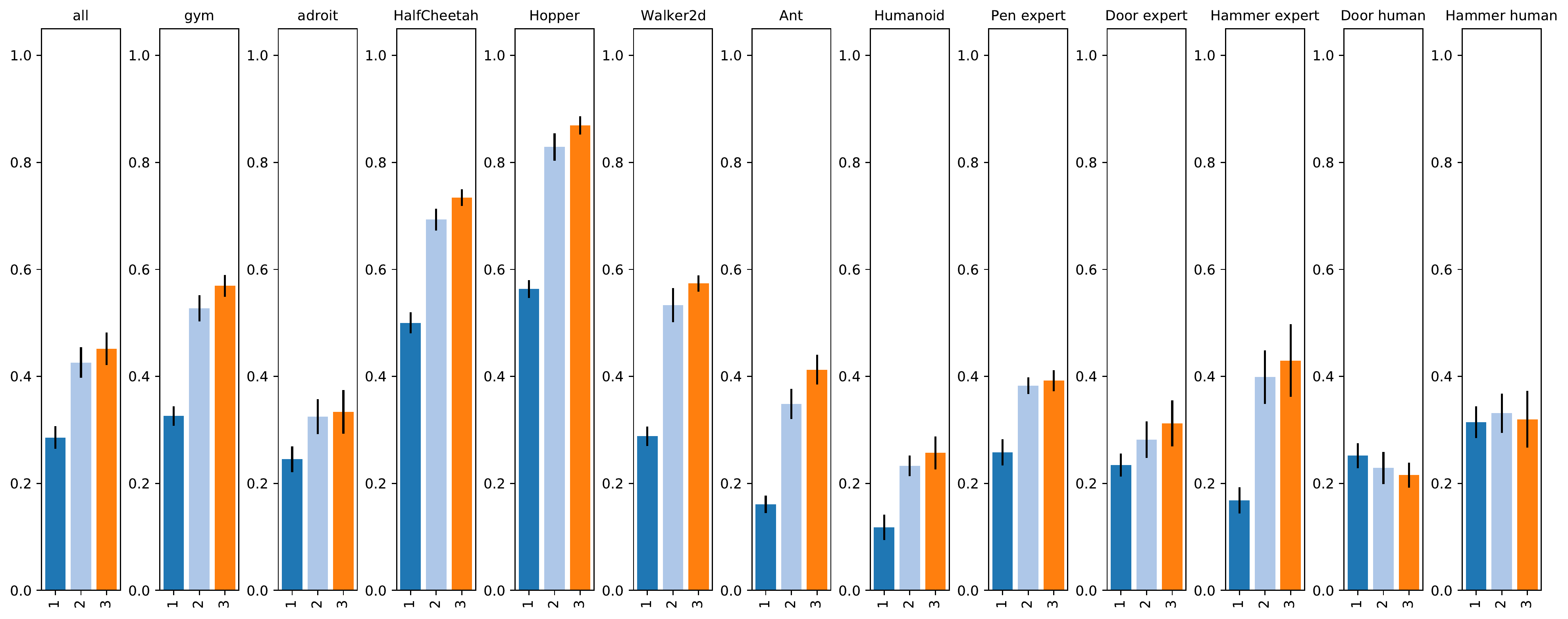}}
\centerline{\includegraphics[height=4.5cm,width=1\textwidth]{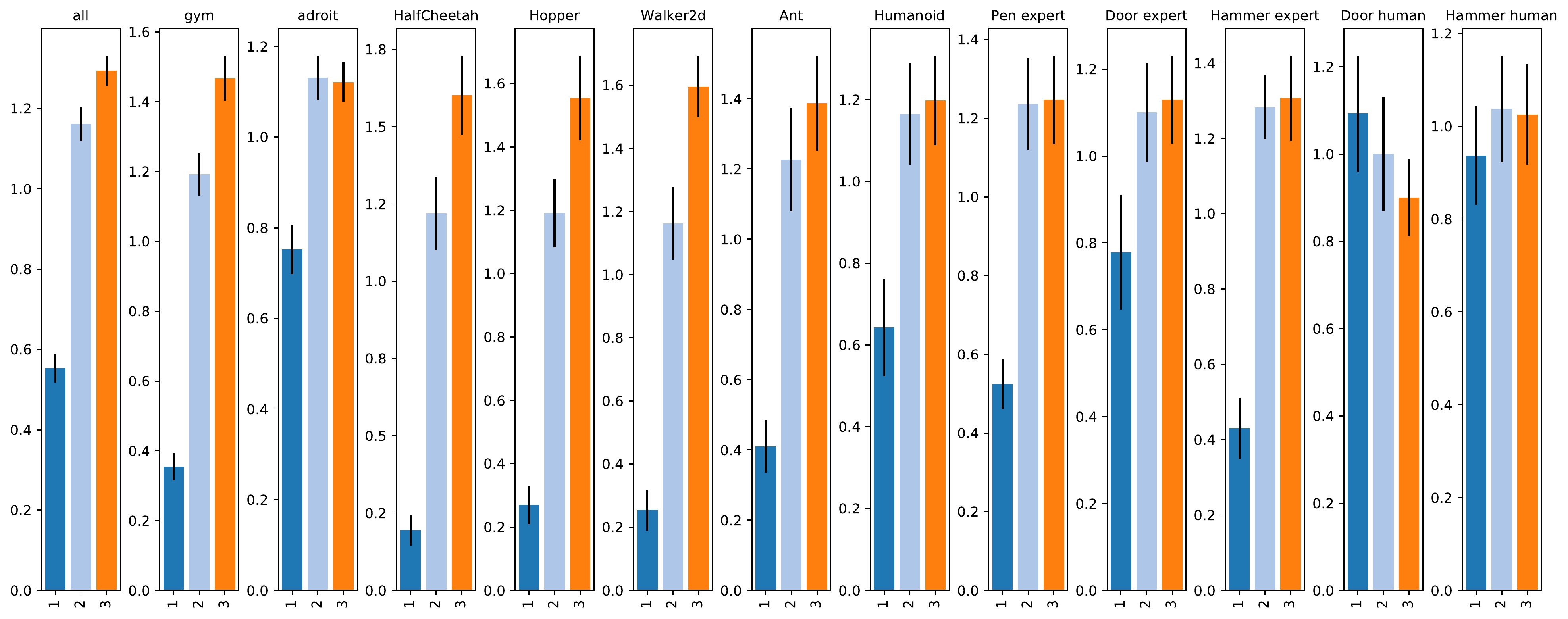}}
\caption{Analysis of choice \choicet{numcriticlayers}: 95th percentile of performance scores conditioned on choice (top) and distribution of choices in top 5\% of configurations (bottom).}
\label{fig:wide_num_critic_layers}
\end{center}
\end{figure}

\begin{figure}[ht]
\begin{center}
\centerline{\includegraphics[height=4.5cm,width=1\textwidth]{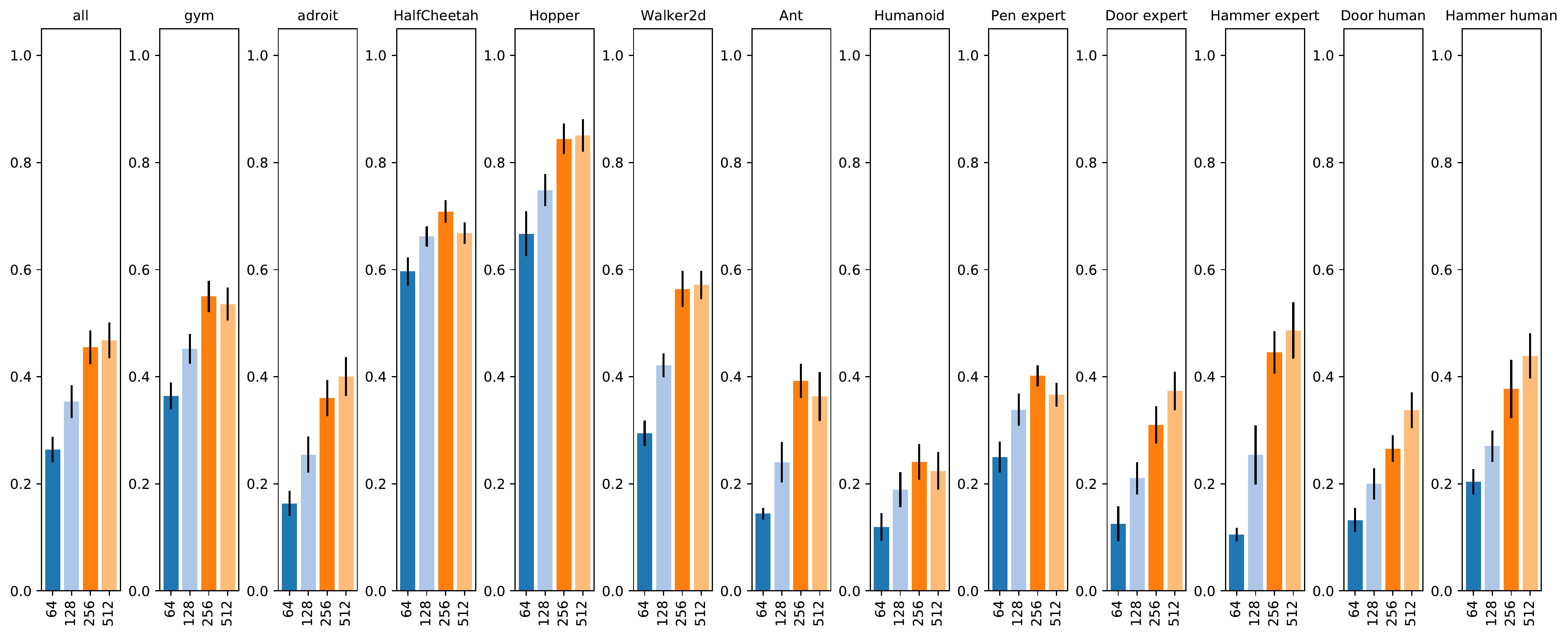}}
\centerline{\includegraphics[height=4.5cm,width=1\textwidth]{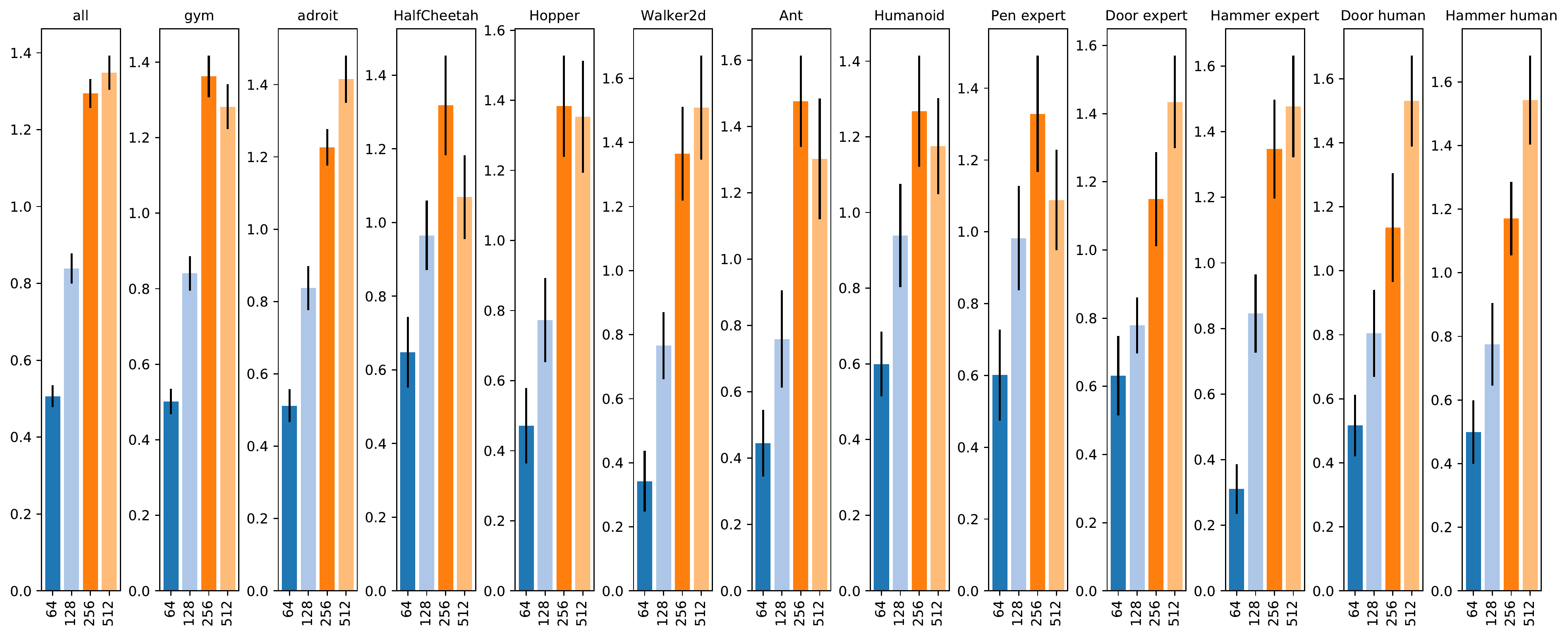}}
\caption{Analysis of choice \choicet{criticlayersize}: 95th percentile of performance scores conditioned on choice (top) and distribution of choices in top 5\% of configurations (bottom).}
\label{fig:wide_critic_layer_size}
\end{center}
\end{figure}

\begin{figure}[ht]
\begin{center}
\centerline{\includegraphics[height=4.5cm,width=1\textwidth]{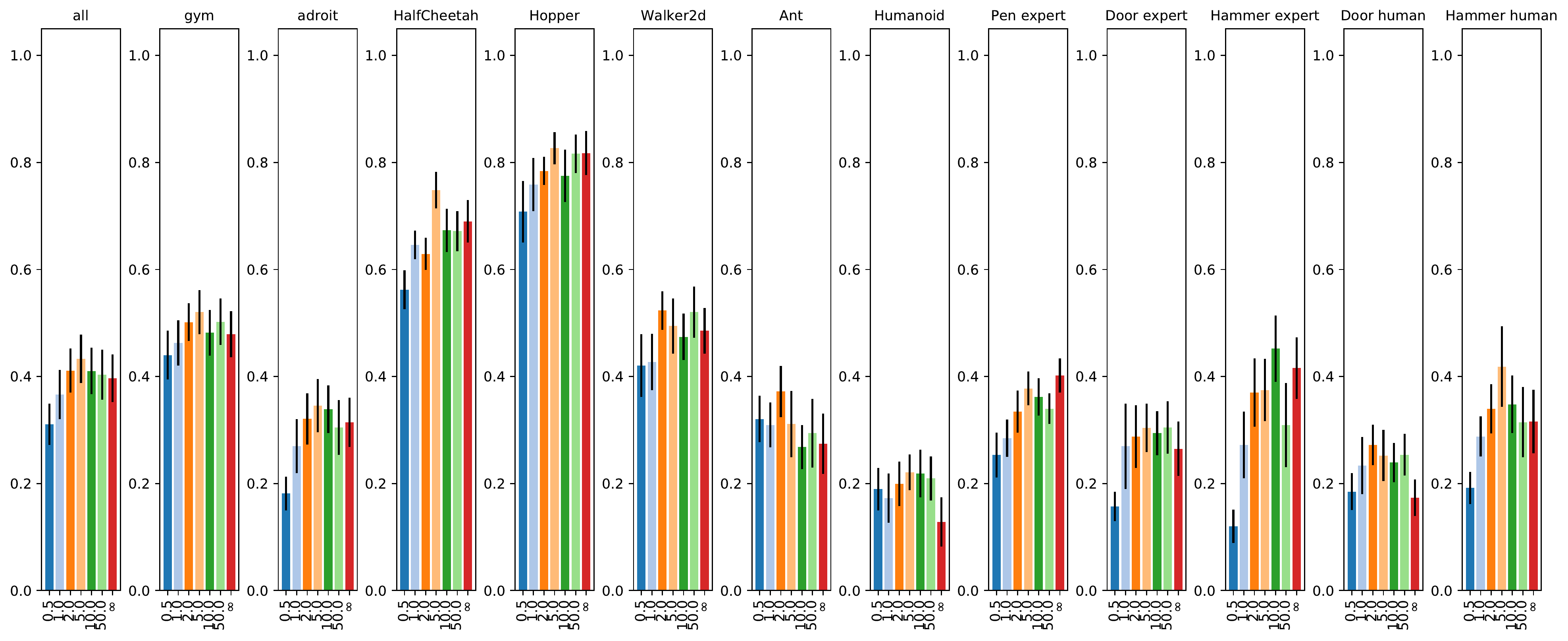}}
\centerline{\includegraphics[height=4.5cm,width=1\textwidth]{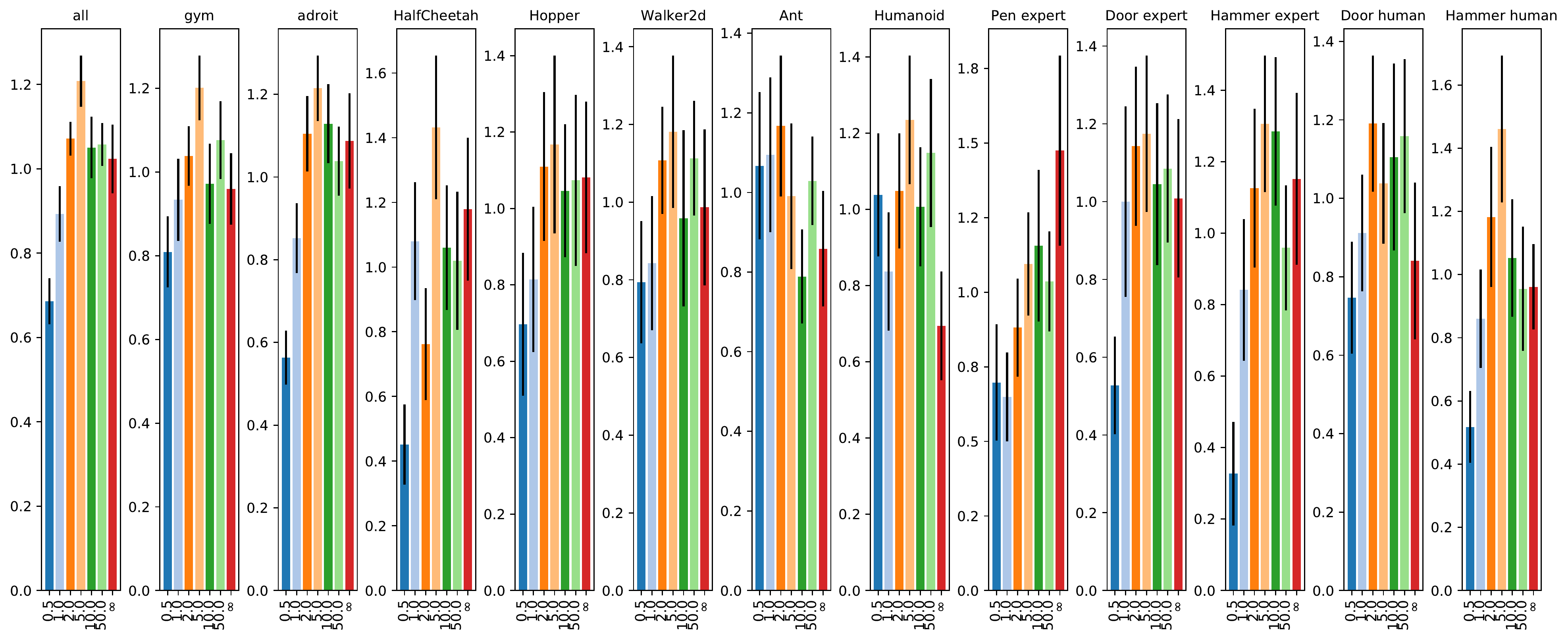}}
\caption{Analysis of choice \choicet{gailmaxrewardmagnitude}: 95th percentile of performance scores conditioned on choice (top) and distribution of choices in top 5\% of configurations (bottom).}
\label{fig:wide__gin_GAILBuilder_max_reward_magnitude}
\end{center}
\end{figure}

\begin{figure}[ht]
\begin{center}
\centerline{\includegraphics[height=4.5cm,width=1\textwidth]{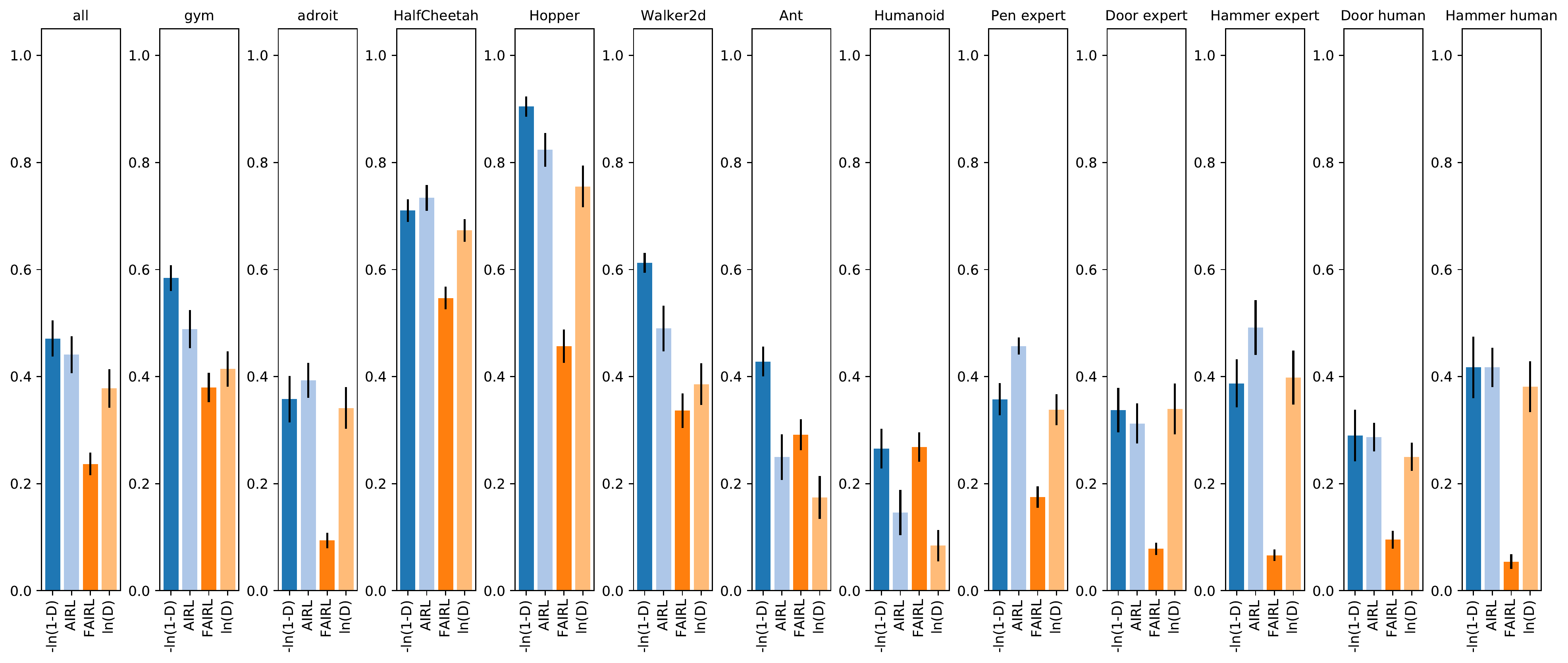}}
\centerline{\includegraphics[height=4.5cm,width=1\textwidth]{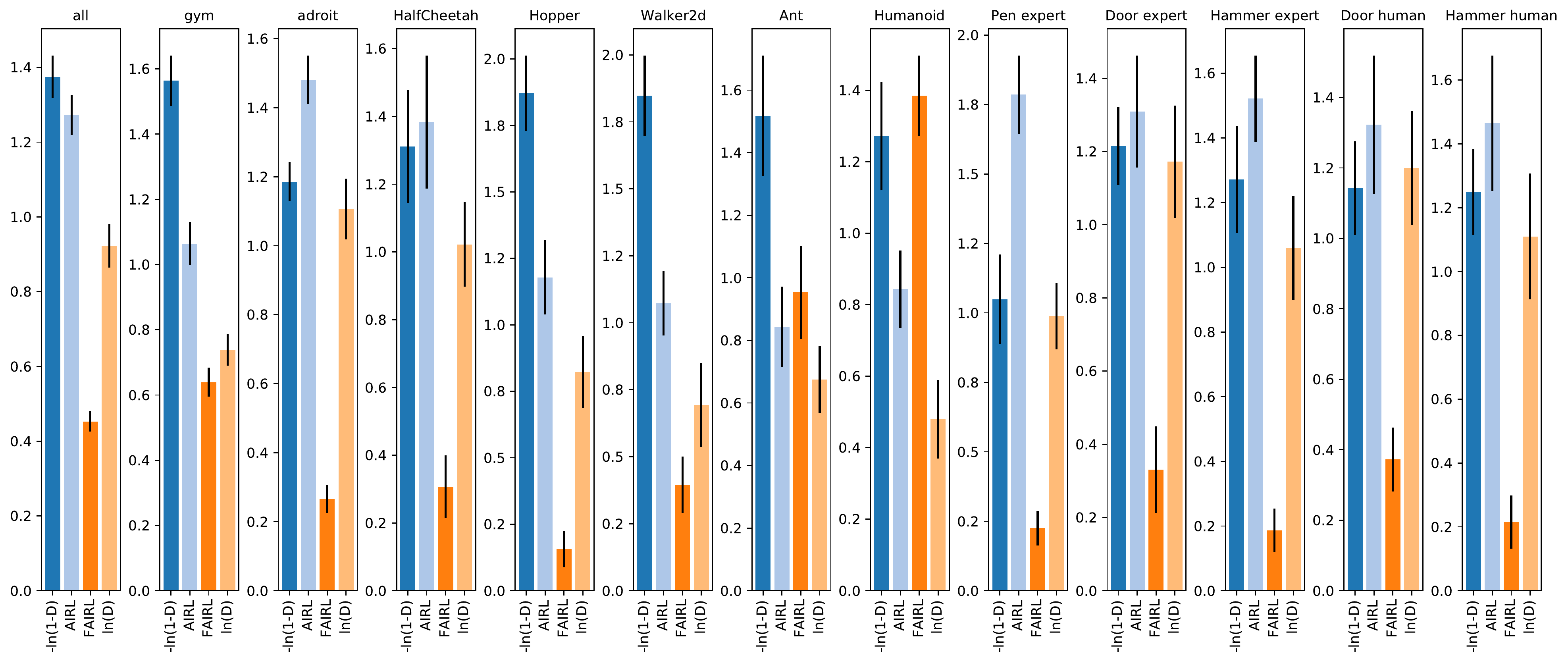}}
\caption{Analysis of choice \choicet{gailreward}: 95th percentile of performance scores conditioned on choice (top) and distribution of choices in top 5\% of configurations (bottom).}
\label{fig:wide__gin_reward_function__macro_value}
\end{center}
\end{figure}

\begin{figure}[ht]
\begin{center}
\centerline{\includegraphics[height=4.5cm,width=1\textwidth]{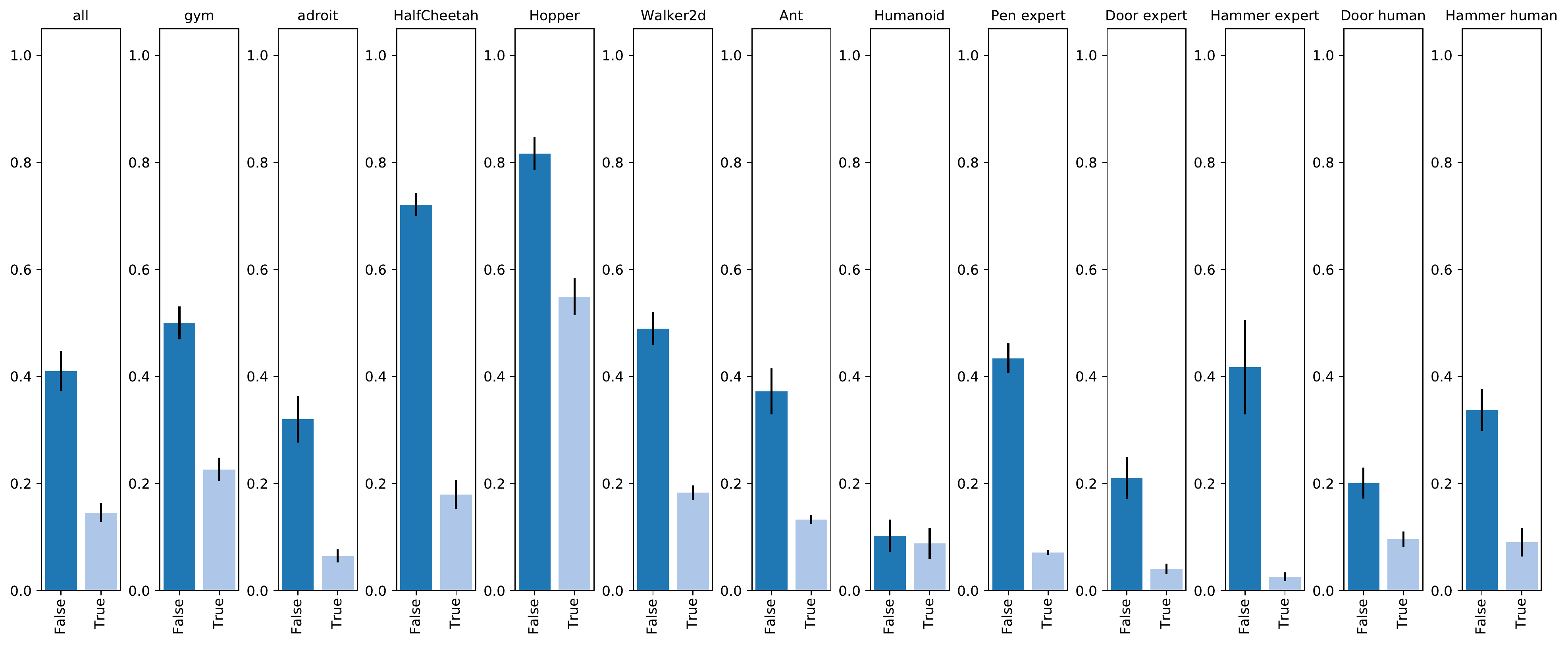}}
\centerline{\includegraphics[height=4.5cm,width=1\textwidth]{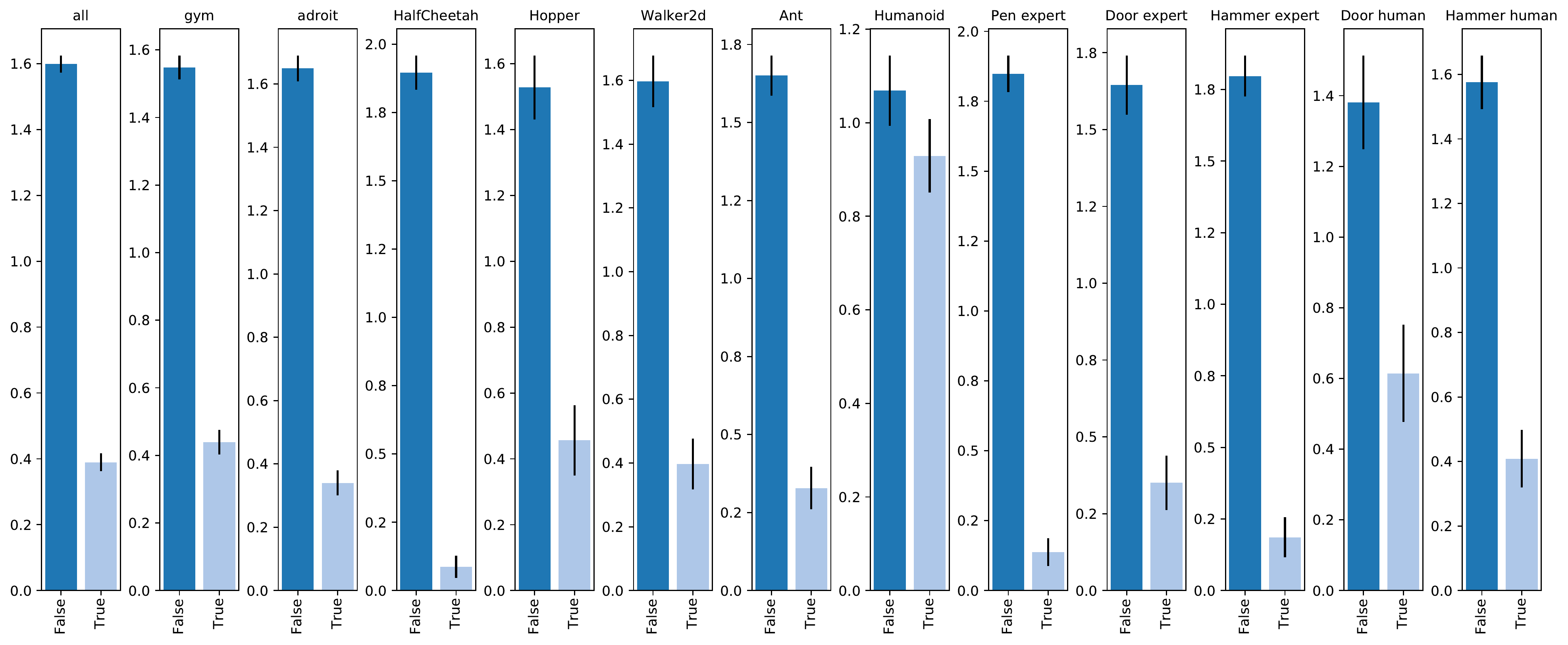}}
\caption{Analysis of choice \choicet{subtractlogp}: 95th percentile of performance scores conditioned on choice (top) and distribution of choices in top 5\% of configurations (bottom).}
\label{fig:wide__gin_make_discriminator_subtract_logpi}
\end{center}
\end{figure}

\begin{figure}[ht]
\begin{center}
\centerline{\includegraphics[height=4.5cm,width=1\textwidth]{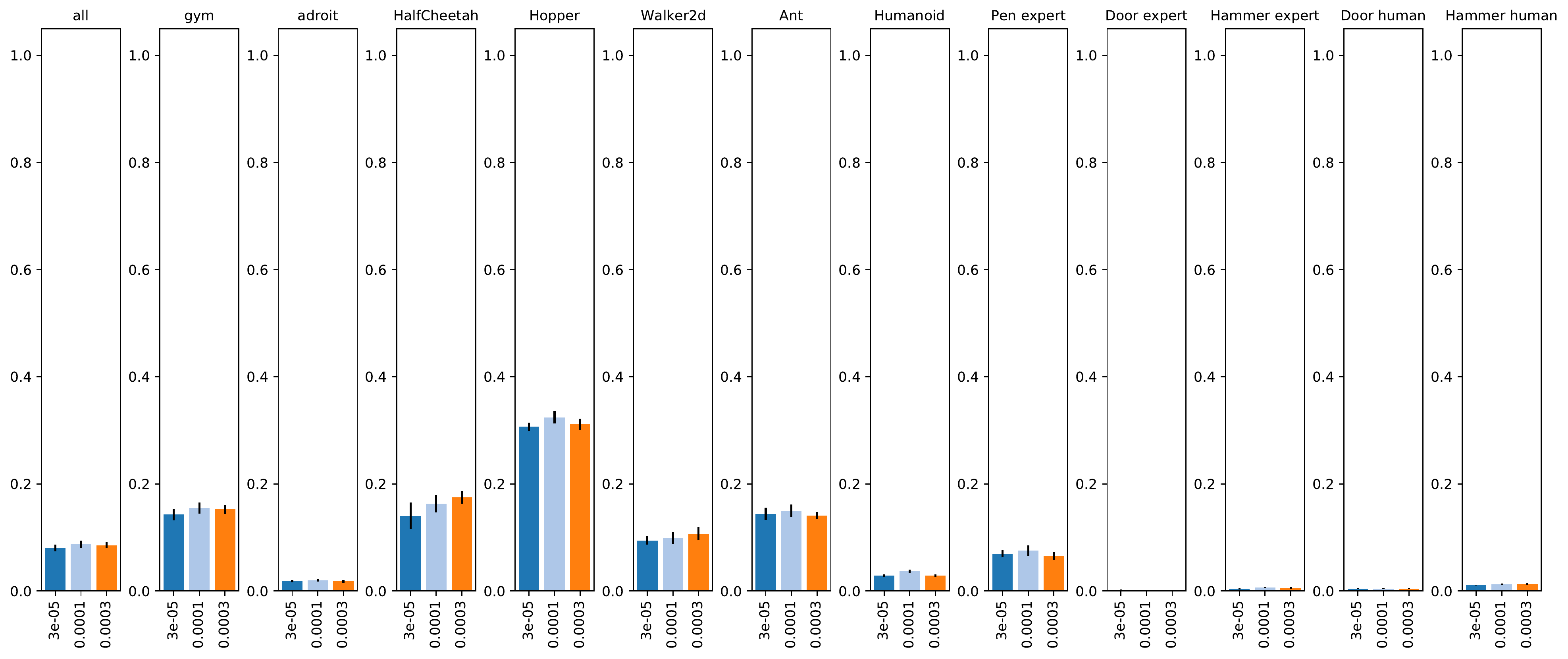}}
\centerline{\includegraphics[height=4.5cm,width=1\textwidth]{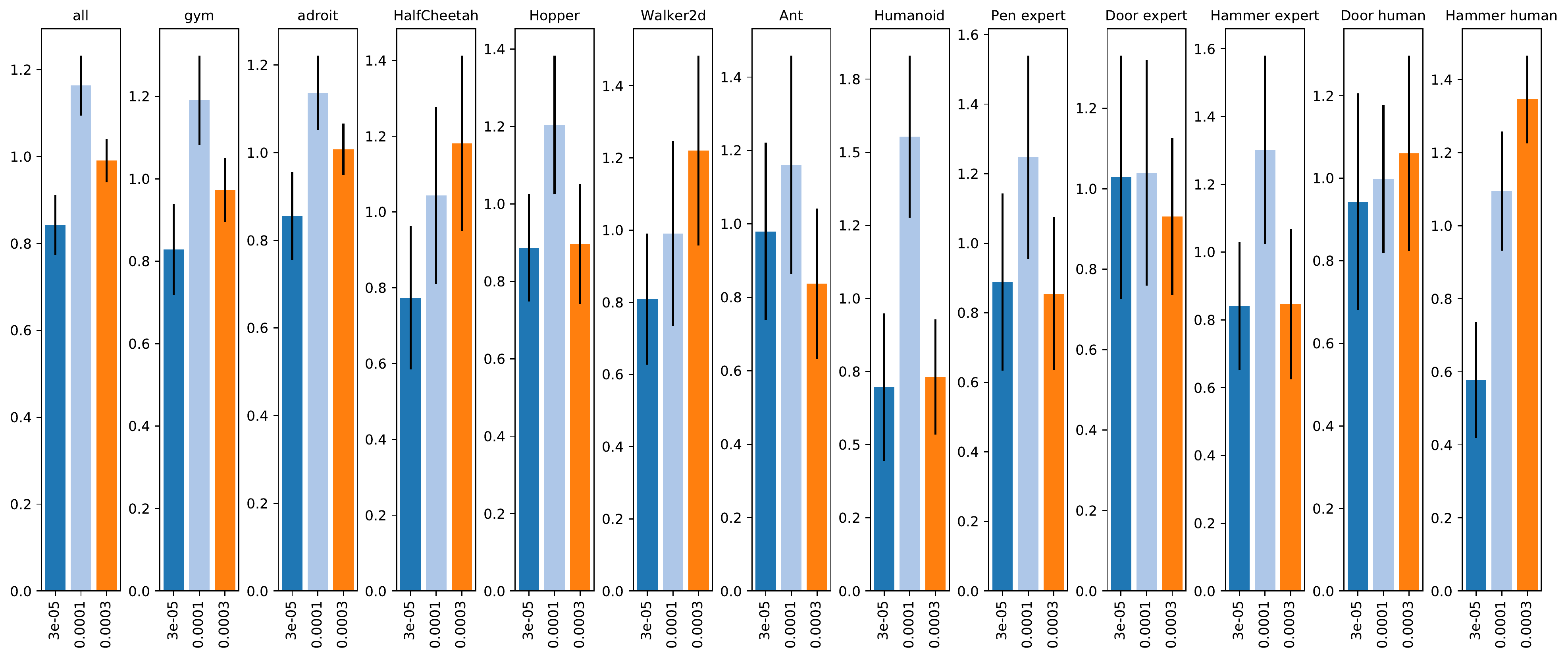}}
\caption{Analysis of choice \choicet{ppolearningrate}: 95th percentile of performance scores conditioned on choice (top) and distribution of choices in top 5\% of configurations (bottom).}
\label{fig:wide_ppo_learning_rate}
\end{center}
\end{figure}

\begin{figure}[ht]
\begin{center}
\centerline{\includegraphics[height=4.5cm,width=1\textwidth]{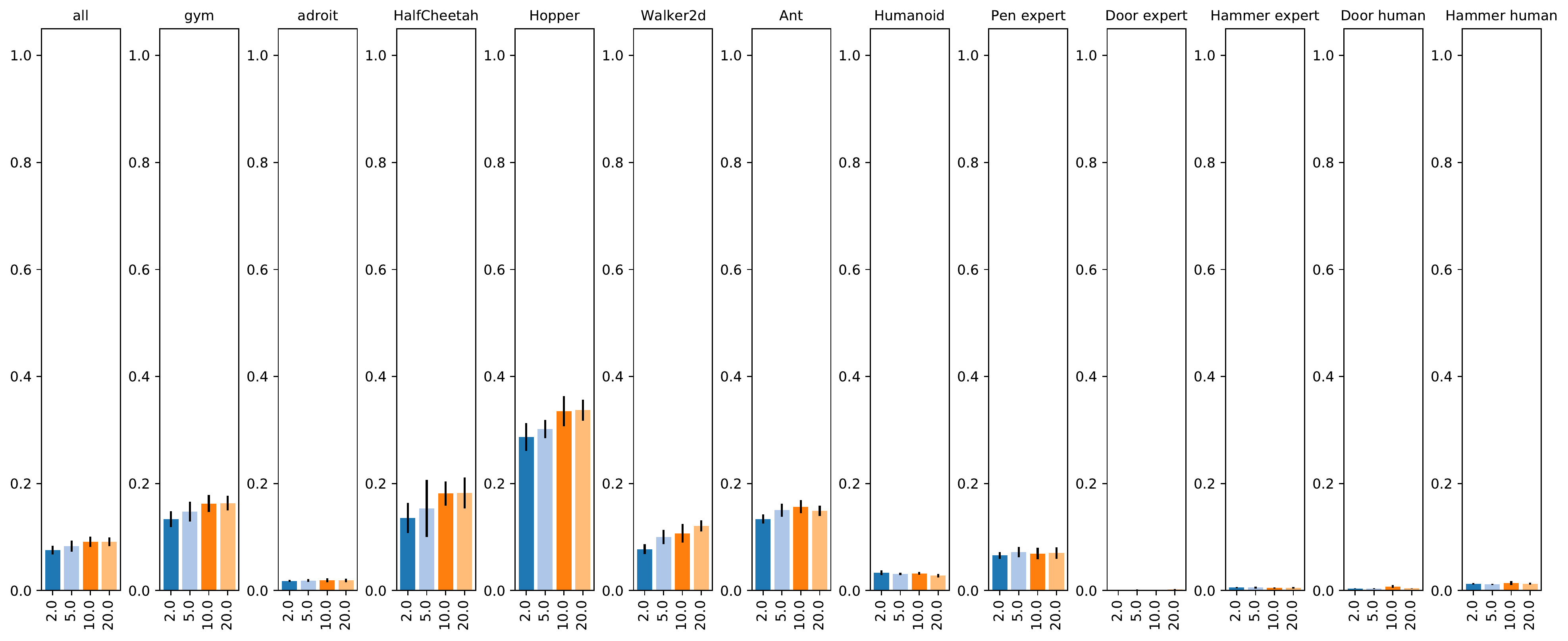}}
\centerline{\includegraphics[height=4.5cm,width=1\textwidth]{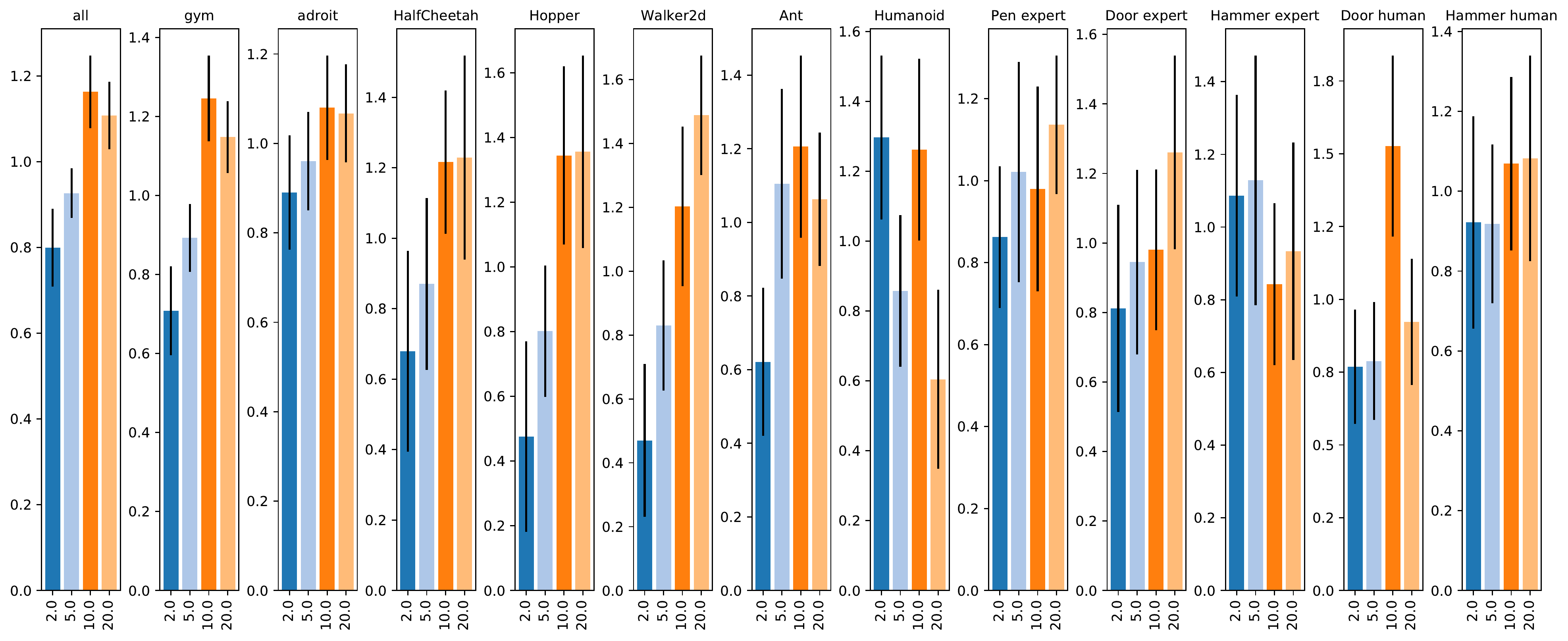}}
\caption{Analysis of choice \choicet{pponumepochs}: 95th percentile of performance scores conditioned on choice (top) and distribution of choices in top 5\% of configurations (bottom).}
\label{fig:wide_ppo_num_epochs}
\end{center}
\end{figure}

\begin{figure}[ht]
\begin{center}
\centerline{\includegraphics[height=4.5cm,width=1\textwidth]{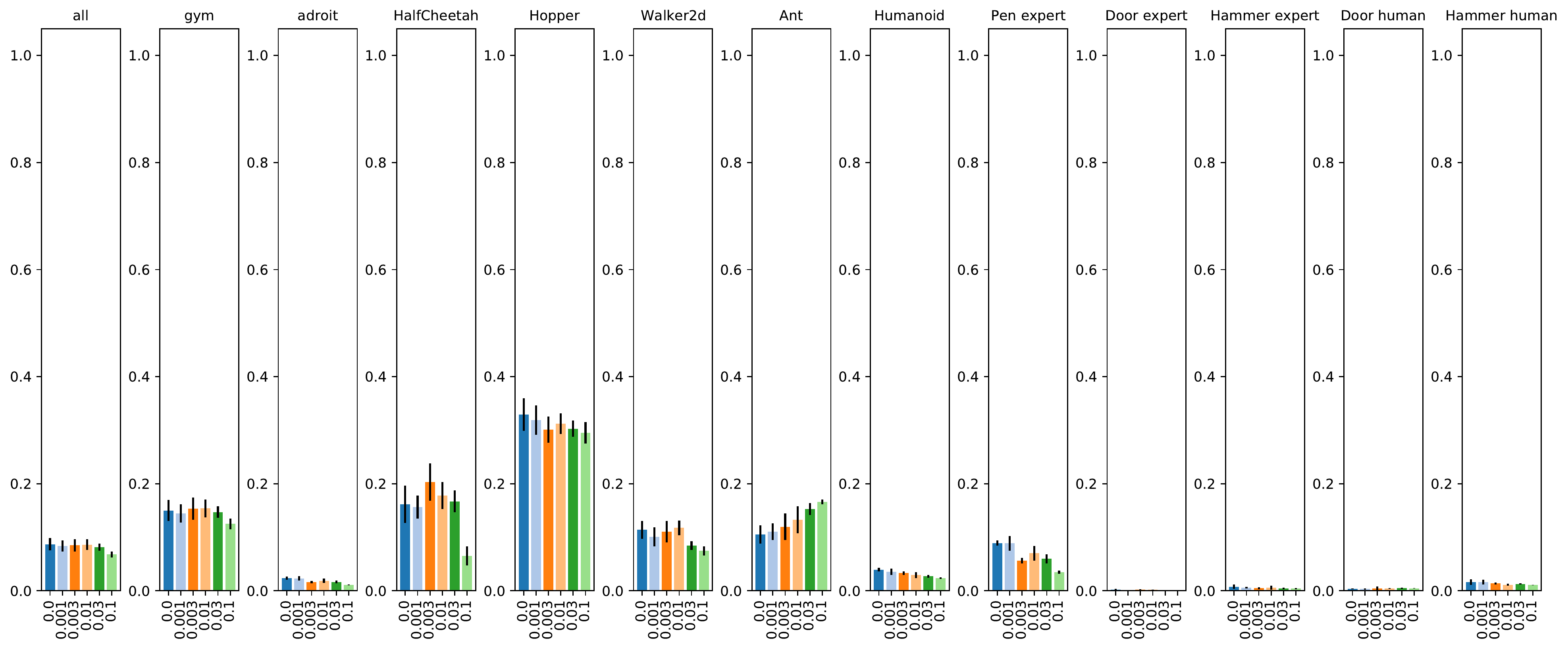}}
\centerline{\includegraphics[height=4.5cm,width=1\textwidth]{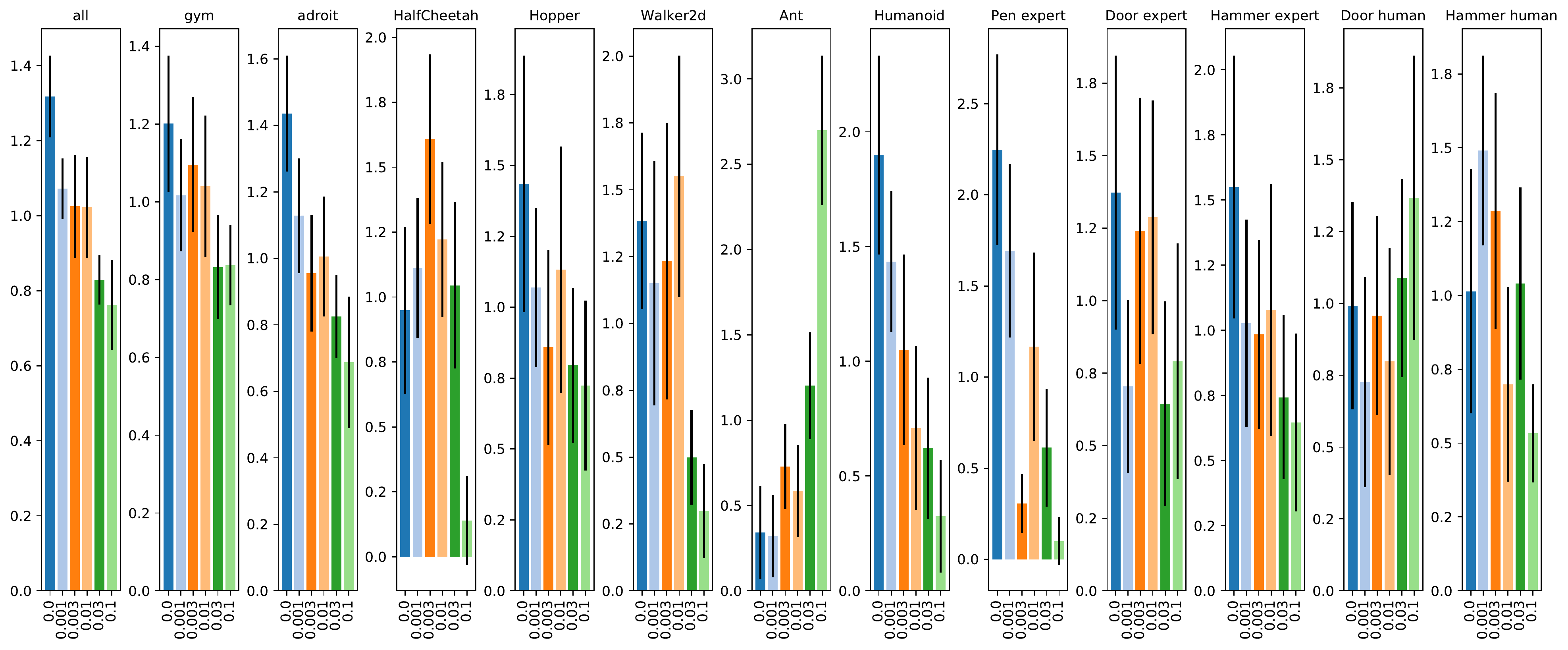}}
\caption{Analysis of choice \choicet{ppoentropycost}: 95th percentile of performance scores conditioned on choice (top) and distribution of choices in top 5\% of configurations (bottom).}
\label{fig:wide_ppo_entropy_cost}
\end{center}
\end{figure}

\begin{figure}[ht]
\begin{center}
\centerline{\includegraphics[height=4.5cm,width=1\textwidth]{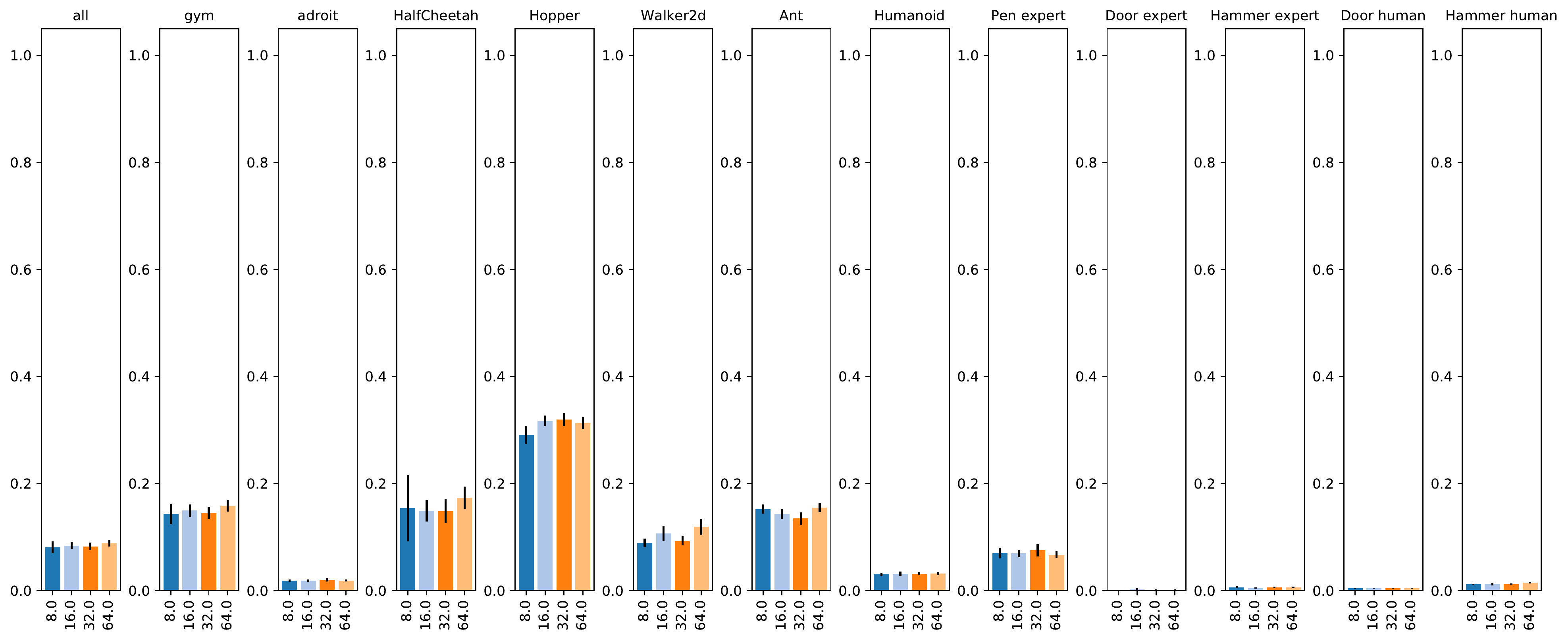}}
\centerline{\includegraphics[height=4.5cm,width=1\textwidth]{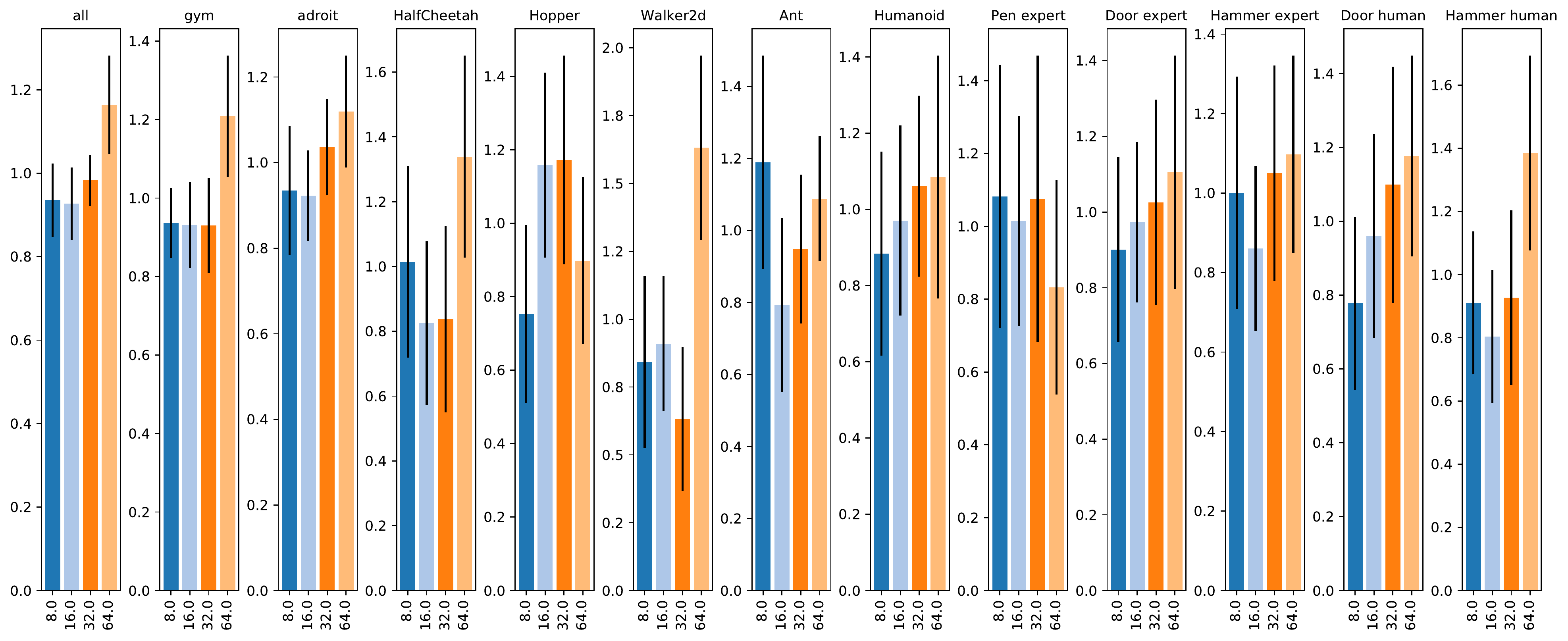}}
\caption{Analysis of choice \choicet{pponumminibatches}: 95th percentile of performance scores conditioned on choice (top) and distribution of choices in top 5\% of configurations (bottom).}
\label{fig:wide_ppo_num_minibatches}
\end{center}
\end{figure}

\begin{figure}[ht]
\begin{center}
\centerline{\includegraphics[height=4.5cm,width=1\textwidth]{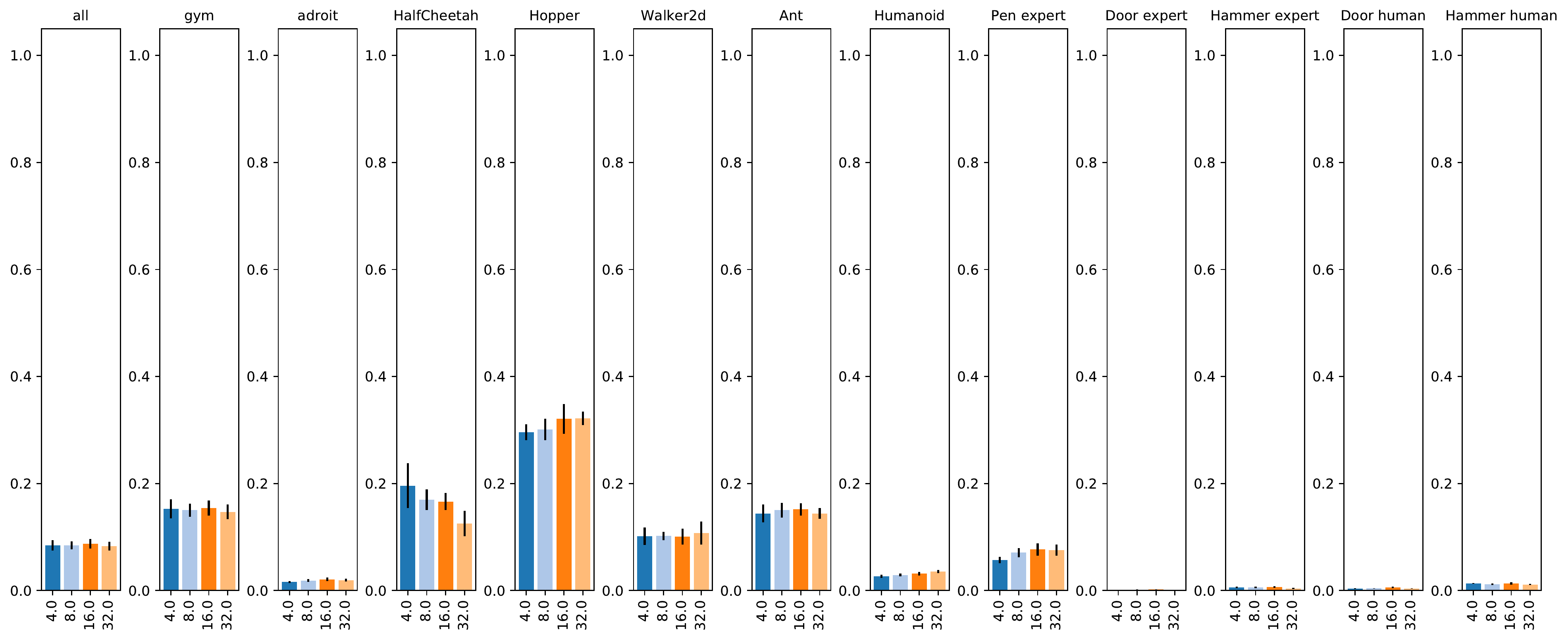}}
\centerline{\includegraphics[height=4.5cm,width=1\textwidth]{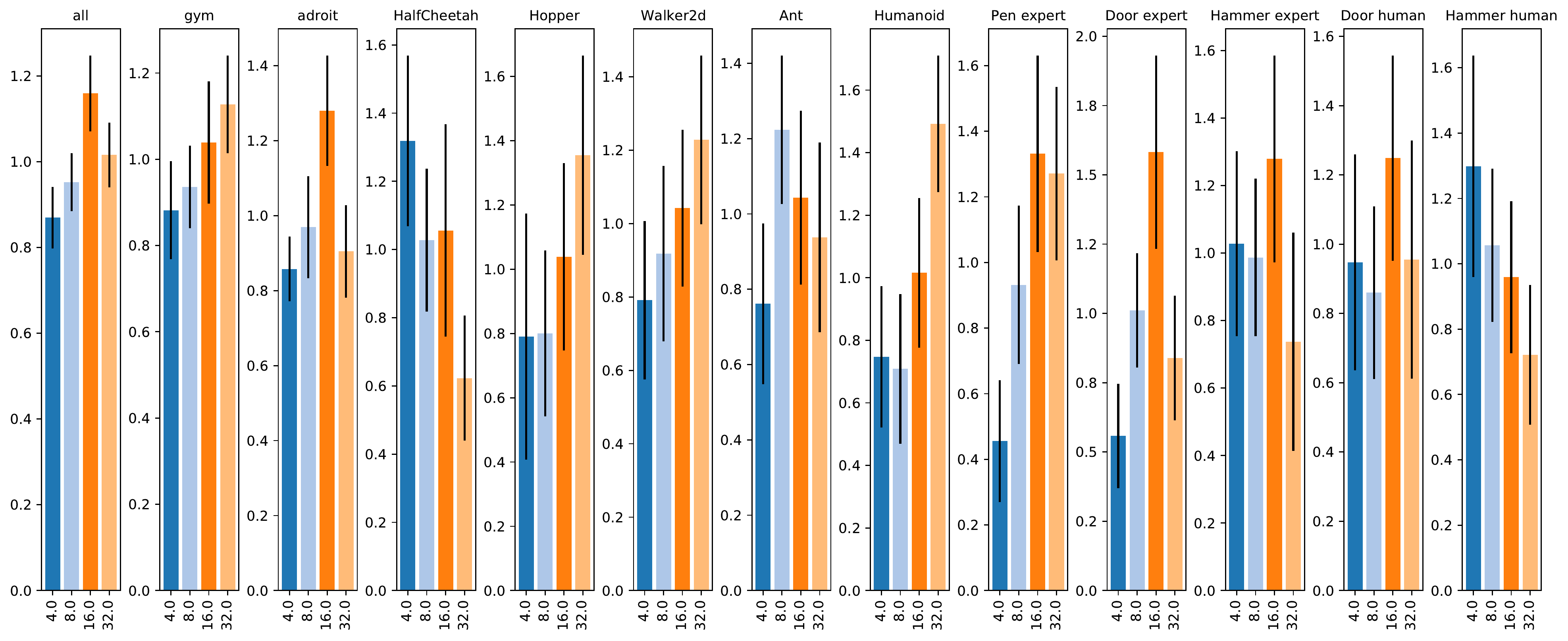}}
\caption{Analysis of choice \choicet{ppounrolllength}: 95th percentile of performance scores conditioned on choice (top) and distribution of choices in top 5\% of configurations (bottom).}
\label{fig:wide_ppo_unroll_length}
\end{center}
\end{figure}

\begin{figure}[ht]
\begin{center}
\centerline{\includegraphics[height=4.5cm,width=1\textwidth]{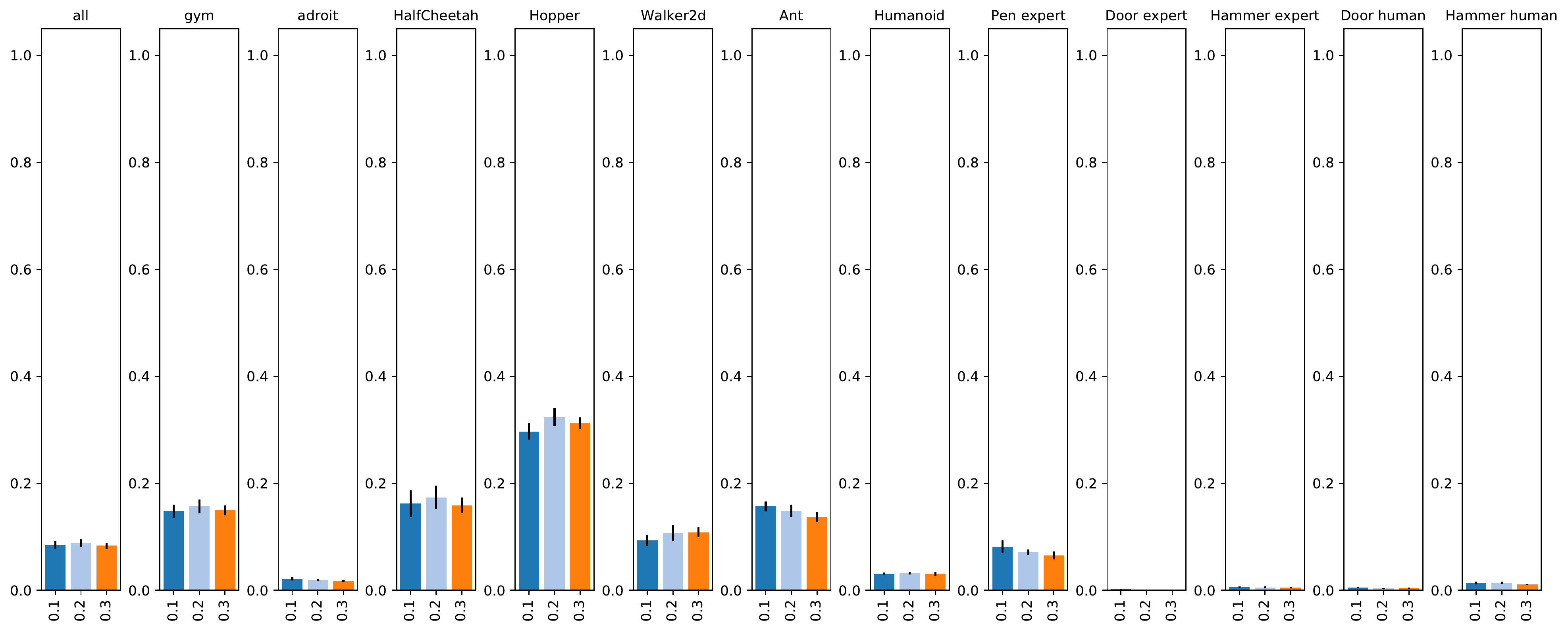}}
\centerline{\includegraphics[height=4.5cm,width=1\textwidth]{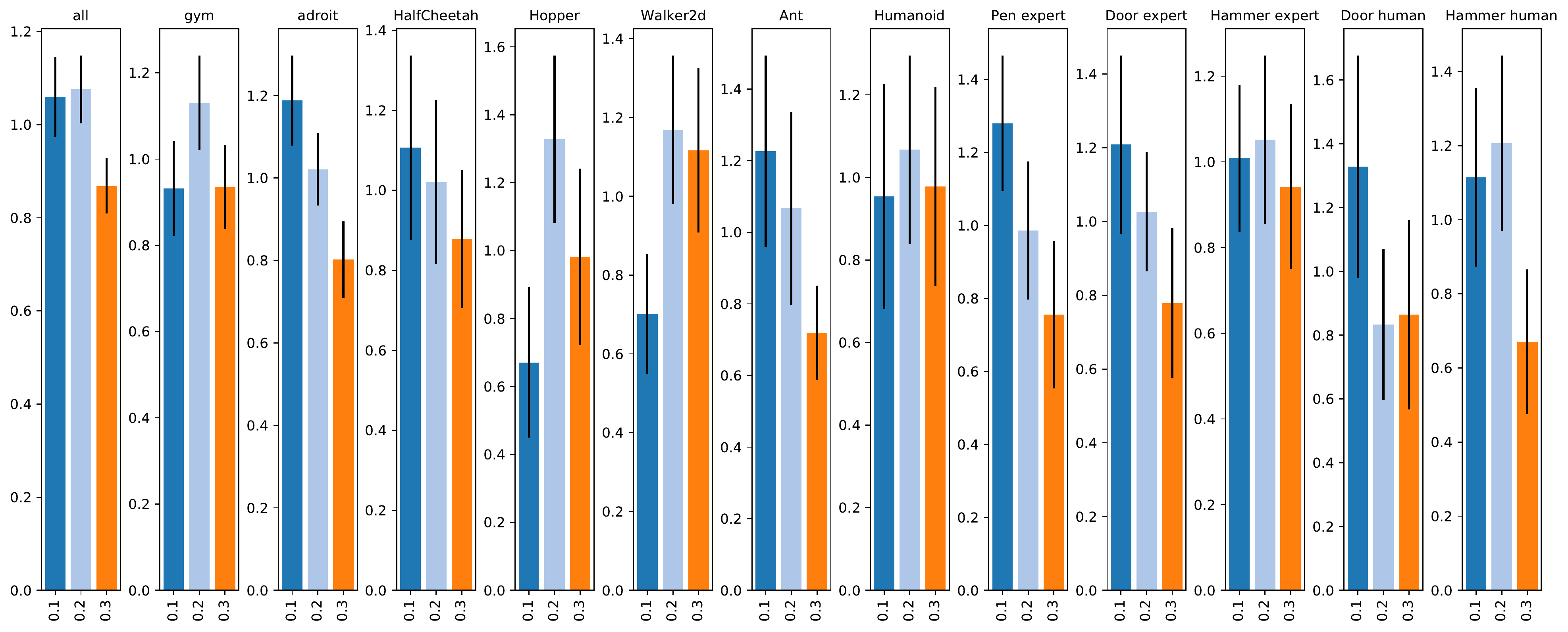}}
\caption{Analysis of choice \choicet{ppoclippingepsilon}: 95th percentile of performance scores conditioned on choice (top) and distribution of choices in top 5\% of configurations (bottom).}
\label{fig:wide_ppo_clipping_epsilon}
\end{center}
\end{figure}

\begin{figure}[ht]
\begin{center}
\centerline{\includegraphics[height=4.5cm,width=1\textwidth]{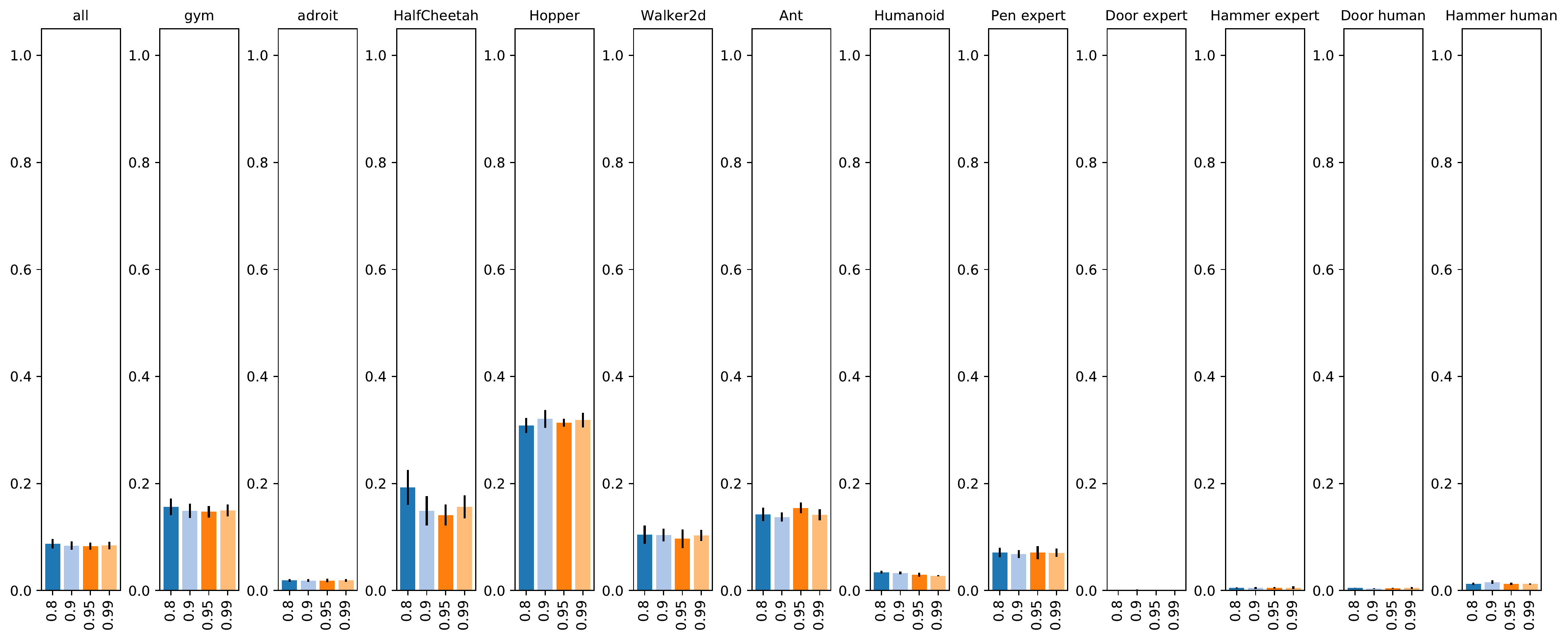}}
\centerline{\includegraphics[height=4.5cm,width=1\textwidth]{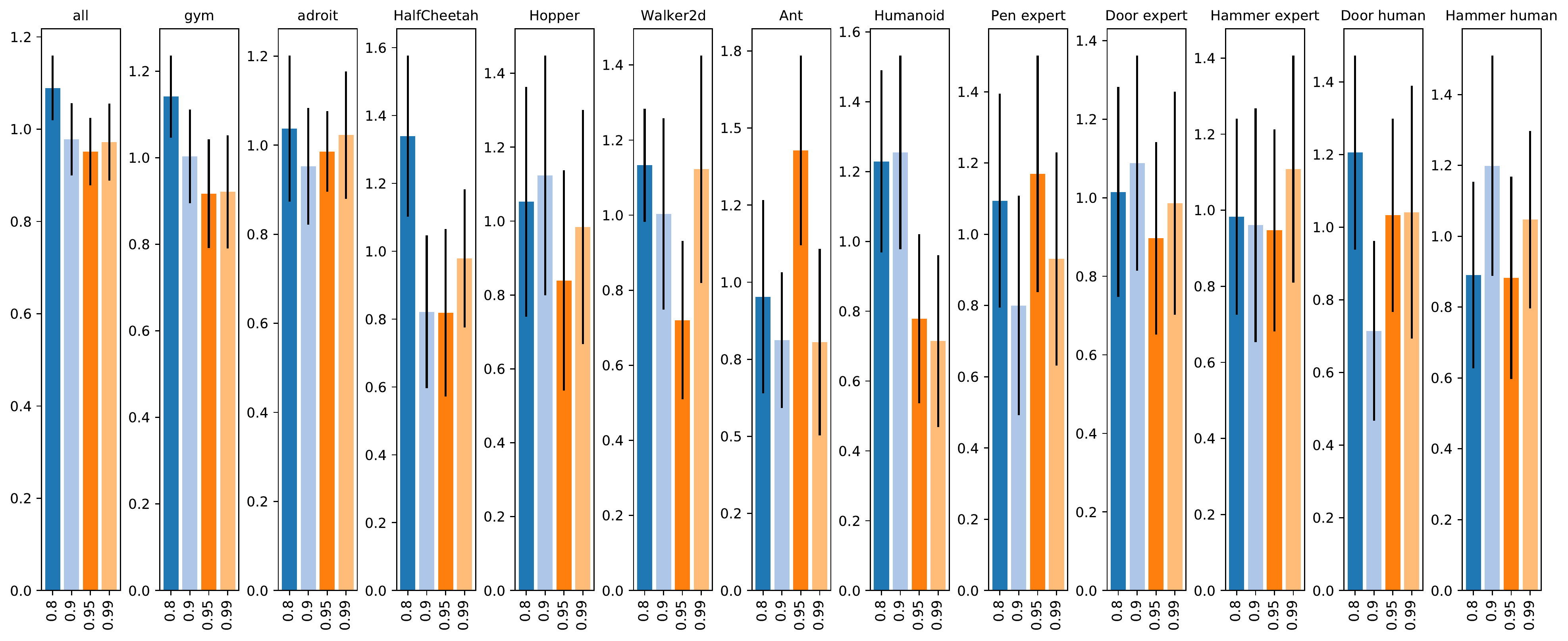}}
\caption{Analysis of choice \choicet{ppogaelambda}: 95th percentile of performance scores conditioned on choice (top) and distribution of choices in top 5\% of configurations (bottom).}
\label{fig:wide_ppo_gae_lambda}
\end{center}
\end{figure}

For each of the sampled choice configurations we compute the performance metric as described in Section~\ref{sec:design}.
We report aggregate statistics of the experiment in Tables~\ref{tab:wide_overview}--\ref{tab:wide_overview4} as well as training curves in Figure~\ref{fig:wide_training_curves}.
We further provide per-choice analyses in Figures~\ref{fig:wide_direct_rl_algorithm}-\ref{fig:wide_ppo_gae_lambda}.
\begin{table}[ht]
\begin{center}
\caption{Quantiles of the \emph{final} agent performance across HP configurations for OpenAI Gym tasks.}
\label{tab:wide_overview}
\begin{tabular}{lrrrrr}
\toprule
{} &  Ant & HalfCheetah & Hopper & Humanoid & Walker2d \\
\midrule
90\% & 0.18 &        0.80 &   0.99 &     0.06 &     0.56 \\
95\% & 0.56 &        0.98 &   1.15 &     0.30 &     0.85 \\
99\% & 0.92 &        1.10 &   1.20 &     0.79 &     0.99 \\
Max & 1.10 &        1.39 &   1.32 &     1.02 &     1.06 \\
\bottomrule
\end{tabular}

\end{center}
\end{table}\begin{table}[ht]
\begin{center}
\caption{Quantiles of the \emph{final} agent performance across HP configurations for Adroit tasks.}
\label{tab:wide_overview2}
\begin{tabular}{lrrrrr}
\toprule
{} & Door expert & Door human & Hammer expert & Hammer human & Pen expert \\
\midrule
90\% &        0.12 &       0.07 &          0.16 &         0.12 &       0.28 \\
95\% &        0.42 &       0.28 &          0.67 &         0.47 &       0.46 \\
99\% &        0.90 &       1.20 &          1.26 &         2.03 &       0.77 \\
Max &        1.11 &       2.82 &          1.42 &         5.39 &       1.12 \\
\bottomrule
\end{tabular}

\end{center}
\end{table}\begin{table}[ht]
\begin{center}
\caption{Quantiles of the \emph{average} agent performance during training across HP configurations for OpenAI Gym tasks.}
\label{tab:wide_overview3}
\begin{tabular}{lrrrrr}
\toprule
{} &  Ant & HalfCheetah & Hopper & Humanoid & Walker2d \\
\midrule
90\% & 0.13 &        0.54 &   0.62 &     0.05 &     0.31 \\
95\% & 0.31 &        0.66 &   0.80 &     0.20 &     0.49 \\
99\% & 0.62 &        0.85 &   0.98 &     0.49 &     0.71 \\
Max & 0.94 &        0.99 &   1.08 &     0.84 &     0.92 \\
\bottomrule
\end{tabular}

\end{center}
\end{table}\begin{table}[ht]
\begin{center}
\caption{Quantiles of the \emph{average} agent performance during training across HP configurations for Adroit tasks.}
\label{tab:wide_overview4}
\begin{tabular}{lrrrrr}
\toprule
{} & Door expert & Door human & Hammer expert & Hammer human & Pen expert \\
\midrule
90\% &        0.11 &       0.11 &          0.11 &         0.15 &       0.21 \\
95\% &        0.27 &       0.24 &          0.34 &         0.33 &       0.35 \\
99\% &        0.55 &       0.61 &          0.78 &         0.85 &       0.59 \\
Max &        0.87 &       1.65 &          1.01 &         1.97 &       0.84 \\
\bottomrule
\end{tabular}

\end{center}
\end{table}


\clearpage

\begin{figure}[ht]
\begin{center}
\centerline{\includegraphics[height=4cm]{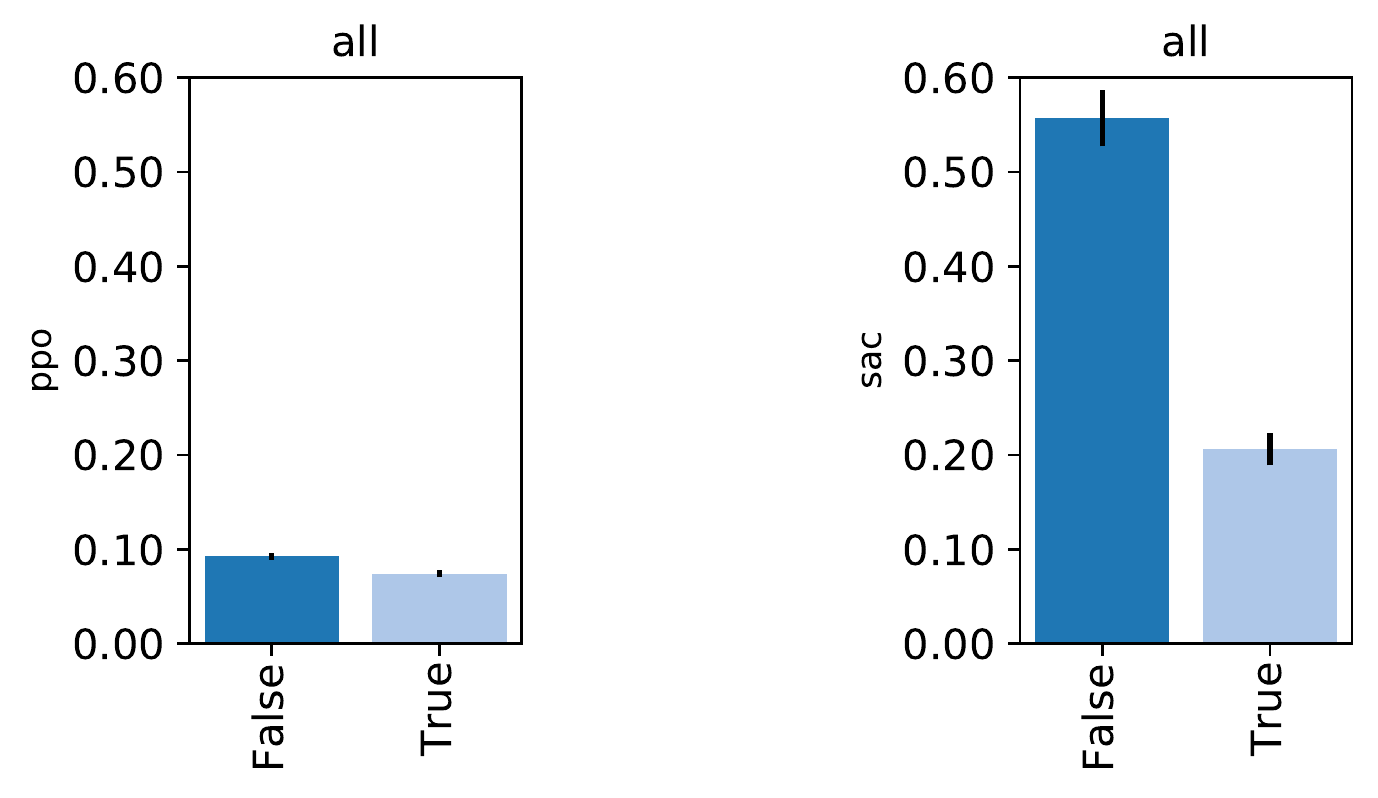}}
\caption{95th percentile of performance scores conditioned on \choicet{directrlalgorithm}(subplots) and \choicet{subtractlogp}(bars).}
\label{fig:corr_rl_logpi}
\end{center}
\end{figure}

\begin{figure}[ht]
\begin{center}
\centerline{\includegraphics[height=4cm]{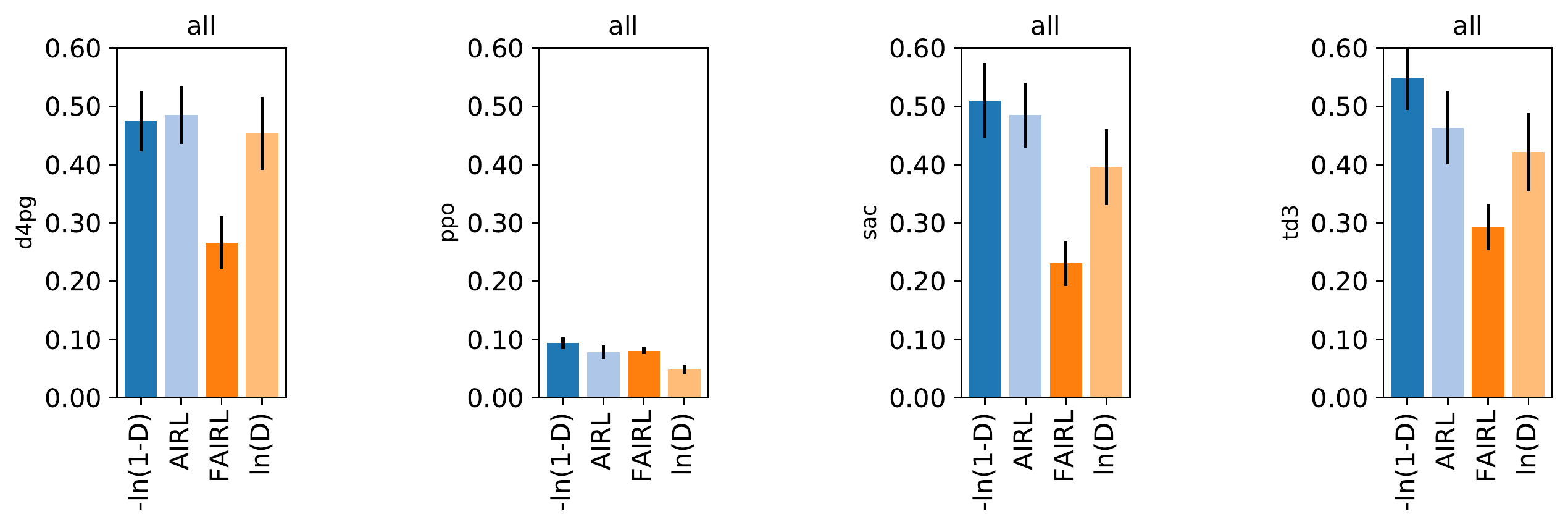}}
\caption{95th percentile of performance scores conditioned on \choicet{directrlalgorithm}(subplots) and \choicet{gailreward}(bars).}
\label{fig:corr_rl_reward}
\end{center}
\end{figure}

\begin{figure}[ht]
\begin{center}
\centerline{\includegraphics[height=4cm]{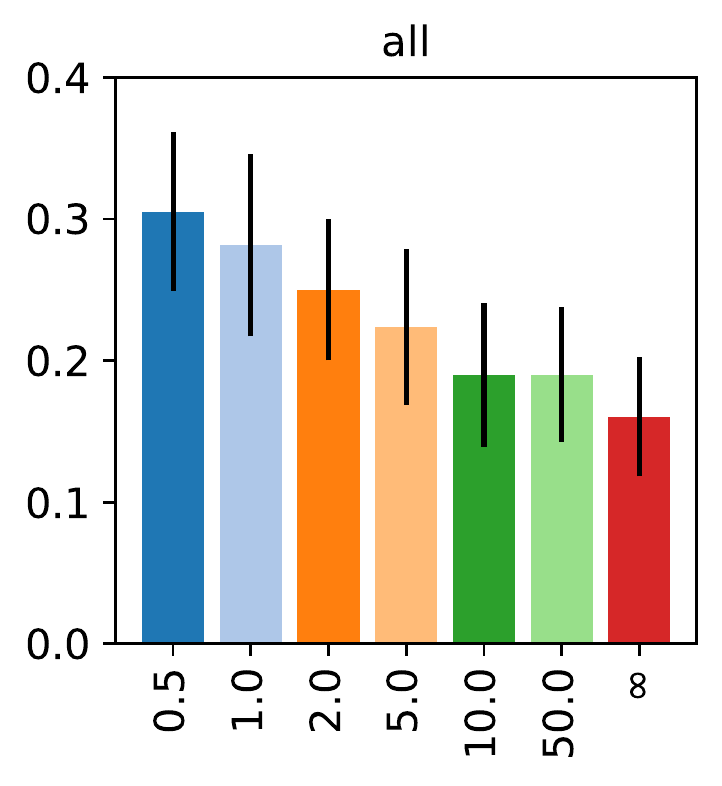}}
\caption{95th percentile of performance scores conditioned on \choicet{gailmaxrewardmagnitude}
and \choicet{gailreward}\texttt{=FAIRL}.}
\label{fig:fairl_clipping}
\end{center}
\end{figure}
\clearpage
\section{Experiment main}
\label{exp_main}
\subsection{Design}
\label{exp_design_main}
For each of the 10 tasks, we sampled 25334 choice configurations where we sampled the following choices independently and uniformly from the following ranges:
\begin{itemize}
    \item \choicet{directrlalgorithm}: \{d4pg, sac, td3\}
    \begin{itemize}
        \item For the case ``\choicet{directrlalgorithm} = sac'', we further sampled the sub-choices:
        \begin{itemize}
            \item \choicet{saclearningrate}: \{0.0001, 0.0003, 0.001\}
            \item \choicet{sactargetentropyperdimension}: \{-2.0, -1.0, -0.5, 0.0\}
            \item \choicet{sactau}: \{0.001, 0.003, 0.01, 0.03\}
        \end{itemize}
        \item For the case ``\choicet{directrlalgorithm} = d4pg'', we further sampled the sub-choices:
        \begin{itemize}
            \item \choicet{dfpglearningrate}: \{3e-05, 0.0001, 0.0003\}
            \item \choicet{rlsigma}: \{0.1, 0.2, 0.3, 0.5\}
            \item \choicet{vmax}: \{150.0, 750.0, 1500.0\}
            \item \choicet{numatoms}: \{51.0, 101.0, 201.0, 401.0\}
            \item \choicet{nstep}: \{1.0, 3.0, 5.0\}
        \end{itemize}
        \item For the case ``\choicet{directrlalgorithm} = td3'', we further sampled the sub-choices:
        \begin{itemize}
            \item \choicet{tdtpolicylearningrate}: \{0.0001, 0.0003, 0.001\}
            \item \choicet{tdtcriticlearningrate}: \{0.0001, 0.0003, 0.001\}
            \item \choicet{tdtgradientclipping}: \{40.0, $\infty$\}
            \item \choicet{rlsigma}: \{0.1, 0.2, 0.3, 0.5\}
        \end{itemize}
    \end{itemize}
    \item \choicet{maxreplaysize}: \{300000, 1000000, 3000000\}
    \item \choicet{numpolicylayers}: \{1, 2, 3\}
    \item \choicet{policylayersize}: \{64, 128, 256, 512\}
    \item \choicet{numcriticlayers}: \{2, 3\}
    \item \choicet{criticlayersize}: \{256, 512\}
    \item \choicet{activation}: \{relu, tanh\}
    \item \choicet{discount}: \{0.97, 0.99\}
    \item \choicet{pretrainwithbc}: \{False, True\}
    \item \choicet{explicitabsorbingstate}: \{False, True\}
    \item \choicet{gailmaxreplaysize}: \{300000, 1000000, 3000000\}
    \item \choicet{gaildiscriminatormodule}: \{False, True\}
    \item \choicet{gailinput}: \{s, sa, sas, ss\}
    \item \choicet{gailmlpnumlayers}: \{1, 2, 3\}
    \item \choicet{gailmlpnumwidth}: \{16, 32, 64, 128, 256, 512\}
    \item \choicet{gailmlpactivation}: \{elu, leaky\_relu, relu, sigmoid, swish, tanh\}
    \item \choicet{gailmlplastlayerinitscale}: \{0.001, 1.0\}
    \item \choicet{regularizer}: \{GP, Mixup, No regularizer, PUGAIL, dropout, entropy, spectral norm, weight decay\}
    \begin{itemize}
        \item For the case ``\choicet{regularizer} = GP'', we further sampled the sub-choices:
        \begin{itemize}
            \item \choicet{gpcoef}: \{0.1, 1.0, 10.0\}
            \item \choicet{gptarget}: \{0.0, 1.0\}
        \end{itemize}
        \item For the case ``\choicet{regularizer} = Mixup'', we further sampled the sub-choices:
        \begin{itemize}
            \item \choicet{mixupalpha}: \{0.1, 0.4, 1.0\}
        \end{itemize}
        \item For the case ``\choicet{regularizer} = PUGAIL'', we further sampled the sub-choices:
        \begin{itemize}
            \item \choicet{pugailpositiveclassprior}: \{0.25, 0.5, 0.7\}
            \item \choicet{pugailbeta}: \{0.0, 0.7, $\infty$\}
        \end{itemize}
        \item For the case ``\choicet{regularizer} = entropy'', we further sampled the sub-choices:
        \begin{itemize}
            \item \choicet{regentropycoef}: \{0.0003, 0.001, 0.003, 0.01, 0.03, 0.1, 0.3\}
        \end{itemize}
        \item For the case ``\choicet{regularizer} = weight decay'', we further sampled the sub-choices:
        \begin{itemize}
            \item \choicet{regweightdecay}: \{0.3, 1.0, 3.0, 10.0, 30.0\}
        \end{itemize}
        \item For the case ``\choicet{regularizer} = dropout'', we further sampled the sub-choices:
        \begin{itemize}
            \item \choicet{dropoutinputrate}: \{0.0, 0.25, 0.5, 0.75\}
            \item \choicet{dropouthiddenrate}: \{0.25, 0.5, 0.75\}
        \end{itemize}
    \end{itemize}
    \item \choicet{obsnormalization}: \{fixed, none, online\}
    \item \choicet{evalbehaviorpolicytype}: \{average, mode, stochastic\}
    \item \choicet{gaildiscriminatorlearningrate}: \{1e-06, 3e-06, 1e-05, 3e-05, 0.0001, 0.0003\}
    \item \choicet{gailreward}: \{-ln(1-D), AIRL, ln(D)\}
    \item \choicet{batchsize}: \{256\}
    \item \choicet{samplesperinsert}: \{256\}
    \item \choicet{discriminatortorlupdatesratio}: \{1\}
    \item \choicet{gradupdatesperbatch}: \{8\}
\end{itemize}

\subsection{Results}
\label{exp_results_main}
\begin{figure}[ht]
\begin{center}
\centerline{\includegraphics[width=1\textwidth]{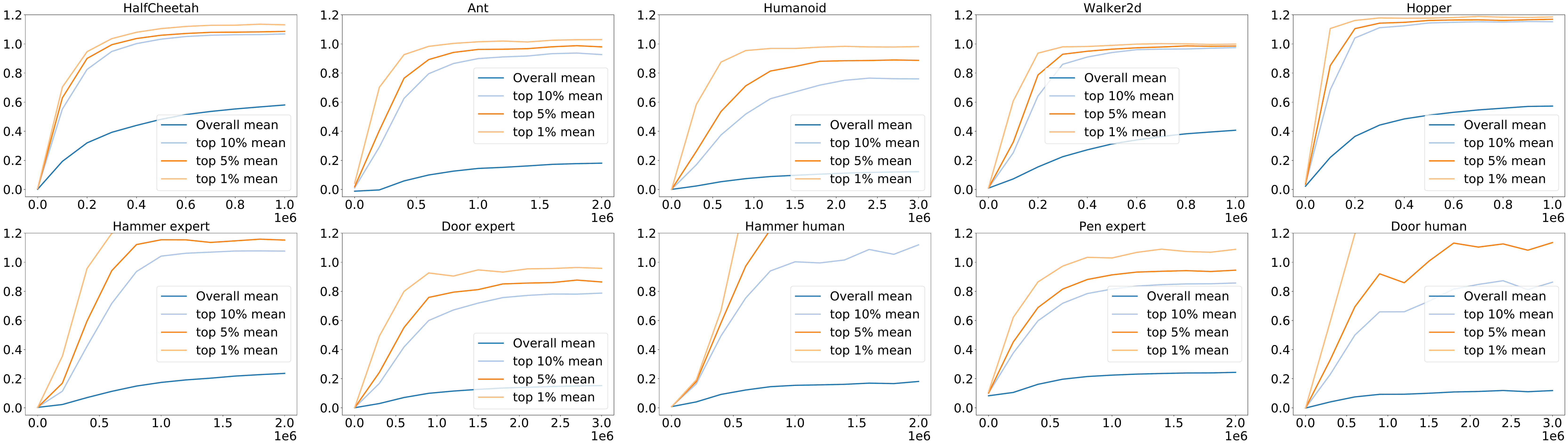}}
\caption{Training curves.}
\label{fig:main_training_curves}
\end{center}
\end{figure}

\begin{figure}[ht]
\begin{center}
\centerline{\includegraphics[height=4.5cm,width=1\textwidth]{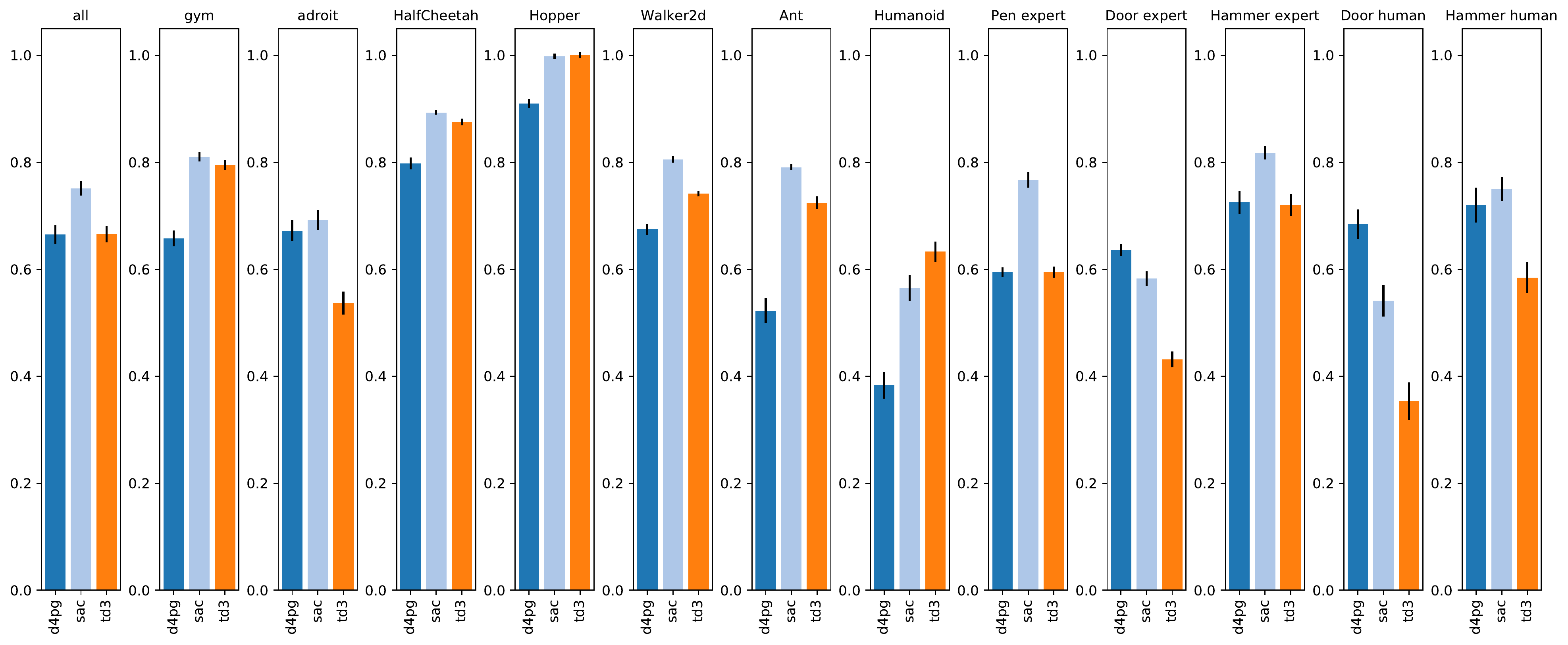}}
\centerline{\includegraphics[height=4.5cm,width=1\textwidth]{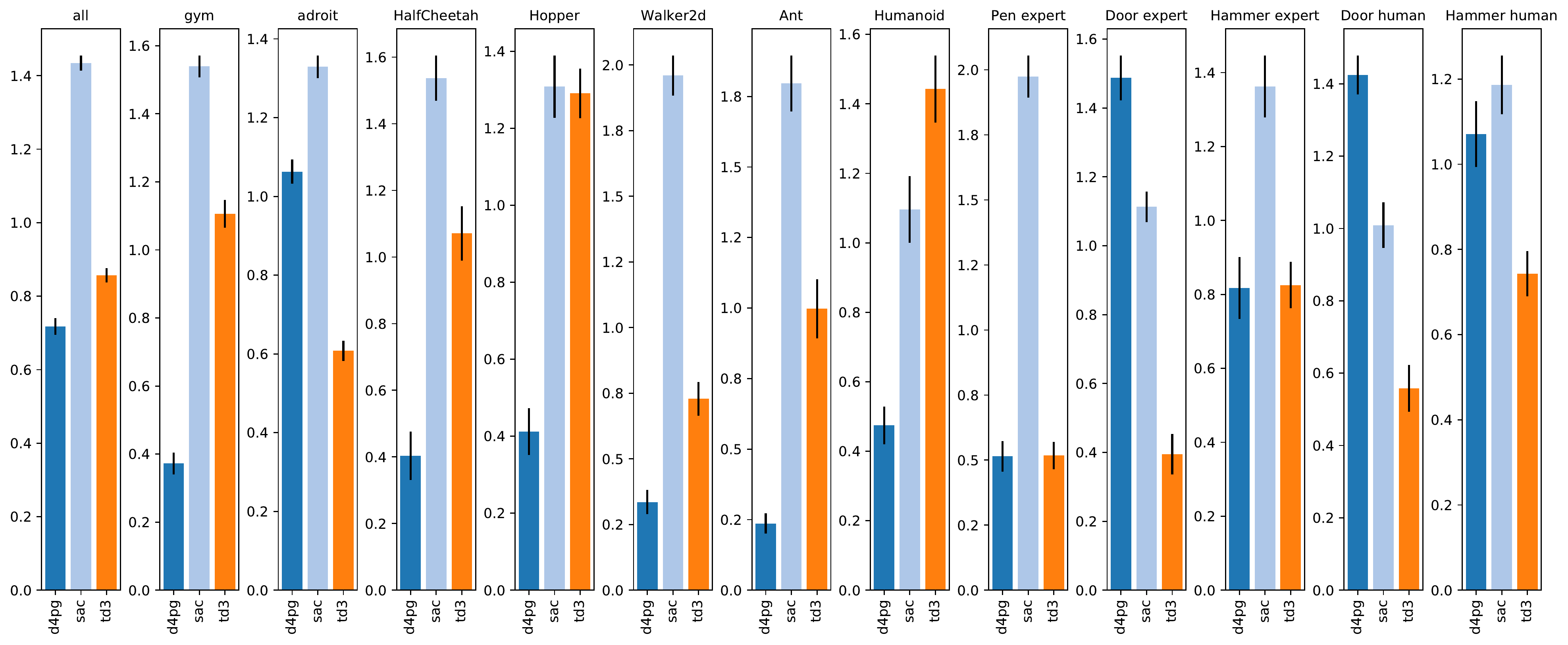}}
\caption{Analysis of choice \choicet{directrlalgorithm}: 95th percentile of performance scores conditioned on choice (top) and distribution of choices in top 5\% of configurations (bottom).}
\label{fig:main_direct_rl_algorithm}
\end{center}
\end{figure}

\begin{figure}[ht]
\begin{center}
\centerline{\includegraphics[height=4.5cm,width=1\textwidth]{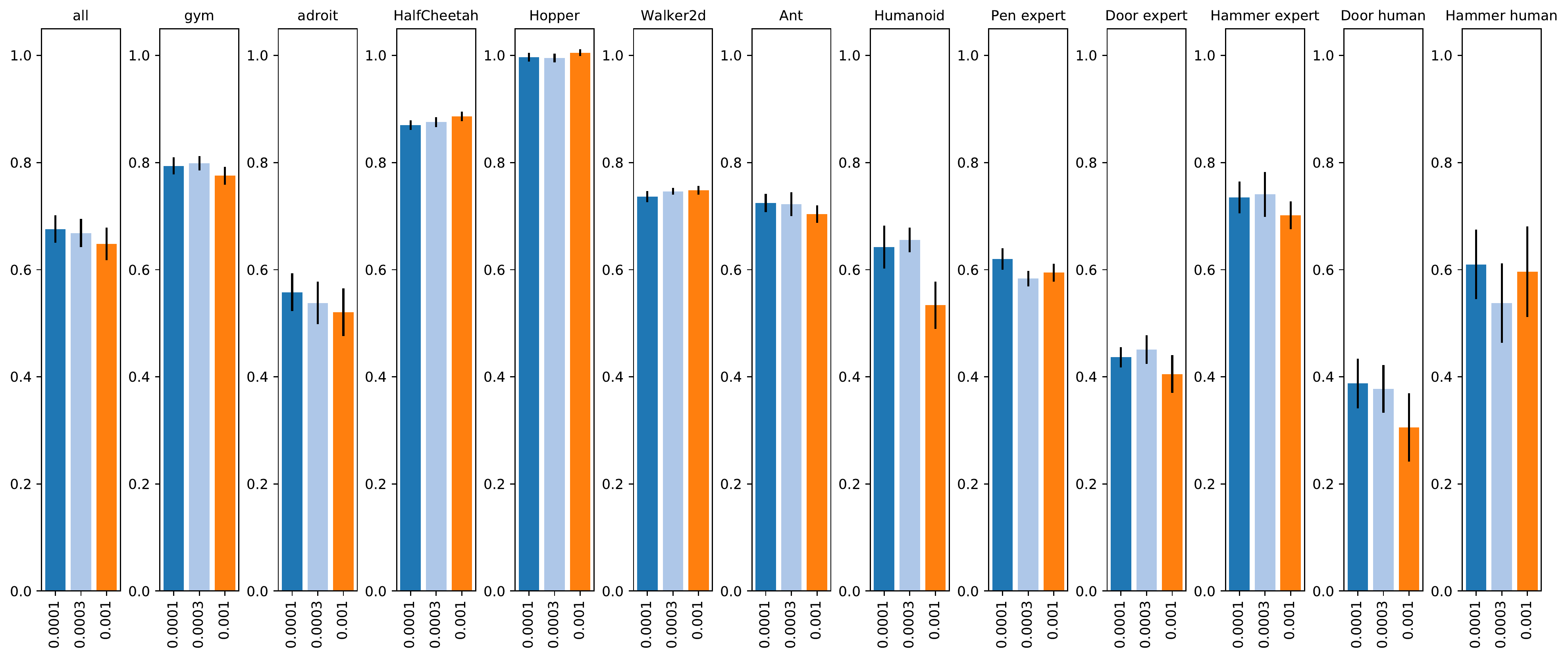}}
\centerline{\includegraphics[height=4.5cm,width=1\textwidth]{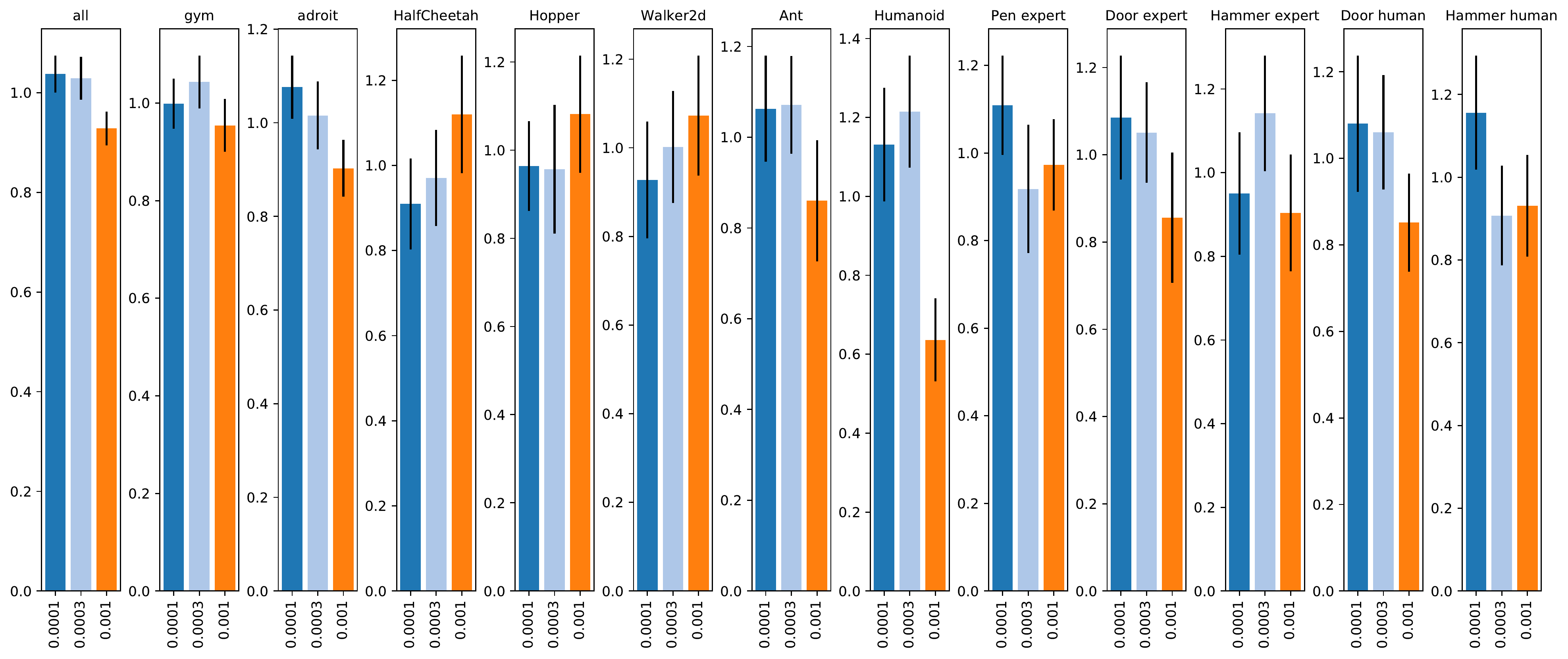}}
\caption{Analysis of choice \choicet{tdtpolicylearningrate}: 95th percentile of performance scores conditioned on choice (top) and distribution of choices in top 5\% of configurations (bottom).}
\label{fig:main_td3_policy_learning_rate}
\end{center}
\end{figure}

\begin{figure}[ht]
\begin{center}
\centerline{\includegraphics[height=4.5cm,width=1\textwidth]{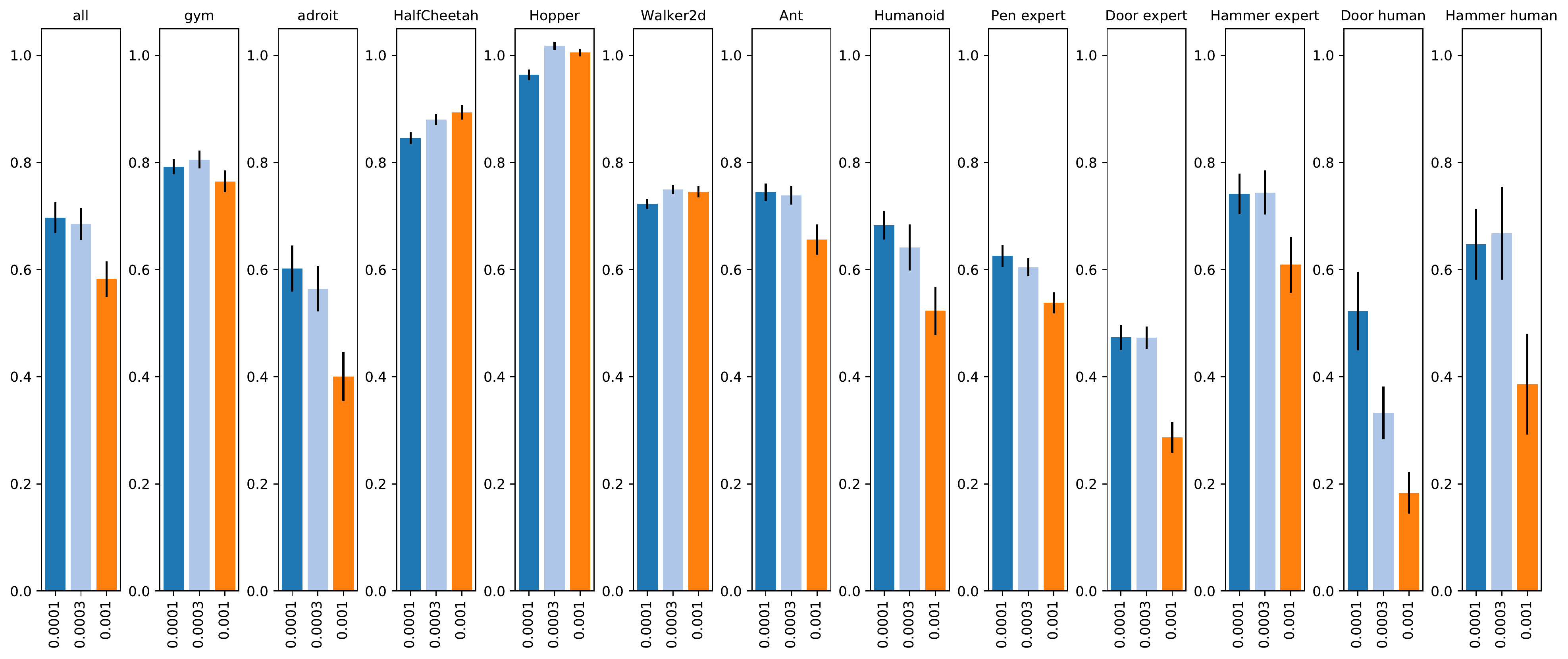}}
\centerline{\includegraphics[height=4.5cm,width=1\textwidth]{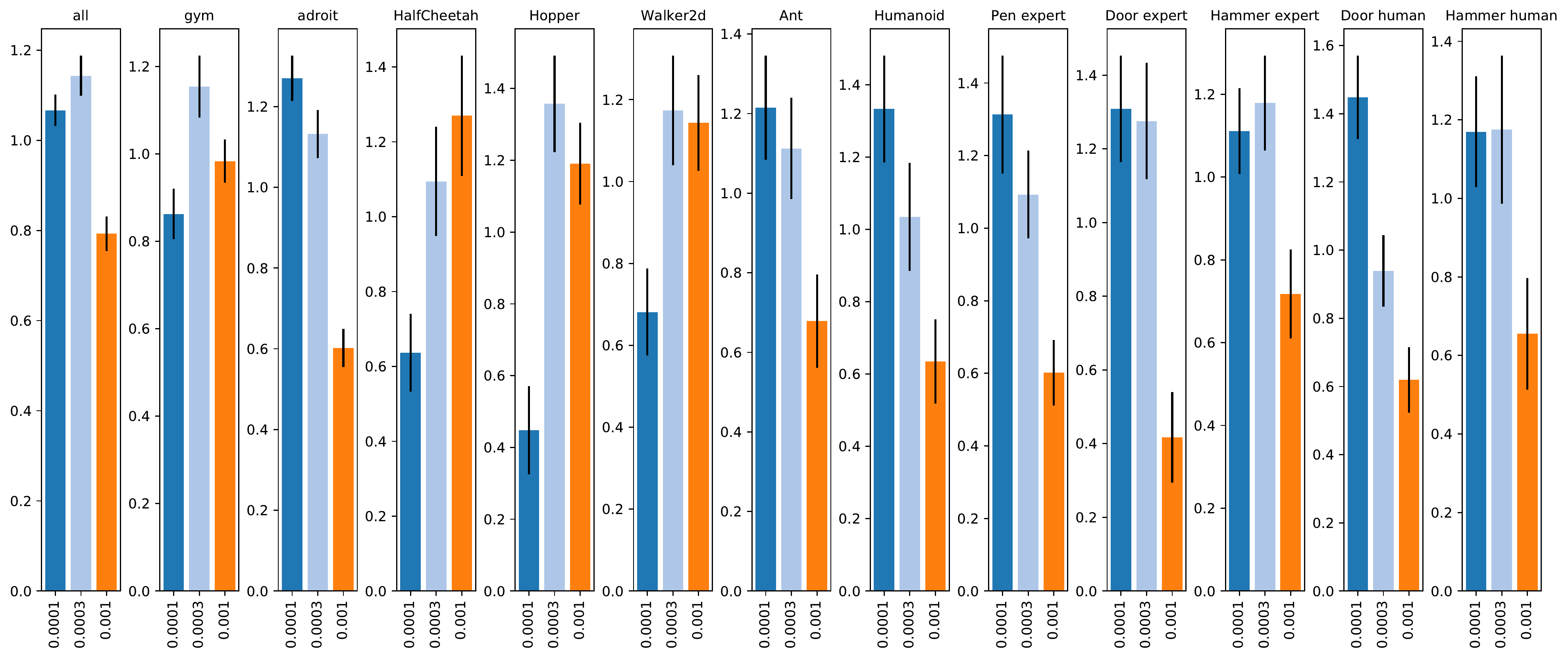}}
\caption{Analysis of choice \choicet{tdtcriticlearningrate}: 95th percentile of performance scores conditioned on choice (top) and distribution of choices in top 5\% of configurations (bottom).}
\label{fig:main_td3_critic_learning_rate}
\end{center}
\end{figure}

\begin{figure}[ht]
\begin{center}
\centerline{\includegraphics[height=4.5cm,width=1\textwidth]{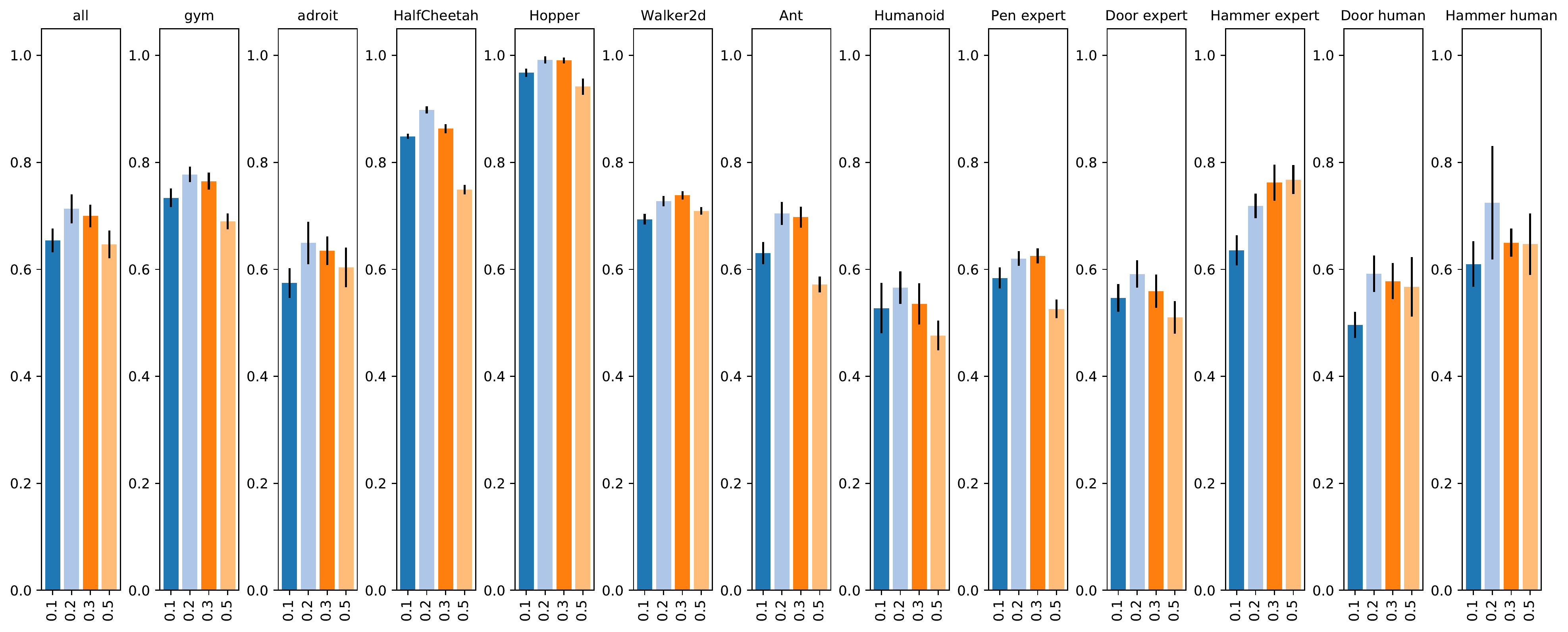}}
\centerline{\includegraphics[height=4.5cm,width=1\textwidth]{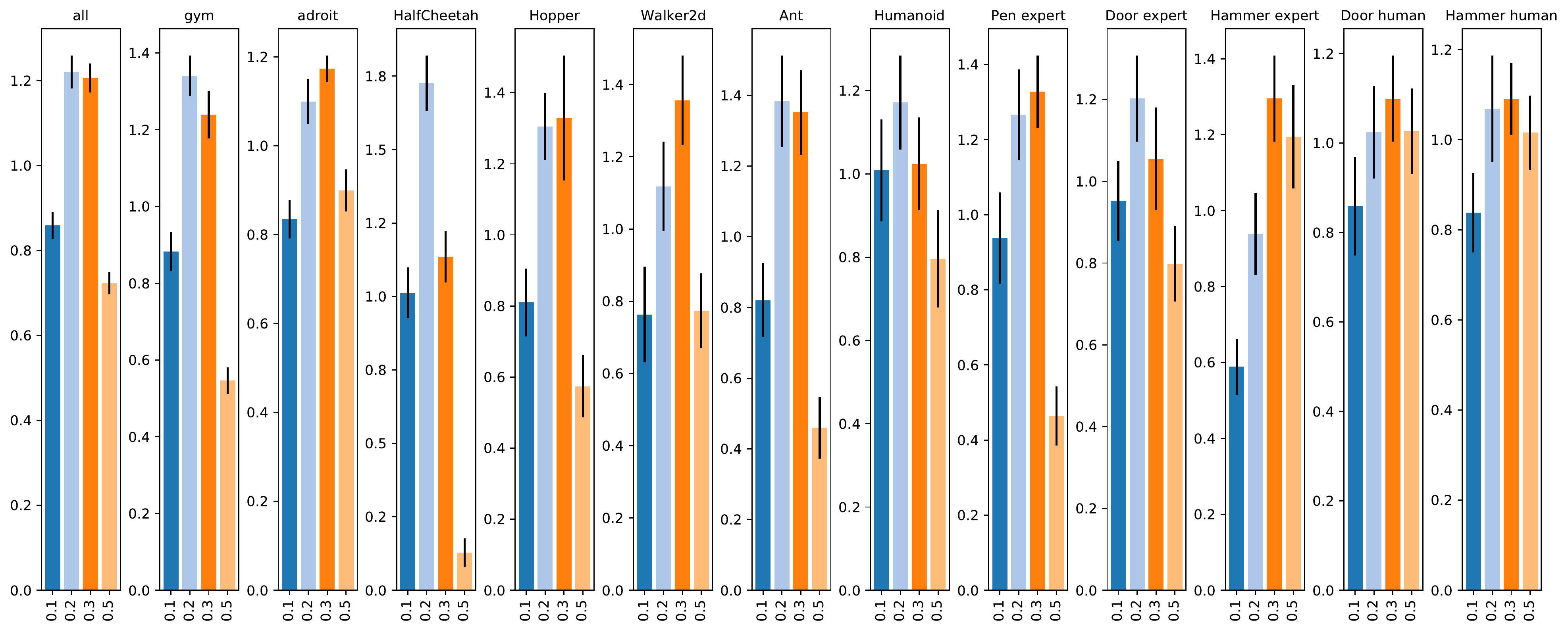}}
\caption{Analysis of choice \choicet{rlsigma}: 95th percentile of performance scores conditioned on choice (top) and distribution of choices in top 5\% of configurations (bottom).}
\label{fig:main_sigma}
\end{center}
\end{figure}

\begin{figure}[ht]
\begin{center}
\centerline{\includegraphics[height=4.5cm,width=1\textwidth]{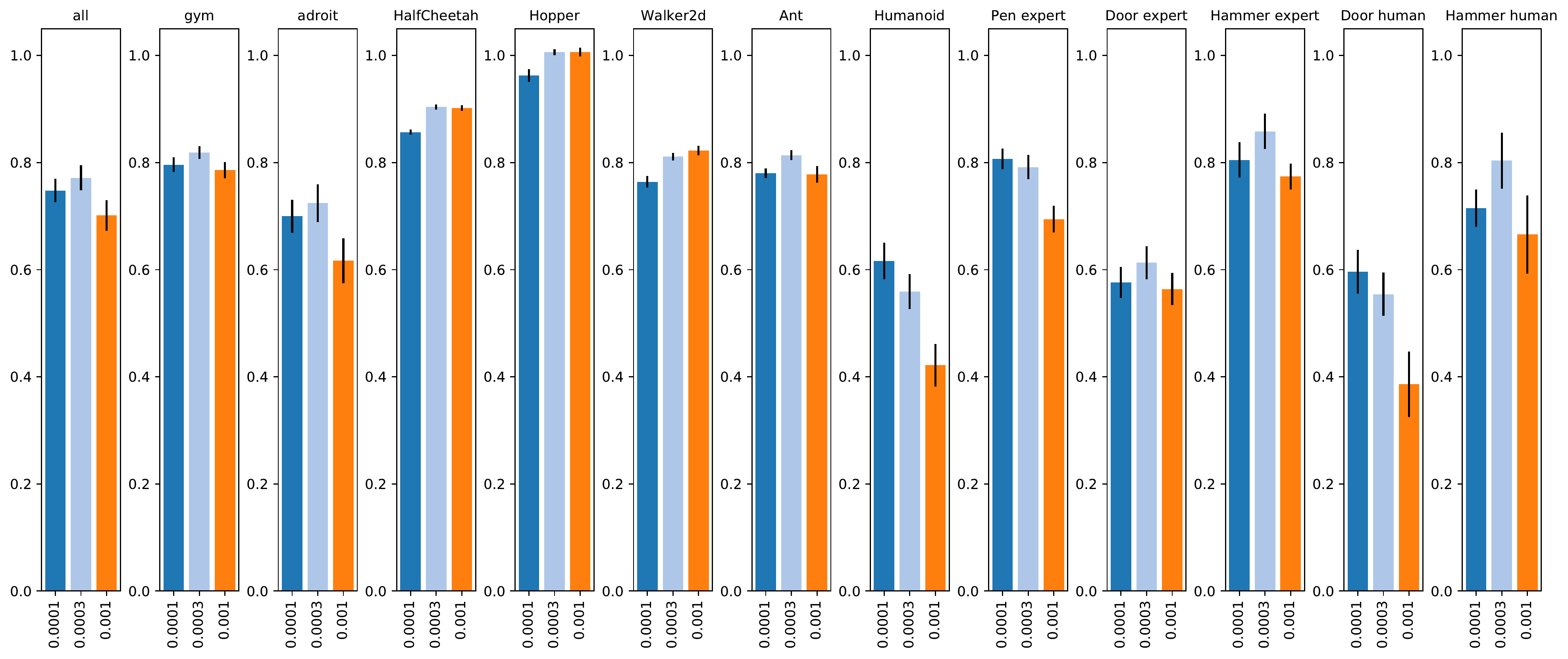}}
\centerline{\includegraphics[height=4.5cm,width=1\textwidth]{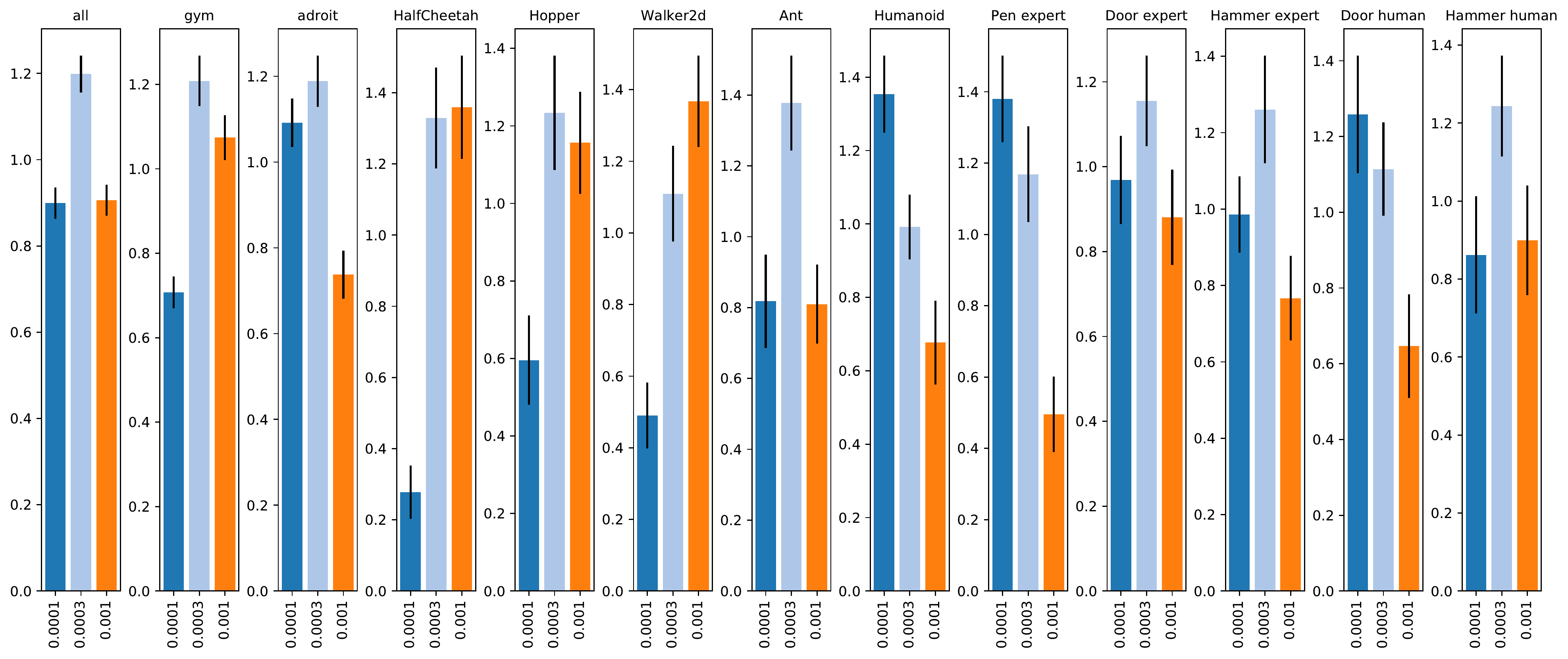}}
\caption{Analysis of choice \choicet{saclearningrate}: 95th percentile of performance scores conditioned on choice (top) and distribution of choices in top 5\% of configurations (bottom).}
\label{fig:main_sac_learning_rate}
\end{center}
\end{figure}

\begin{figure}[ht]
\begin{center}
\centerline{\includegraphics[height=4.5cm,width=1\textwidth]{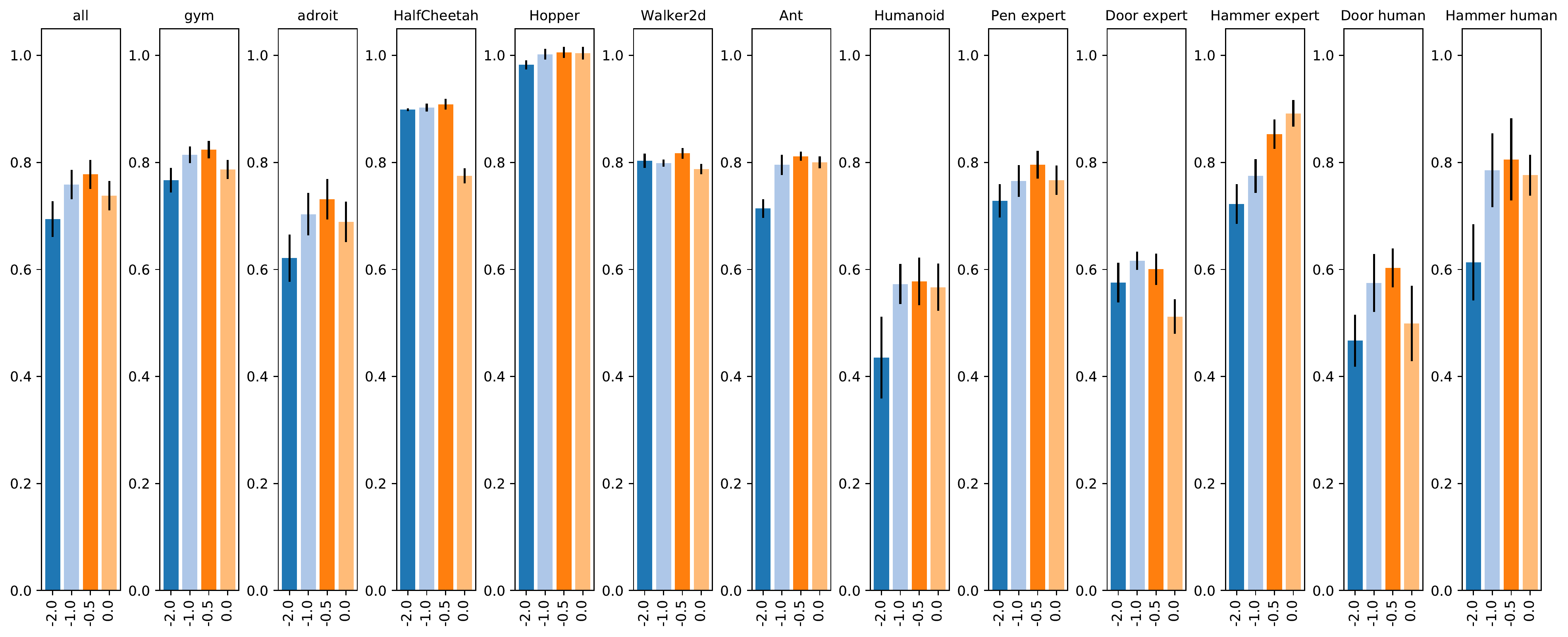}}
\centerline{\includegraphics[height=4.5cm,width=1\textwidth]{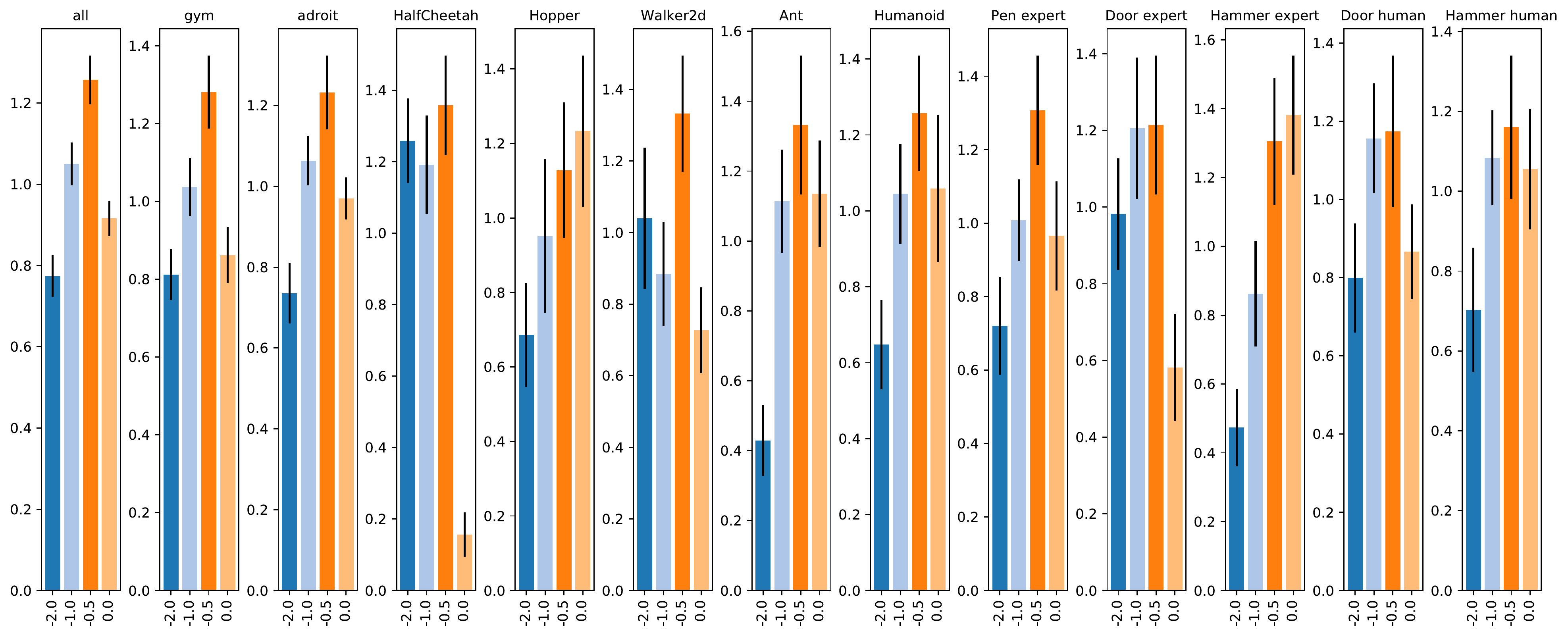}}
\caption{Analysis of choice \choicet{sactargetentropyperdimension}: 95th percentile of performance scores conditioned on choice (top) and distribution of choices in top 5\% of configurations (bottom).}
\label{fig:main_sac_target_entropy_per_dimension}
\end{center}
\end{figure}

\begin{figure}[ht]
\begin{center}
\centerline{\includegraphics[height=4.5cm,width=1\textwidth]{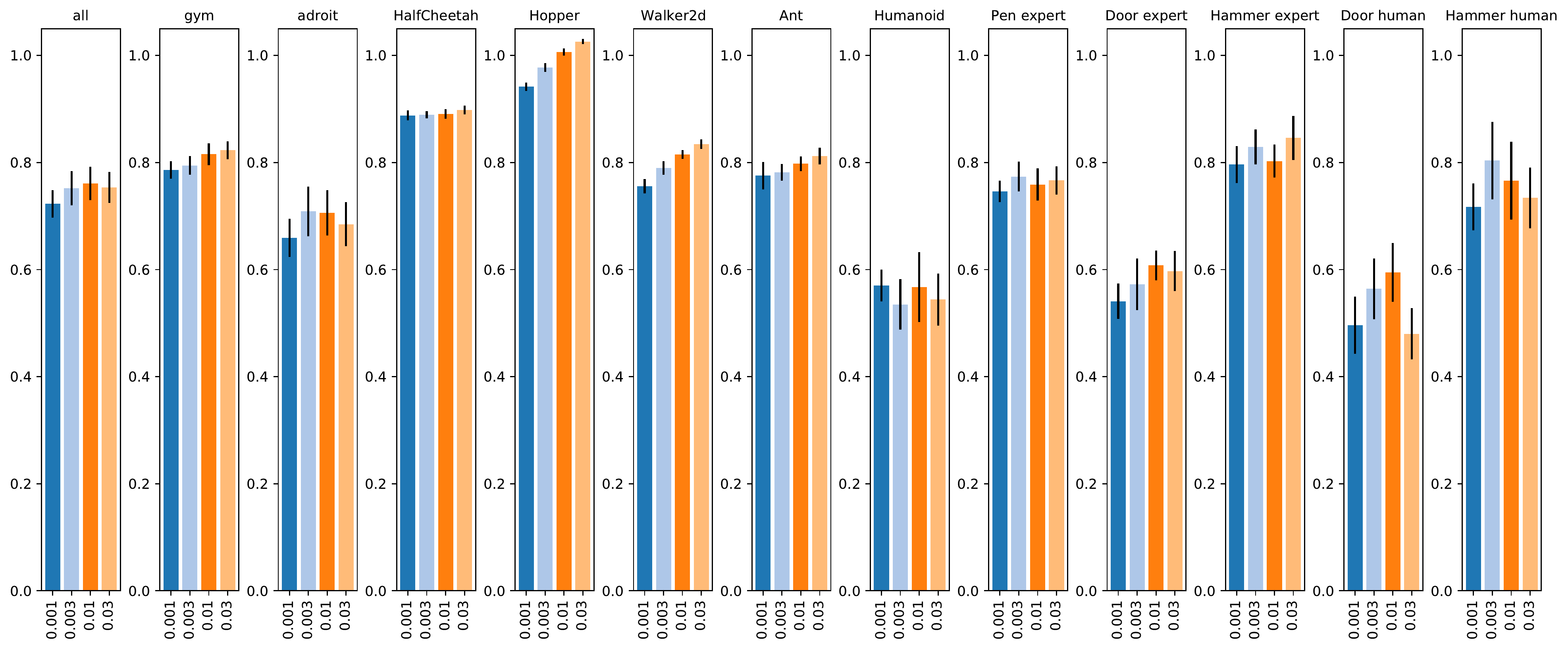}}
\centerline{\includegraphics[height=4.5cm,width=1\textwidth]{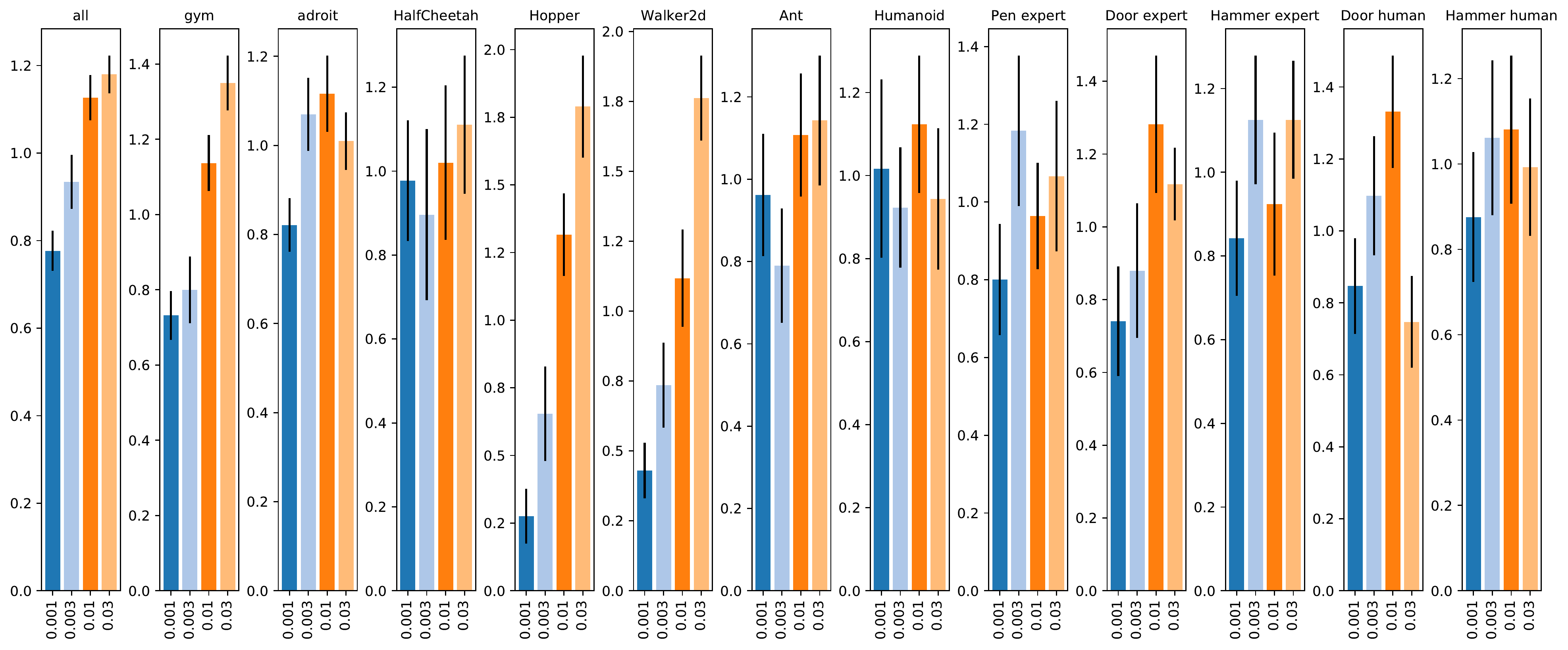}}
\caption{Analysis of choice \choicet{sactau}: 95th percentile of performance scores conditioned on choice (top) and distribution of choices in top 5\% of configurations (bottom).}
\label{fig:main_tau}
\end{center}
\end{figure}

\begin{figure}[ht]
\begin{center}
\centerline{\includegraphics[height=4.5cm,width=1\textwidth]{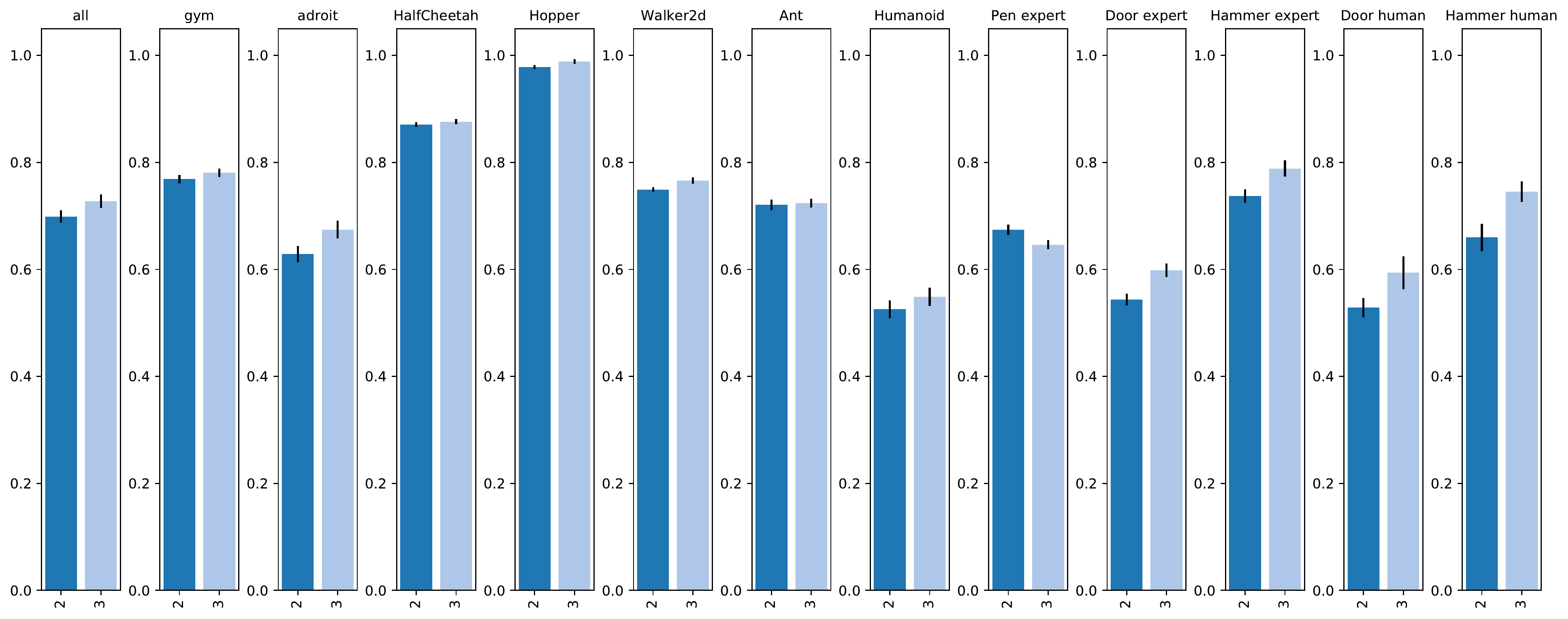}}
\centerline{\includegraphics[height=4.5cm,width=1\textwidth]{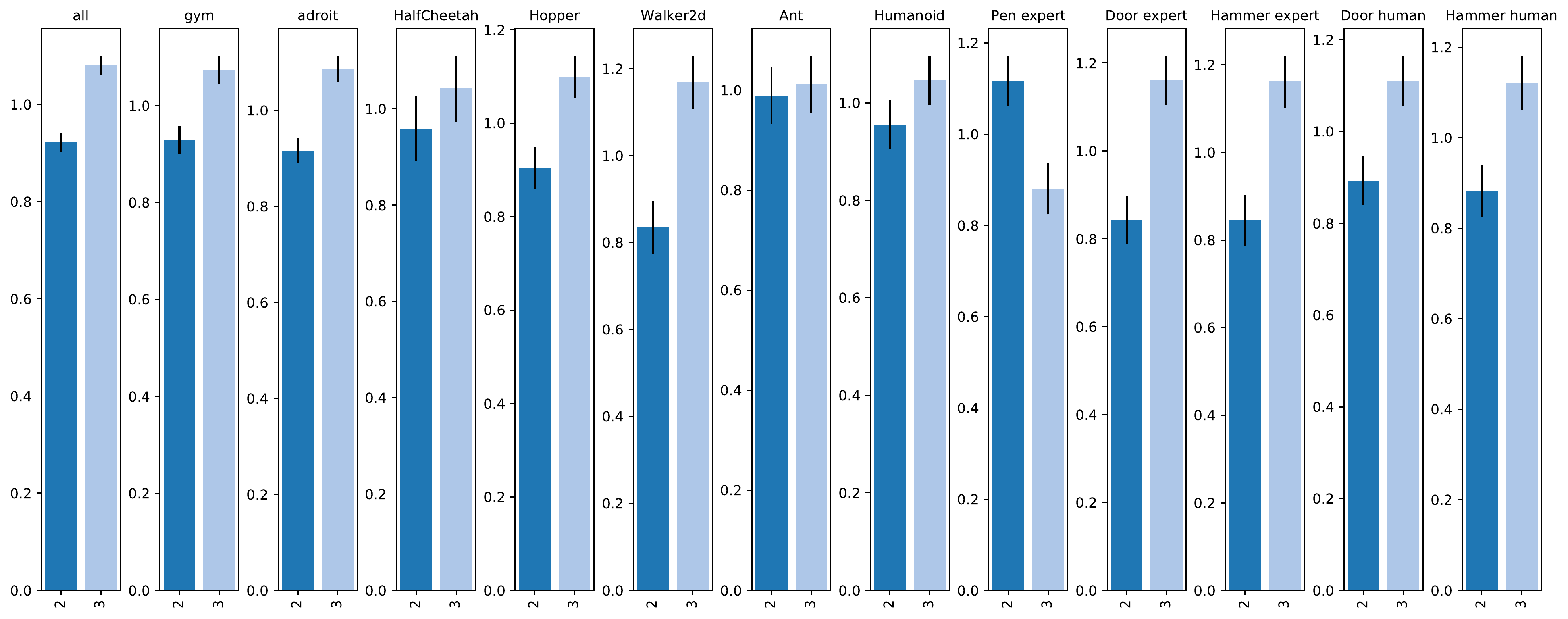}}
\caption{Analysis of choice \choicet{numcriticlayers}: 95th percentile of performance scores conditioned on choice (top) and distribution of choices in top 5\% of configurations (bottom).}
\label{fig:main_num_critic_layers}
\end{center}
\end{figure}

\begin{figure}[ht]
\begin{center}
\centerline{\includegraphics[height=4.5cm,width=1\textwidth]{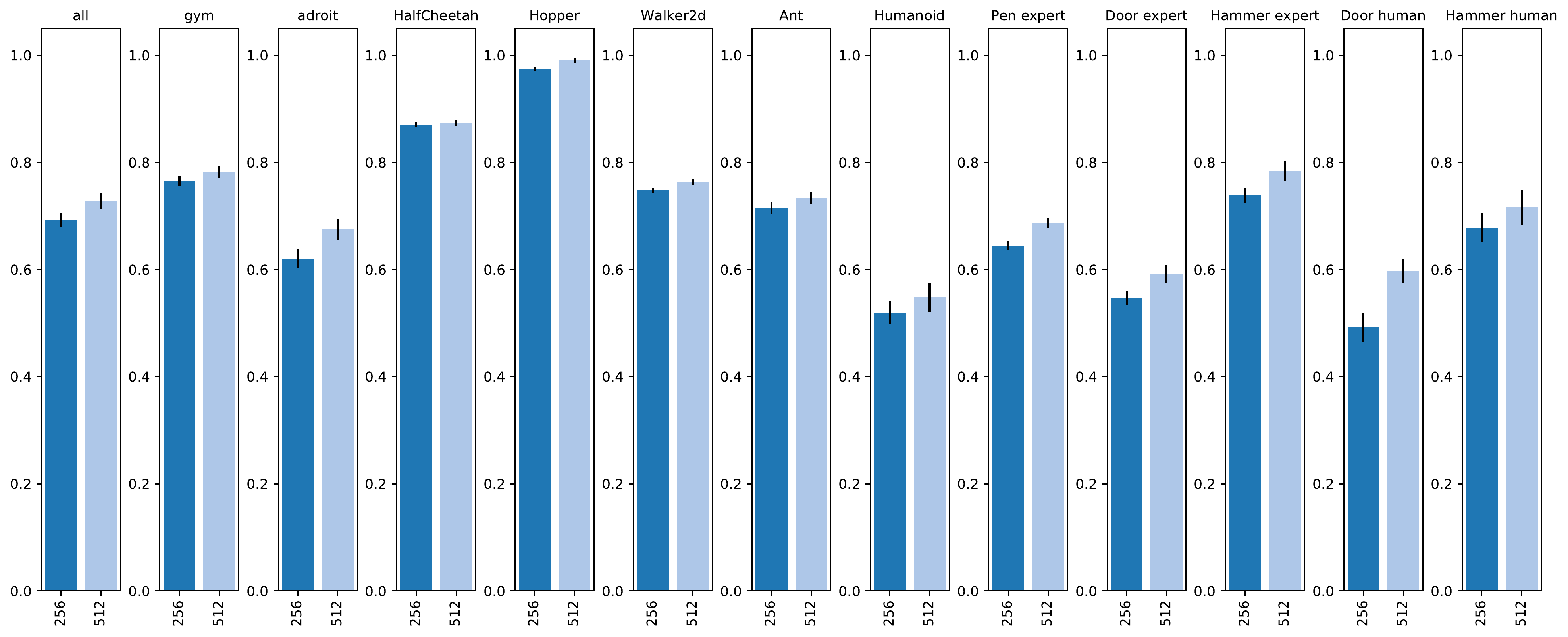}}
\centerline{\includegraphics[height=4.5cm,width=1\textwidth]{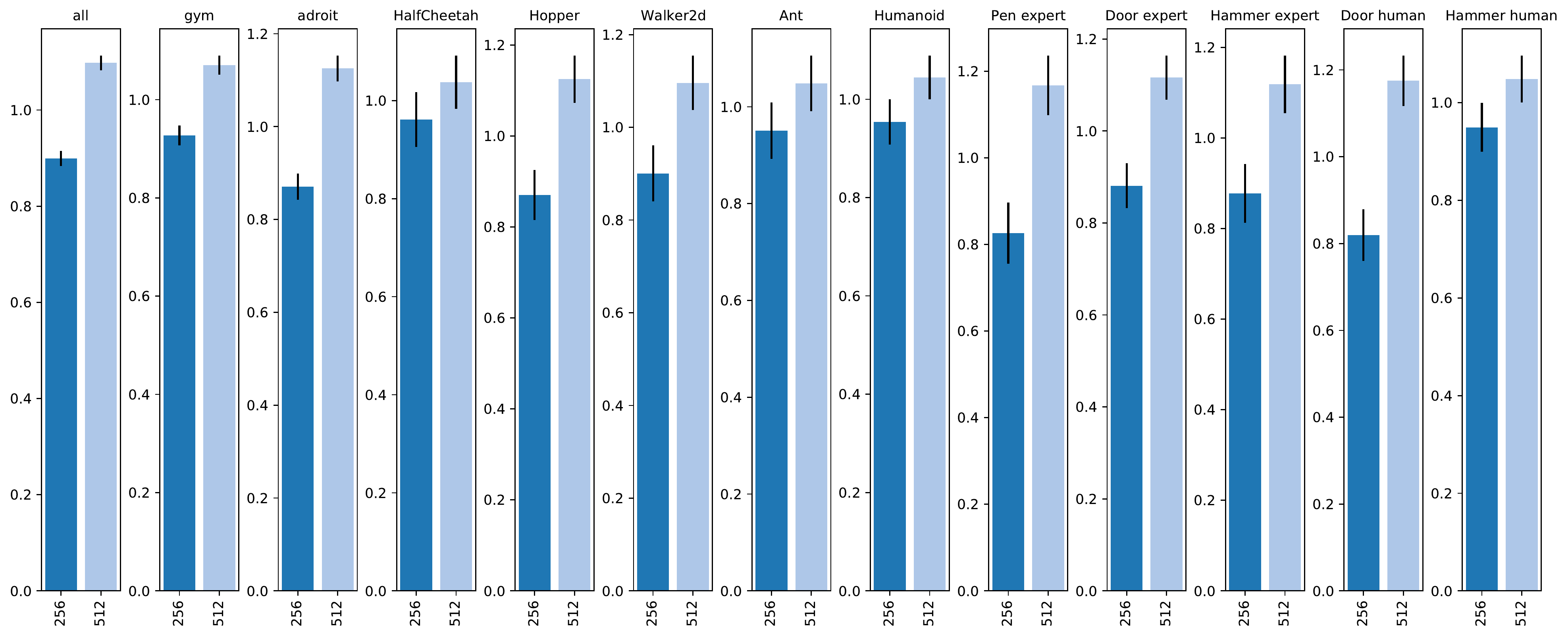}}
\caption{Analysis of choice \choicet{criticlayersize}: 95th percentile of performance scores conditioned on choice (top) and distribution of choices in top 5\% of configurations (bottom).}
\label{fig:main_critic_layer_size}
\end{center}
\end{figure}

\begin{figure}[ht]
\begin{center}
\centerline{\includegraphics[height=4.5cm,width=1\textwidth]{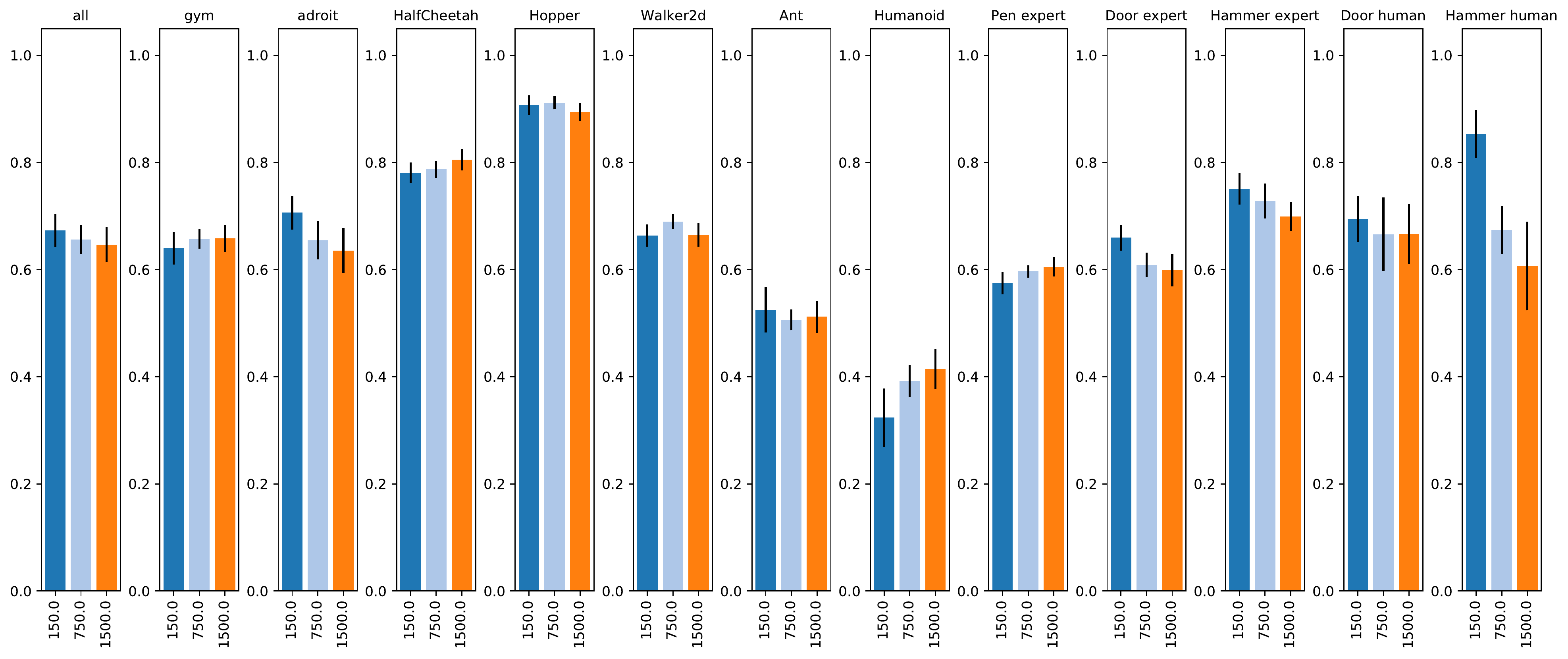}}
\centerline{\includegraphics[height=4.5cm,width=1\textwidth]{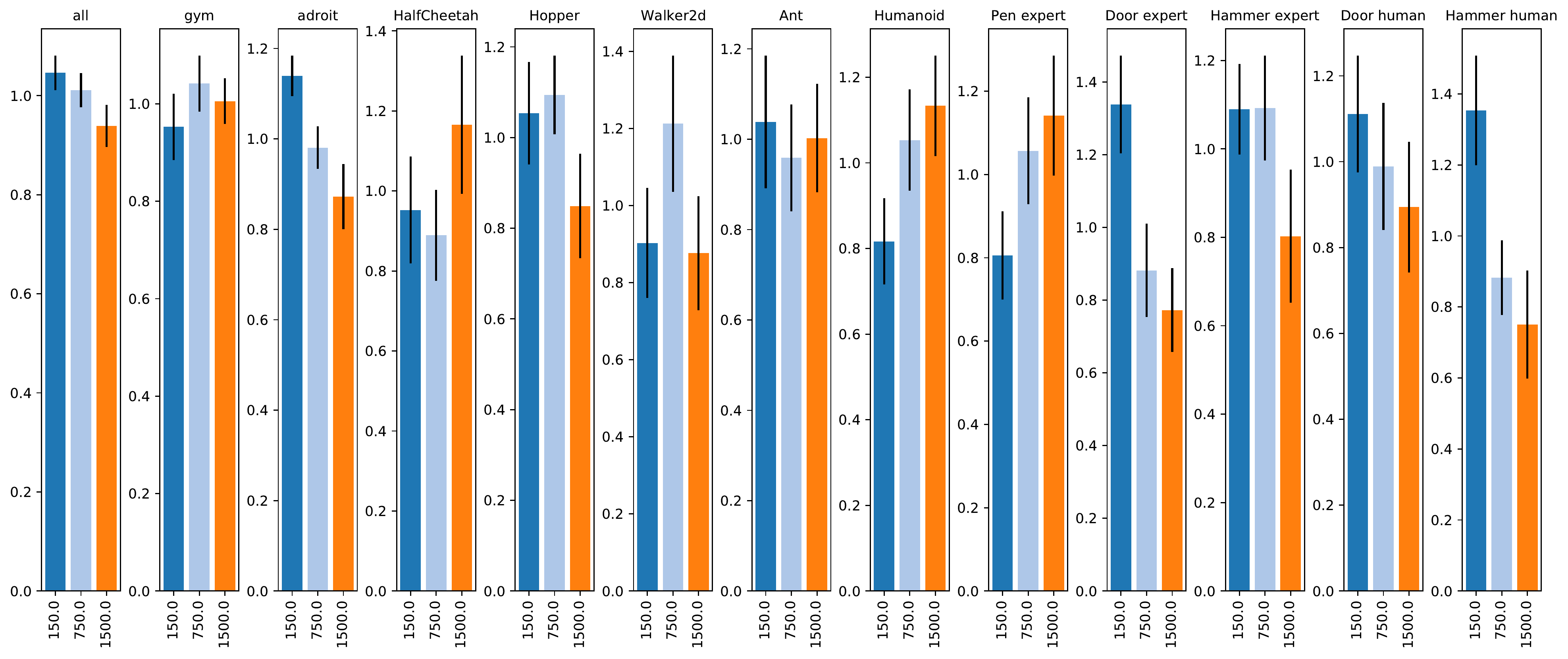}}
\caption{Analysis of choice \choicet{vmax}: 95th percentile of performance scores conditioned on choice (top) and distribution of choices in top 5\% of configurations (bottom).}
\label{fig:main_vmax}
\end{center}
\end{figure}

\begin{figure}[ht]
\begin{center}
\centerline{\includegraphics[height=4.5cm,width=1\textwidth]{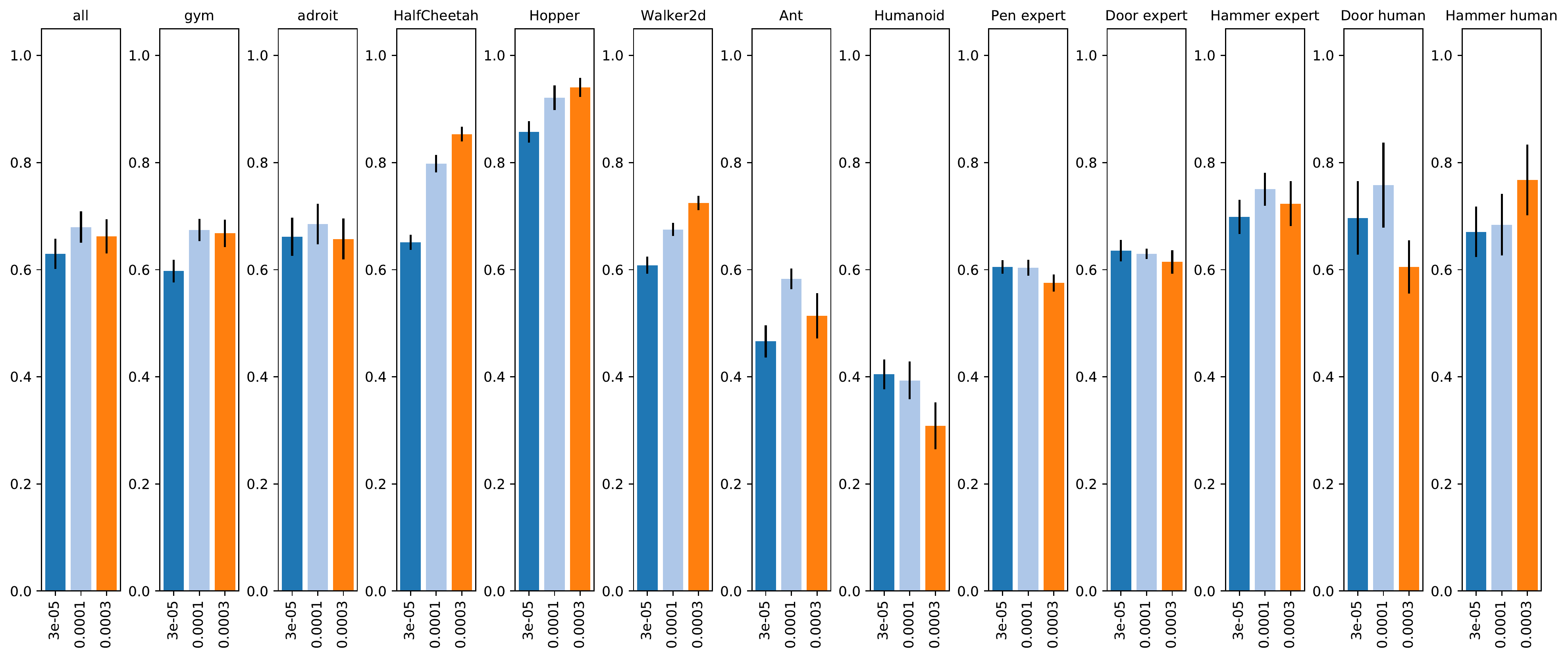}}
\centerline{\includegraphics[height=4.5cm,width=1\textwidth]{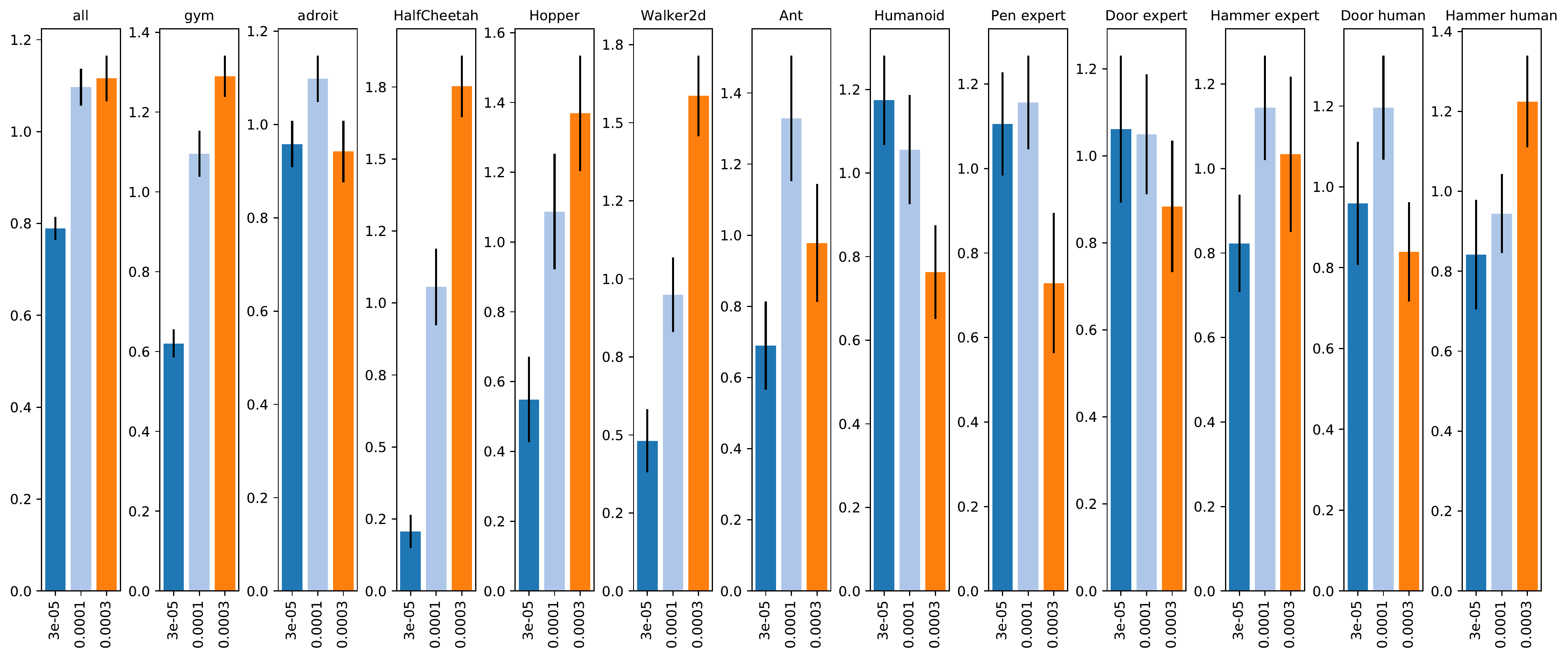}}
\caption{Analysis of choice \choicet{dfpglearningrate}: 95th percentile of performance scores conditioned on choice (top) and distribution of choices in top 5\% of configurations (bottom).}
\label{fig:main_d4pg_learning_rate}
\end{center}
\end{figure}

\begin{figure}[ht]
\begin{center}
\centerline{\includegraphics[height=4.5cm,width=1\textwidth]{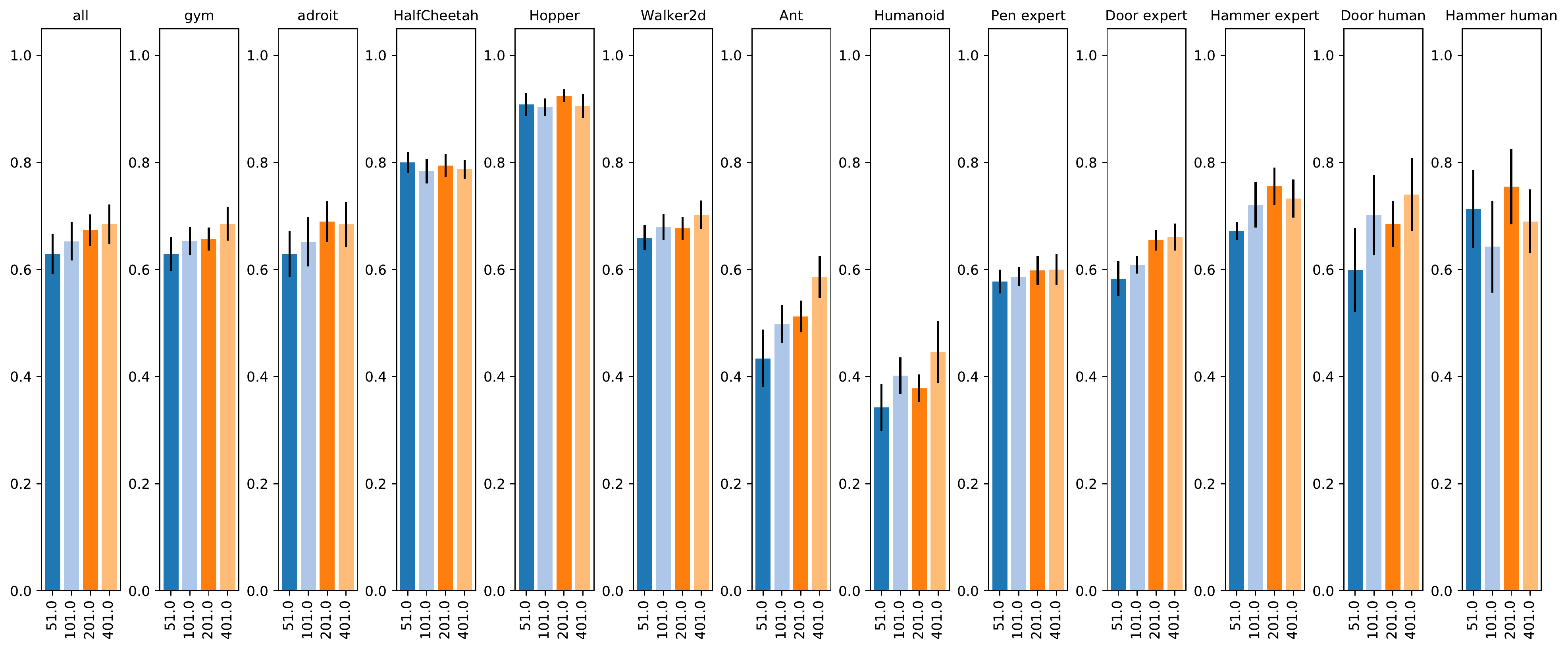}}
\centerline{\includegraphics[height=4.5cm,width=1\textwidth]{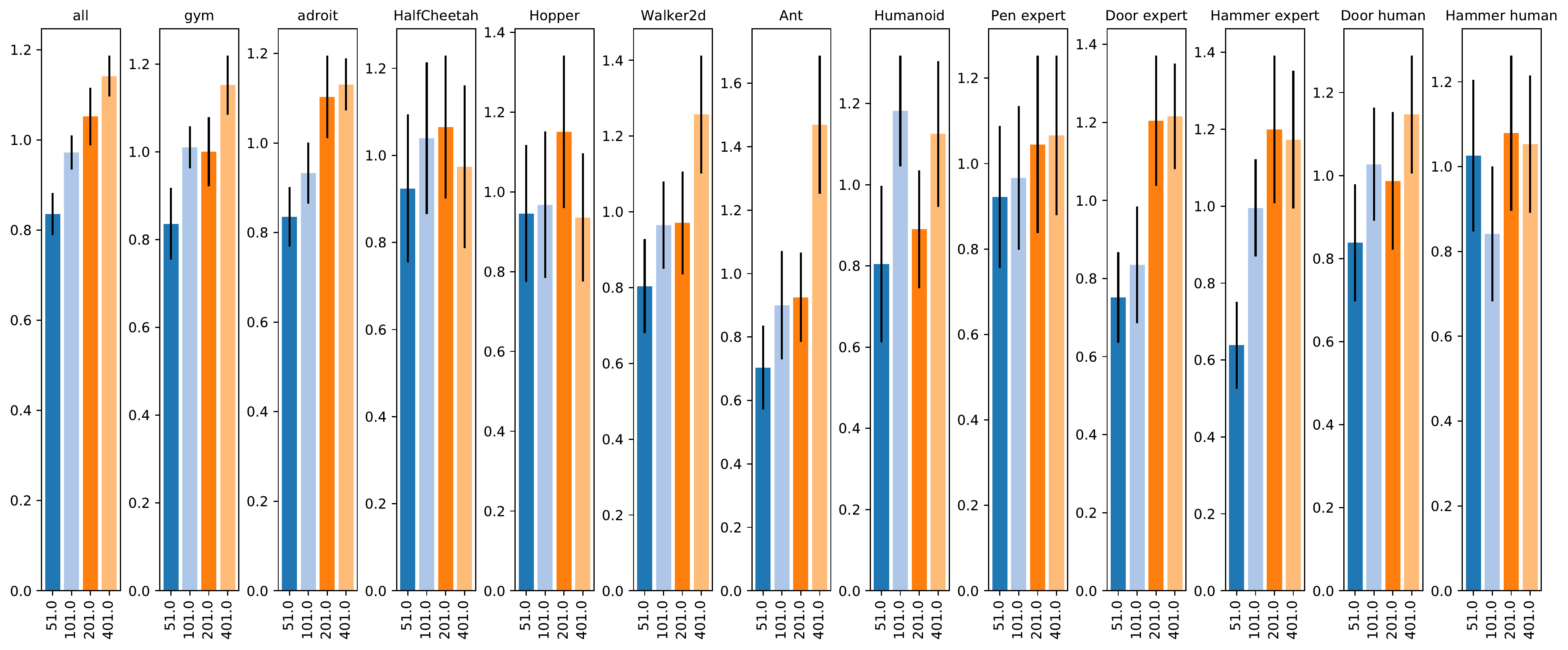}}
\caption{Analysis of choice \choicet{numatoms}: 95th percentile of performance scores conditioned on choice (top) and distribution of choices in top 5\% of configurations (bottom).}
\label{fig:main_num_atoms}
\end{center}
\end{figure}

\begin{figure}[ht]
\begin{center}
\centerline{\includegraphics[height=4.5cm,width=1\textwidth]{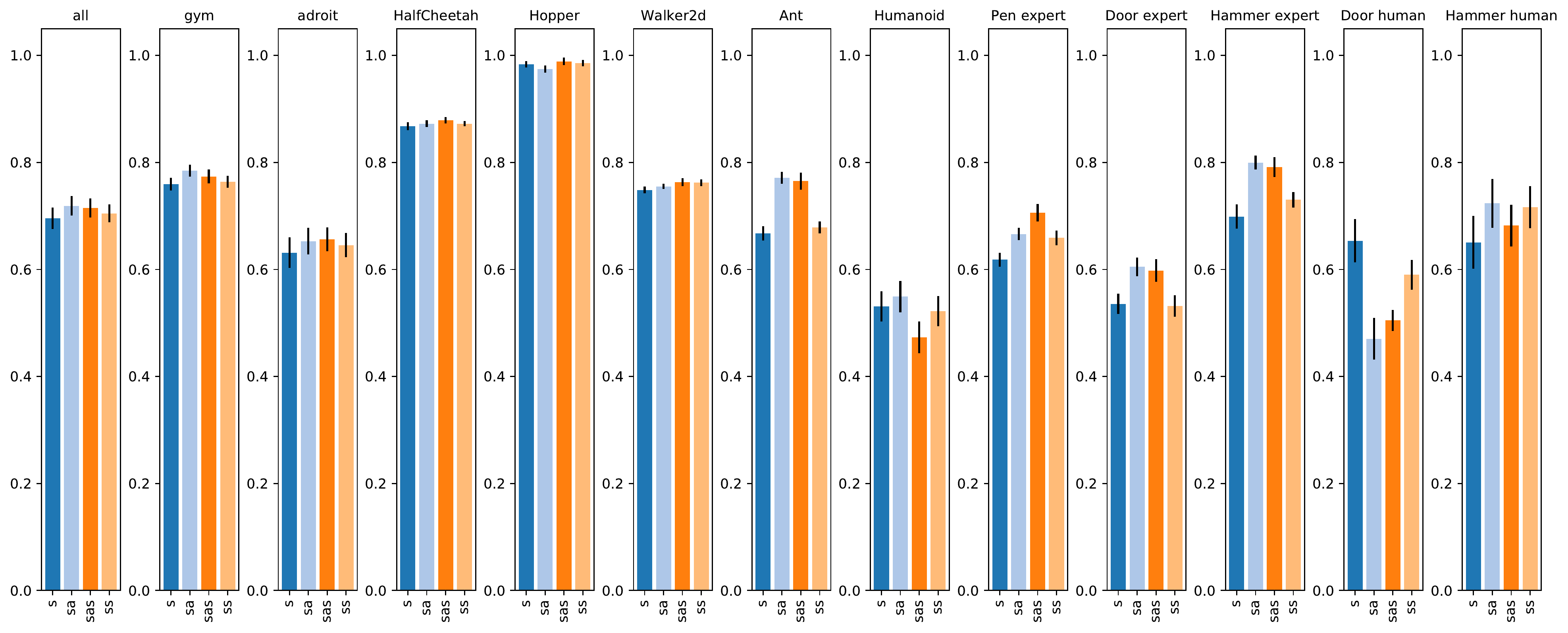}}
\centerline{\includegraphics[height=4.5cm,width=1\textwidth]{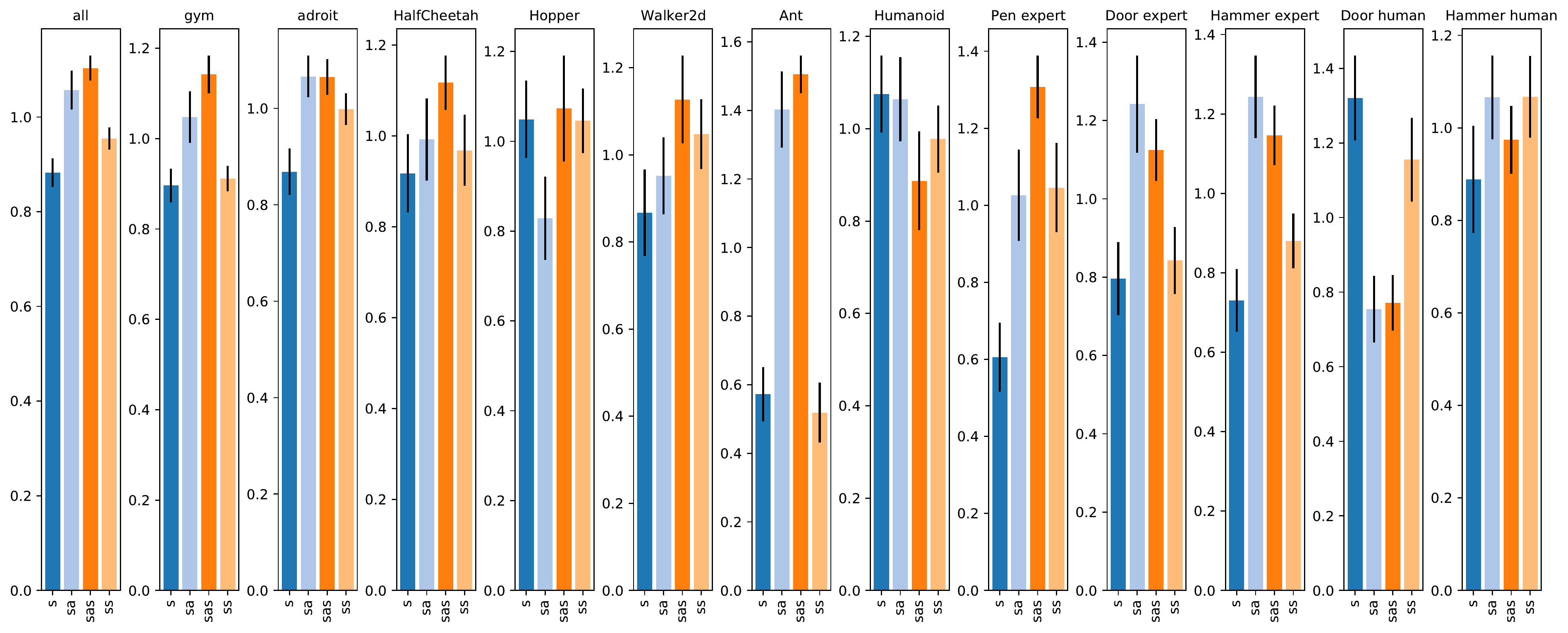}}
\caption{Analysis of choice \choicet{gailinput}: 95th percentile of performance scores conditioned on choice (top) and distribution of choices in top 5\% of configurations (bottom).}
\label{fig:main__gin_discriminator_input__macro_value}
\end{center}
\end{figure}

\begin{figure}[ht]
\begin{center}
\centerline{\includegraphics[height=4.5cm,width=1\textwidth]{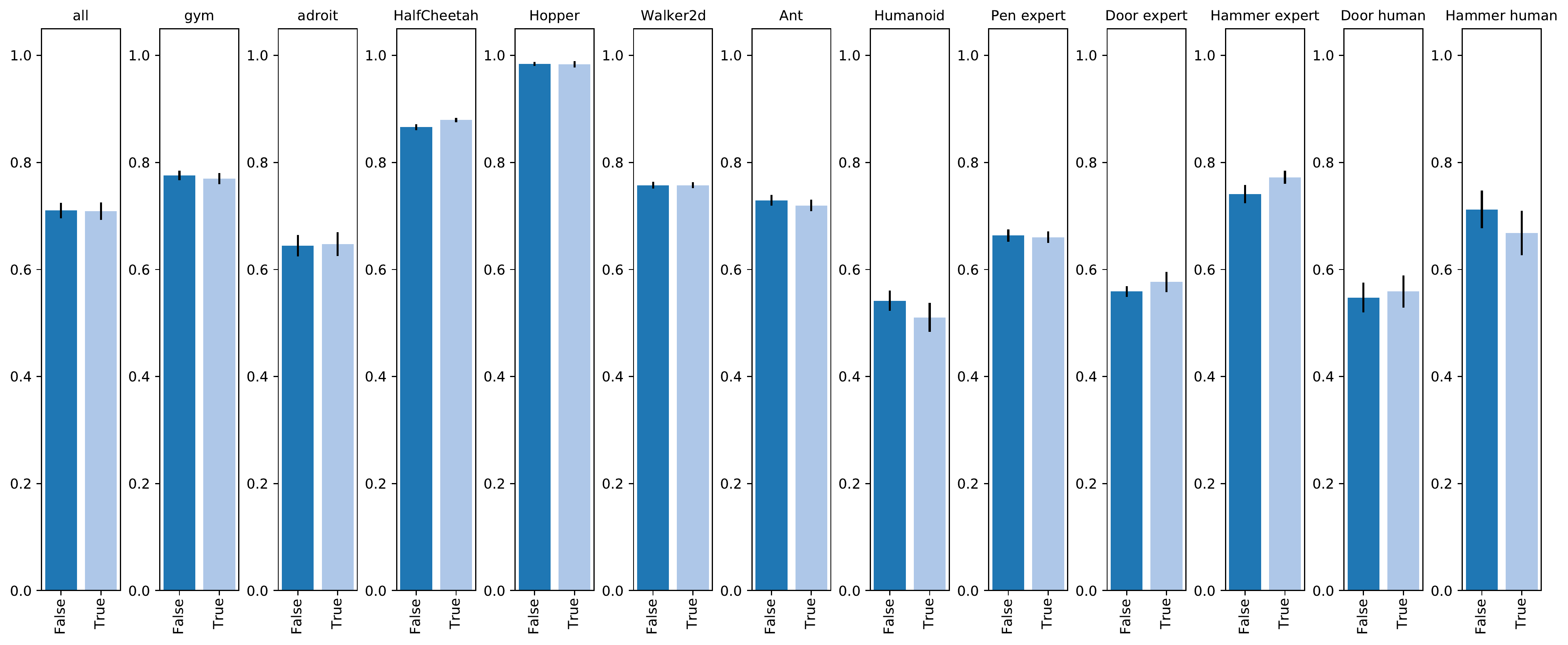}}
\centerline{\includegraphics[height=4.5cm,width=1\textwidth]{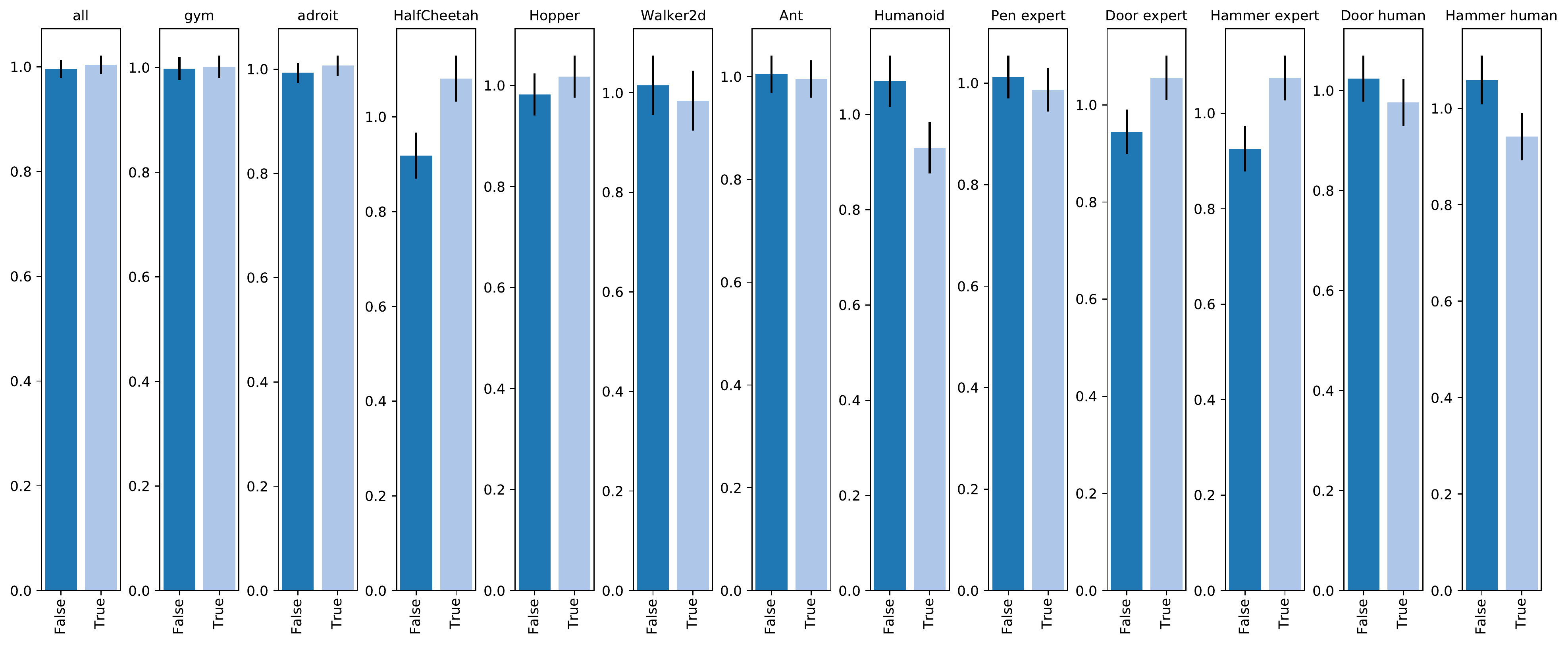}}
\caption{Analysis of choice \choicet{gaildiscriminatormodule}: 95th percentile of performance scores conditioned on choice (top) and distribution of choices in top 5\% of configurations (bottom).}
\label{fig:main__gin_make_discriminator_discriminator_module}
\end{center}
\end{figure}

\begin{figure}[ht]
\begin{center}
\centerline{\includegraphics[height=4.5cm,width=1\textwidth]{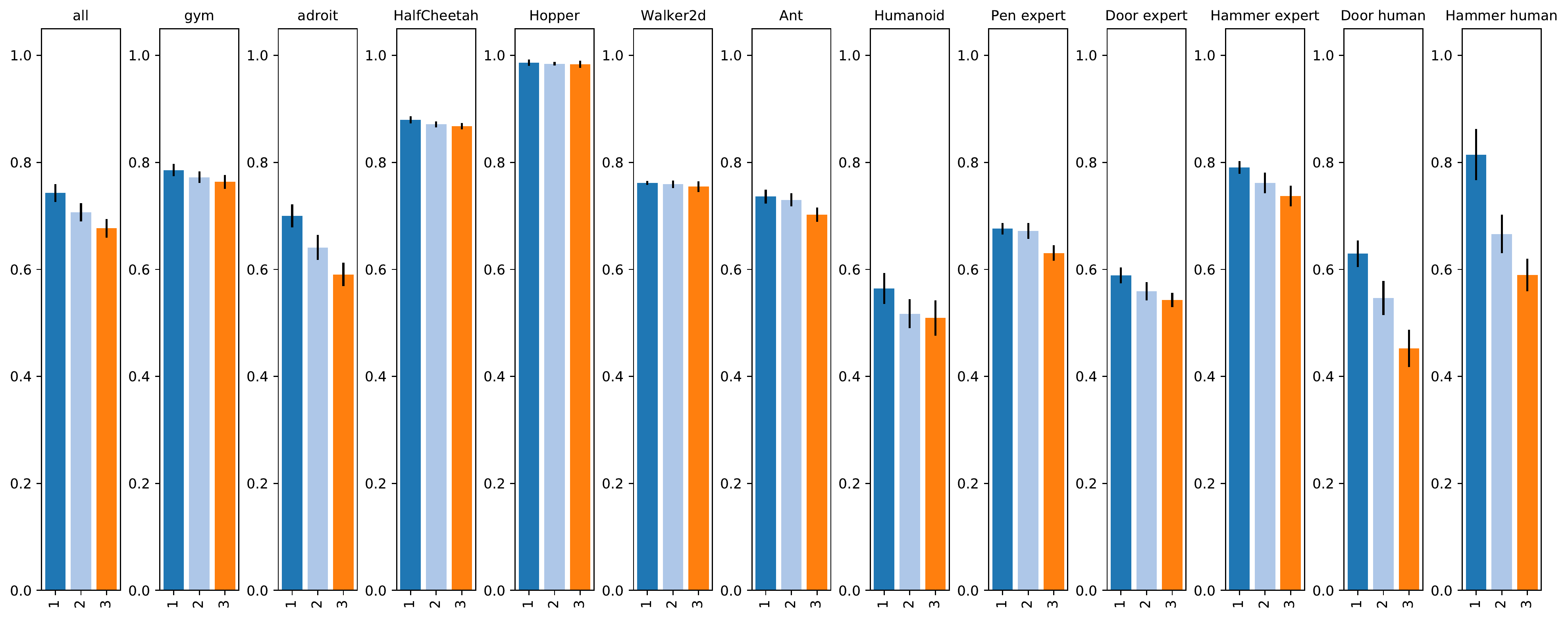}}
\centerline{\includegraphics[height=4.5cm,width=1\textwidth]{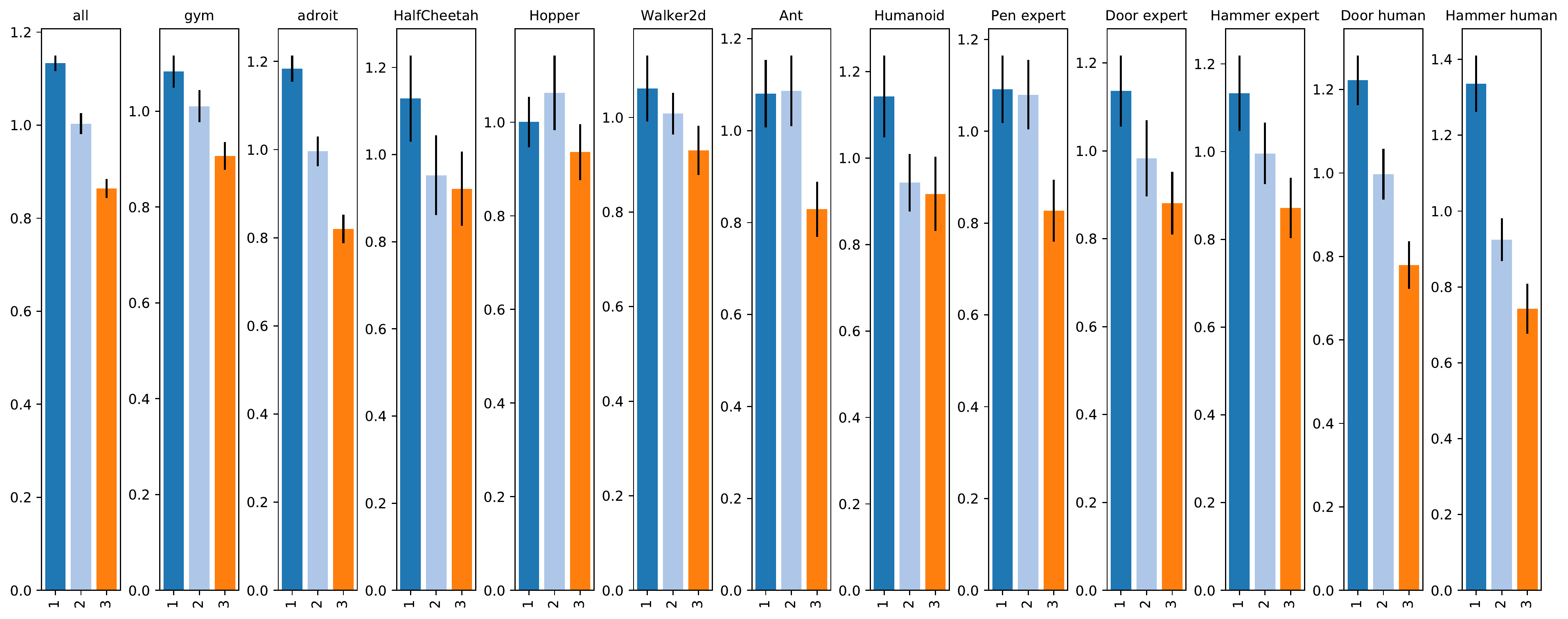}}
\caption{Analysis of choice \choicet{gailmlpnumlayers}: 95th percentile of performance scores conditioned on choice (top) and distribution of choices in top 5\% of configurations (bottom).}
\label{fig:main__gin_discriminator__MLP_num_layers}
\end{center}
\end{figure}

\begin{figure}[ht]
\begin{center}
\centerline{\includegraphics[height=4.5cm,width=1\textwidth]{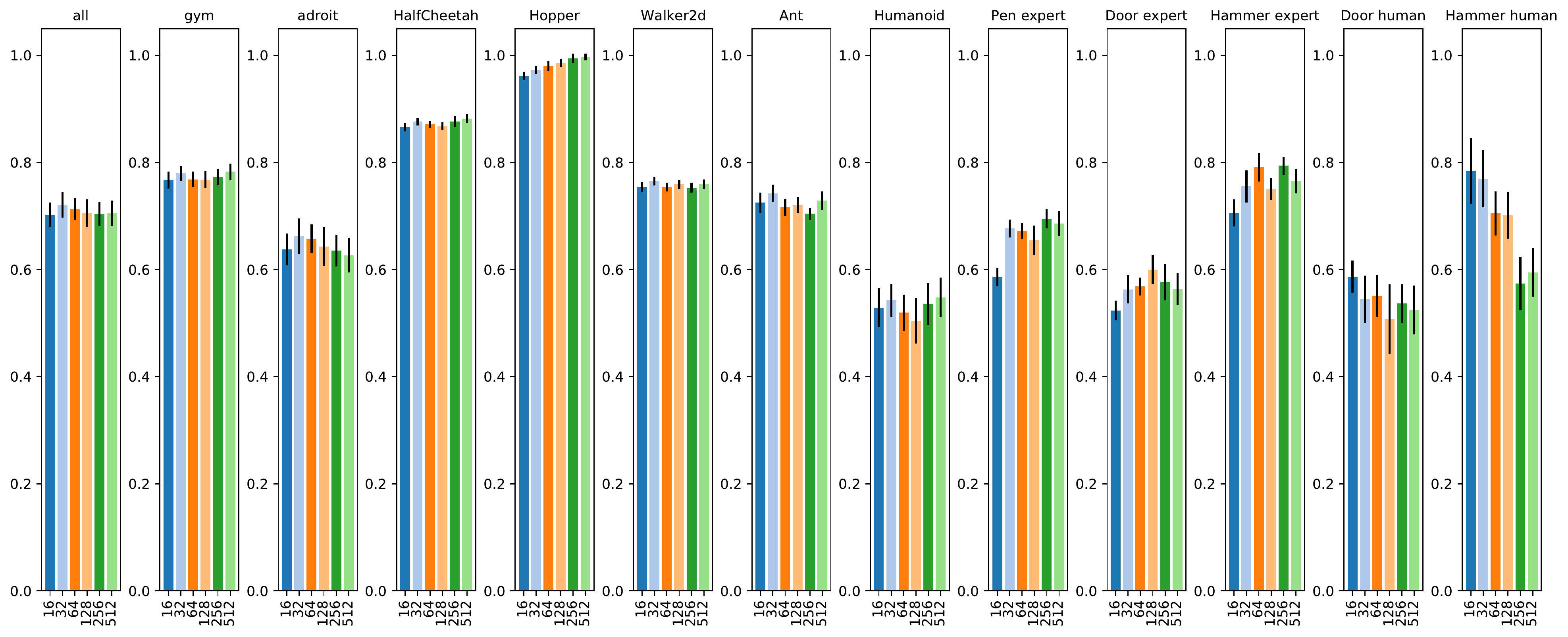}}
\centerline{\includegraphics[height=4.5cm,width=1\textwidth]{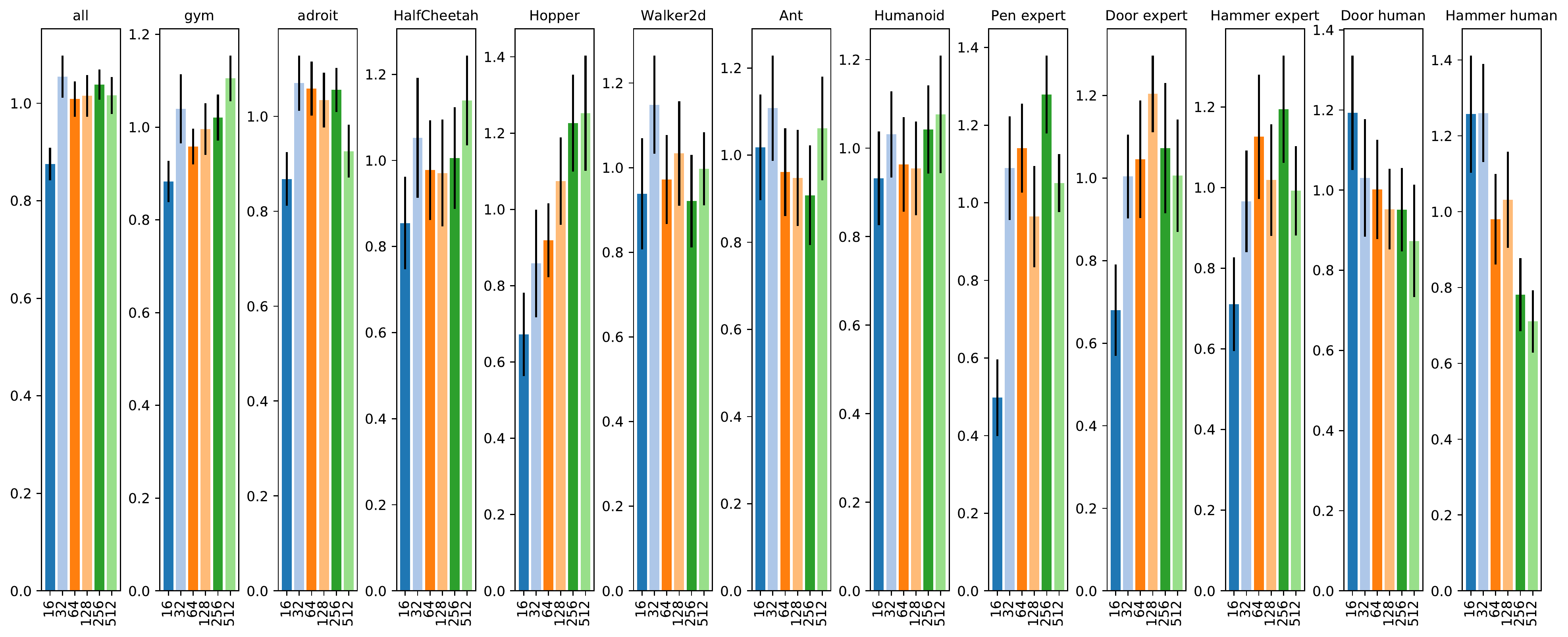}}
\caption{Analysis of choice \choicet{gailmlpnumwidth}: 95th percentile of performance scores conditioned on choice (top) and distribution of choices in top 5\% of configurations (bottom).}
\label{fig:main__gin_discriminator__MLP_num_units}
\end{center}
\end{figure}

\begin{figure}[ht]
\begin{center}
\centerline{\includegraphics[height=4.5cm,width=1\textwidth]{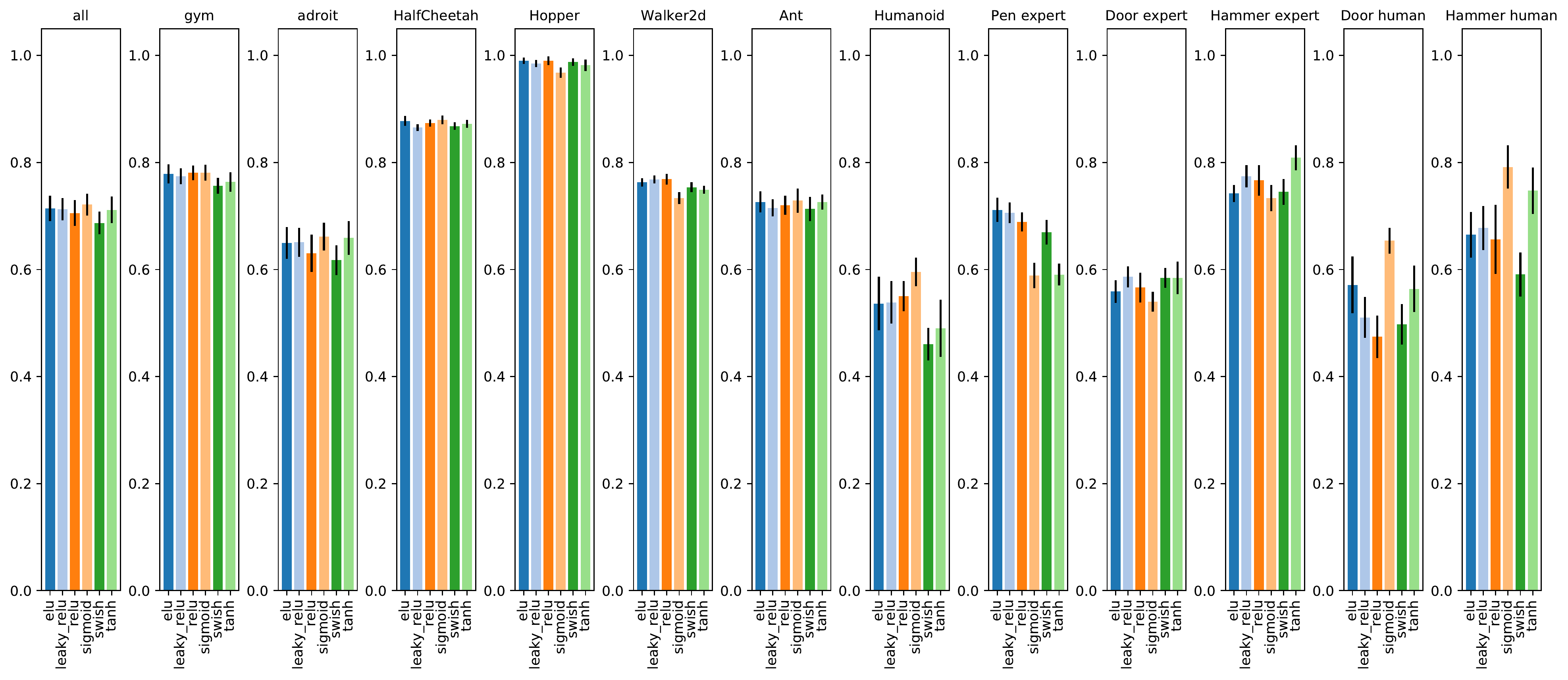}}
\centerline{\includegraphics[height=4.5cm,width=1\textwidth]{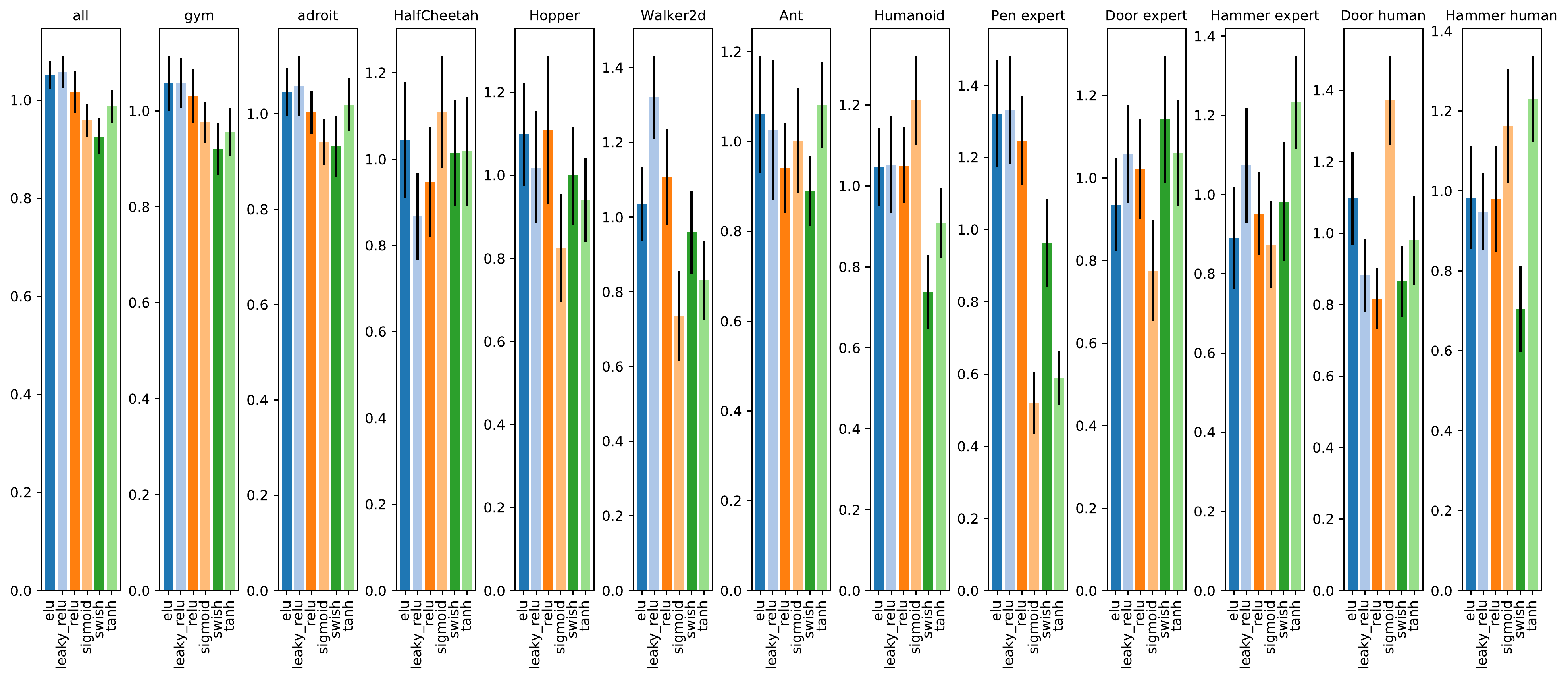}}
\caption{Analysis of choice \choicet{gailmlpactivation}: 95th percentile of performance scores conditioned on choice (top) and distribution of choices in top 5\% of configurations (bottom).}
\label{fig:main__gin_discriminator__MLP_activation}
\end{center}
\end{figure}

\begin{figure}[ht]
\begin{center}
\centerline{\includegraphics[height=4.5cm,width=1\textwidth]{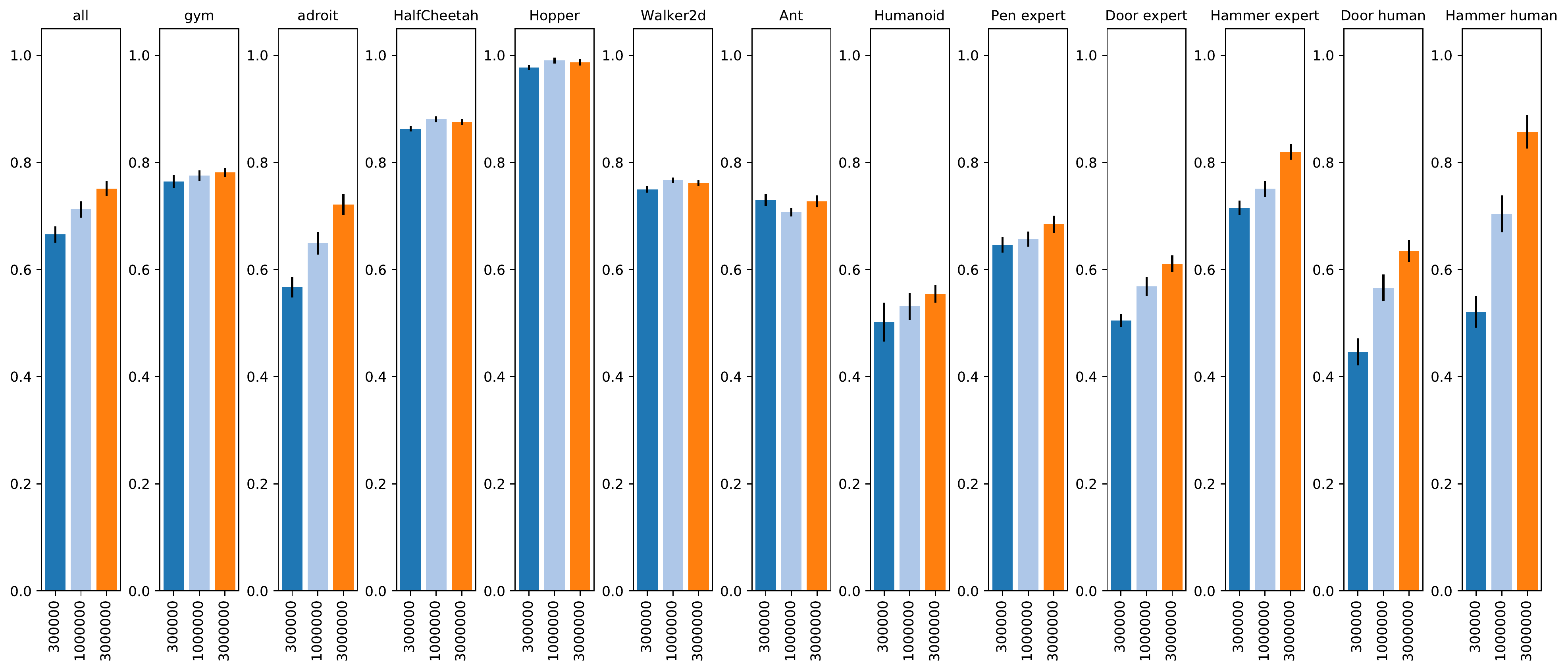}}
\centerline{\includegraphics[height=4.5cm,width=1\textwidth]{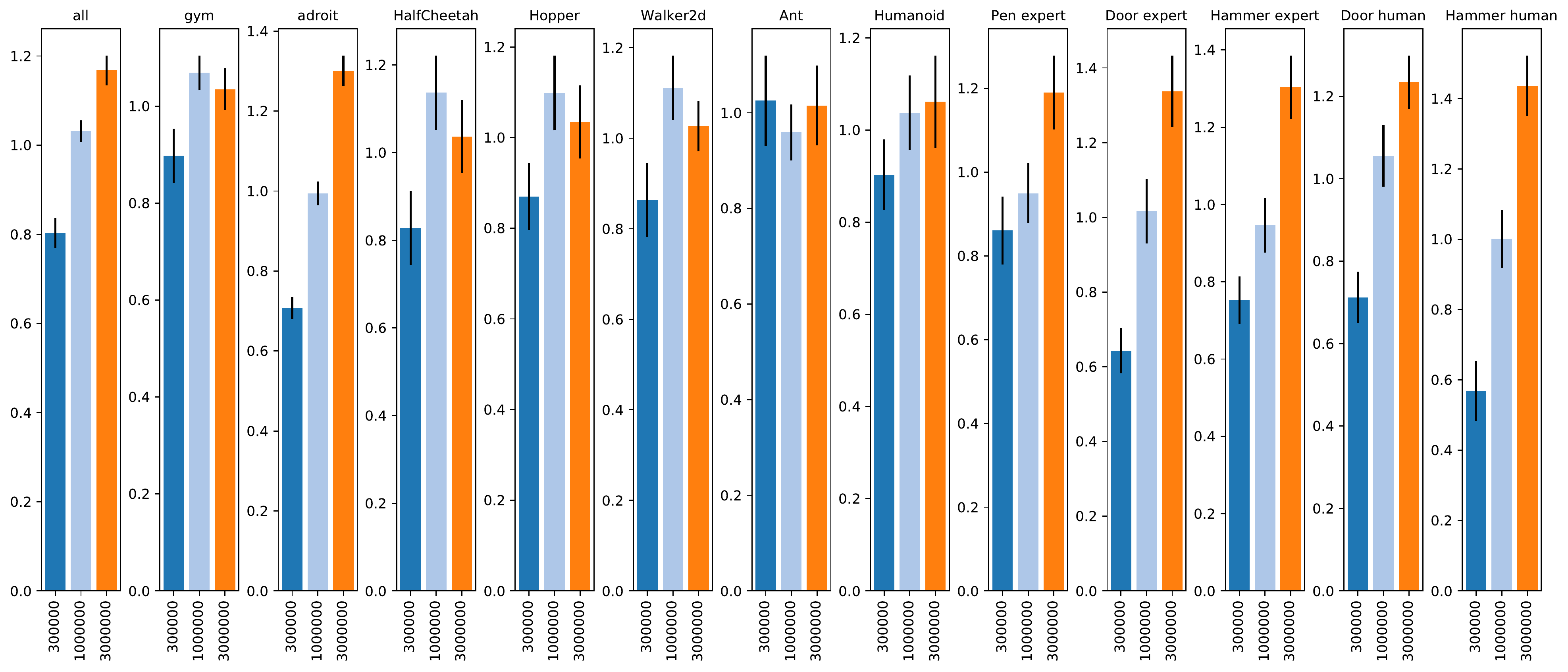}}
\caption{Analysis of choice \choicet{gailmaxreplaysize}: 95th percentile of performance scores conditioned on choice (top) and distribution of choices in top 5\% of configurations (bottom).}
\label{fig:main__gin_GAILBuilder_max_replay_size}
\end{center}
\end{figure}

\begin{figure}[ht]
\begin{center}
\centerline{\includegraphics[height=4.5cm,width=1\textwidth]{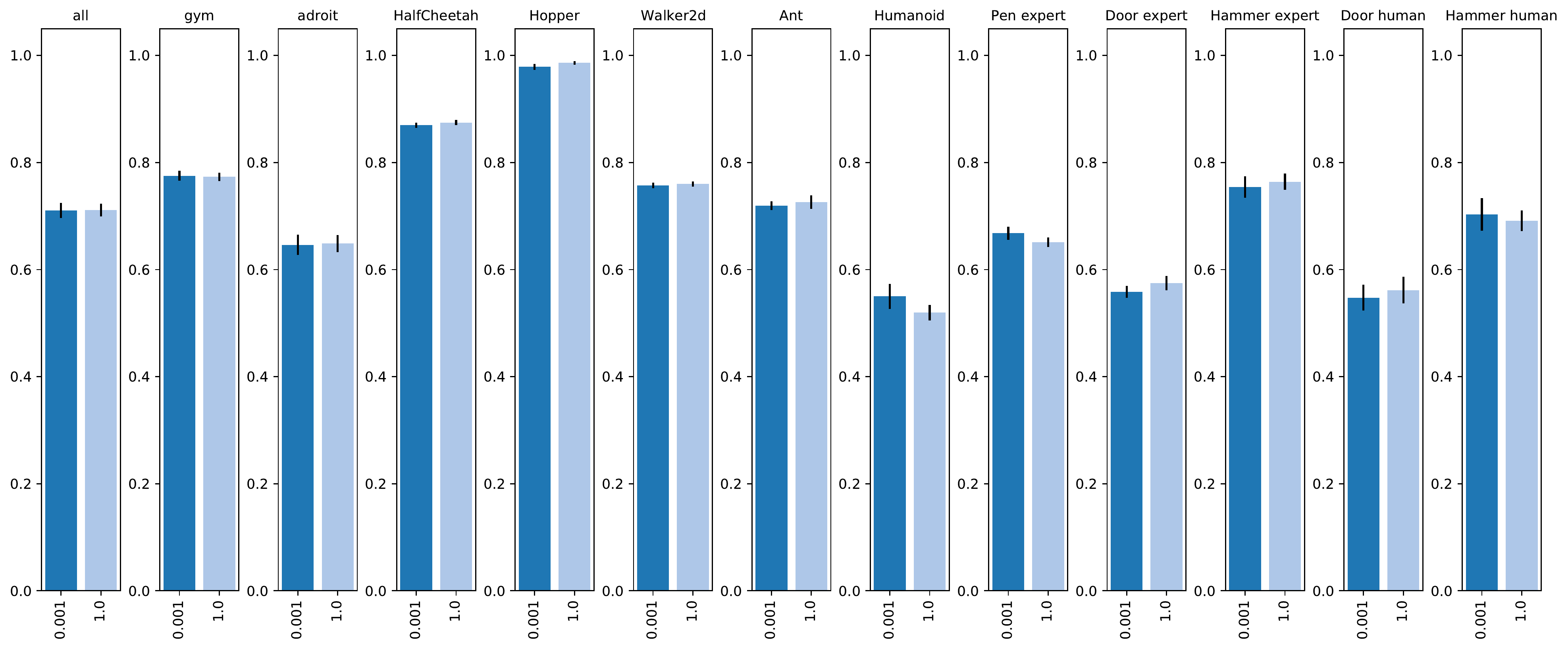}}
\centerline{\includegraphics[height=4.5cm,width=1\textwidth]{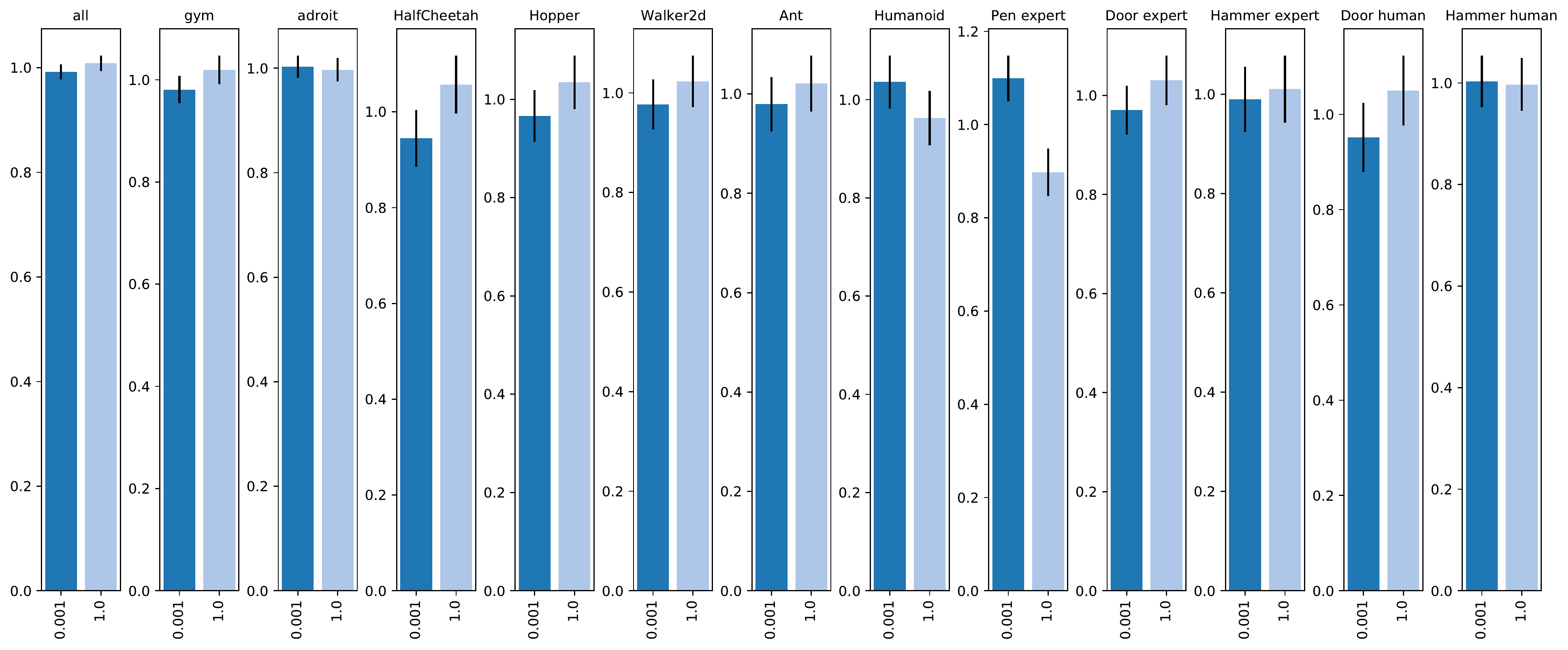}}
\caption{Analysis of choice \choicet{gailmlplastlayerinitscale}: 95th percentile of performance scores conditioned on choice (top) and distribution of choices in top 5\% of configurations (bottom).}
\label{fig:main__gin_discriminator__MLP_last_layer_kernel_init_scale}
\end{center}
\end{figure}

\begin{figure}[ht]
\begin{center}
\centerline{\includegraphics[height=4.5cm,width=1\textwidth]{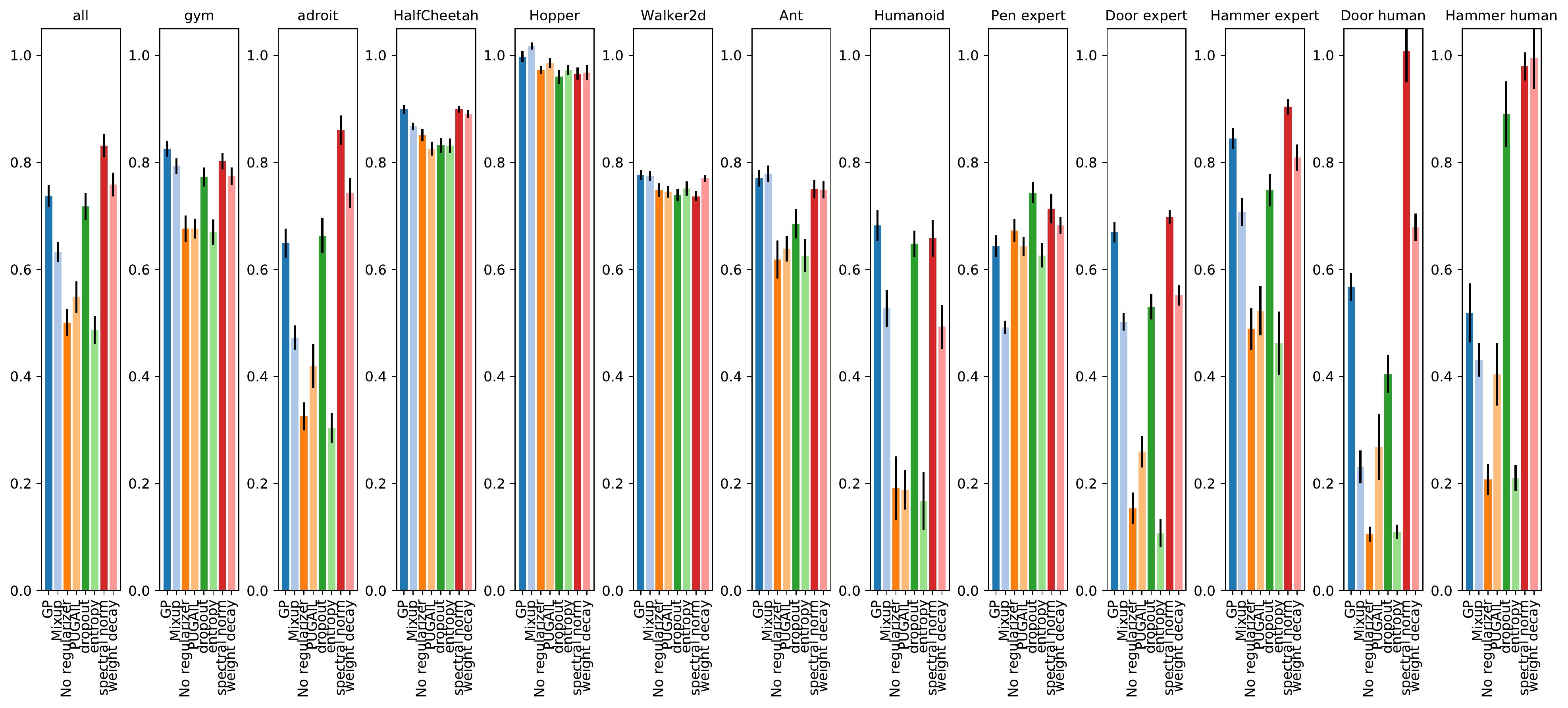}}
\centerline{\includegraphics[height=4.5cm,width=1\textwidth]{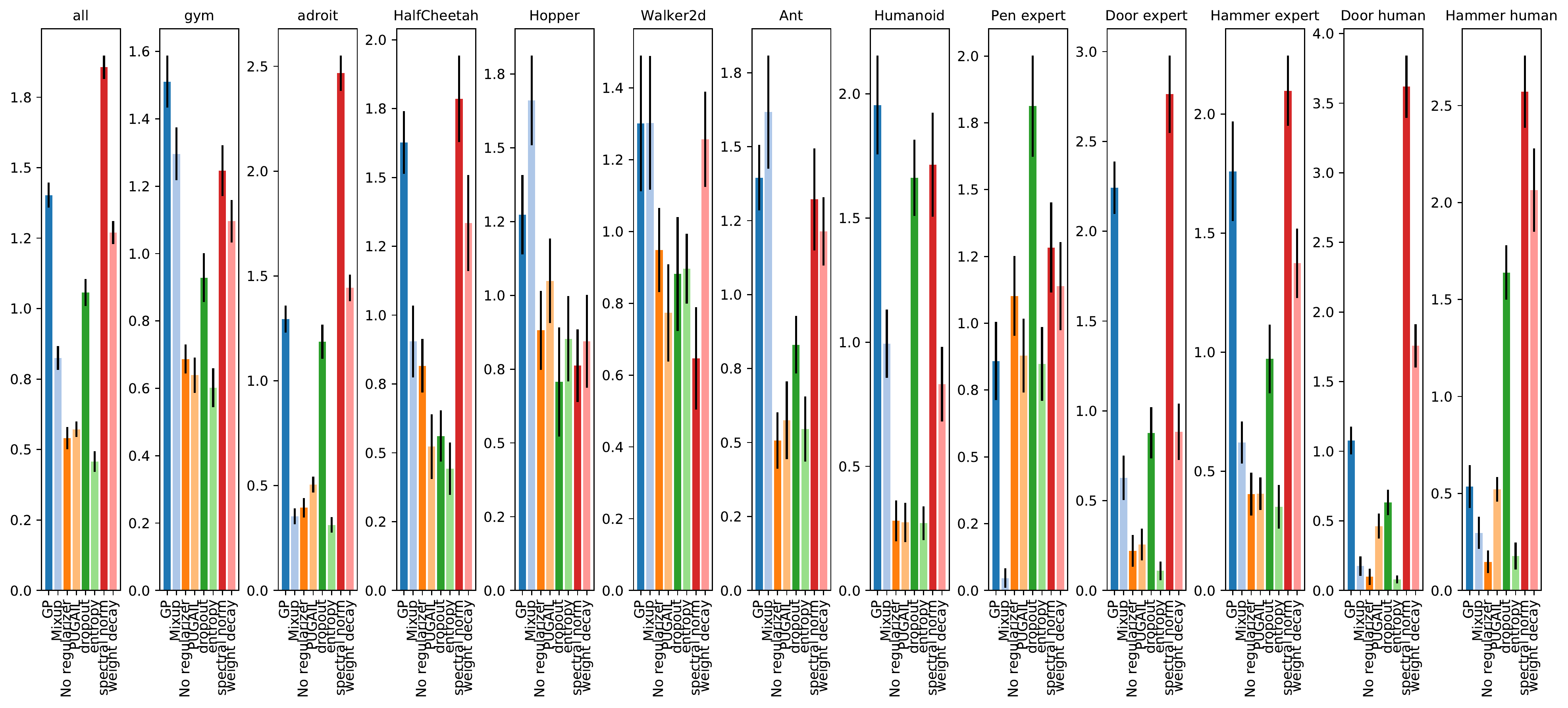}}
\caption{Analysis of choice \choicet{regularizer}: 95th percentile of performance scores conditioned on choice (top) and distribution of choices in top 5\% of configurations (bottom).}
\label{fig:main__gin_regularizer__macro_value}
\end{center}
\end{figure}

\begin{figure}[ht]
\begin{center}
\centerline{\includegraphics[height=4.5cm,width=1\textwidth]{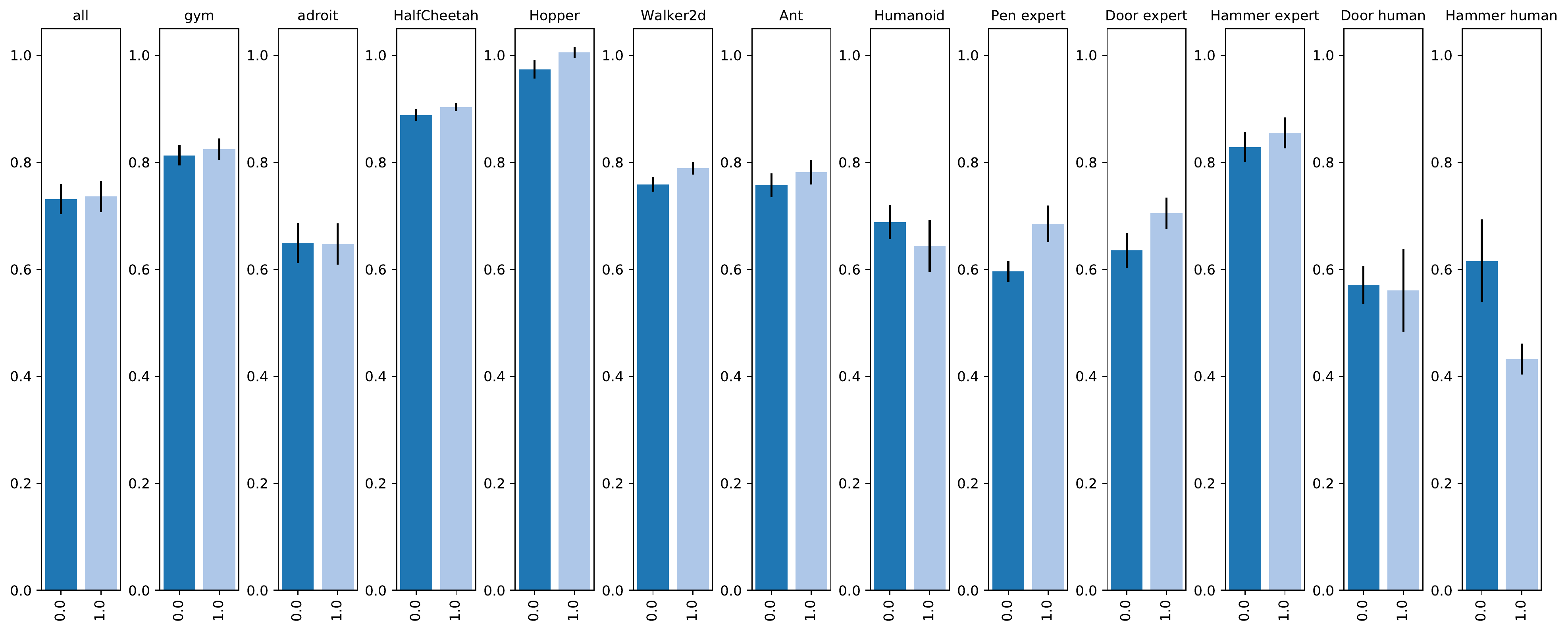}}
\centerline{\includegraphics[height=4.5cm,width=1\textwidth]{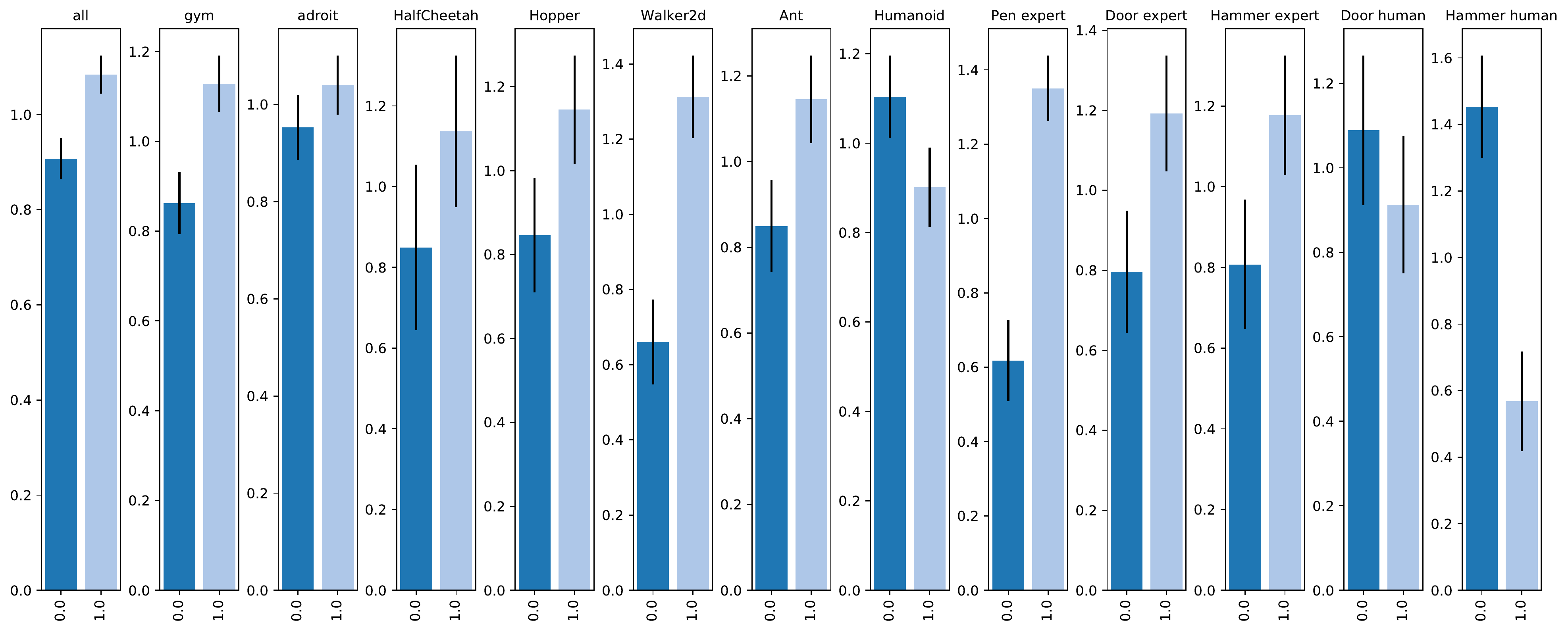}}
\caption{Analysis of choice \choicet{gptarget}: 95th percentile of performance scores conditioned on choice (top) and distribution of choices in top 5\% of configurations (bottom).}
\label{fig:main__gin_add_gradient_penalty_gradient_penalty_target}
\end{center}
\end{figure}

\begin{figure}[ht]
\begin{center}
\centerline{\includegraphics[height=4.5cm,width=1\textwidth]{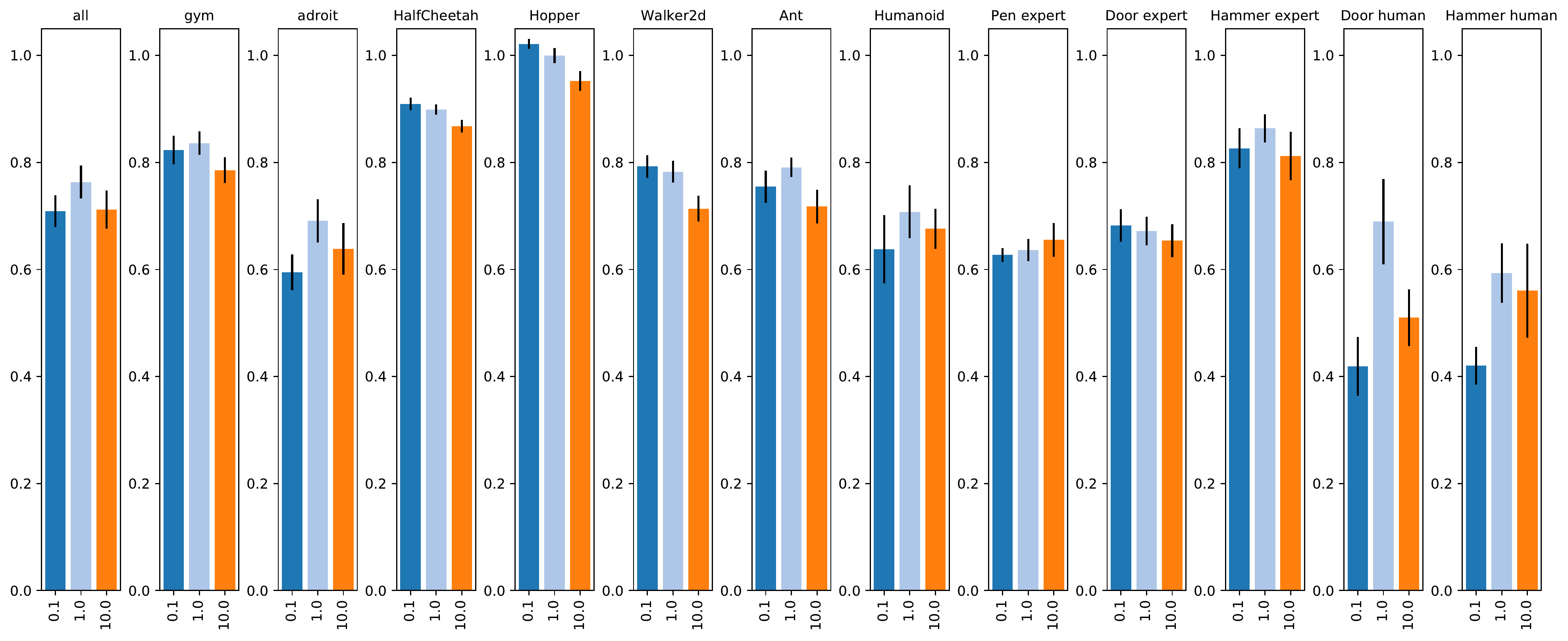}}
\centerline{\includegraphics[height=4.5cm,width=1\textwidth]{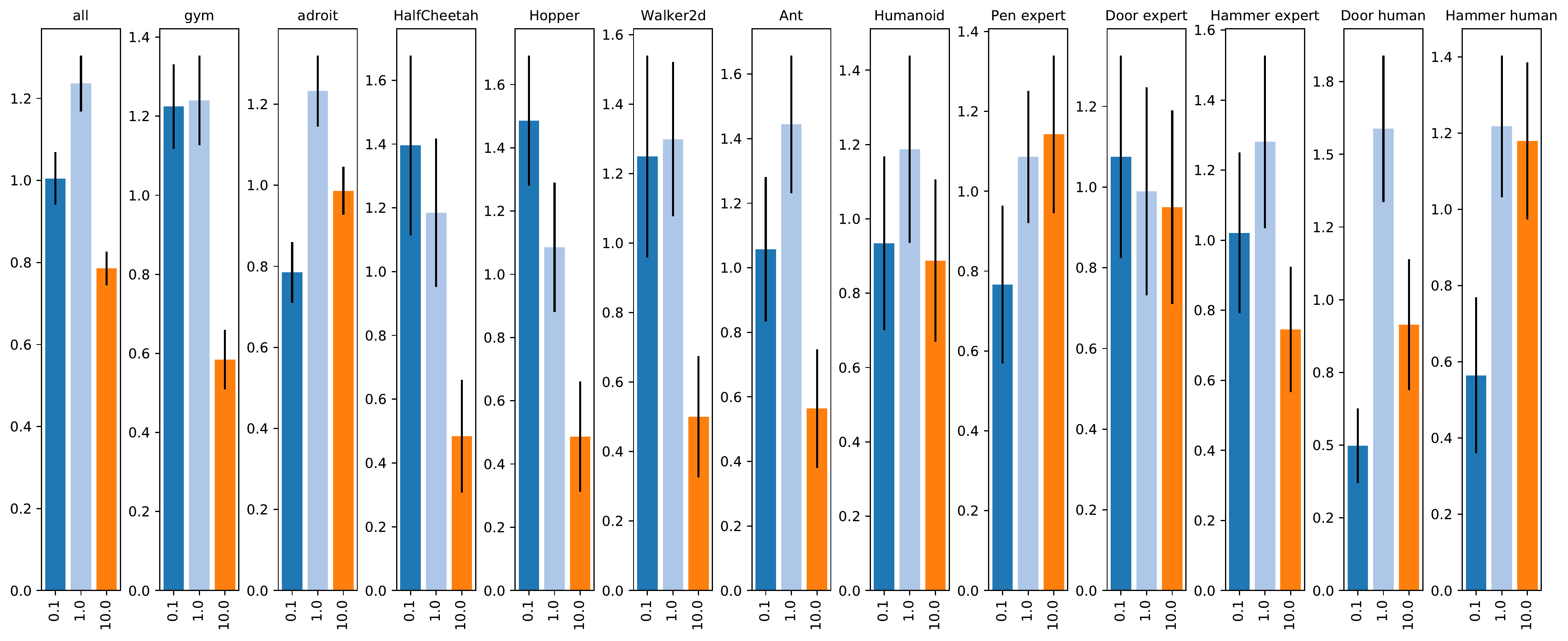}}
\caption{Analysis of choice \choicet{gpcoef}: 95th percentile of performance scores conditioned on choice (top) and distribution of choices in top 5\% of configurations (bottom).}
\label{fig:main__gin_add_gradient_penalty_gradient_penalty_coefficient}
\end{center}
\end{figure}

\begin{figure}[ht]
\begin{center}
\centerline{\includegraphics[height=4.5cm,width=1\textwidth]{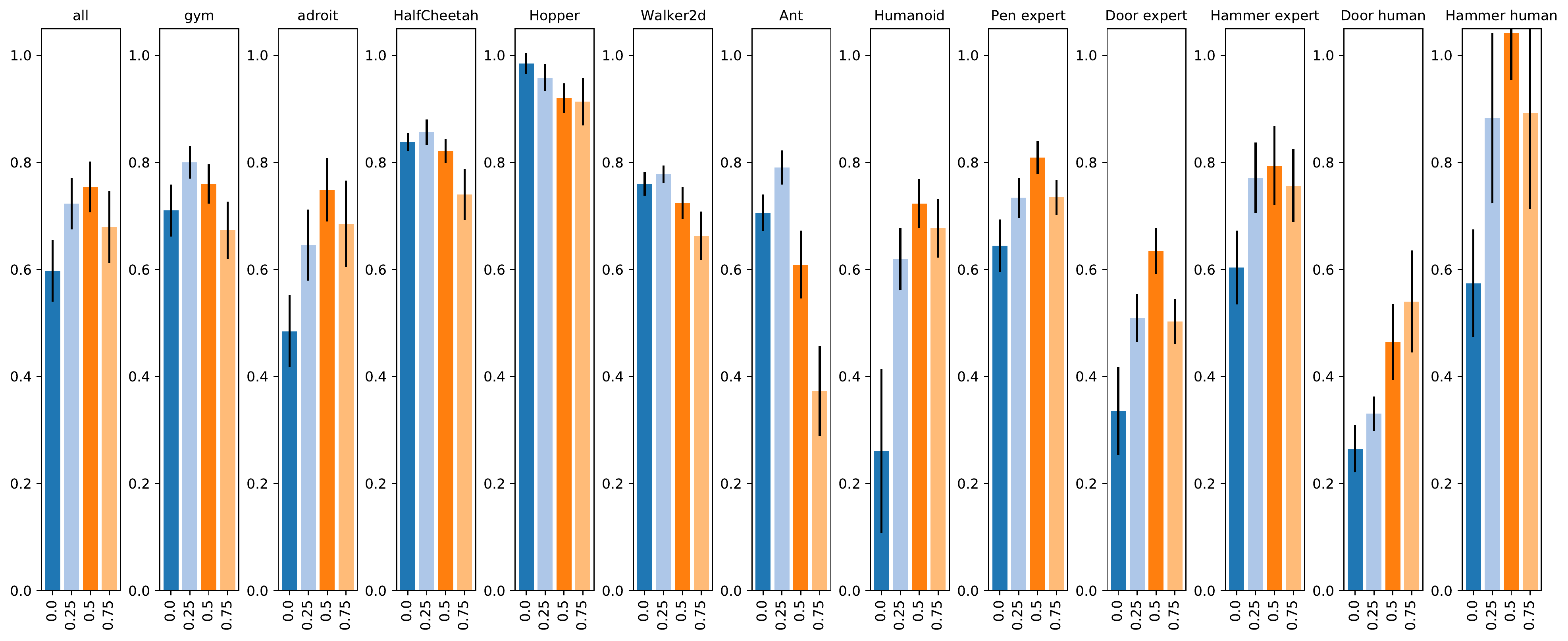}}
\centerline{\includegraphics[height=4.5cm,width=1\textwidth]{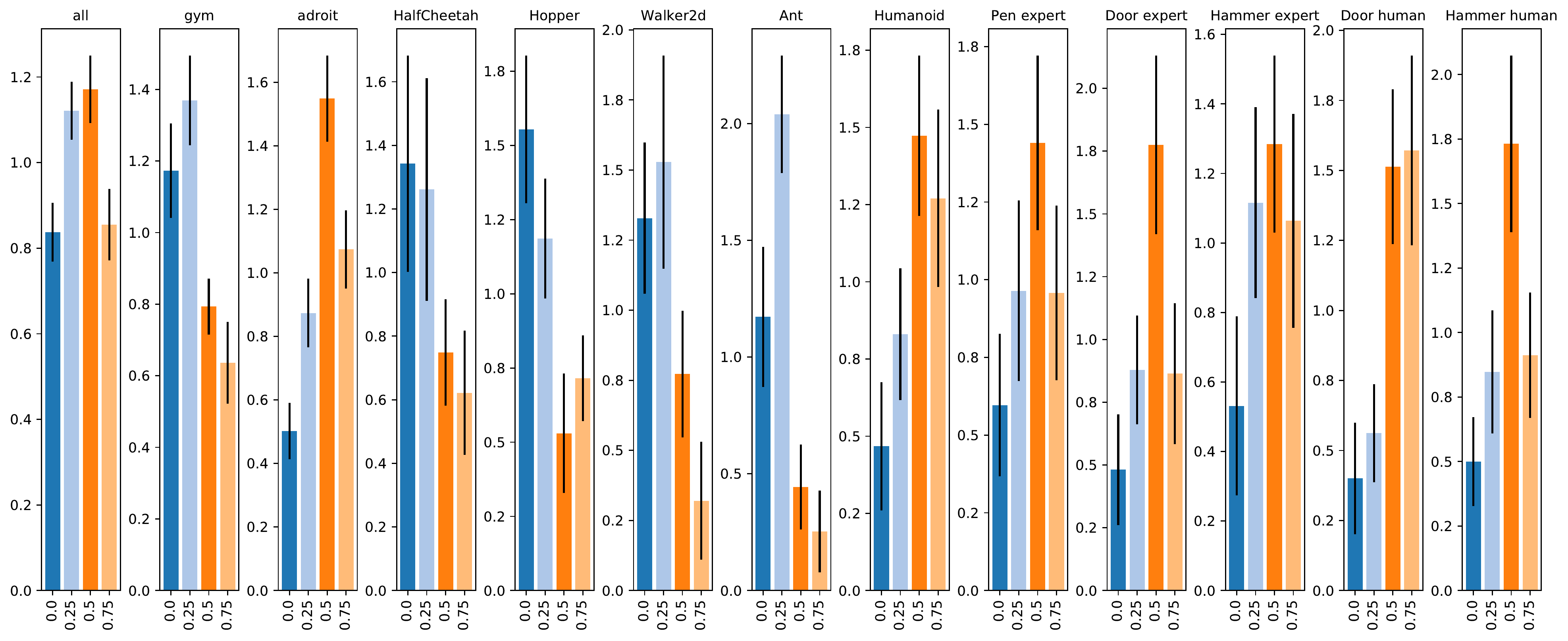}}
\caption{Analysis of choice \choicet{dropoutinputrate}: 95th percentile of performance scores conditioned on choice (top) and distribution of choices in top 5\% of configurations (bottom).}
\label{fig:main__gin_discriminator__MLP_input_dropout_rate}
\end{center}
\end{figure}

\begin{figure}[ht]
\begin{center}
\centerline{\includegraphics[height=4.5cm,width=1\textwidth]{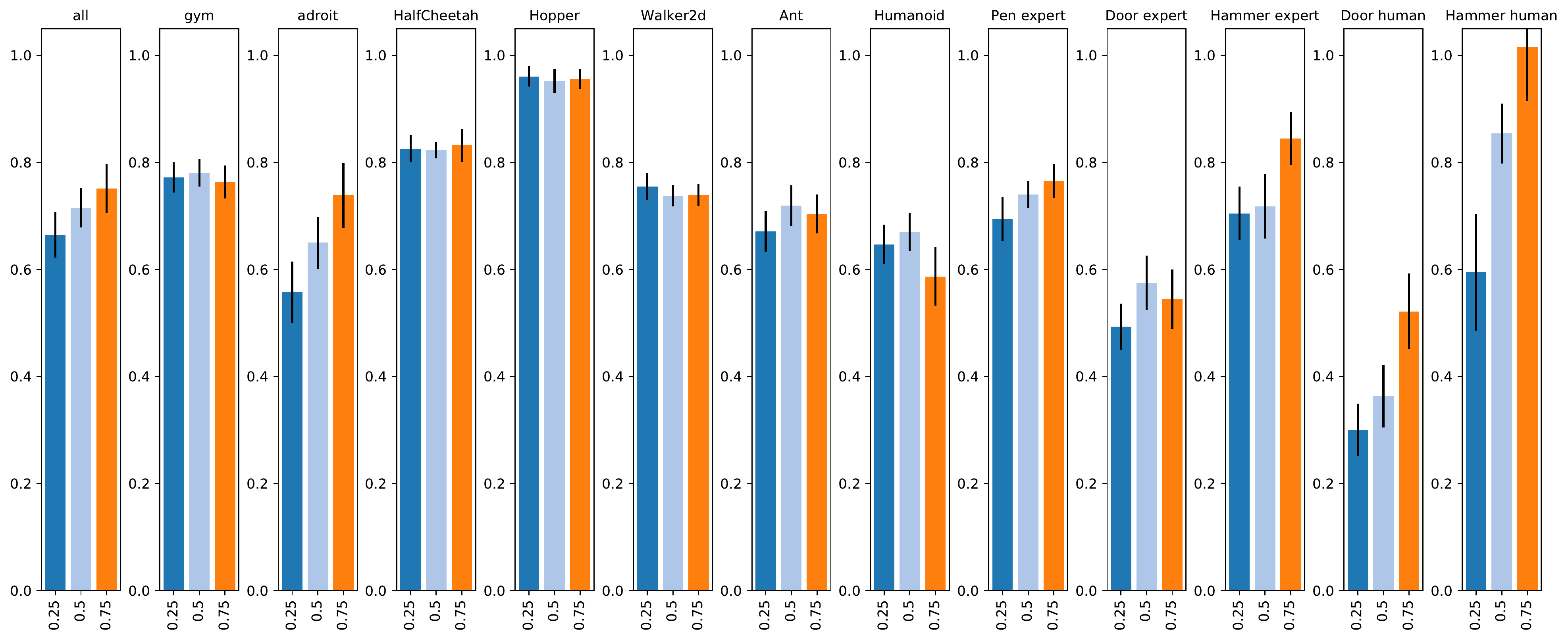}}
\centerline{\includegraphics[height=4.5cm,width=1\textwidth]{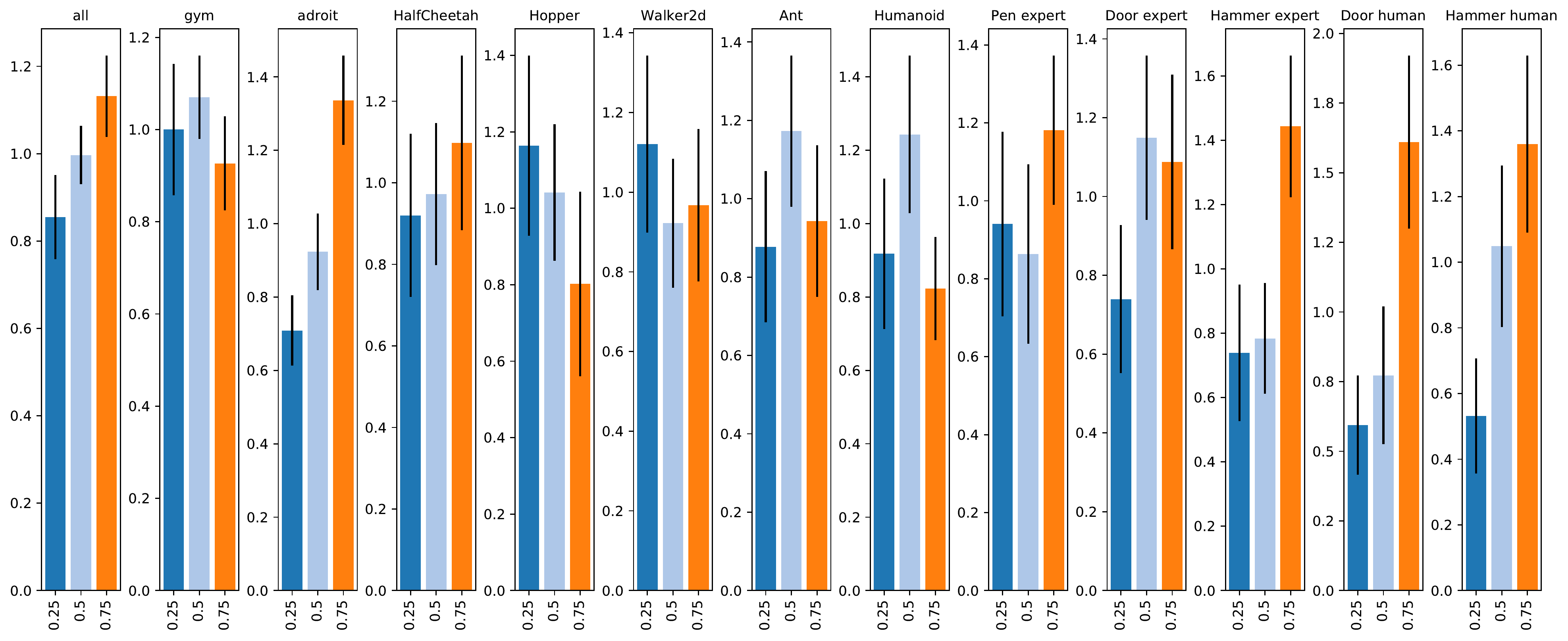}}
\caption{Analysis of choice \choicet{dropouthiddenrate}: 95th percentile of performance scores conditioned on choice (top) and distribution of choices in top 5\% of configurations (bottom).}
\label{fig:main__gin_discriminator__MLP_hidden_dropout_rate}
\end{center}
\end{figure}

\begin{figure}[ht]
\begin{center}
\centerline{\includegraphics[height=4.5cm,width=1\textwidth]{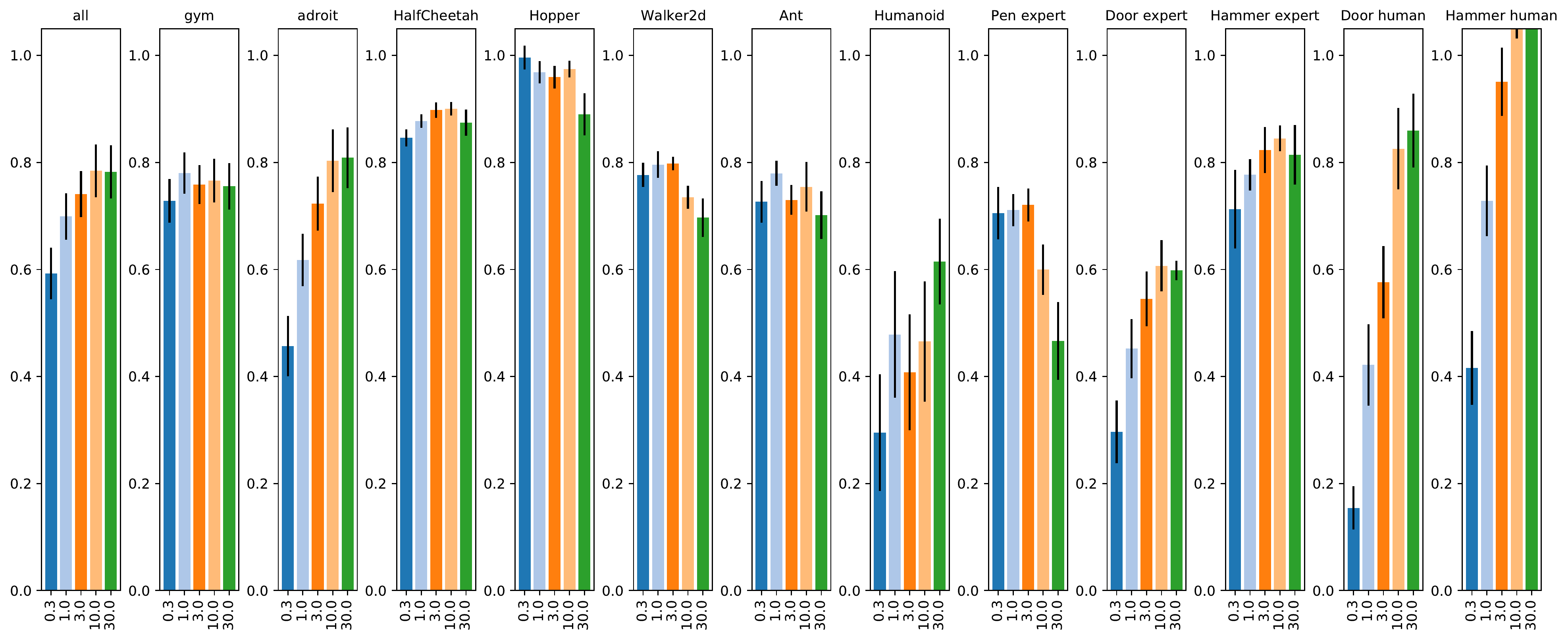}}
\centerline{\includegraphics[height=4.5cm,width=1\textwidth]{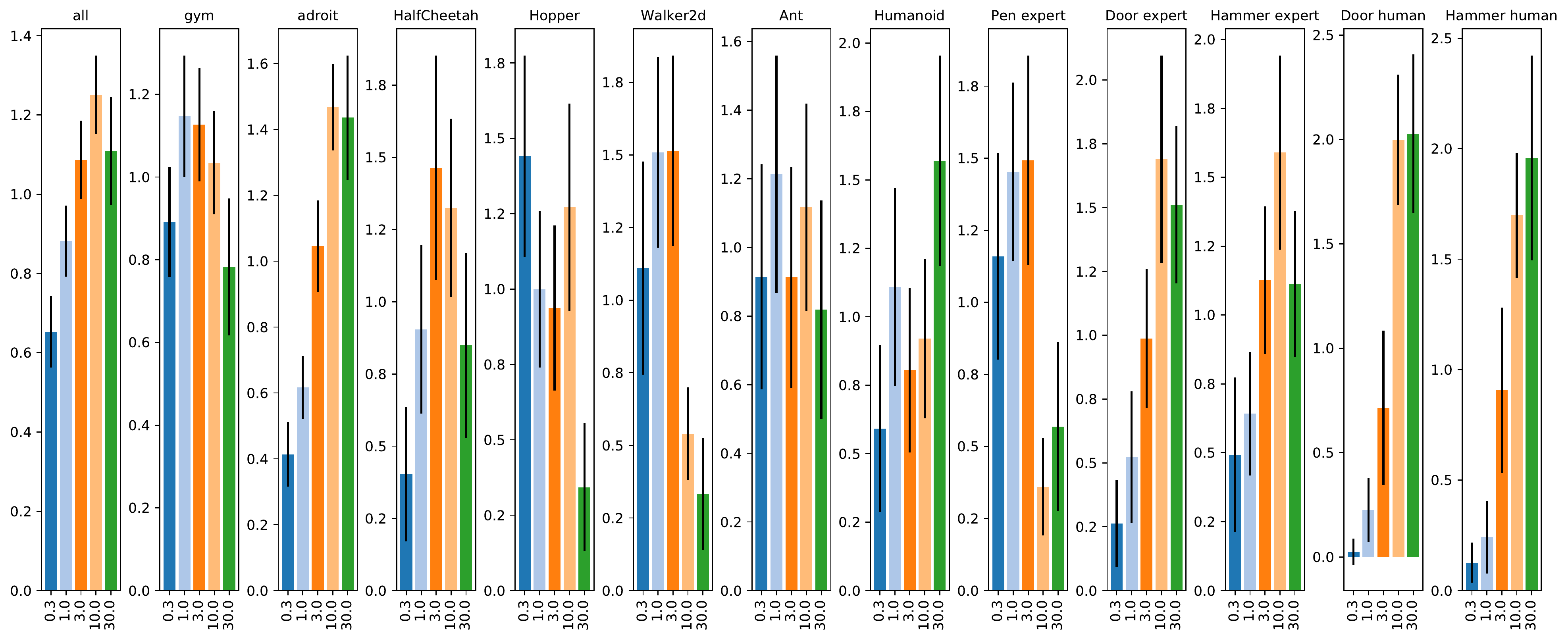}}
\caption{Analysis of choice \choicet{regweightdecay}: 95th percentile of performance scores conditioned on choice (top) and distribution of choices in top 5\% of configurations (bottom).}
\label{fig:main__gin_discriminator__optax_adamw_weight_decay}
\end{center}
\end{figure}

\begin{figure}[ht]
\begin{center}
\centerline{\includegraphics[height=4.5cm,width=1\textwidth]{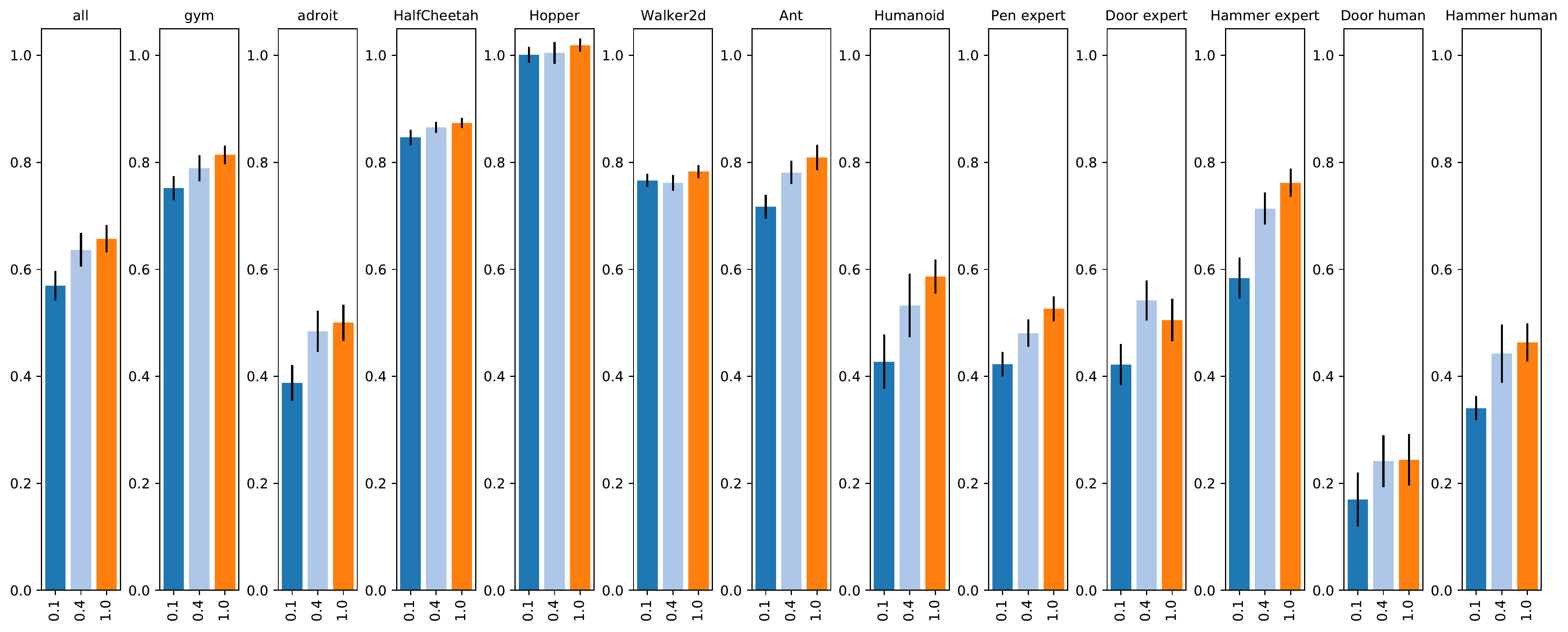}}
\centerline{\includegraphics[height=4.5cm,width=1\textwidth]{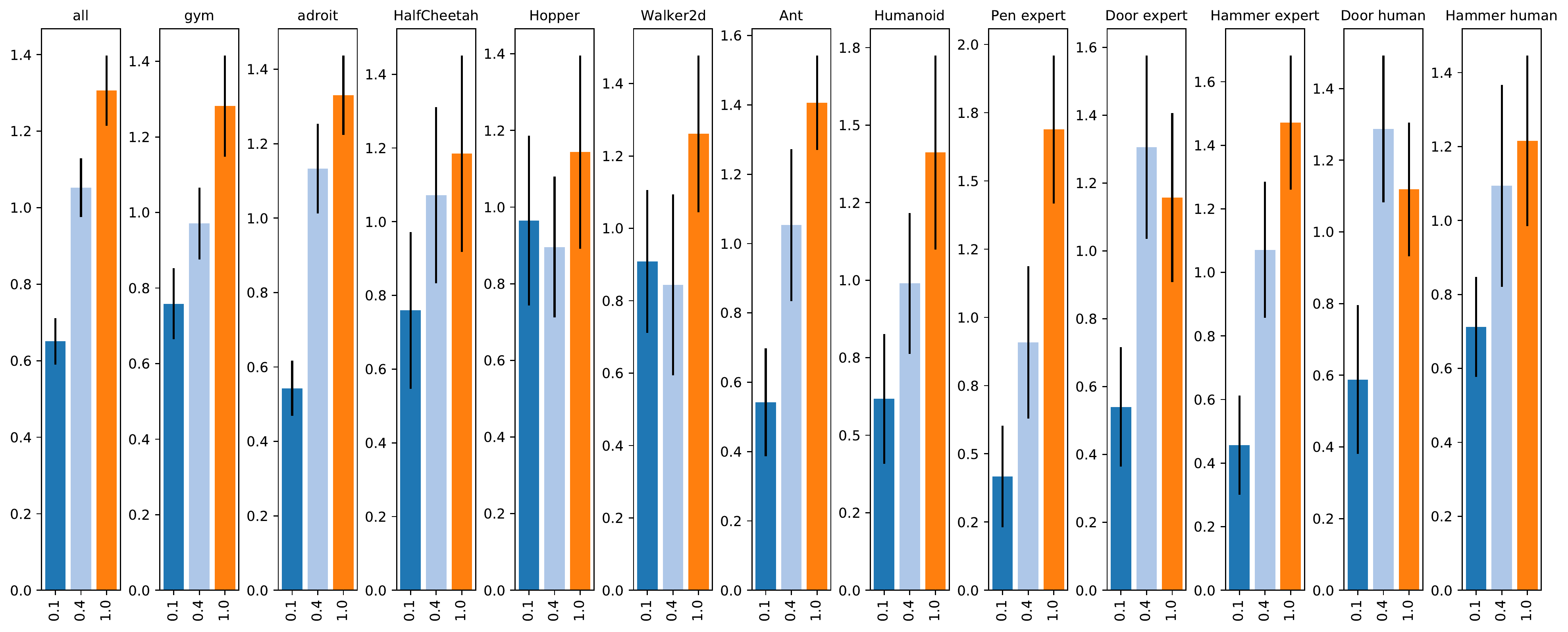}}
\caption{Analysis of choice \choicet{mixupalpha}: 95th percentile of performance scores conditioned on choice (top) and distribution of choices in top 5\% of configurations (bottom).}
\label{fig:main__gin_gail_loss_mixup_alpha}
\end{center}
\end{figure}

\begin{figure}[ht]
\begin{center}
\centerline{\includegraphics[height=4.5cm,width=1\textwidth]{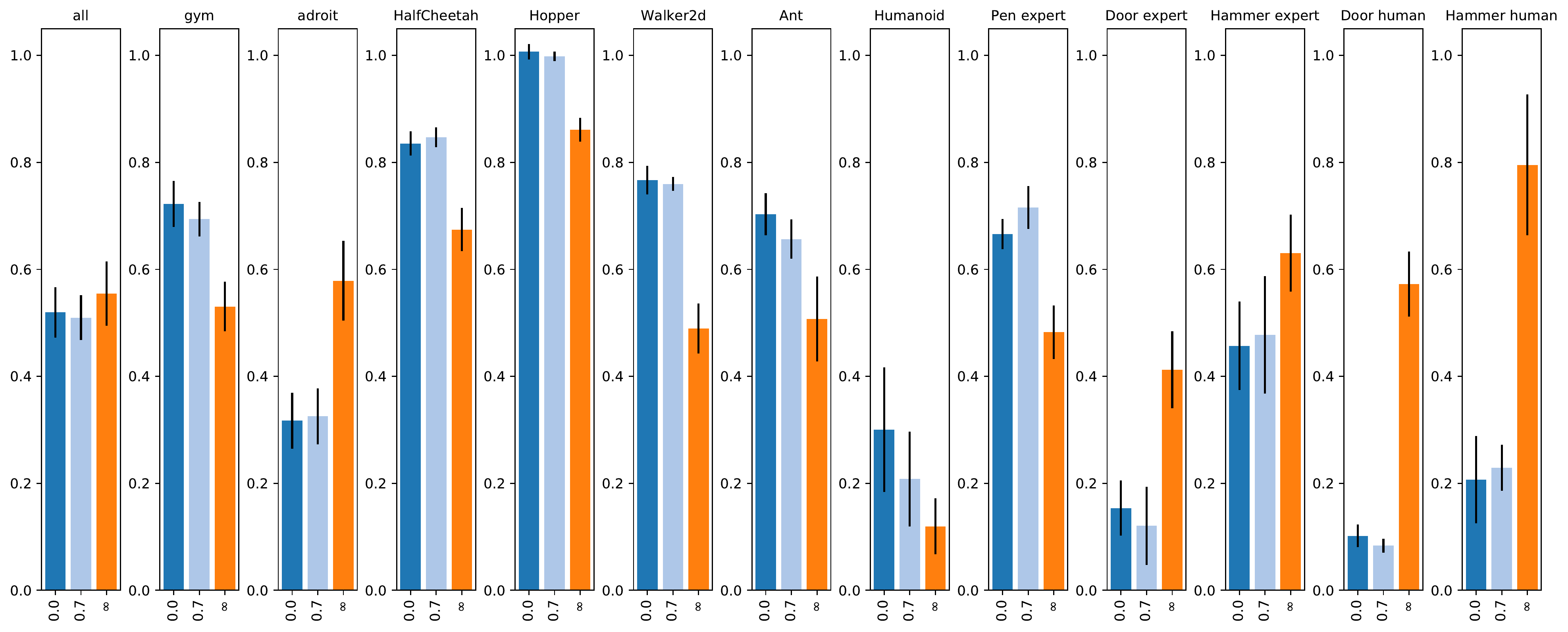}}
\centerline{\includegraphics[height=4.5cm,width=1\textwidth]{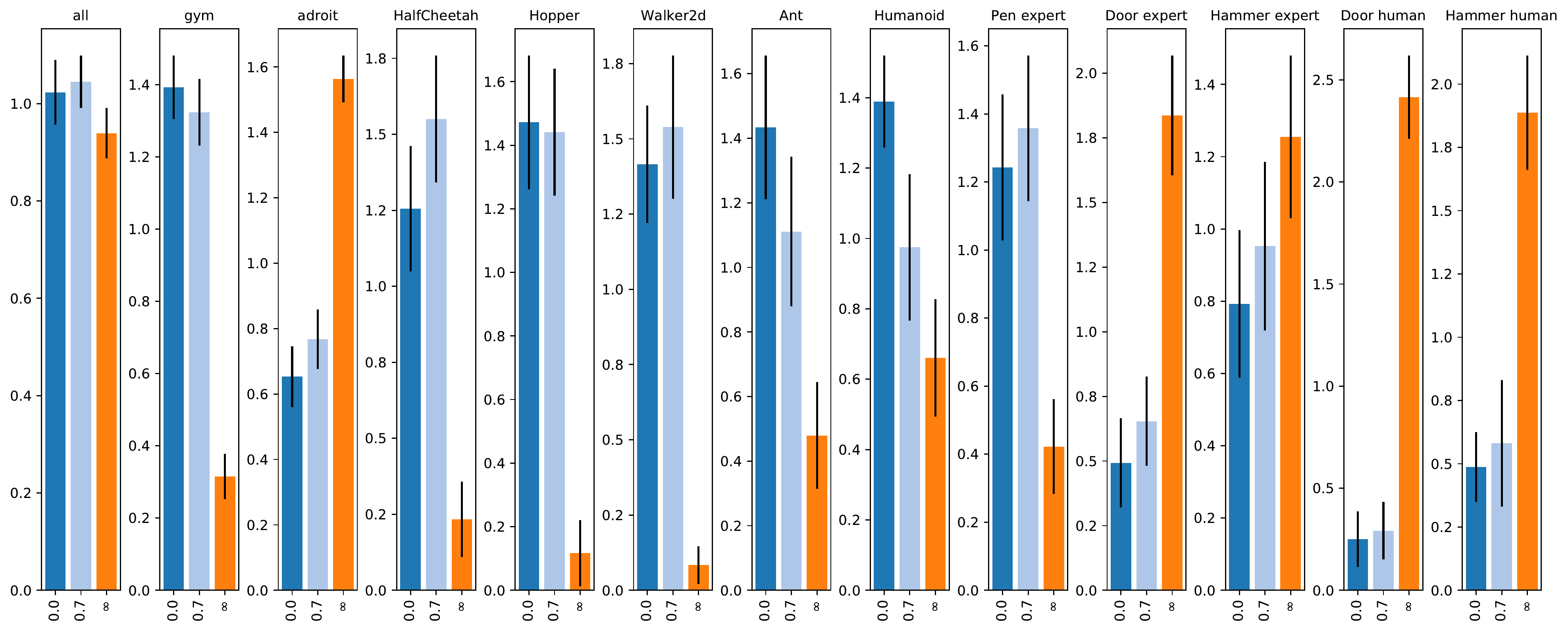}}
\caption{Analysis of choice \choicet{pugailbeta}: 95th percentile of performance scores conditioned on choice (top) and distribution of choices in top 5\% of configurations (bottom).}
\label{fig:main__gin_pugail_loss_pugail_beta}
\end{center}
\end{figure}

\begin{figure}[ht]
\begin{center}
\centerline{\includegraphics[height=4.5cm,width=1\textwidth]{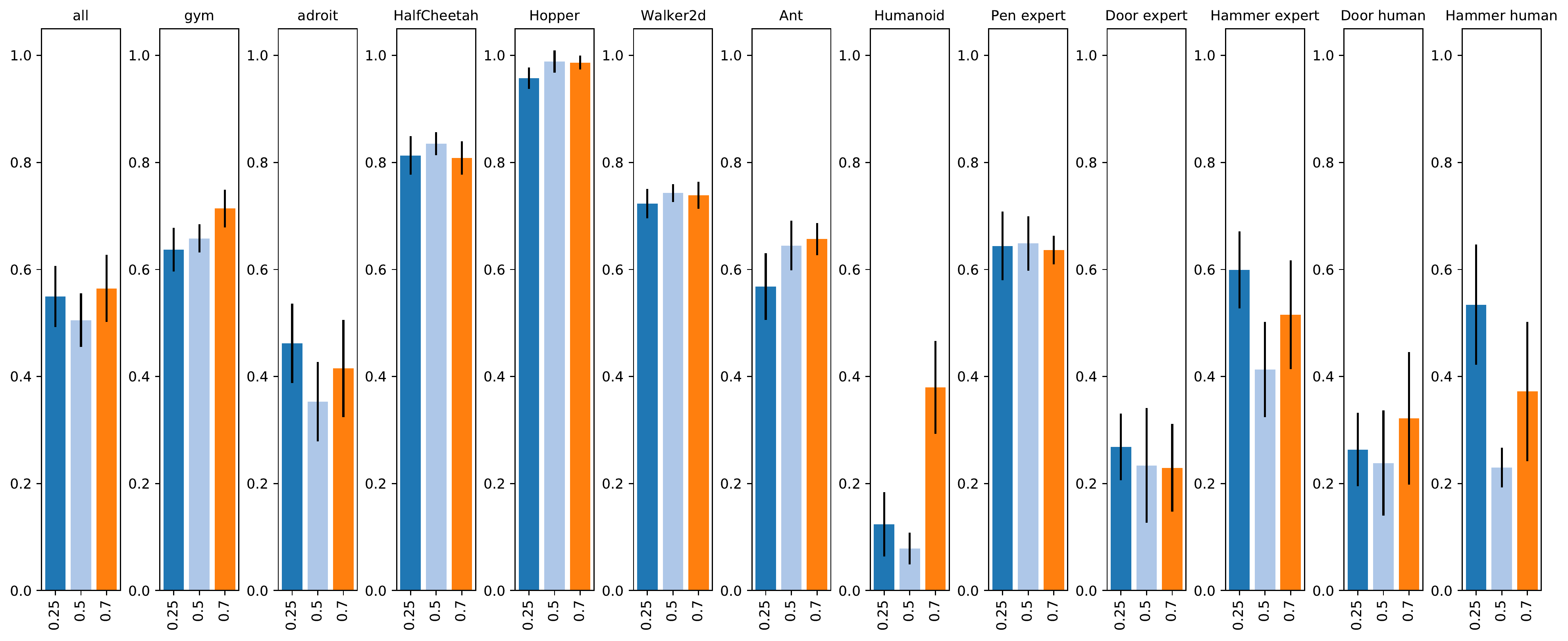}}
\centerline{\includegraphics[height=4.5cm,width=1\textwidth]{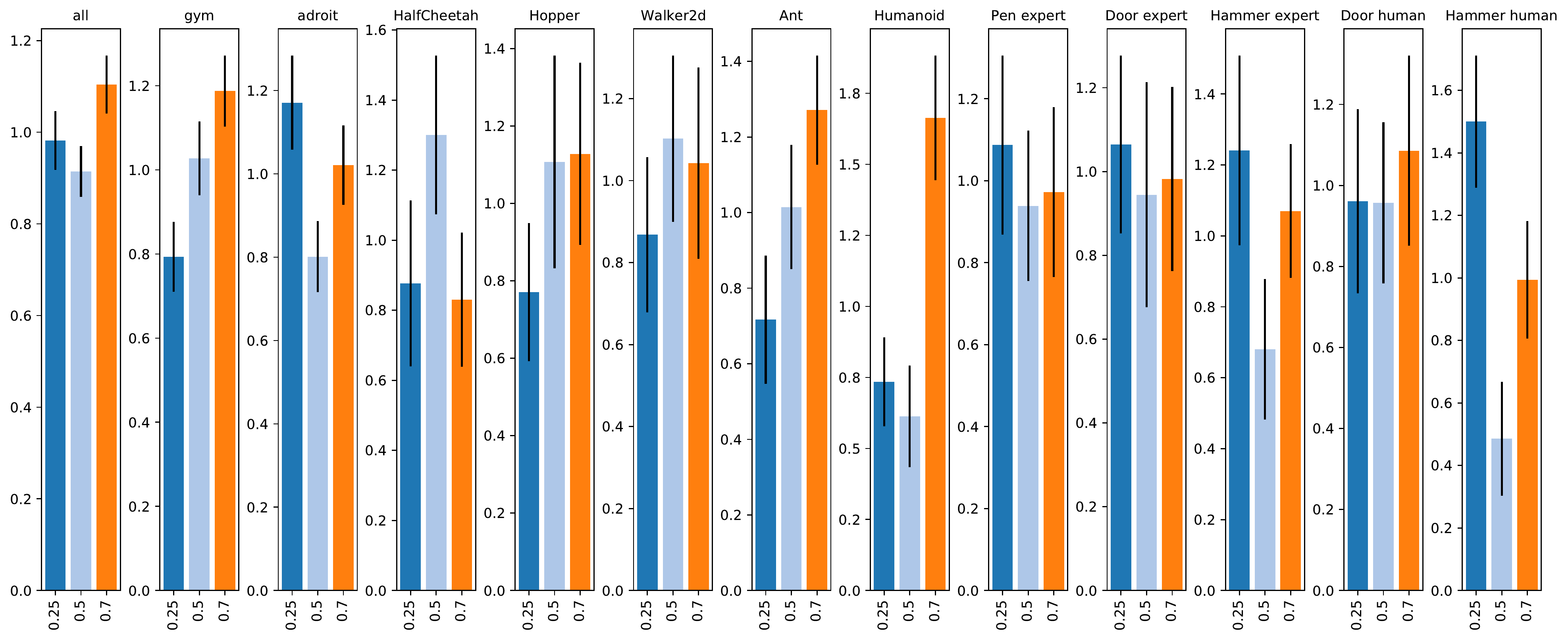}}
\caption{Analysis of choice \choicet{pugailpositiveclassprior}: 95th percentile of performance scores conditioned on choice (top) and distribution of choices in top 5\% of configurations (bottom).}
\label{fig:main__gin_pugail_loss_positive_class_prior}
\end{center}
\end{figure}

\begin{figure}[ht]
\begin{center}
\centerline{\includegraphics[height=4.5cm,width=1\textwidth]{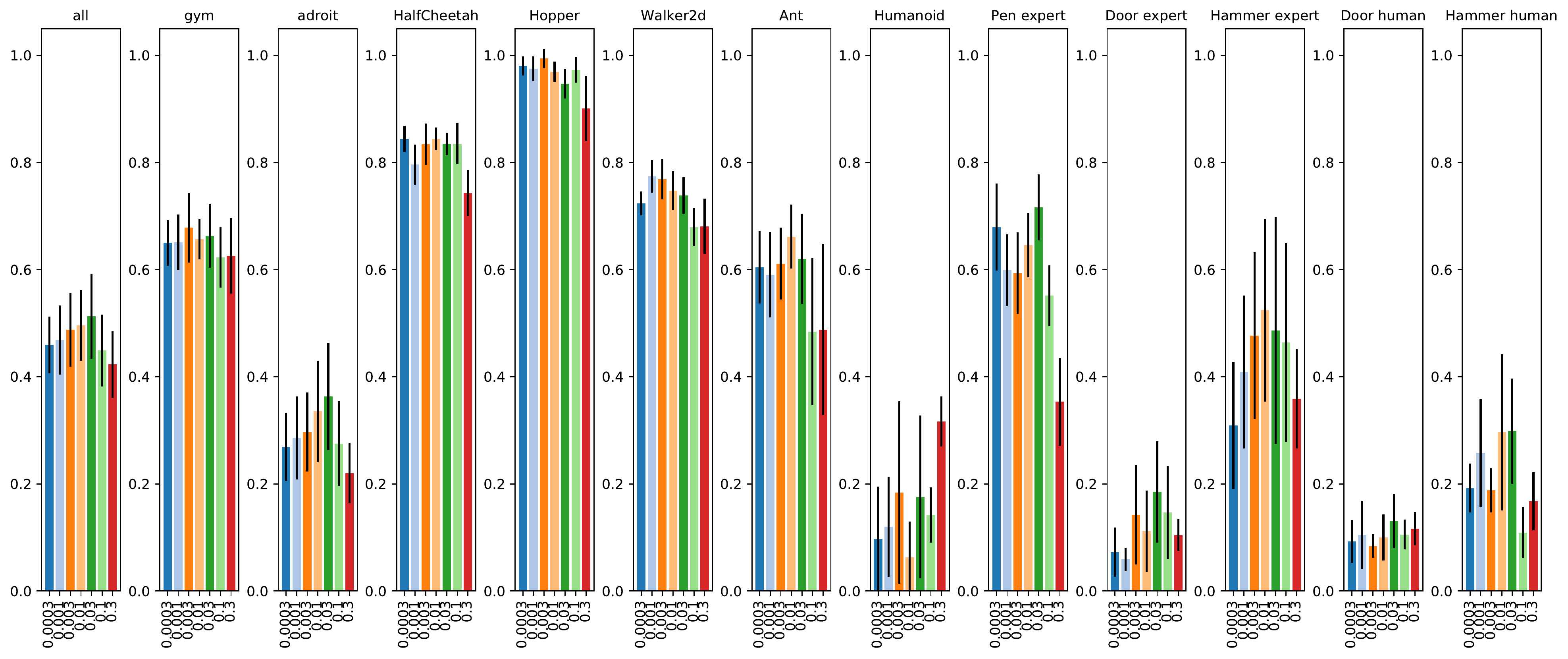}}
\centerline{\includegraphics[height=4.5cm,width=1\textwidth]{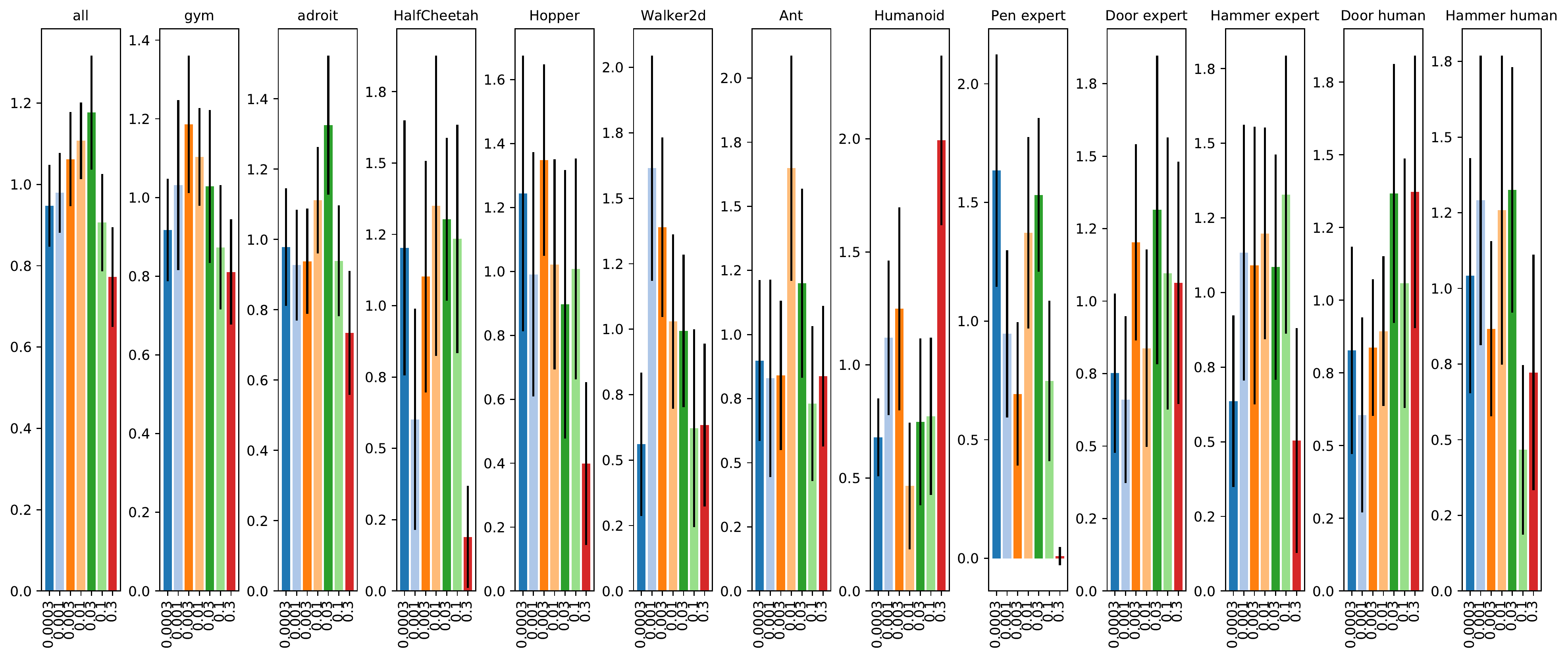}}
\caption{Analysis of choice \choicet{regentropycoef}: 95th percentile of performance scores conditioned on choice (top) and distribution of choices in top 5\% of configurations (bottom).}
\label{fig:main__gin_gail_loss_entropy_coefficient}
\end{center}
\end{figure}

\begin{figure}[ht]
\begin{center}
\centerline{\includegraphics[height=4.5cm,width=1\textwidth]{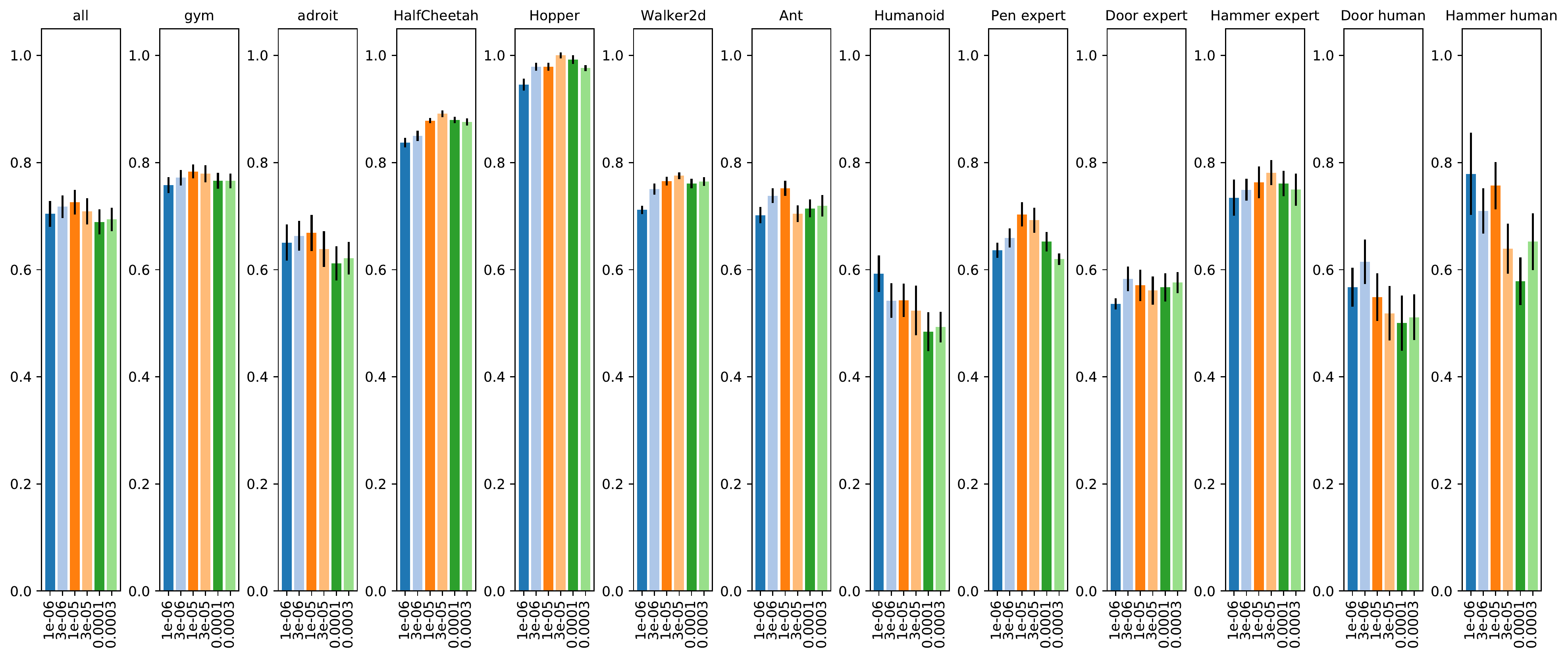}}
\centerline{\includegraphics[height=4.5cm,width=1\textwidth]{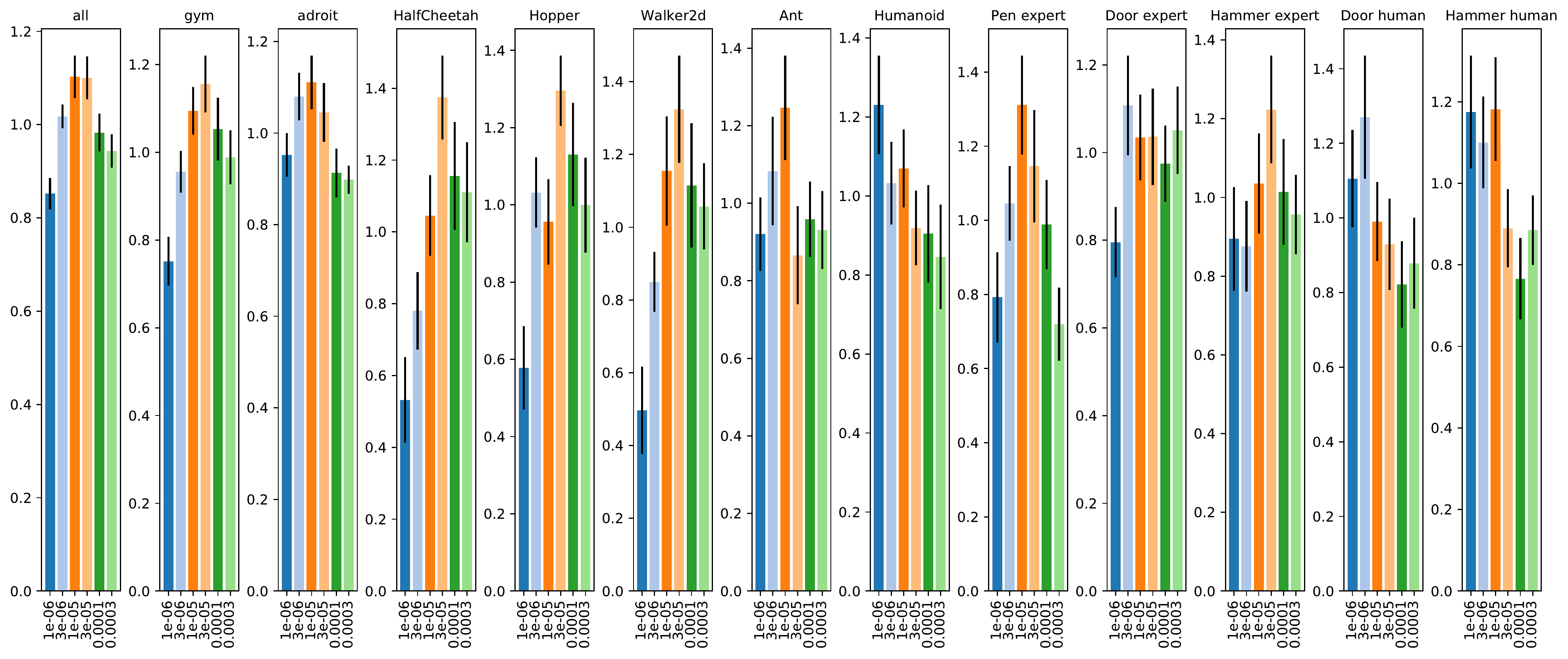}}
\caption{Analysis of choice \choicet{gaildiscriminatorlearningrate}: 95th percentile of performance scores conditioned on choice (top) and distribution of choices in top 5\% of configurations (bottom).}
\label{fig:main__gin_discriminator__optax_adamw_learning_rate}
\end{center}
\end{figure}

\begin{figure}[ht]
\begin{center}
\centerline{\includegraphics[width=1\textwidth]{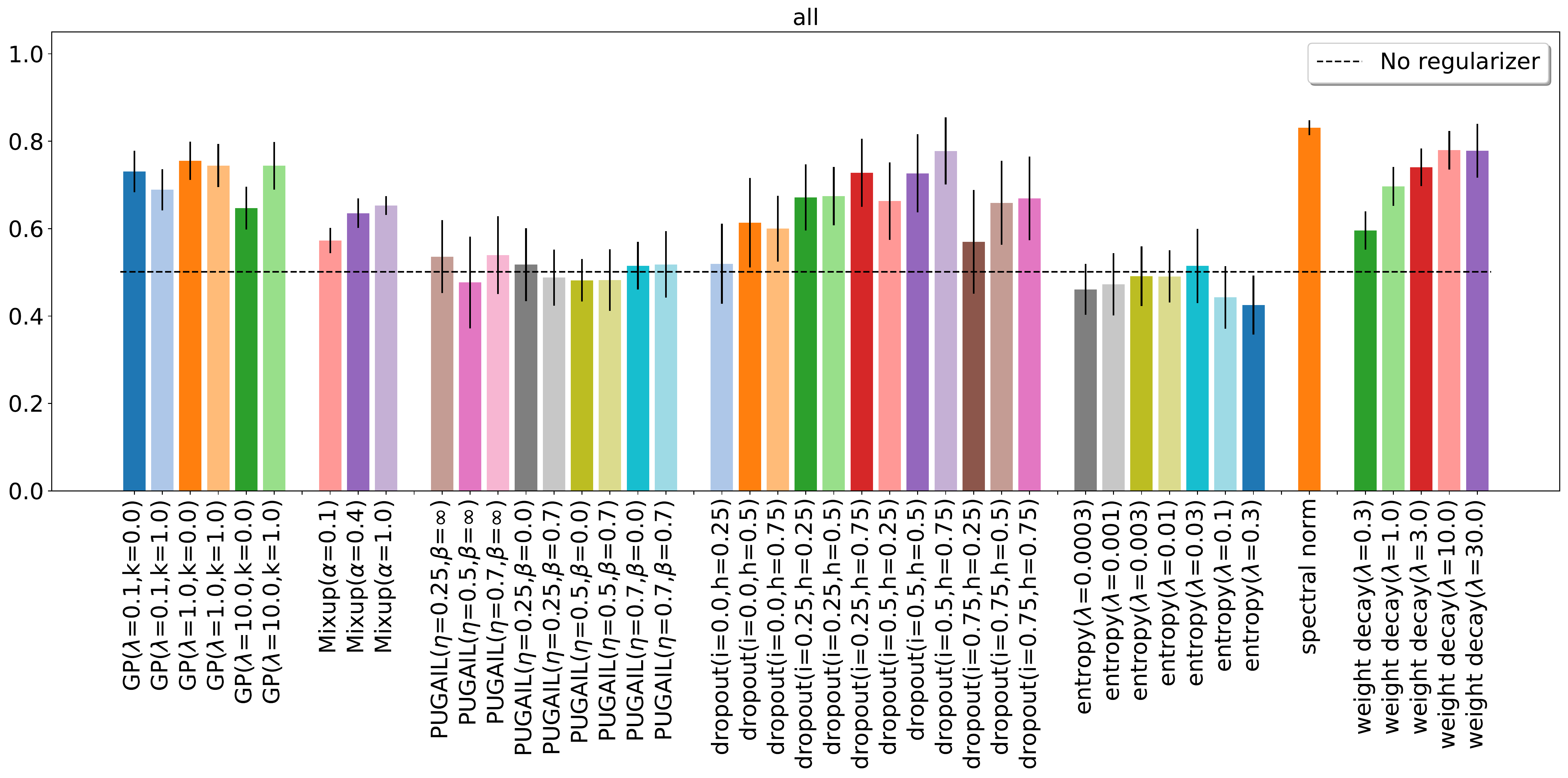}}
\caption{95th percentile of performance scores conditioned on \choicet{regularizer} and regularizers' HPs averaged across all environments.}
\label{fig:main_perf_new_regularizer_0}
\end{center}
\end{figure}

\begin{figure}[ht]
\begin{center}
\centerline{\includegraphics[width=1\textwidth]{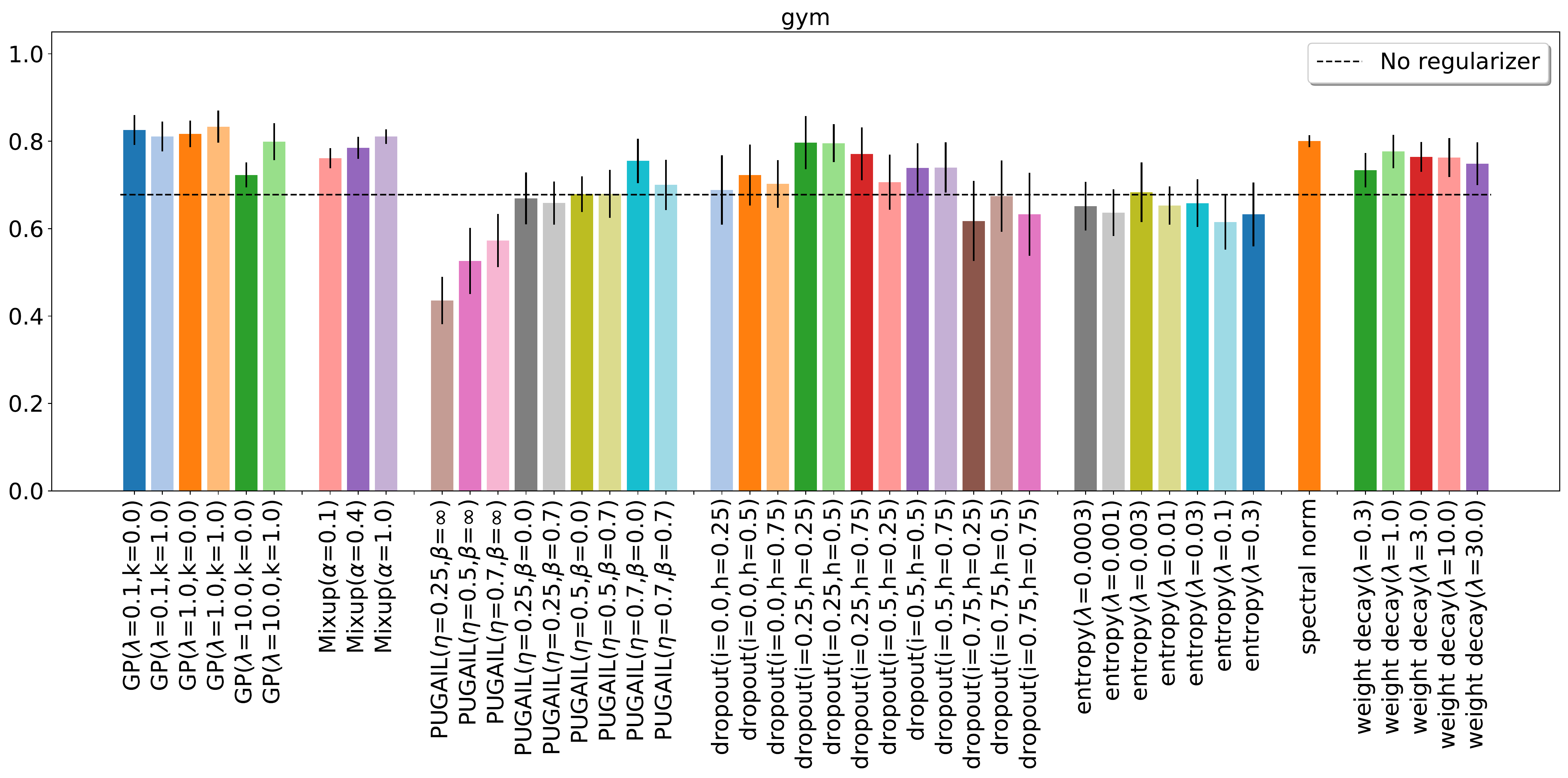}}
\caption{95th percentile of performance scores conditioned on \choicet{regularizer} and regularizers' HPs averaged across OpenAI Gym environments.}
\label{fig:main_perf_new_regularizer_1}
\end{center}
\end{figure}

\begin{figure}[ht]
\begin{center}
\centerline{\includegraphics[width=1\textwidth]{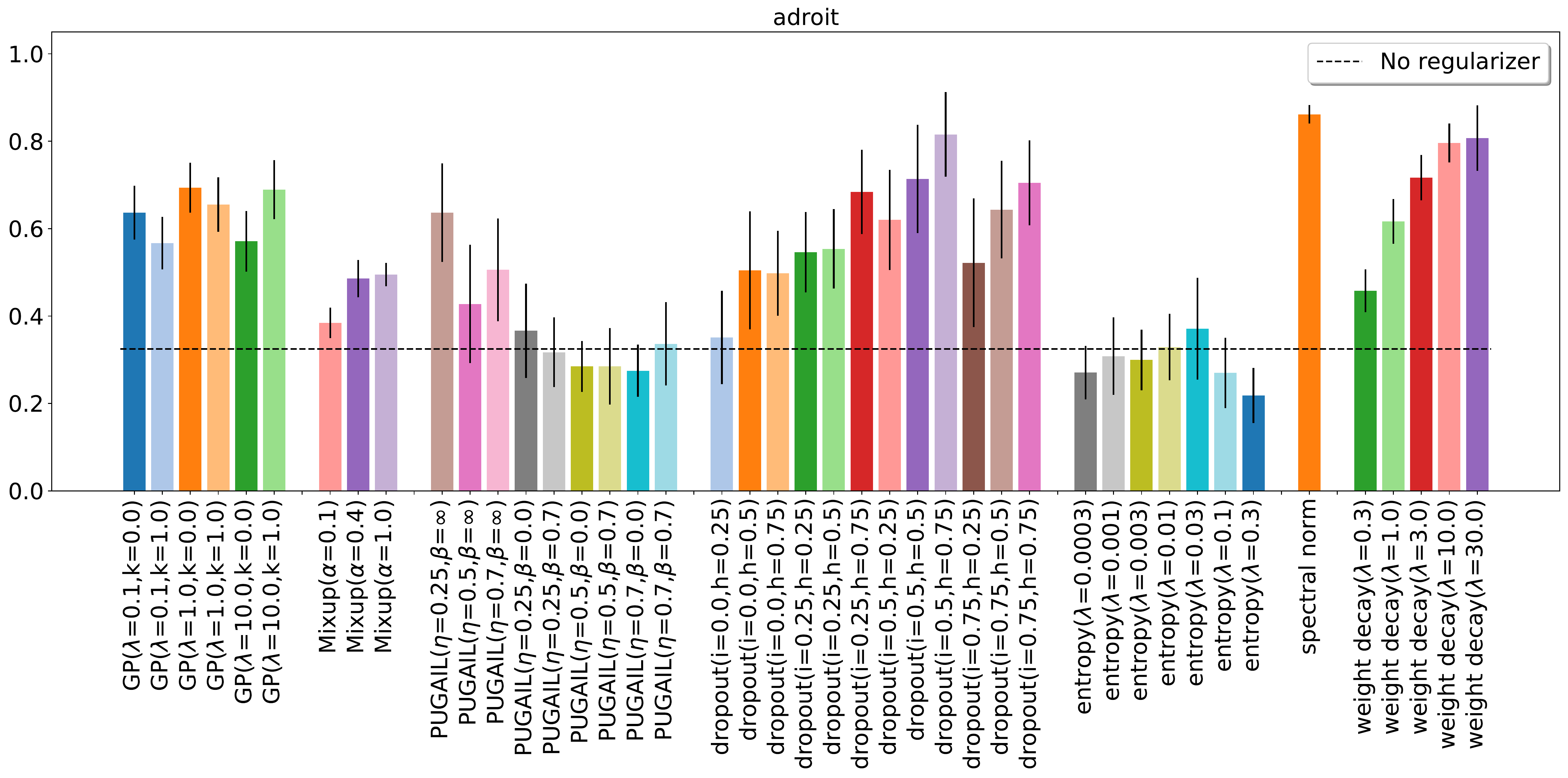}}
\caption{95th percentile of performance scores conditioned on \choicet{regularizer} and regularizers' HPs averaged across Adroit environments.}
\label{fig:main_perf_new_regularizer_2}
\end{center}
\end{figure}

\begin{figure}[ht]
\begin{center}
\centerline{\includegraphics[width=1\textwidth]{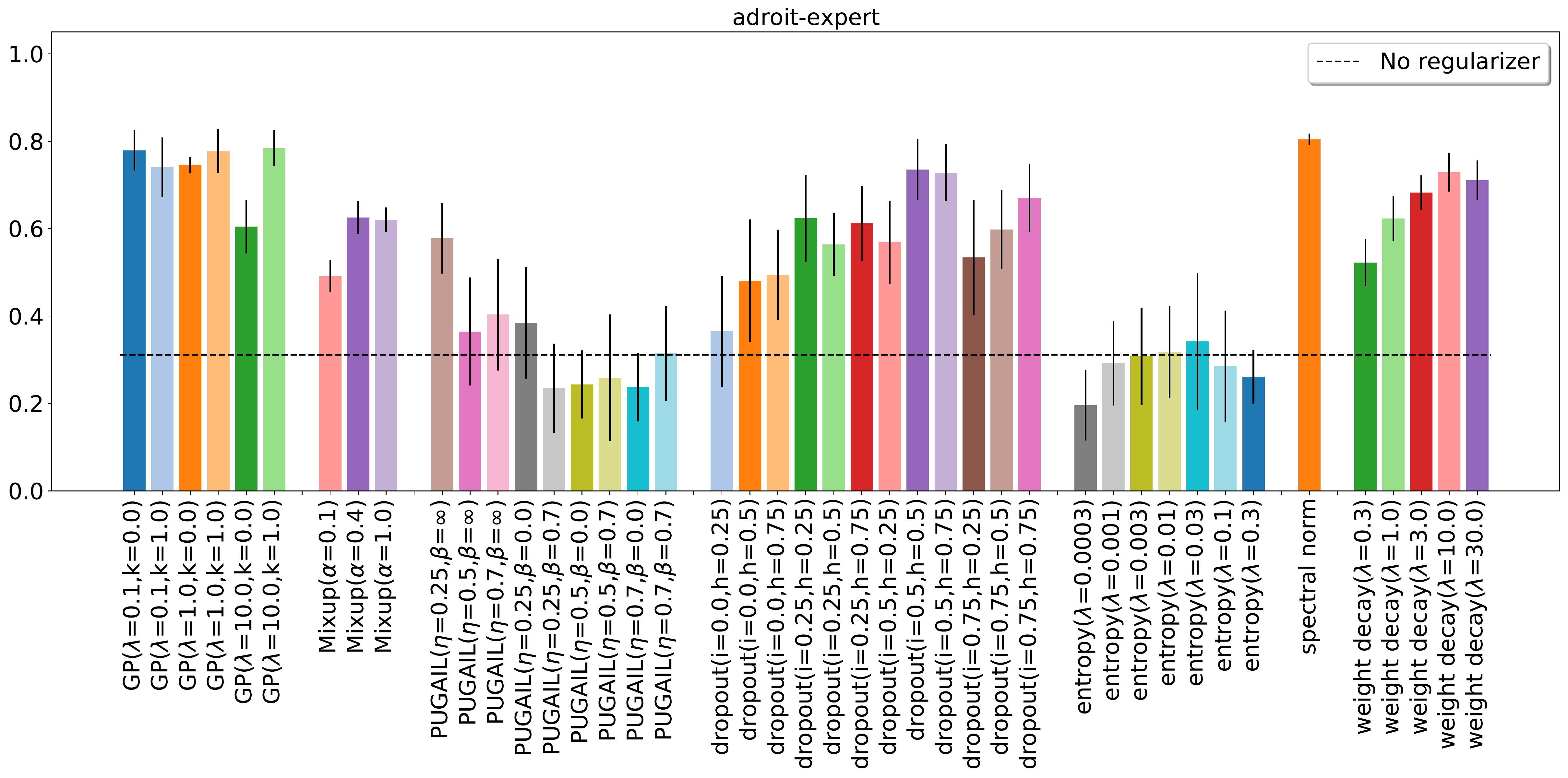}}
\caption{95th percentile of performance scores conditioned on \choicet{regularizer} and regularizers' HPs averaged across \texttt{door-expert} and \texttt{hammer-expert} tasks.}
\label{fig:main_perf_new_regularizer_3}
\end{center}
\end{figure}

\begin{figure}[ht]
\begin{center}
\centerline{\includegraphics[width=1\textwidth]{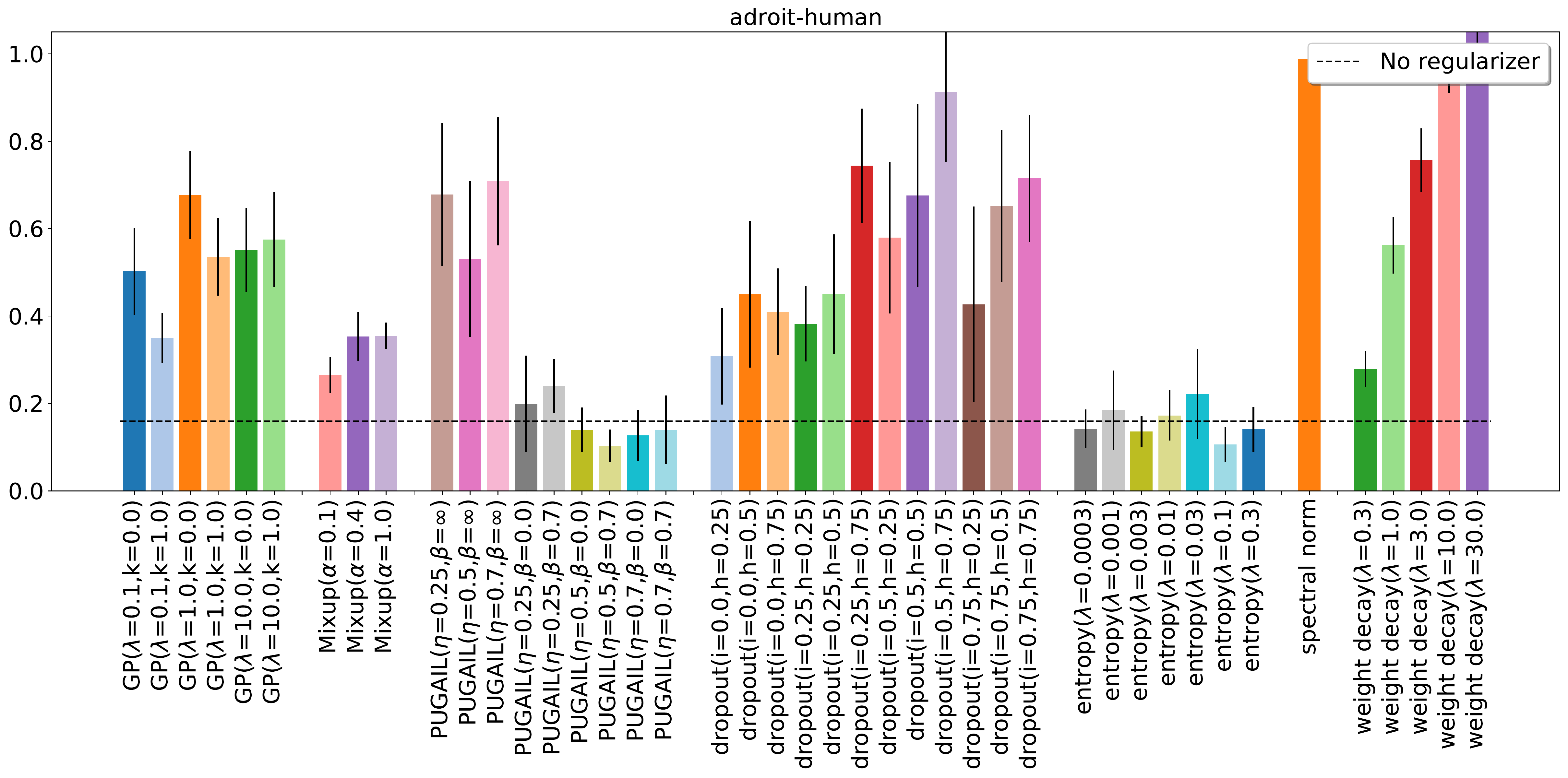}}
\caption{95th percentile of performance scores conditioned on \choicet{regularizer} and regularizers' HPs averaged across \texttt{door-human} and \texttt{hammer-human} tasks.}
\label{fig:main_perf_new_regularizer_4}
\end{center}
\end{figure}

For each of the sampled choice configurations we compute the performance metric as described in Section~\ref{sec:design}.
We report aggregate statistics of the experiment in Tables~\ref{tab:main_overview}--\ref{tab:main_overview4} as well as training curves in Figure~\ref{fig:main_training_curves}.
We further provide per-choice analyses in Figures~\ref{fig:main_direct_rl_algorithm}-\ref{fig:main_perf_new_regularizer_4}.
\begin{table}[ht]
\begin{center}
\caption{Quantiles of the \emph{final} agent performance across HP configurations for OpenAI Gym tasks.}
\label{tab:main_overview}
\begin{tabular}{lrrrrr}
\toprule
{} &  Ant & HalfCheetah & Hopper & Humanoid & Walker2d \\
\midrule
90\% & 0.90 &        1.07 &   1.18 &     0.51 &     0.99 \\
95\% & 0.99 &        1.11 &   1.20 &     0.87 &     1.01 \\
99\% & 1.07 &        1.17 &   1.23 &     1.01 &     1.04 \\
Max & 1.18 &        1.37 &   1.34 &     1.06 &     1.21 \\
\bottomrule
\end{tabular}

\end{center}
\end{table}\begin{table}[ht]
\begin{center}
\caption{Quantiles of the \emph{final} agent performance across HP configurations for Adroit tasks.}
\label{tab:main_overview2}
\begin{tabular}{lrrrrr}
\toprule
{} & Door expert & Door human & Hammer expert & Hammer human & Pen expert \\
\midrule
90\% &        0.72 &       0.25 &          1.08 &         0.46 &       0.74 \\
95\% &        0.91 &       0.83 &          1.26 &         1.15 &       0.89 \\
99\% &        1.04 &       2.29 &          1.37 &         3.04 &       1.11 \\
Max &        1.16 &       3.73 &          1.45 &         5.55 &       1.44 \\
\bottomrule
\end{tabular}

\end{center}
\end{table}\begin{table}[ht]
\begin{center}
\caption{Quantiles of the \emph{average} agent performance during training across HP configurations for OpenAI Gym tasks.}
\label{tab:main_overview3}
\begin{tabular}{lrrrrr}
\toprule
{} &  Ant & HalfCheetah & Hopper & Humanoid & Walker2d \\
\midrule
90\% & 0.61 &        0.82 &   0.93 &     0.29 &     0.70 \\
95\% & 0.72 &        0.87 &   0.98 &     0.53 &     0.76 \\
99\% & 0.85 &        0.94 &   1.06 &     0.79 &     0.84 \\
Max & 0.96 &        1.05 &   1.10 &     0.92 &     0.92 \\
\bottomrule
\end{tabular}

\end{center}
\end{table}\begin{table}[ht]
\begin{center}
\caption{Quantiles of the \emph{average} agent performance during training across HP configurations for Adroit tasks.}
\label{tab:main_overview4}
\begin{tabular}{lrrrrr}
\toprule
{} & Door expert & Door human & Hammer expert & Hammer human & Pen expert \\
\midrule
90\% &        0.42 &       0.30 &          0.59 &         0.42 &       0.56 \\
95\% &        0.57 &       0.56 &          0.77 &         0.70 &       0.66 \\
99\% &        0.74 &       1.04 &          0.96 &         1.23 &       0.84 \\
Max &        0.92 &       2.08 &          1.18 &         3.42 &       1.09 \\
\bottomrule
\end{tabular}

\end{center}
\end{table}


\begin{figure}[ht]
\begin{center}
\centerline{\includegraphics[height=4.5cm,width=1\textwidth]{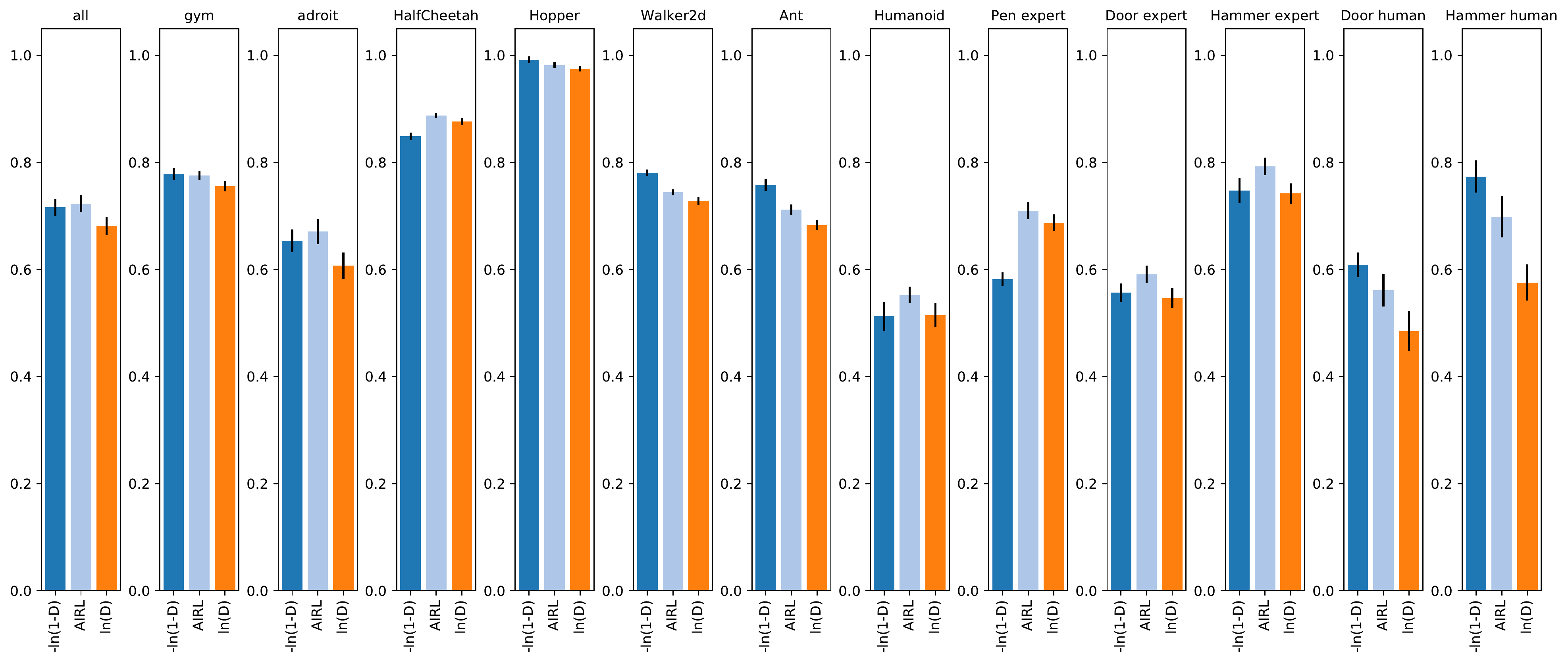}}
\centerline{\includegraphics[height=4.5cm,width=1\textwidth]{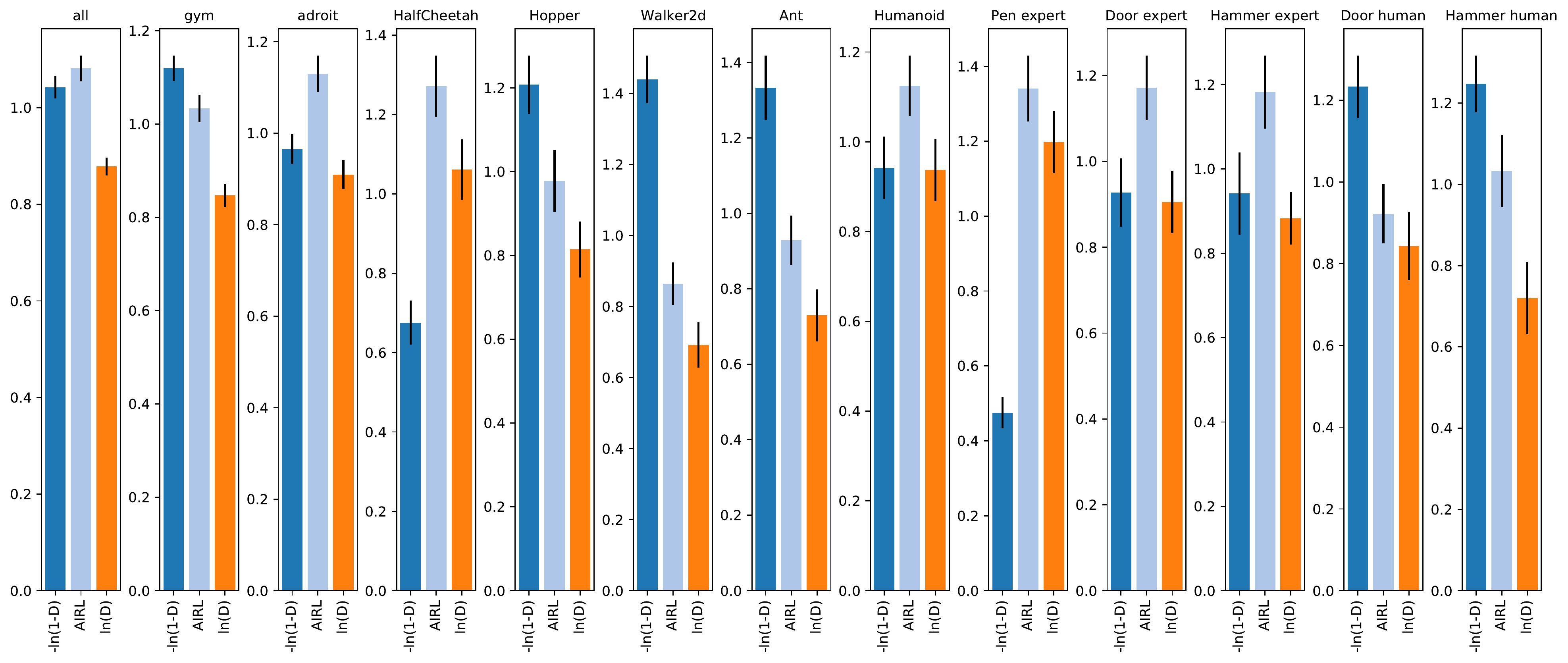}}
\caption{Analysis of choice \choicet{gailreward}: 95th percentile of performance scores conditioned on choice (top) and distribution of choices in top 5\% of configurations (bottom).}
\label{fig:main__gin_reward_function__macro_value}
\end{center}
\end{figure}

\begin{figure}[ht]
\begin{center}
\centerline{\includegraphics[height=4.5cm,width=1\textwidth]{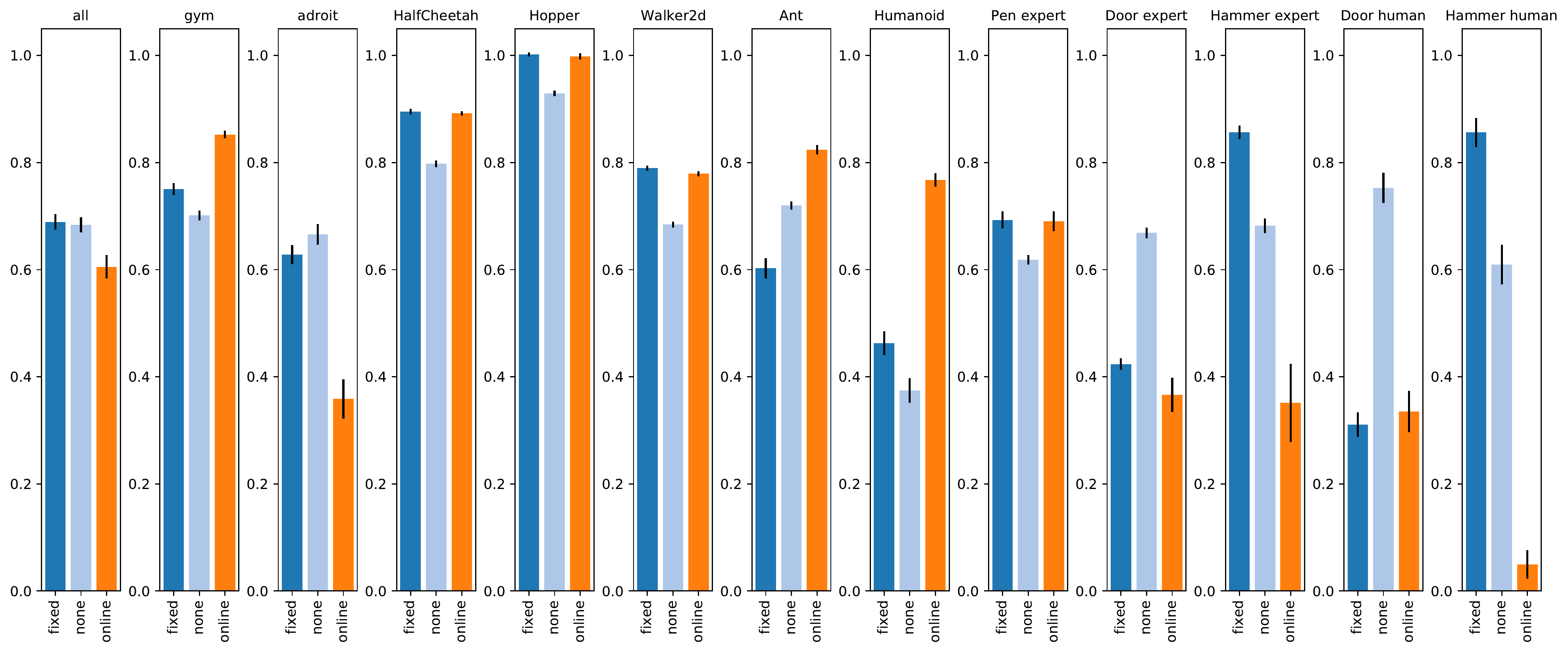}}
\centerline{\includegraphics[height=4.5cm,width=1\textwidth]{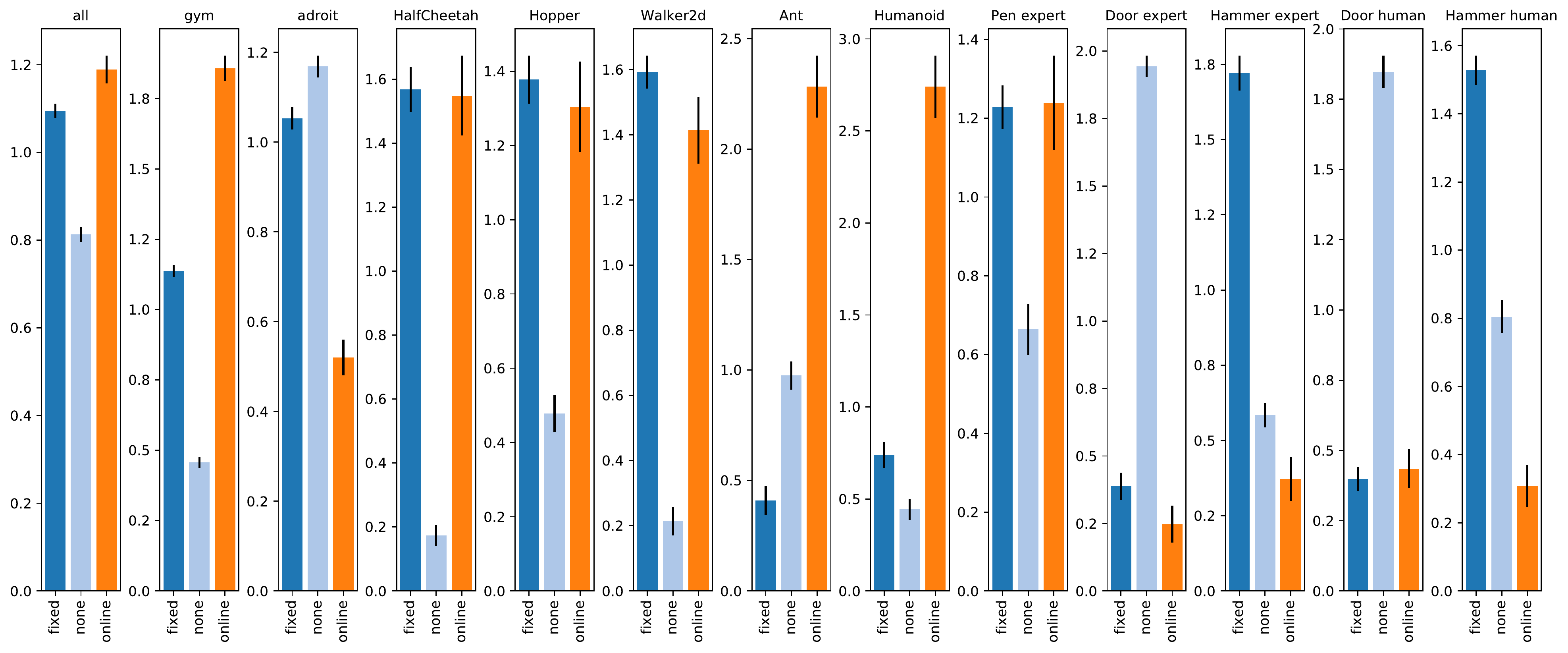}}
\caption{Analysis of choice \choicet{obsnormalization}: 95th percentile of performance scores conditioned on choice (top) and distribution of choices in top 5\% of configurations (bottom).}
\label{fig:main_obs_normalization}
\end{center}
\end{figure}

\begin{figure}[ht]
\begin{center}
\centerline{\includegraphics[height=4.5cm,width=1\textwidth]{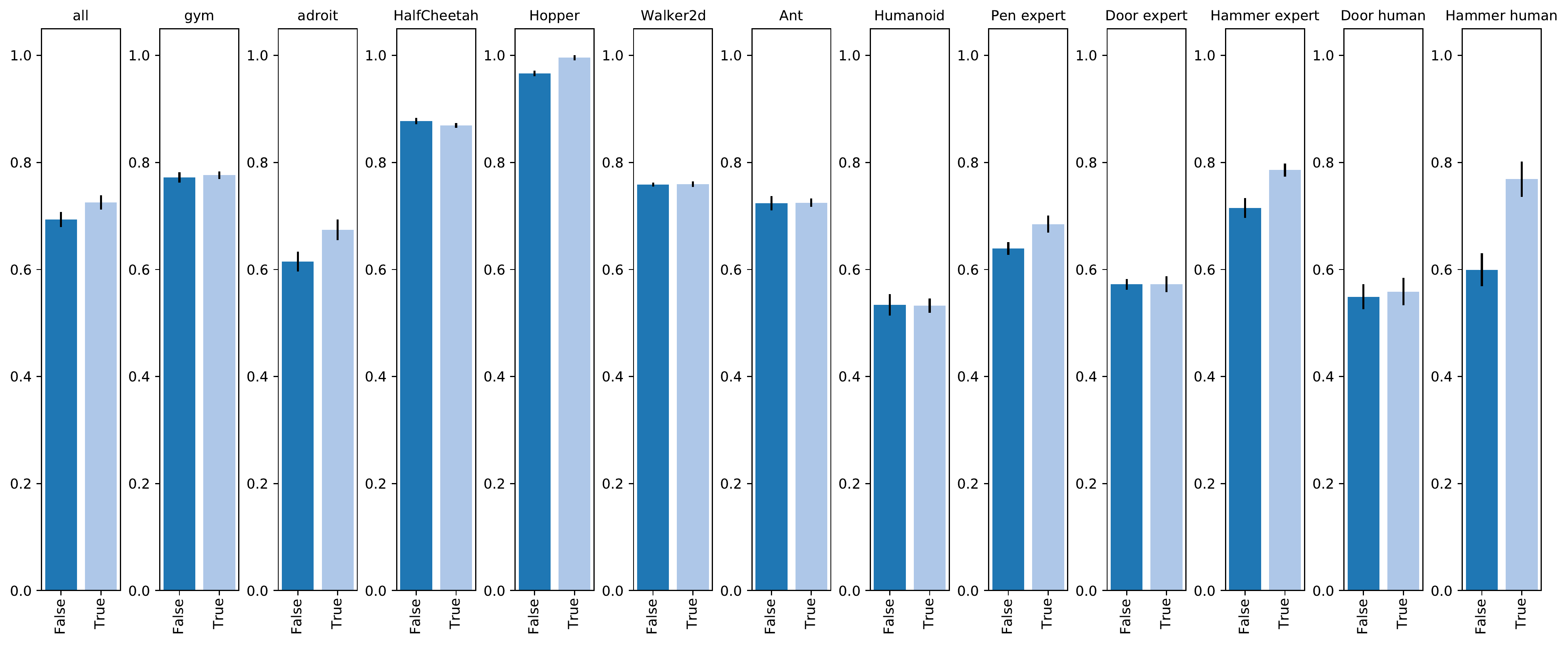}}
\centerline{\includegraphics[height=4.5cm,width=1\textwidth]{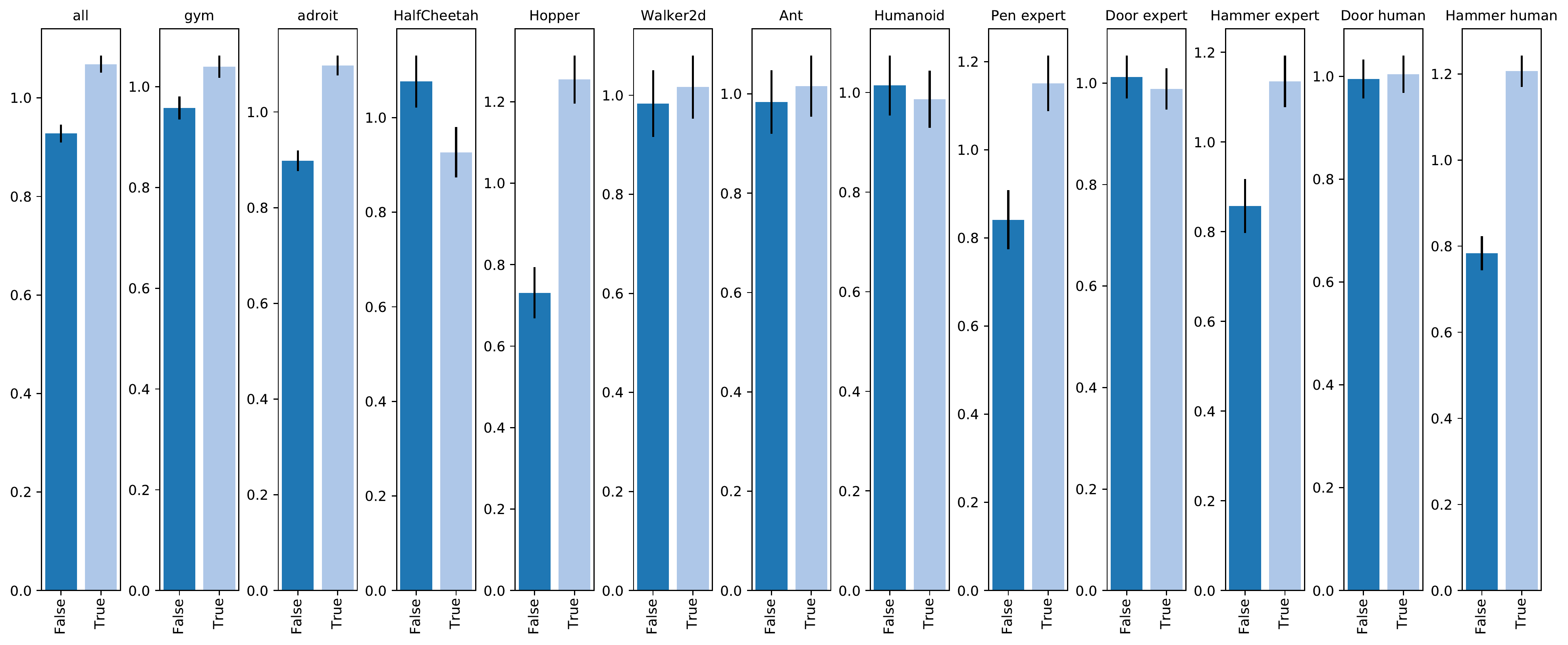}}
\caption{Analysis of choice \choicet{pretrainwithbc}: 95th percentile of performance scores conditioned on choice (top) and distribution of choices in top 5\% of configurations (bottom).}
\label{fig:main_pretrain_with_bc}
\end{center}
\end{figure}

\begin{figure}[ht]
\begin{center}
\centerline{\includegraphics[height=4.5cm,width=1\textwidth]{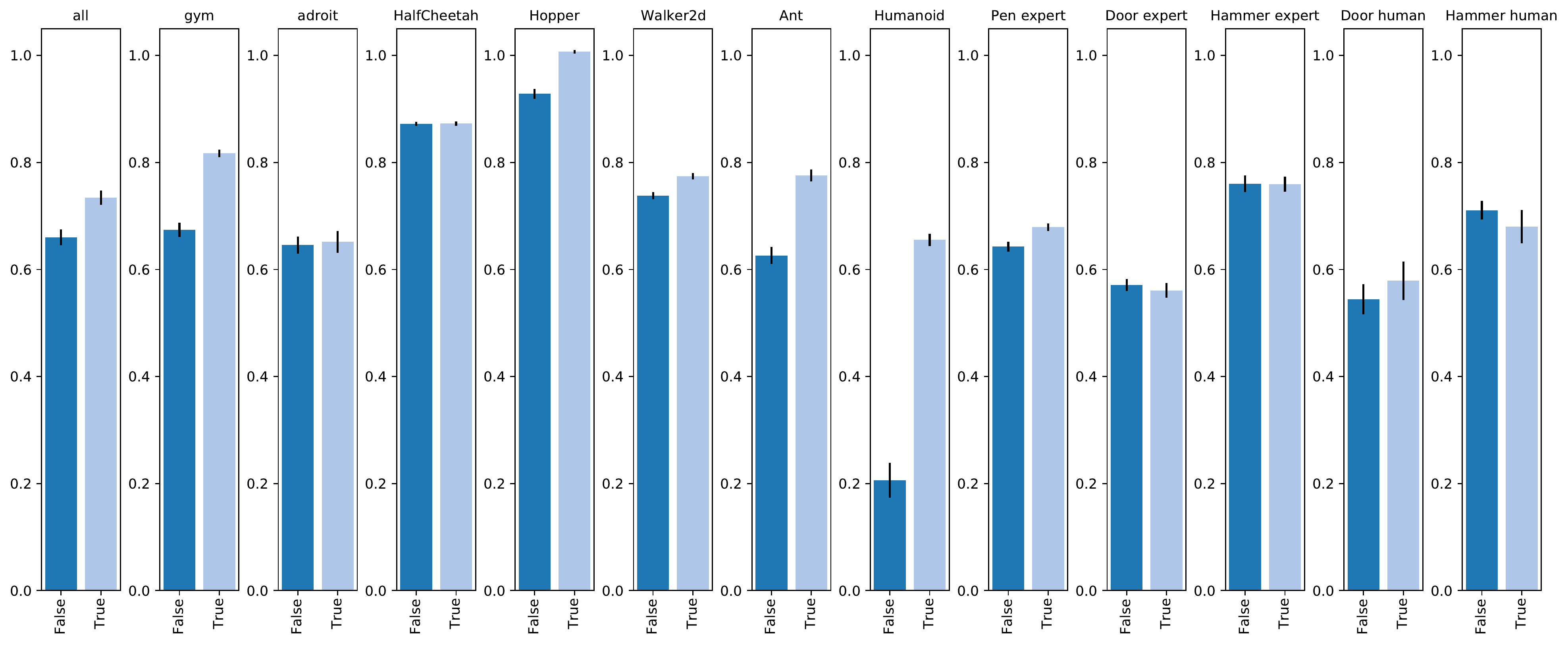}}
\centerline{\includegraphics[height=4.5cm,width=1\textwidth]{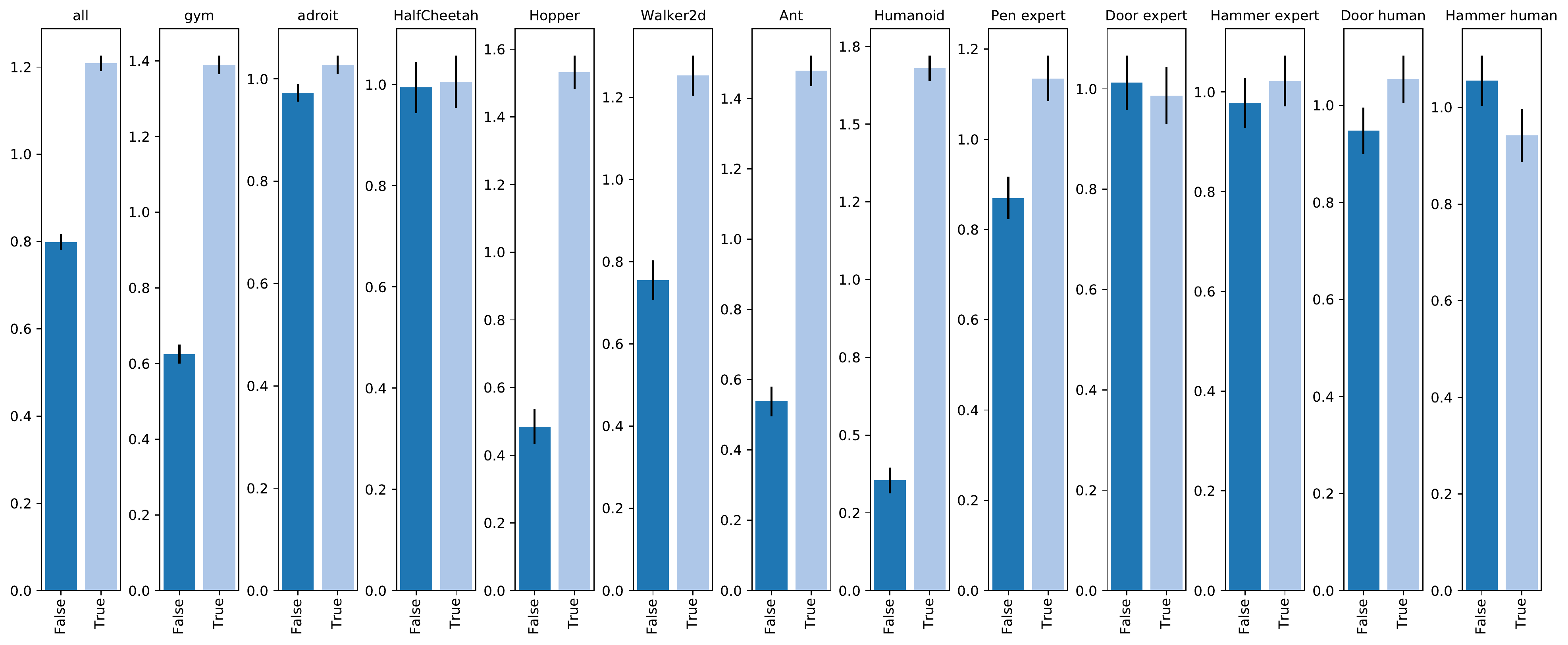}}
\caption{Analysis of choice \choicet{explicitabsorbingstate}: 95th percentile of performance scores conditioned on choice (top) and distribution of choices in top 5\% of configurations (bottom).}
\label{fig:main_explicit_absorbing_state}
\end{center}
\end{figure}


\begin{figure}[ht]
\begin{center}
\centerline{\includegraphics[height=4.5cm,width=1\textwidth]{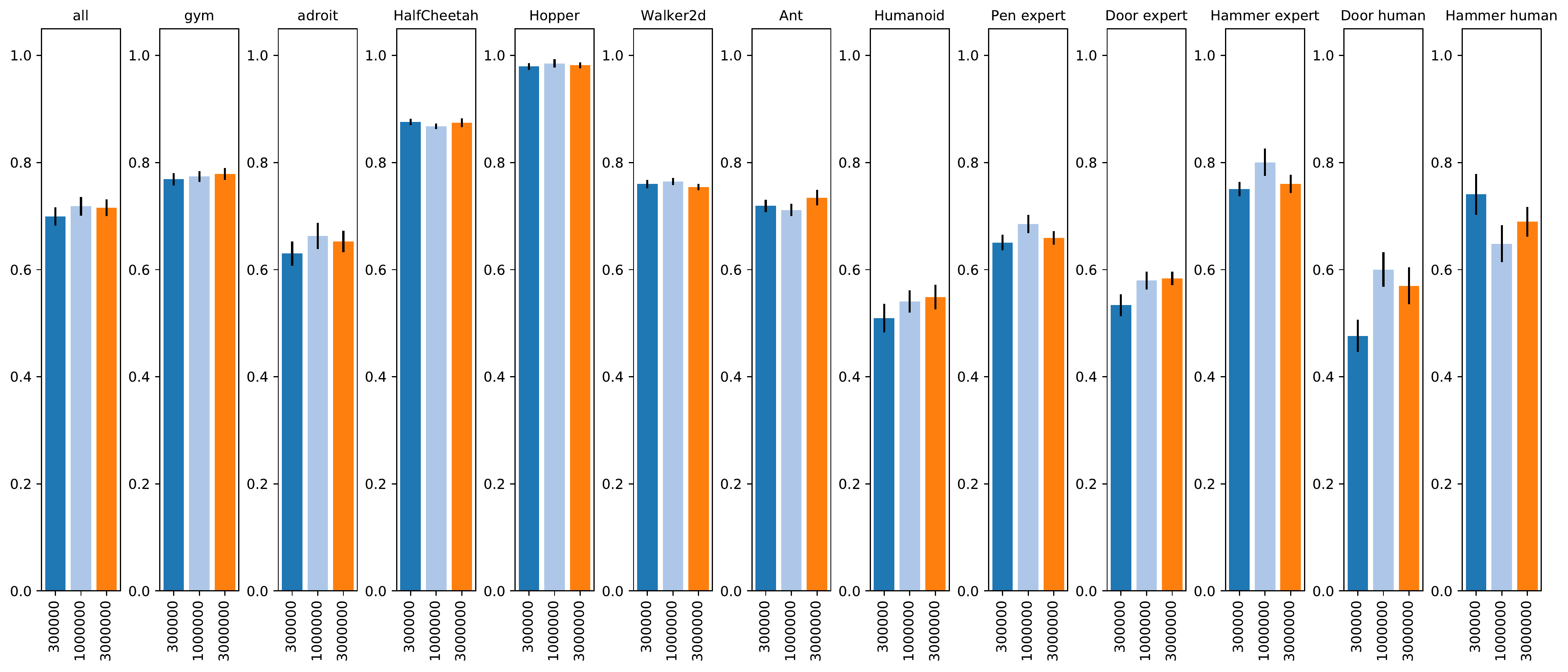}}
\centerline{\includegraphics[height=4.5cm,width=1\textwidth]{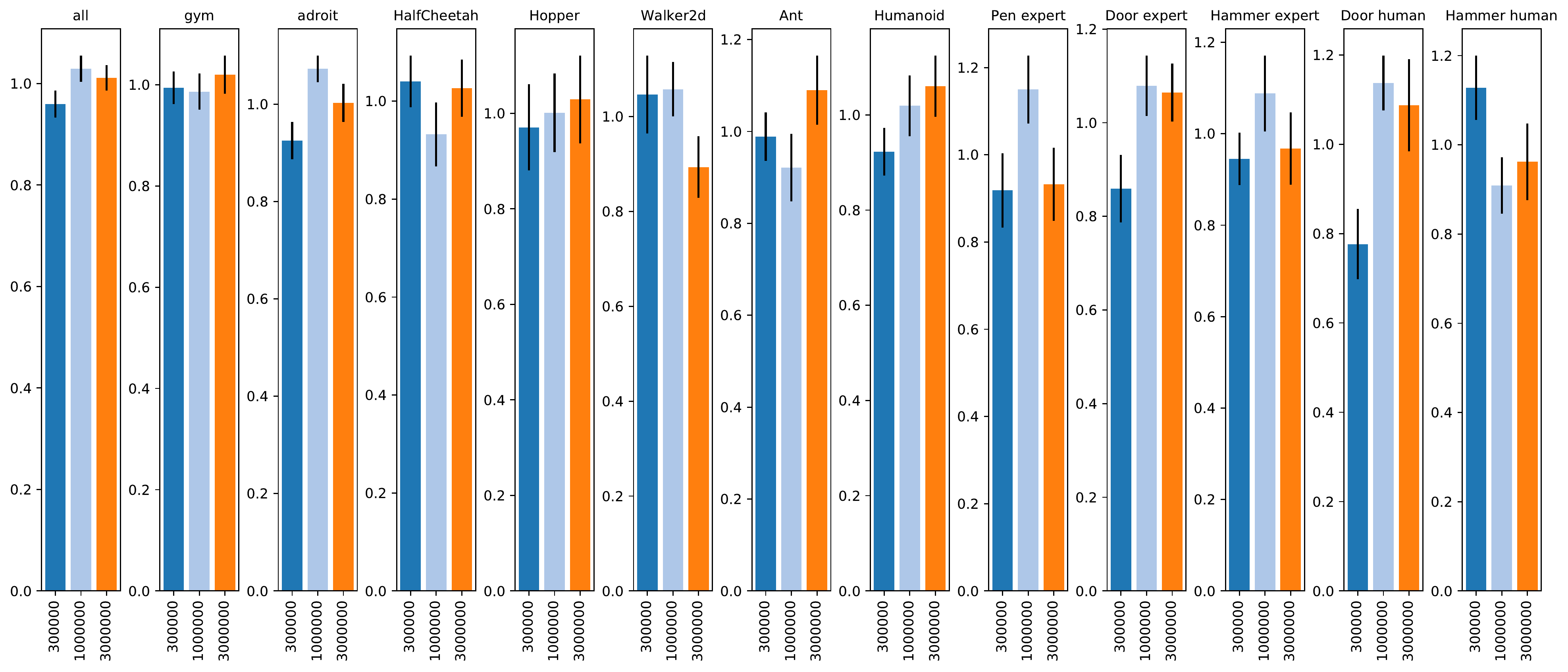}}
\caption{Analysis of choice \choicet{maxreplaysize}: 95th percentile of performance scores conditioned on choice (top) and distribution of choices in top 5\% of configurations (bottom).}
\label{fig:main_max_replay_size}
\end{center}
\end{figure}

\begin{figure}[ht]
\begin{center}
\centerline{\includegraphics[height=4.5cm,width=1\textwidth]{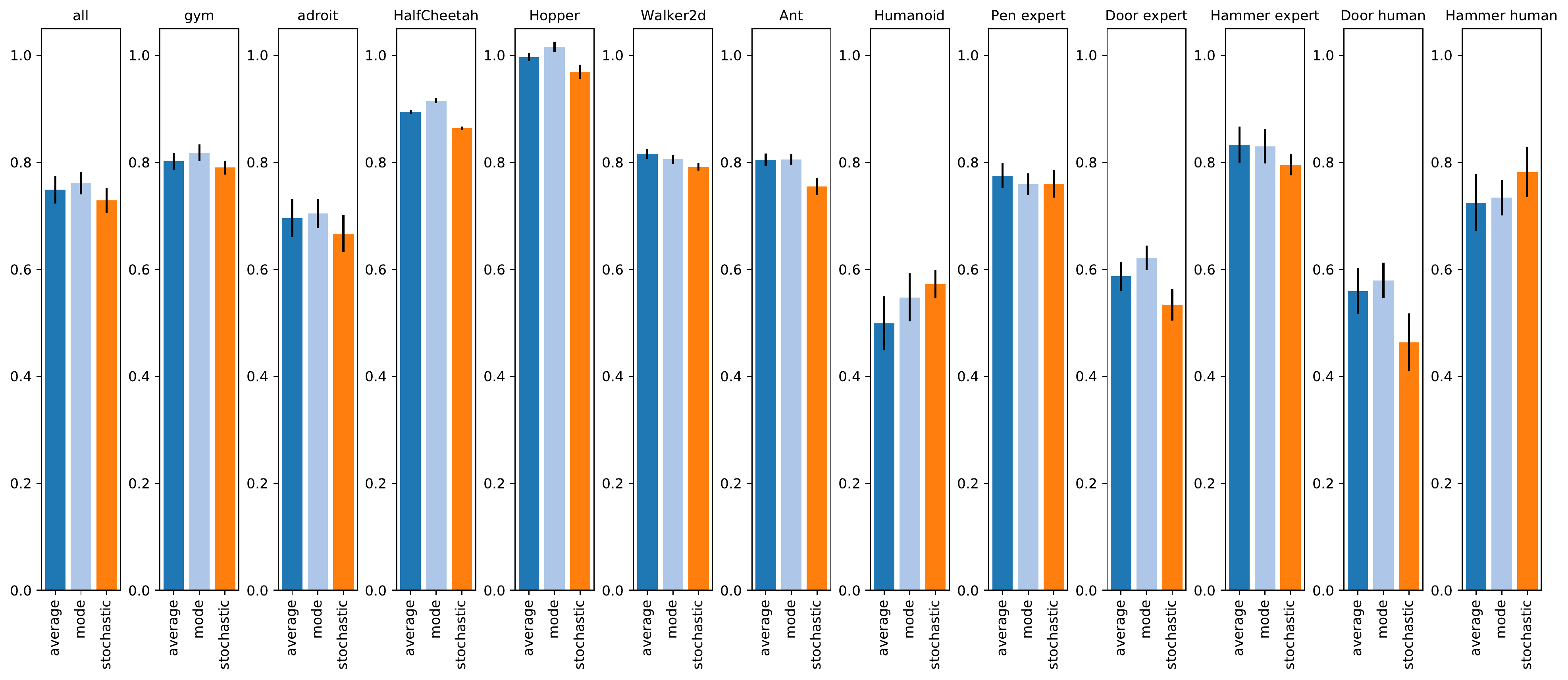}}
\centerline{\includegraphics[height=4.5cm,width=1\textwidth]{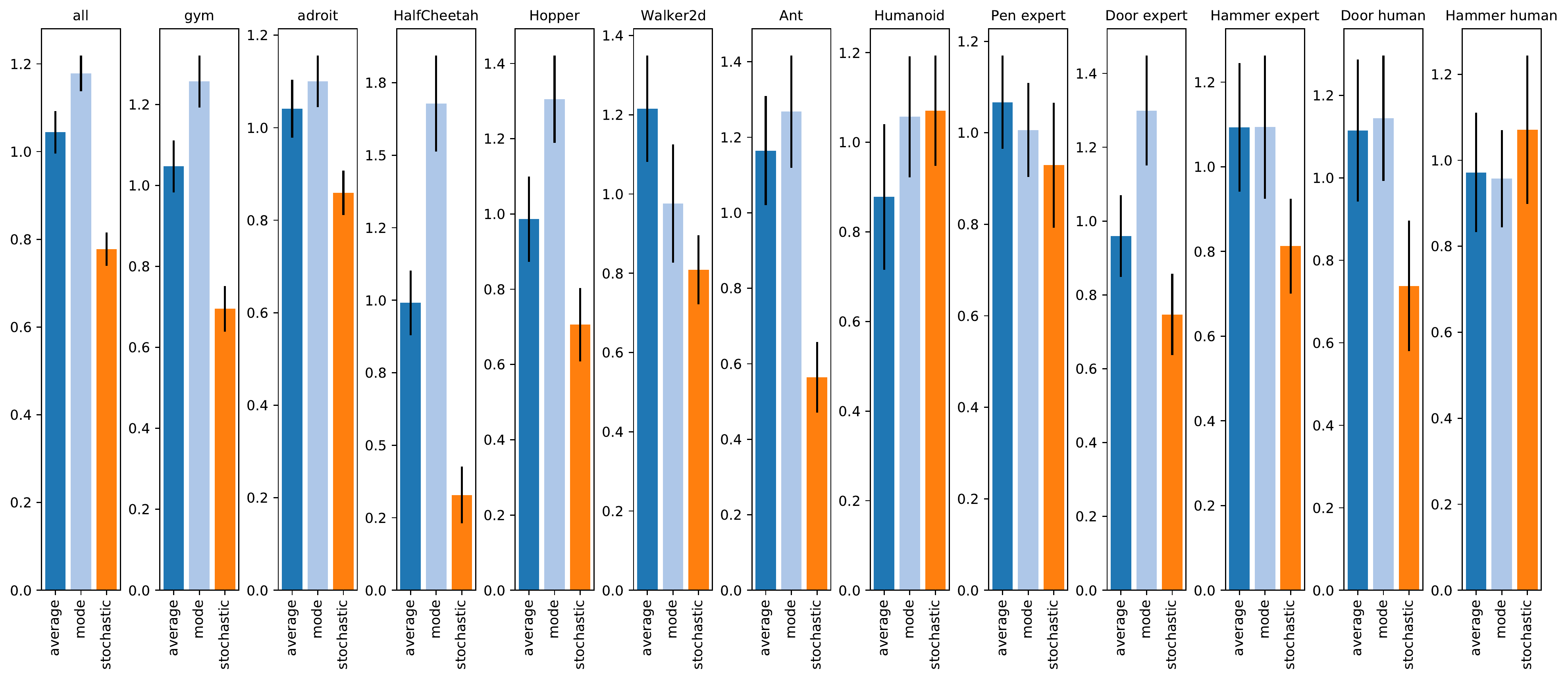}}
\caption{Analysis of choice \choicet{evalbehaviorpolicytype}: 95th percentile of performance scores conditioned on choice (top) and distribution of choices in top 5\% of configurations (bottom).}
\label{fig:main_eval_behavior_policy_type}
\end{center}
\end{figure}

\begin{figure}[ht]
\begin{center}
\centerline{\includegraphics[height=4.5cm,width=1\textwidth]{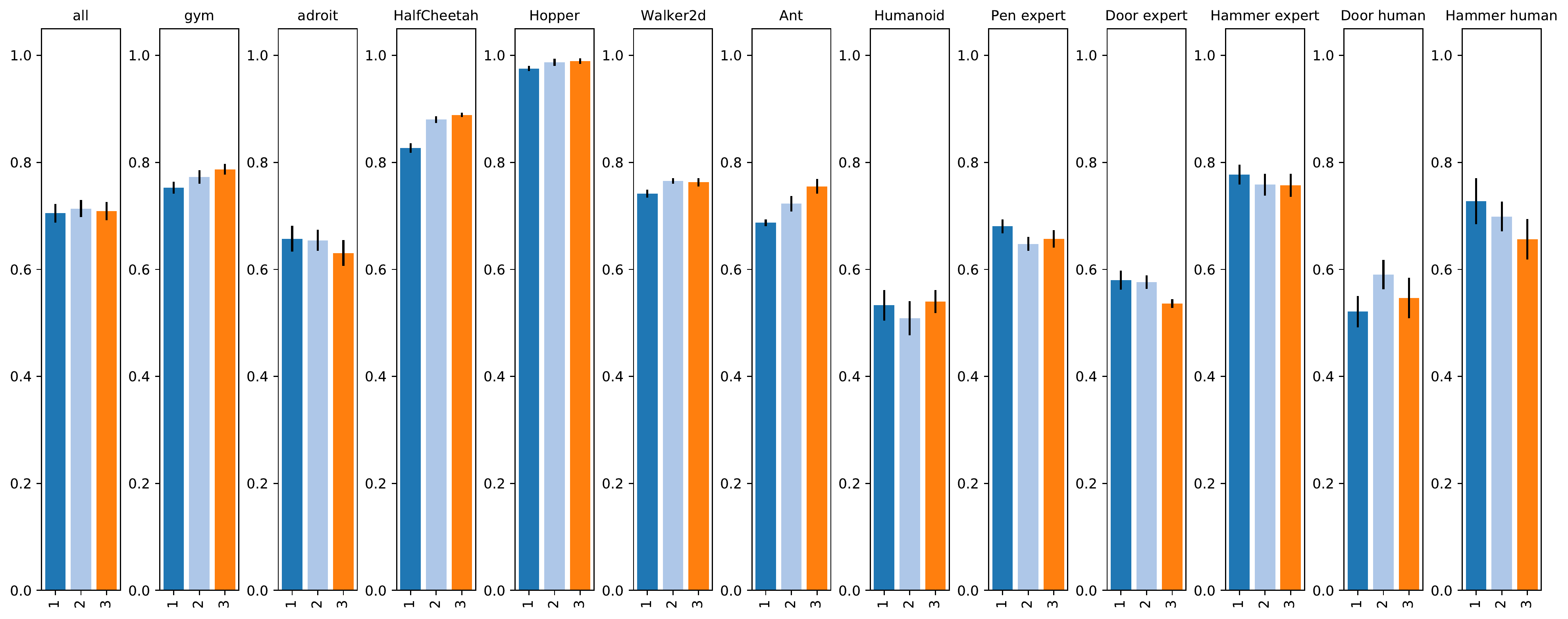}}
\centerline{\includegraphics[height=4.5cm,width=1\textwidth]{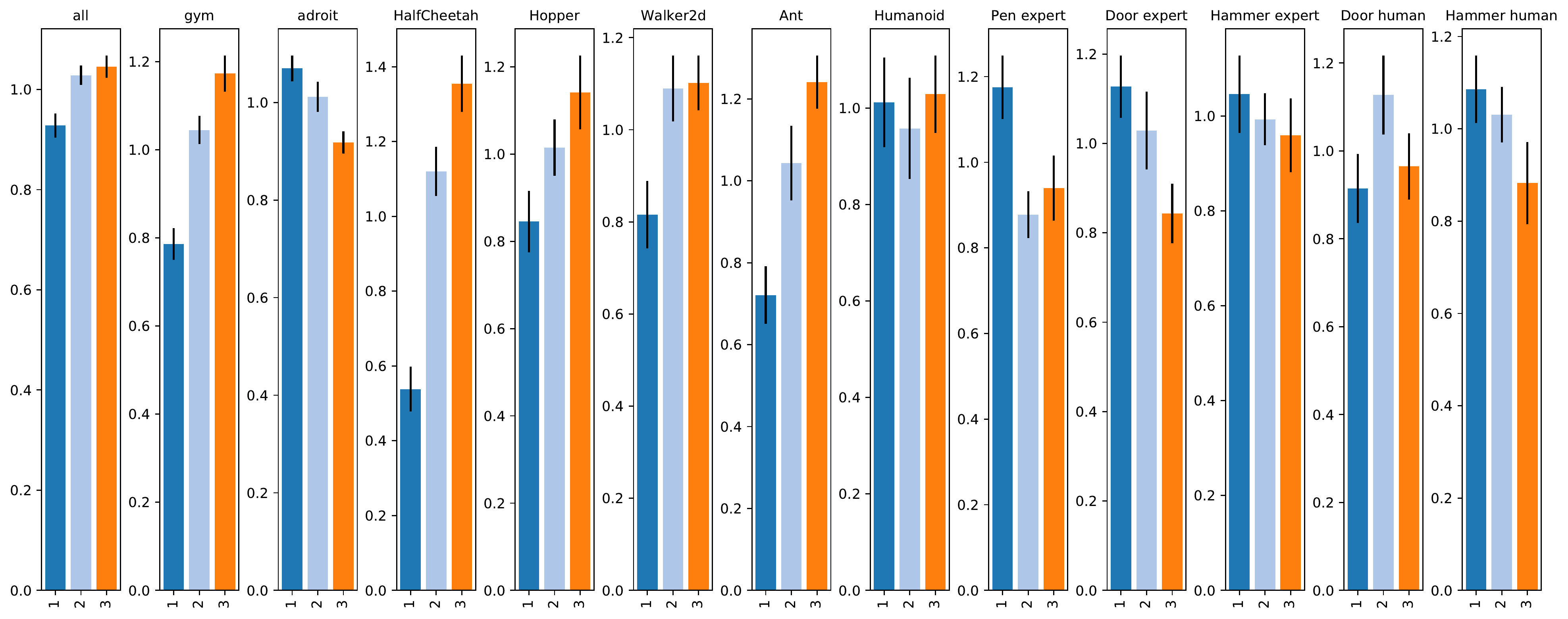}}
\caption{Analysis of choice \choicet{numpolicylayers}: 95th percentile of performance scores conditioned on choice (top) and distribution of choices in top 5\% of configurations (bottom).}
\label{fig:main_num_policy_layers}
\end{center}
\end{figure}

\begin{figure}[ht]
\begin{center}
\centerline{\includegraphics[height=4.5cm,width=1\textwidth]{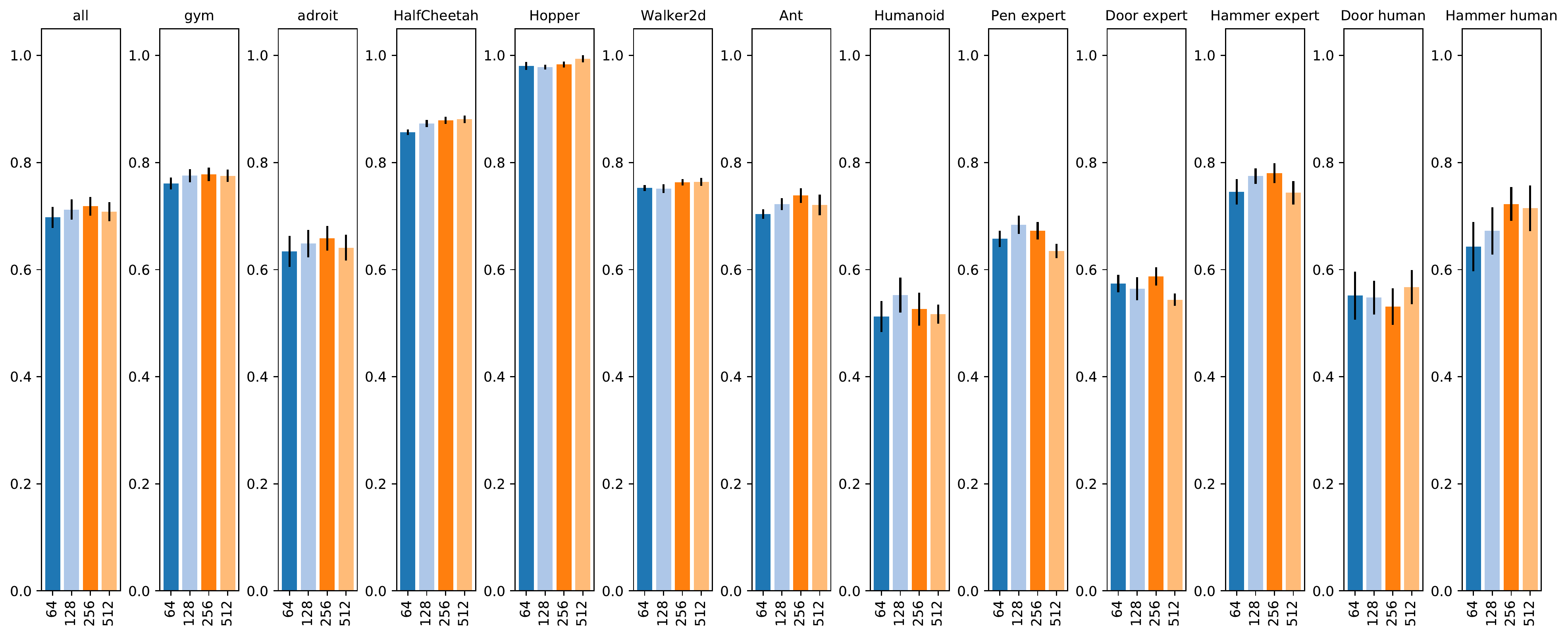}}
\centerline{\includegraphics[height=4.5cm,width=1\textwidth]{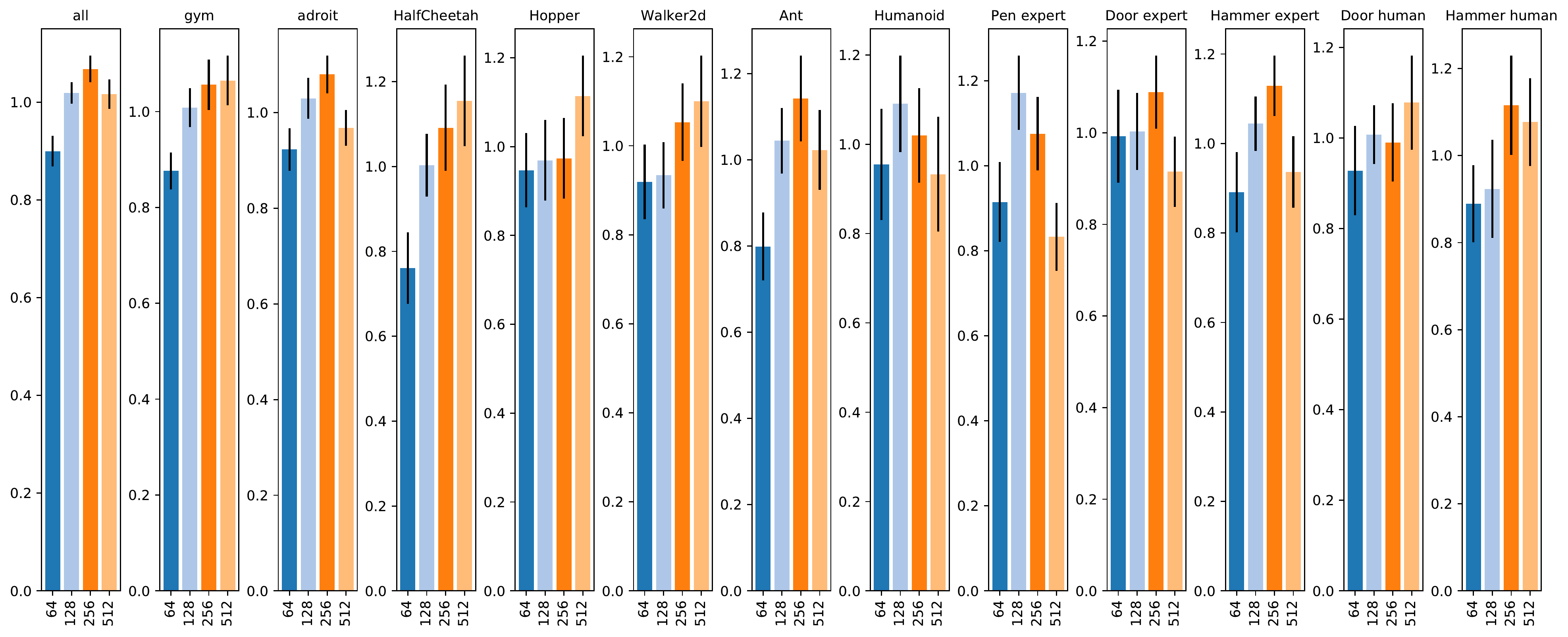}}
\caption{Analysis of choice \choicet{policylayersize}: 95th percentile of performance scores conditioned on choice (top) and distribution of choices in top 5\% of configurations (bottom).}
\label{fig:main_policy_layer_size}
\end{center}
\end{figure}

\begin{figure}[ht]
\begin{center}
\centerline{\includegraphics[height=4.5cm,width=1\textwidth]{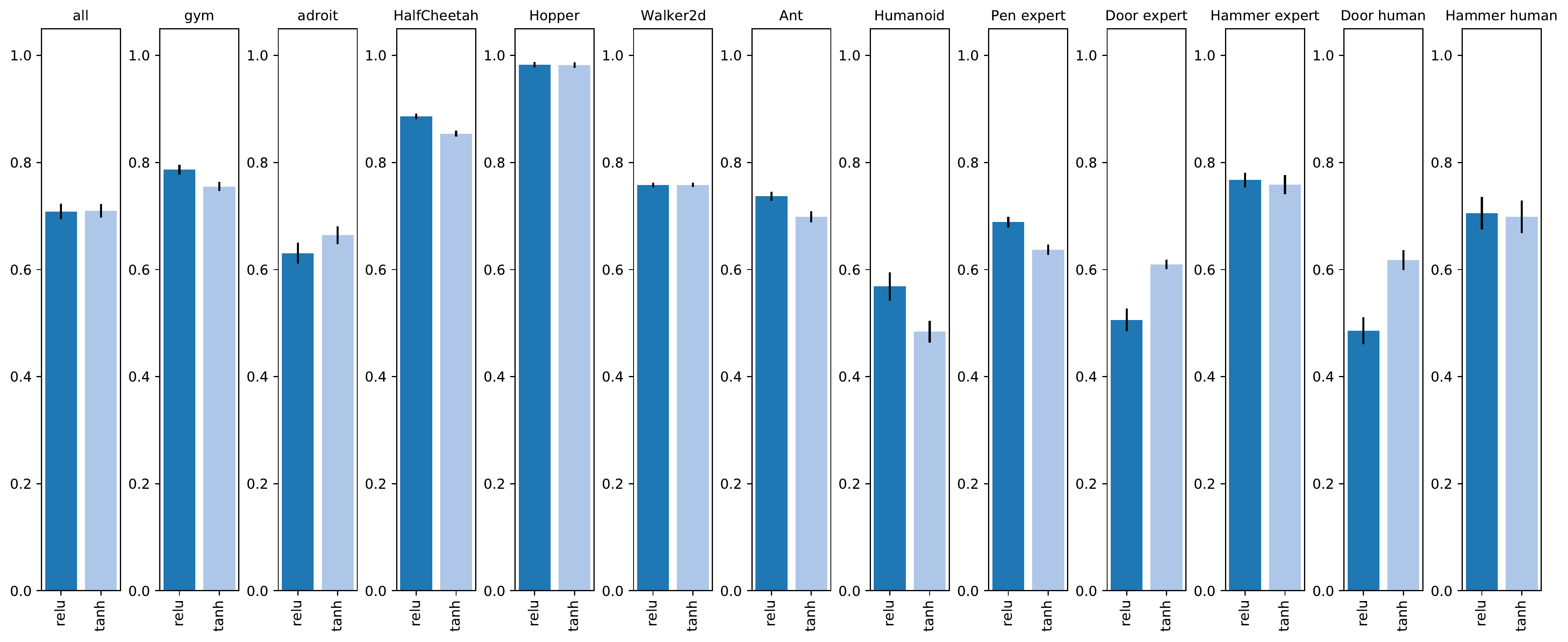}}
\centerline{\includegraphics[height=4.5cm,width=1\textwidth]{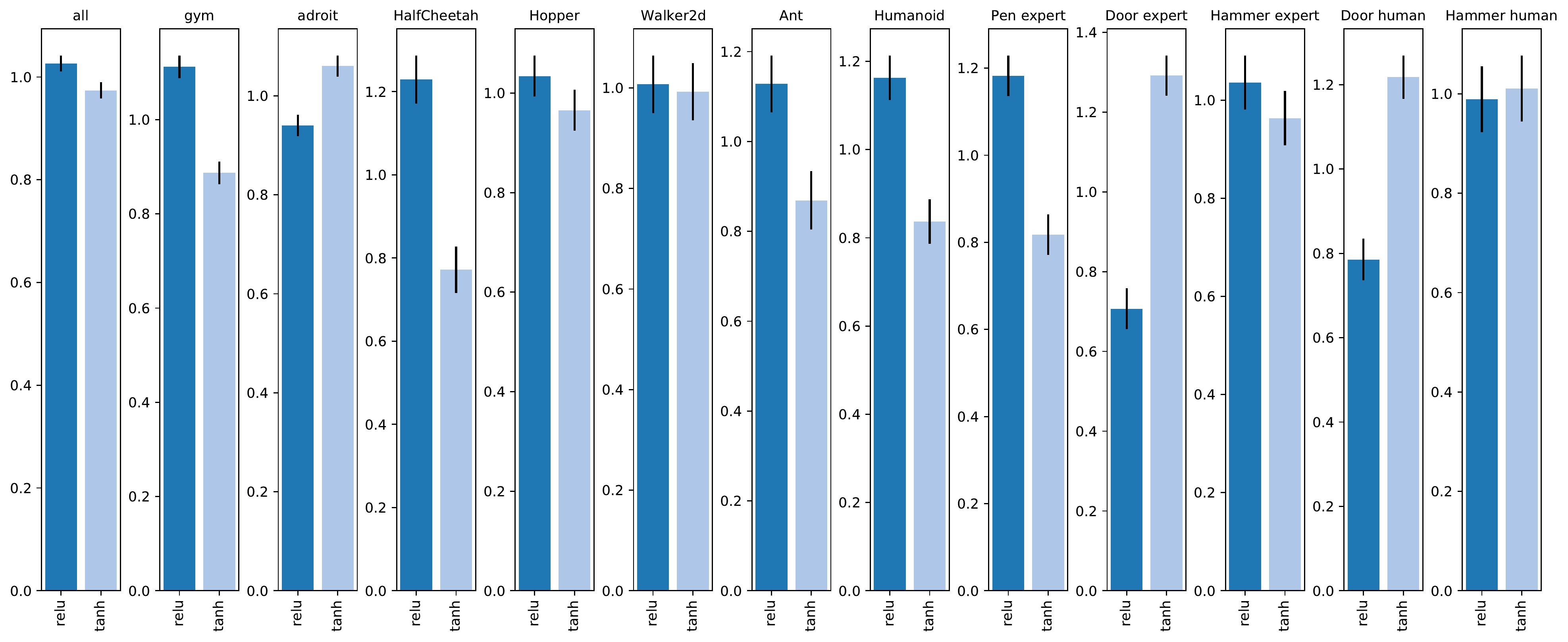}}
\caption{Analysis of choice \choicet{activation}: 95th percentile of performance scores conditioned on choice (top) and distribution of choices in top 5\% of configurations (bottom).}
\label{fig:main_activation}
\end{center}
\end{figure}

\begin{figure}[ht]
\begin{center}
\centerline{\includegraphics[height=4.5cm,width=1\textwidth]{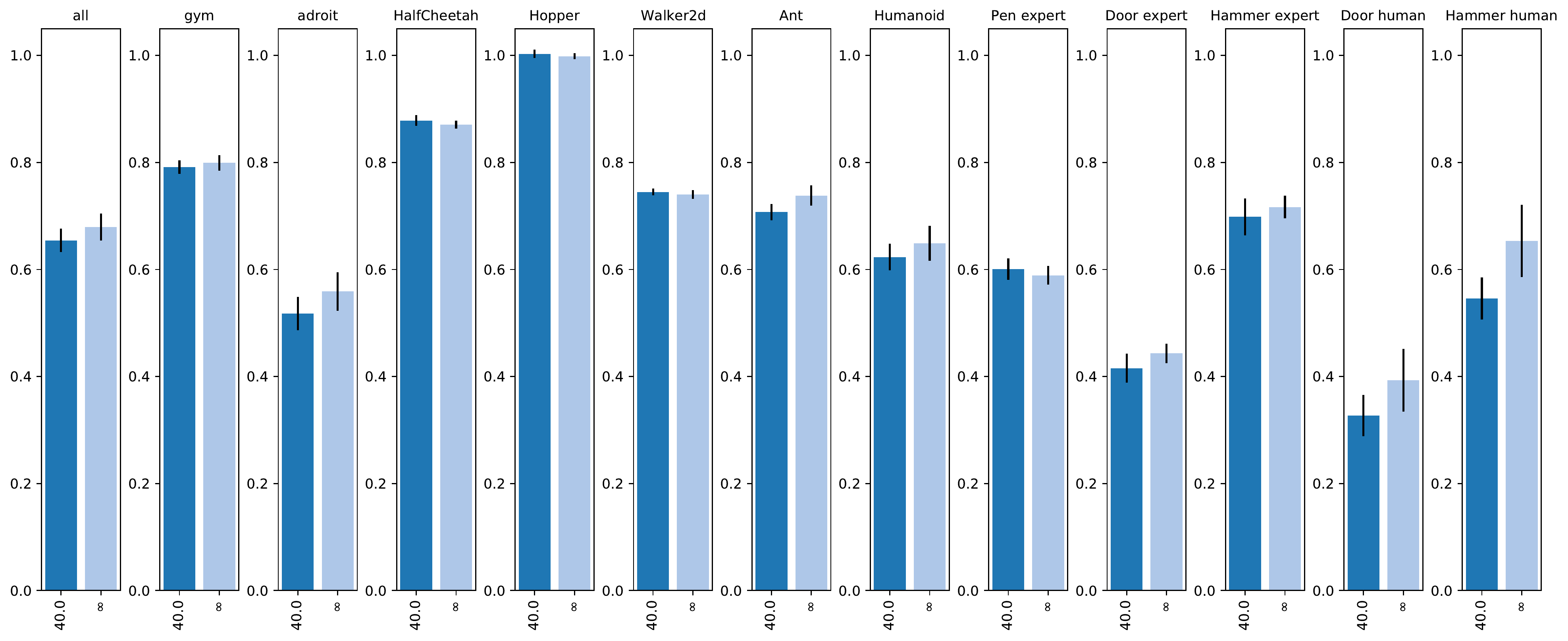}}
\centerline{\includegraphics[height=4.5cm,width=1\textwidth]{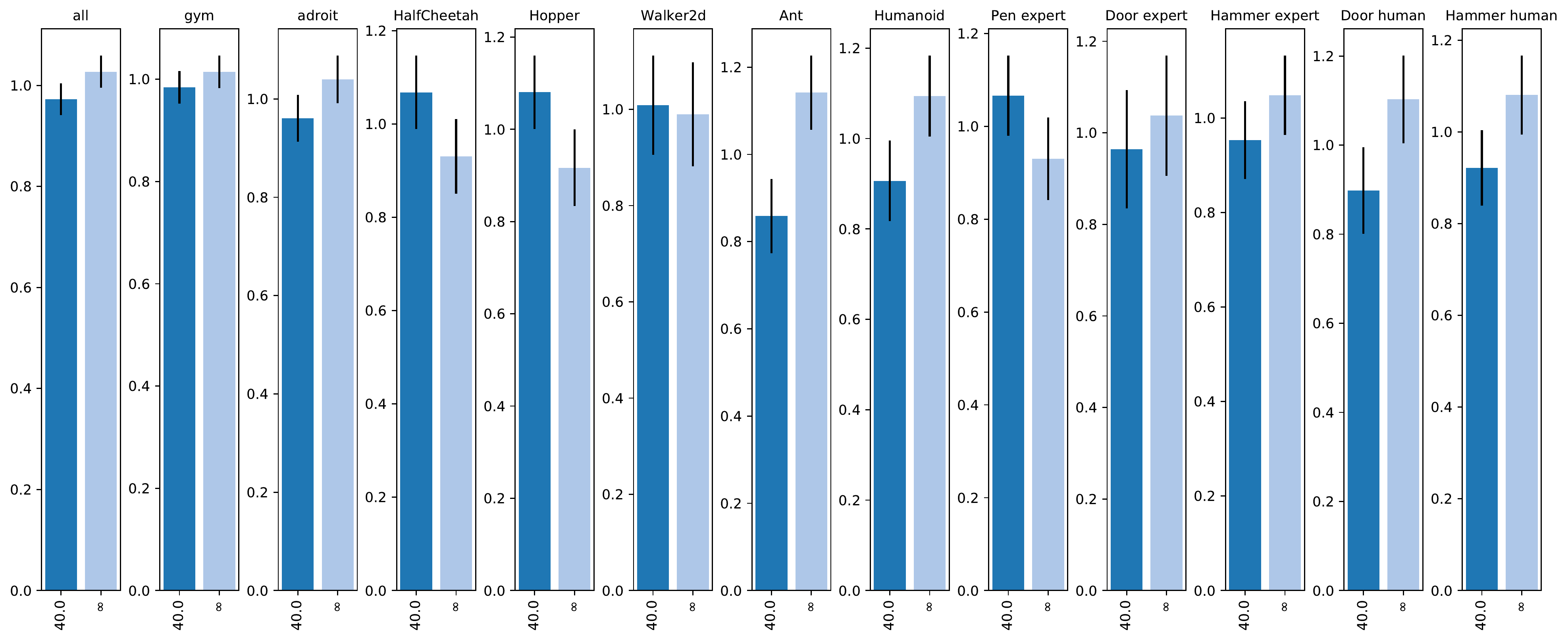}}
\caption{Analysis of choice \choicet{tdtgradientclipping}: 95th percentile of performance scores conditioned on choice (top) and distribution of choices in top 5\% of configurations (bottom).}
\label{fig:main_td3_gradient_clipping}
\end{center}
\end{figure}

\begin{figure}[ht]
\begin{center}
\centerline{\includegraphics[height=4.5cm,width=1\textwidth]{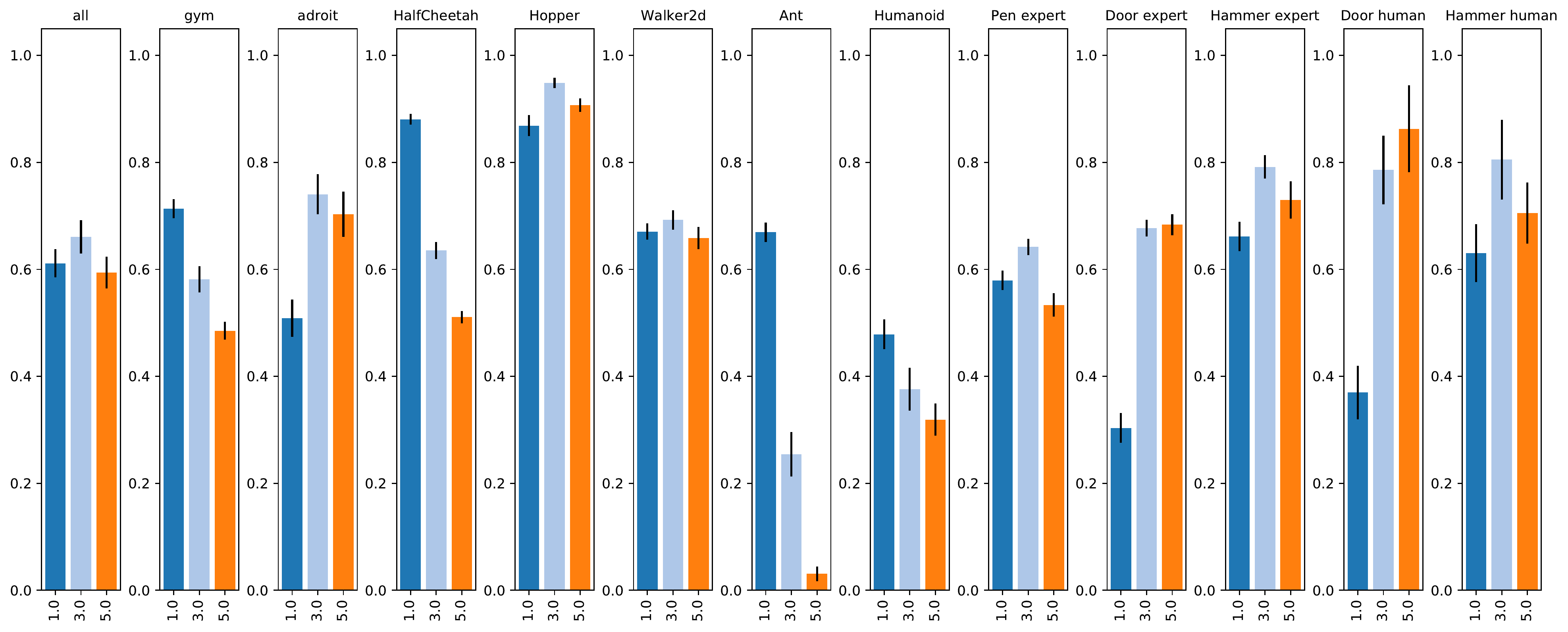}}
\centerline{\includegraphics[height=4.5cm,width=1\textwidth]{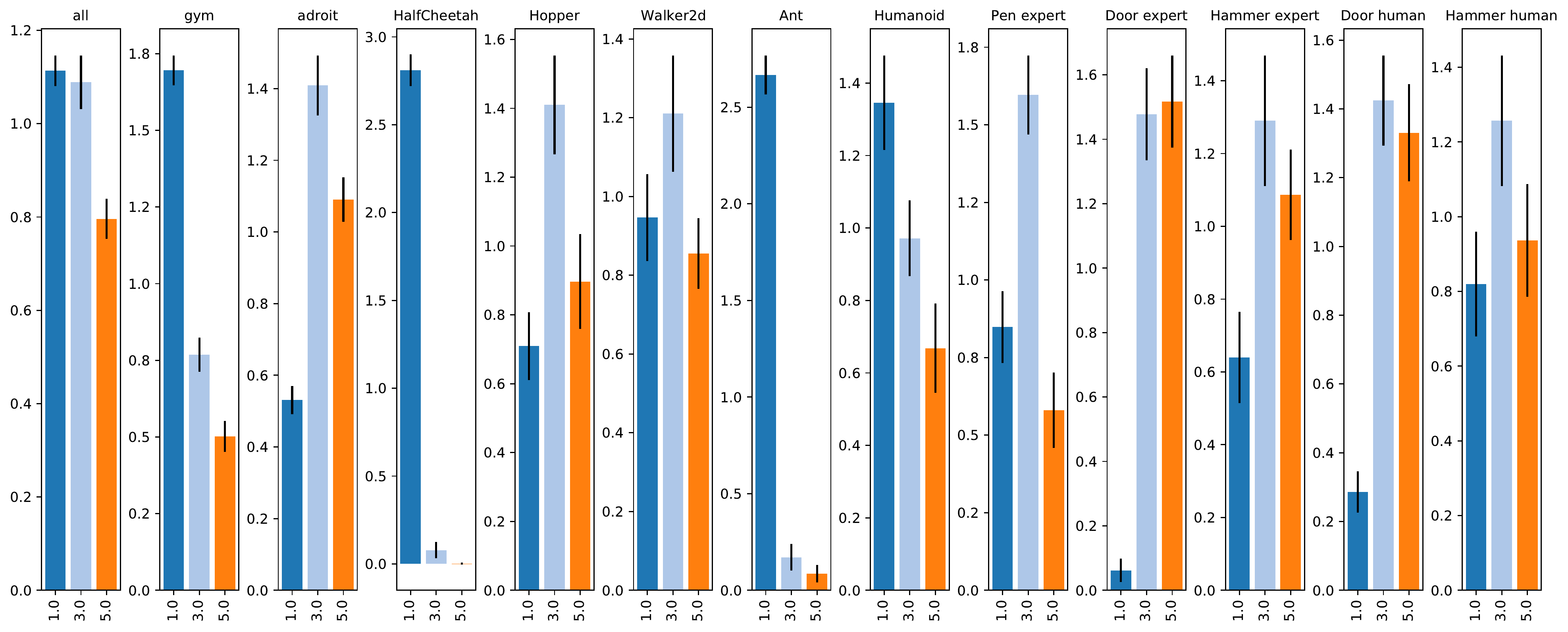}}
\caption{Analysis of choice \choicet{nstep}: 95th percentile of performance scores conditioned on choice (top) and distribution of choices in top 5\% of configurations (bottom).}
\label{fig:main_n_step}
\end{center}
\end{figure}


\begin{figure}[ht]
\begin{center}
\centerline{\includegraphics[height=4.5cm,width=1\textwidth]{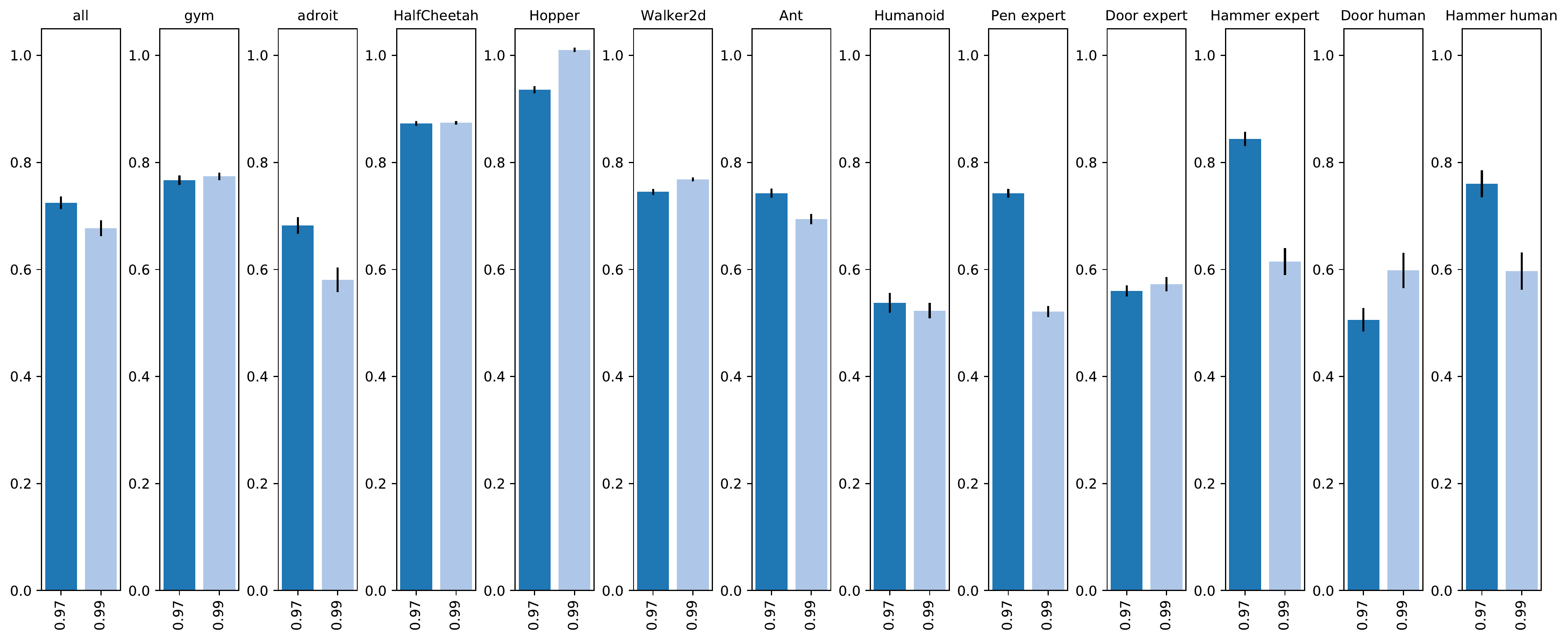}}
\centerline{\includegraphics[height=4.5cm,width=1\textwidth]{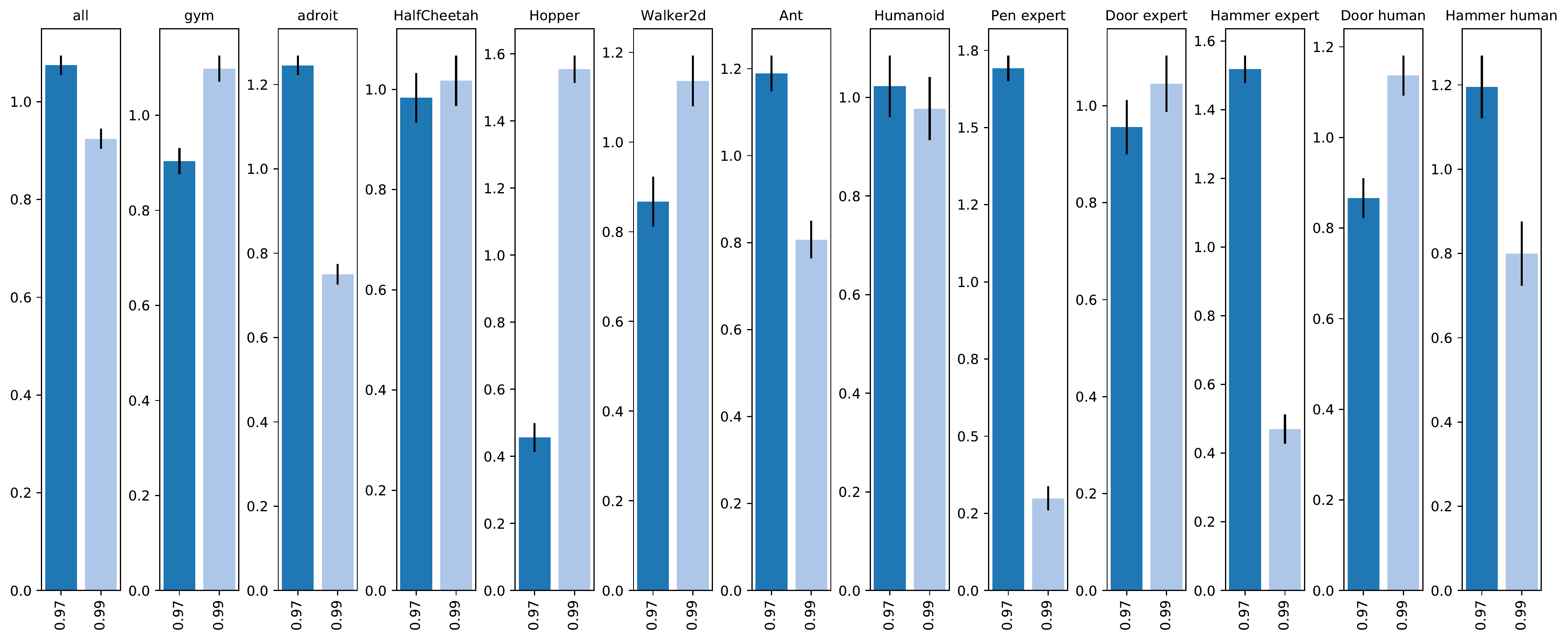}}
\caption{Analysis of choice \choicet{discount}: 95th percentile of performance scores conditioned on choice (top) and distribution of choices in top 5\% of configurations (bottom).}
\label{fig:main_discount}
\end{center}
\end{figure}

\clearpage

\begin{figure}[ht]
\begin{center}
\centerline{\includegraphics[width=\textwidth]{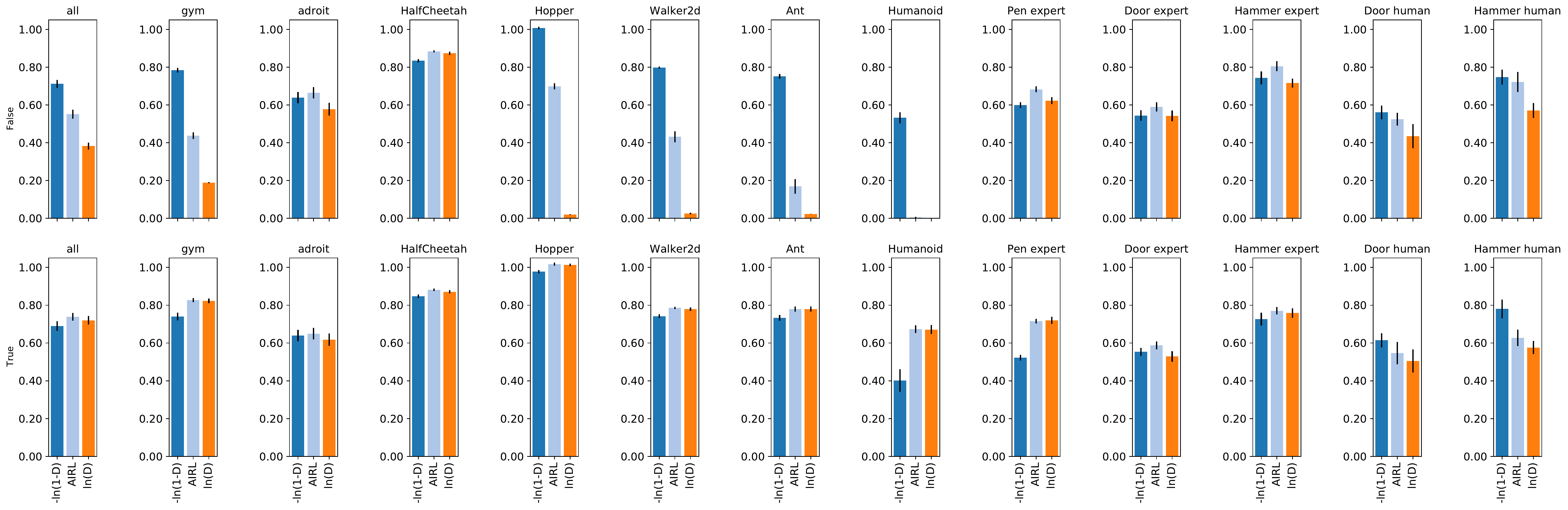}}
\caption{95th percentile of performance scores conditioned on \choicet{explicitabsorbingstate}(rows) and \choicet{gailreward}(bars).}
\label{fig:corr_abs_reward}
\end{center}
\end{figure}

\begin{figure}[ht]
\begin{center}
\centerline{\includegraphics[width=\textwidth]{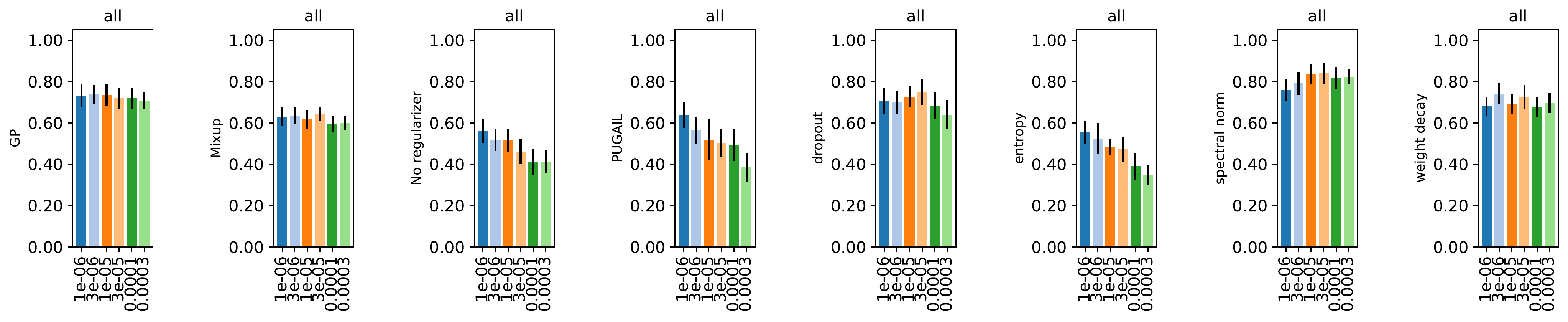}}
\caption{95th percentile of performance scores conditioned on \choicet{regularizer}(subplots) and \choicet{gaildiscriminatorlearningrate}(bars).}
\label{fig:corr_reg_lr_main}
\end{center}
\end{figure}

\begin{figure}[ht]
\begin{center}
\centerline{\includegraphics[width=\textwidth]{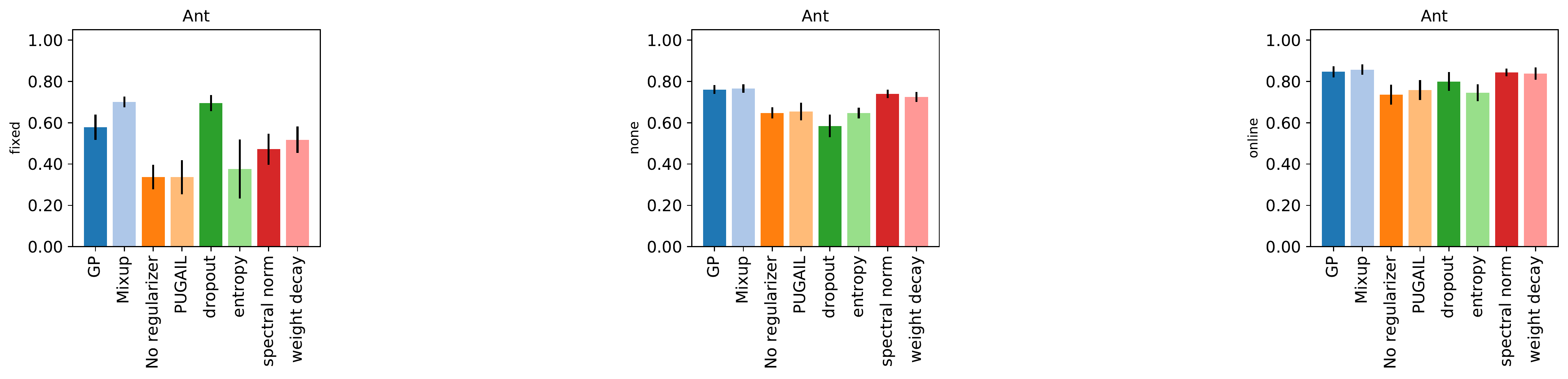}}
\caption{95th percentile of performance scores conditioned on \choicet{obsnormalization}(subplots) and \choicet{regularizer}(bars)
in the \texttt{Ant} environment.}
\label{fig:corr_reg_obs_ant}
\end{center}
\end{figure}
\clearpage
\section{Experiment trade-offs}
\label{exp_tradeoffs}
\subsection{Design}
\label{exp_design_tradeoffs}
For each of the 10 tasks, we sampled 7991 choice configurations where we sampled the following choices independently and uniformly from the following ranges:
\begin{itemize}
    \item \choicet{directrlalgorithm}: \{d4pg, sac, td3\}
    \begin{itemize}
        \item For the case ``\choicet{directrlalgorithm} = sac'', we further sampled the sub-choices:
        \begin{itemize}
            \item \choicet{saclearningrate}: \{0.0001, 0.0003, 0.001\}
            \item \choicet{sactargetentropyperdimension}: \{-2.0, -1.0, -0.5, 0.0\}
            \item \choicet{sactau}: \{0.001, 0.003, 0.01, 0.03\}
        \end{itemize}
        \item For the case ``\choicet{directrlalgorithm} = d4pg'', we further sampled the sub-choices:
        \begin{itemize}
            \item \choicet{dfpglearningrate}: \{3e-05, 0.0001, 0.0003\}
            \item \choicet{rlsigma}: \{0.1, 0.2, 0.3, 0.5\}
            \item \choicet{vmax}: \{150.0, 750.0, 1500.0\}
            \item \choicet{numatoms}: \{51.0, 101.0, 201.0, 401.0\}
            \item \choicet{nstep}: \{1.0, 3.0, 5.0\}
        \end{itemize}
        \item For the case ``\choicet{directrlalgorithm} = td3'', we further sampled the sub-choices:
        \begin{itemize}
            \item \choicet{tdtpolicylearningrate}: \{0.0001, 0.0003, 0.001\}
            \item \choicet{tdtcriticlearningrate}: \{0.0001, 0.0003, 0.001\}
            \item \choicet{tdtgradientclipping}: \{40.0, $\infty$\}
            \item \choicet{rlsigma}: \{0.1, 0.2, 0.3, 0.5\}
        \end{itemize}
    \end{itemize}
    \item \choicet{maxreplaysize}: \{300000, 1000000, 3000000\}
    \item \choicet{numpolicylayers}: \{1, 2, 3\}
    \item \choicet{policylayersize}: \{64, 128, 256, 512\}
    \item \choicet{numcriticlayers}: \{2, 3\}
    \item \choicet{criticlayersize}: \{256, 512\}
    \item \choicet{activation}: \{relu, tanh\}
    \item \choicet{discount}: \{0.97, 0.99\}
    \item \choicet{pretrainwithbc}: \{False, True\}
    \item \choicet{explicitabsorbingstate}: \{False, True\}
    \item \choicet{gailmaxreplaysize}: \{300000, 1000000, 3000000\}
    \item \choicet{gaildiscriminatormodule}: \{False, True\}
    \item \choicet{gailinput}: \{s, sa, sas, ss\}
    \item \choicet{gailmlpnumlayers}: \{1, 2, 3\}
    \item \choicet{gailmlpnumwidth}: \{16, 32, 64, 128, 256, 512\}
    \item \choicet{gailmlpactivation}: \{elu, leaky\_relu, relu, sigmoid, swish, tanh\}
    \item \choicet{gailmlplastlayerinitscale}: \{0.001, 1.0\}
    \item \choicet{regularizer}: \{GP, Mixup, No regularizer, PUGAIL, dropout, entropy, spectral norm, weight decay\}
    \begin{itemize}
        \item For the case ``\choicet{regularizer} = GP'', we further sampled the sub-choices:
        \begin{itemize}
            \item \choicet{gpcoef}: \{0.1, 1.0, 10.0\}
            \item \choicet{gptarget}: \{0.0, 1.0\}
        \end{itemize}
        \item For the case ``\choicet{regularizer} = Mixup'', we further sampled the sub-choices:
        \begin{itemize}
            \item \choicet{mixupalpha}: \{0.1, 0.4, 1.0\}
        \end{itemize}
        \item For the case ``\choicet{regularizer} = PUGAIL'', we further sampled the sub-choices:
        \begin{itemize}
            \item \choicet{pugailpositiveclassprior}: \{0.25, 0.5, 0.7\}
            \item \choicet{pugailbeta}: \{0.0, 0.7, $\infty$\}
        \end{itemize}
        \item For the case ``\choicet{regularizer} = entropy'', we further sampled the sub-choices:
        \begin{itemize}
            \item \choicet{regentropycoef}: \{0.0003, 0.001, 0.003, 0.01, 0.03, 0.1, 0.3\}
        \end{itemize}
        \item For the case ``\choicet{regularizer} = weight decay'', we further sampled the sub-choices:
        \begin{itemize}
            \item \choicet{regweightdecay}: \{0.3, 1.0, 3.0, 10.0, 30.0\}
        \end{itemize}
        \item For the case ``\choicet{regularizer} = dropout'', we further sampled the sub-choices:
        \begin{itemize}
            \item \choicet{dropoutinputrate}: \{0.0, 0.25, 0.5, 0.75\}
            \item \choicet{dropouthiddenrate}: \{0.25, 0.5, 0.75\}
        \end{itemize}
    \end{itemize}
    \item \choicet{obsnormalization}: \{fixed, none\}
    \item \choicet{evalbehaviorpolicytype}: \{average, mode, stochastic\}
    \item \choicet{gaildiscriminatorlearningrate}: \{1e-06, 3e-06, 1e-05, 3e-05, 0.0001, 0.0003\}
    \item \choicet{samplesperinsert}: \{64, 128, 256, 512, 1024\}
    \item \choicet{batchsize}: \{64, 128, 256, 512, 1024\}
    \item \choicet{discriminatortorlupdatesratio}: \{1, 2\}
    \item \choicet{gradupdatesperbatch}: \{1, 2, 4, 8, 16, 32, 64\}
    \item \choicet{gailreward}: \{-ln(1-D), AIRL, ln(D)\}
\end{itemize}

\subsection{Results}
\label{exp_results_tradeoffs}
\begin{figure}[ht]
\begin{center}
\centerline{\includegraphics[width=1\textwidth]{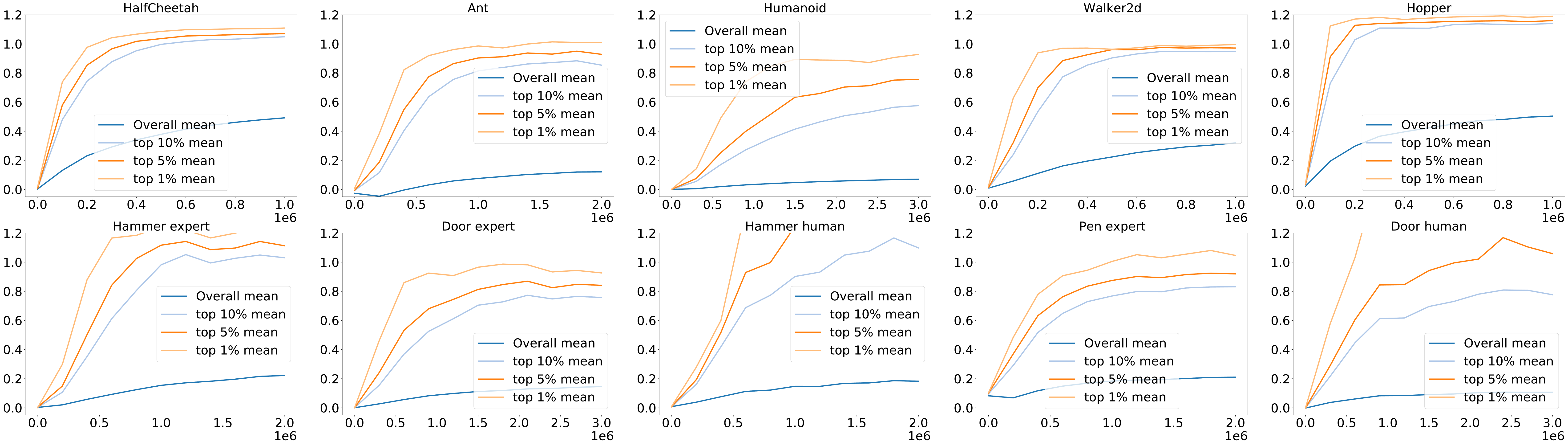}}
\caption{Training curves.}
\label{fig:tradeoffs_training_curves}
\end{center}
\end{figure}

\begin{figure}[ht]
\begin{center}
\centerline{\includegraphics[height=4.5cm,width=1\textwidth]{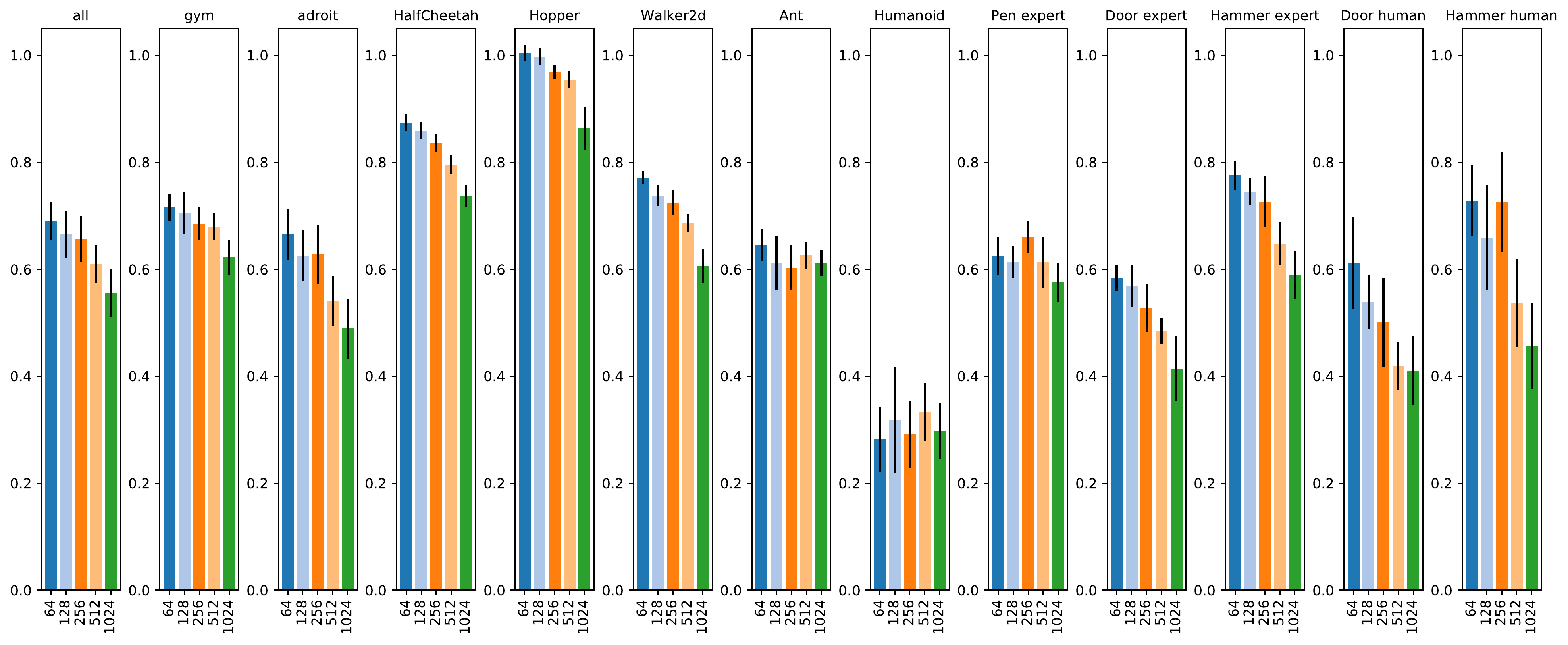}}
\centerline{\includegraphics[height=4.5cm,width=1\textwidth]{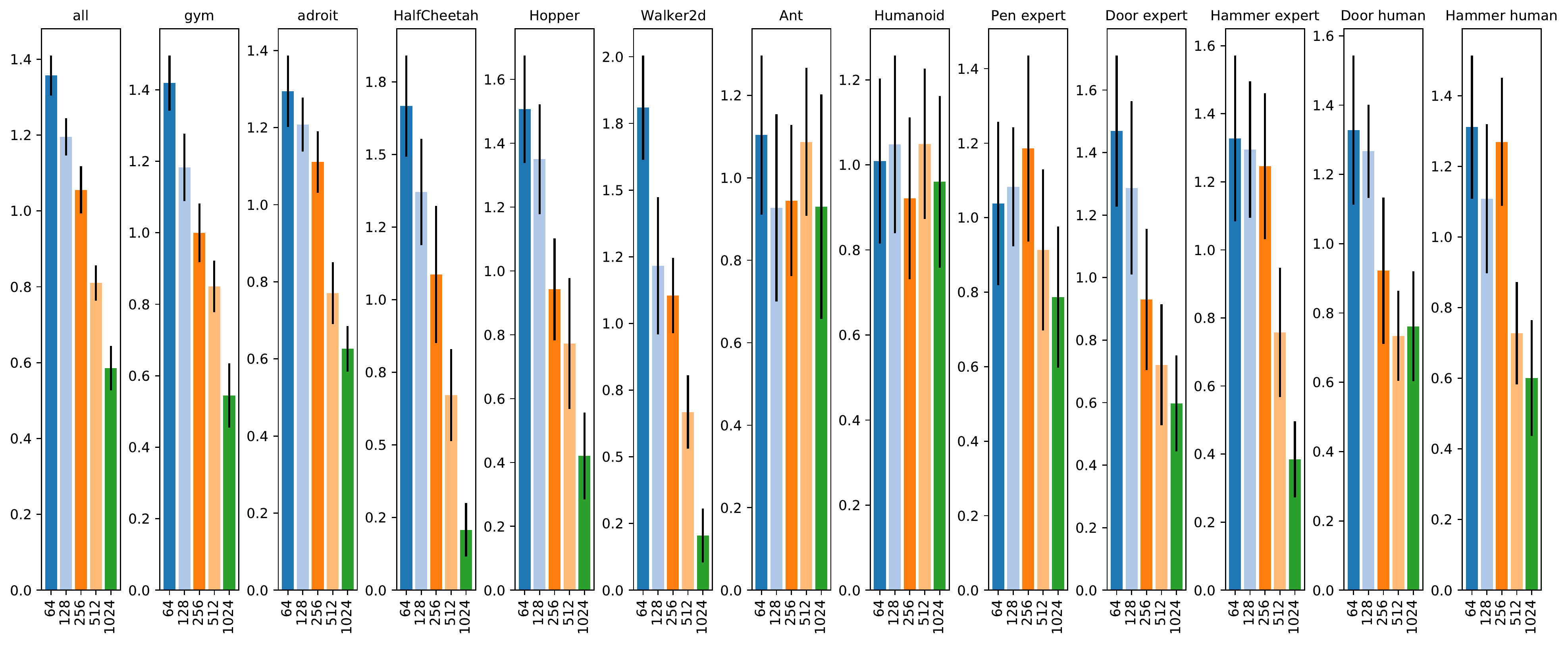}}
\caption{Analysis of choice \choicet{batchsize}: 95th percentile of performance scores conditioned on choice (top) and distribution of choices in top 5\% of configurations (bottom).}
\label{fig:tradeoffs_batch_size}
\end{center}
\end{figure}

\begin{figure}[ht]
\begin{center}
\centerline{\includegraphics[height=4.5cm,width=1\textwidth]{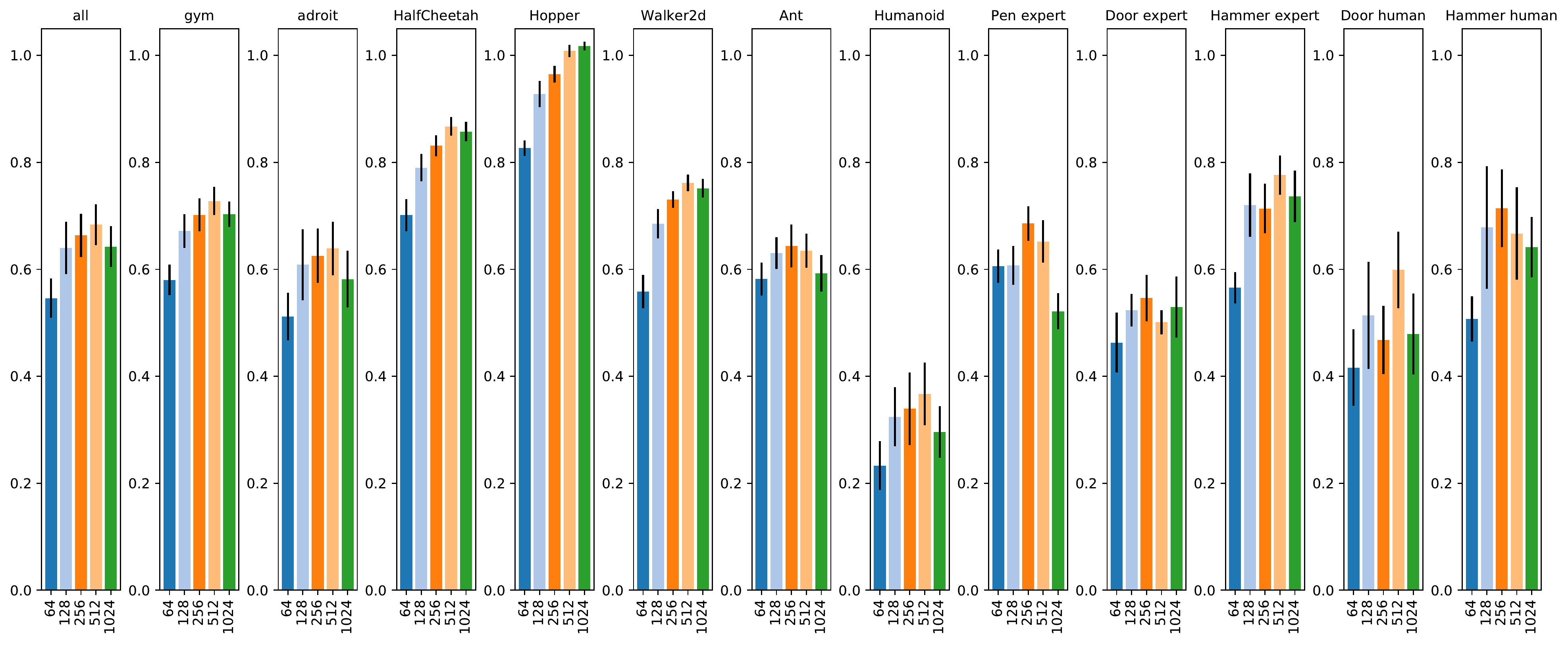}}
\centerline{\includegraphics[height=4.5cm,width=1\textwidth]{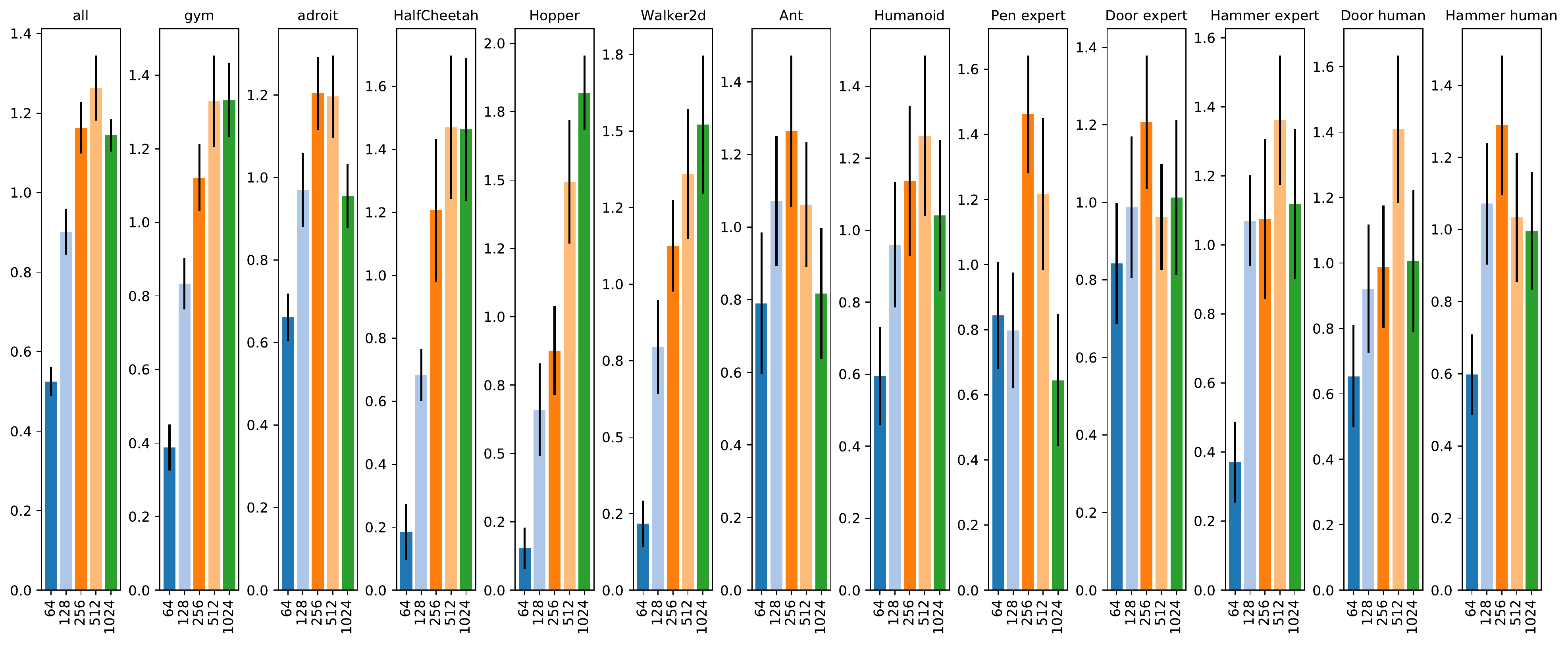}}
\caption{Analysis of choice \choicet{samplesperinsert}: 95th percentile of performance scores conditioned on choice (top) and distribution of choices in top 5\% of configurations (bottom).}
\label{fig:tradeoffs_samples_per_insert}
\end{center}
\end{figure}

\begin{figure}[ht]
\begin{center}
\centerline{\includegraphics[height=4.5cm,width=1\textwidth]{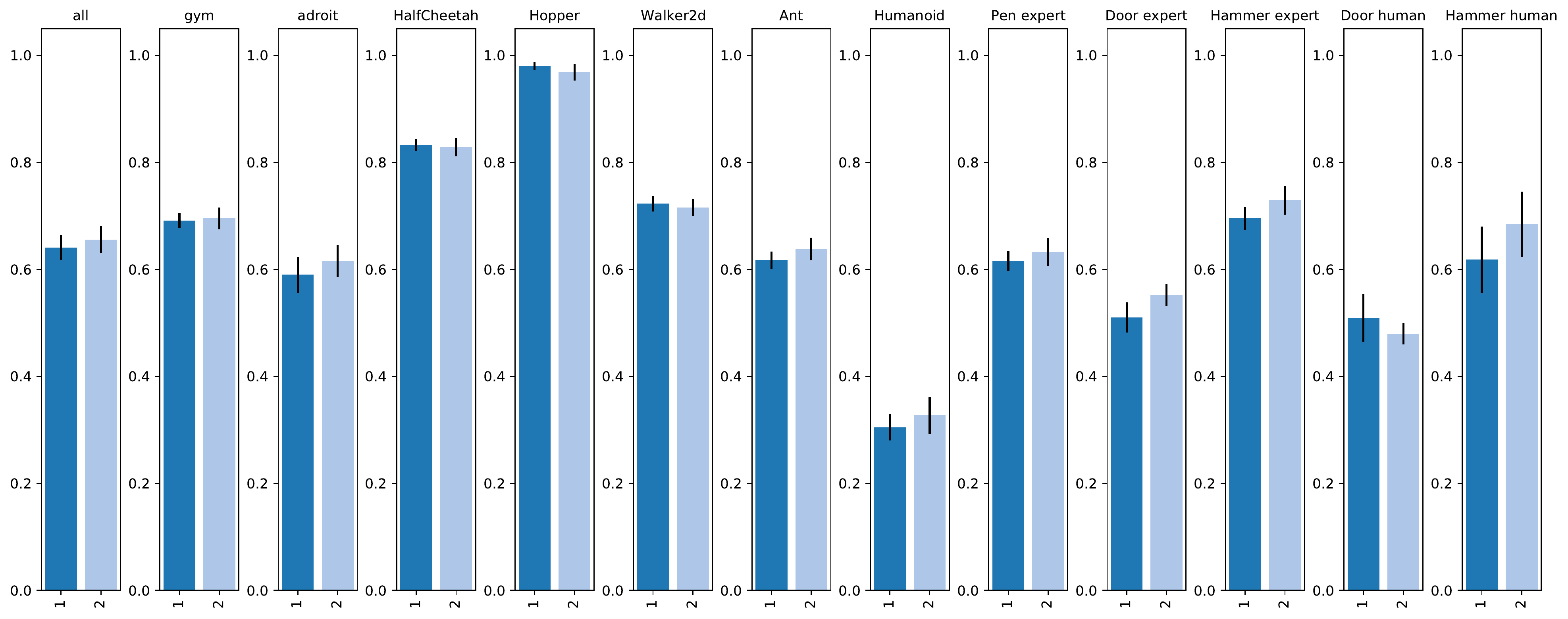}}
\centerline{\includegraphics[height=4.5cm,width=1\textwidth]{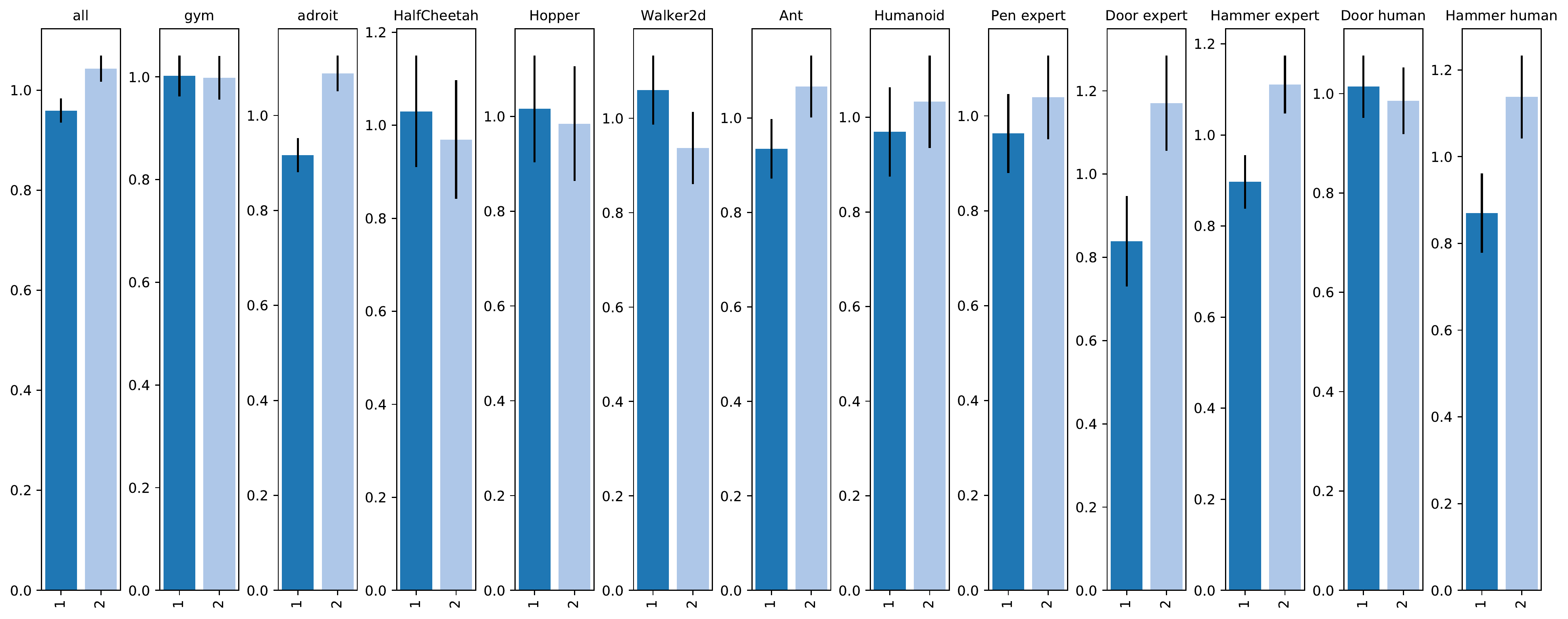}}
\caption{Analysis of choice \choicet{discriminatortorlupdatesratio}: 95th percentile of performance scores conditioned on choice (top) and distribution of choices in top 5\% of configurations (bottom).}
\label{fig:tradeoffs_discriminator_to_rl_updates_ratio}
\end{center}
\end{figure}

\begin{figure}[ht]
\begin{center}
\centerline{\includegraphics[height=4.5cm,width=1\textwidth]{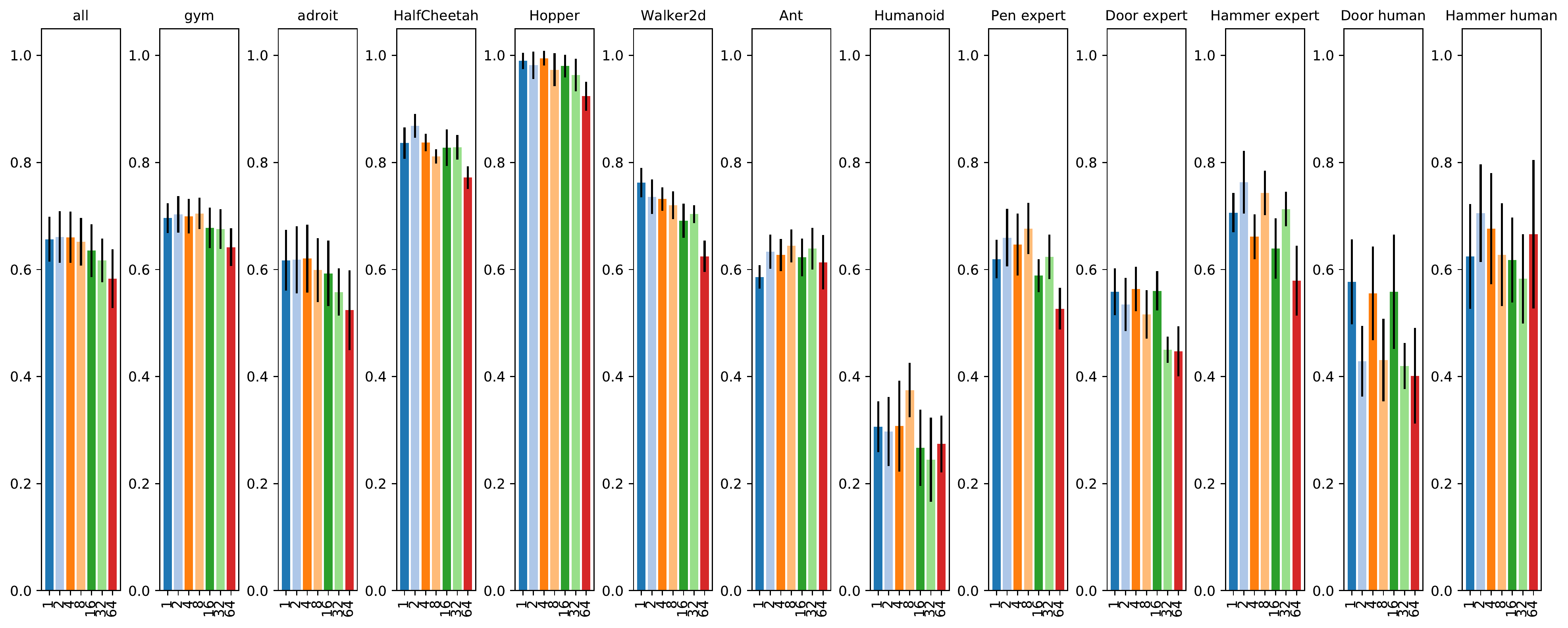}}
\centerline{\includegraphics[height=4.5cm,width=1\textwidth]{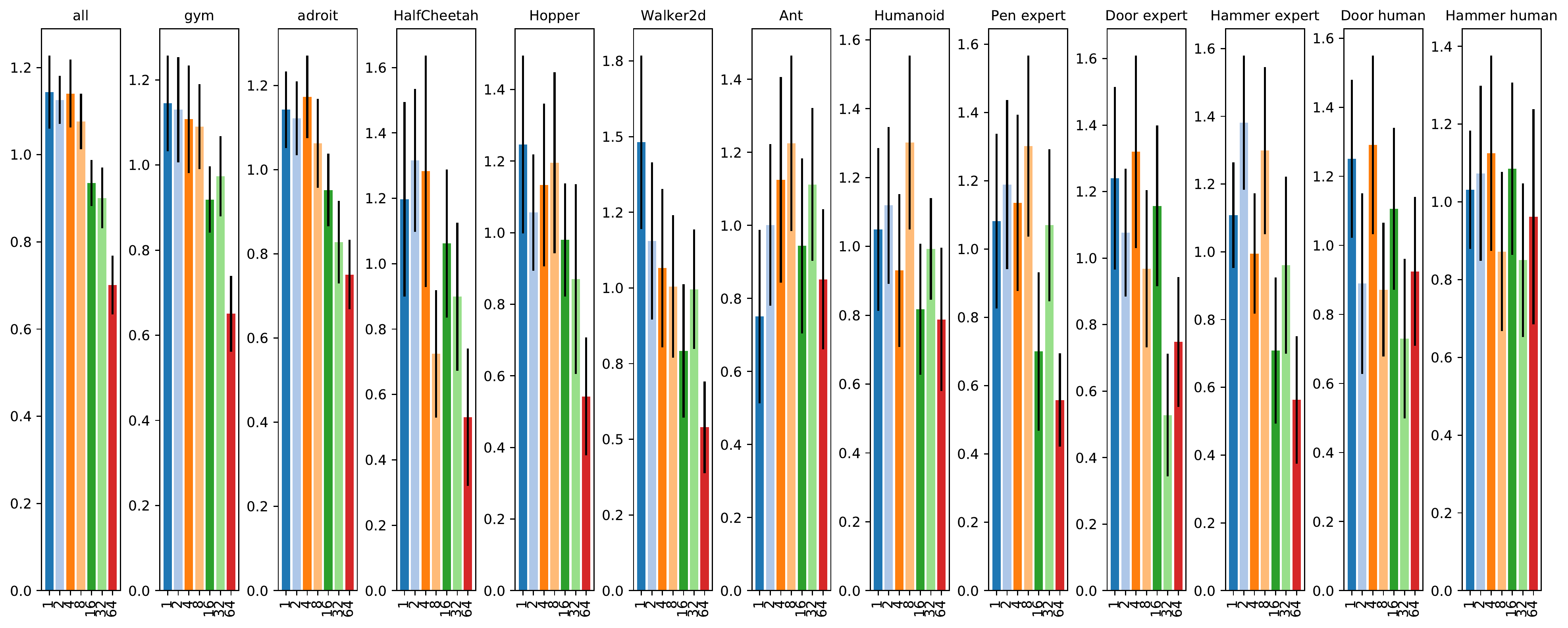}}
\caption{Analysis of choice \choicet{gradupdatesperbatch}: 95th percentile of performance scores conditioned on choice (top) and distribution of choices in top 5\% of configurations (bottom).}
\label{fig:tradeoffs_grad_updates_per_batch}
\end{center}
\end{figure}

For each of the sampled choice configurations we compute the performance metric as described in Section~\ref{sec:design}.
We report aggregate statistics of the experiment in Tables~\ref{tab:tradeoffs_overview}--\ref{tab:tradeoffs_overview4} as well as training curves in Figure~\ref{fig:tradeoffs_training_curves}.
We further provide per-choice analyses in Figures~\ref{fig:tradeoffs_batch_size}-\ref{fig:tradeoffs_grad_updates_per_batch}.

\begin{table}[ht]
\begin{center}
\caption{Quantiles of the \emph{final} agent performance across HP configurations for OpenAI Gym tasks.}
\label{tab:tradeoffs_overview}
\begin{tabular}{lrrrrr}
\toprule
{} &  Ant & HalfCheetah & Hopper & Humanoid & Walker2d \\
\midrule
90\% & 0.81 &        1.04 &   1.18 &     0.14 &     0.97 \\
95\% & 0.94 &        1.08 &   1.19 &     0.62 &     1.00 \\
99\% & 1.04 &        1.15 &   1.22 &     0.98 &     1.03 \\
Max & 1.15 &        1.41 &   1.31 &     1.05 &     1.16 \\
\bottomrule
\end{tabular}

\end{center}
\end{table}\begin{table}[ht]
\begin{center}
\caption{Quantiles of the \emph{final} agent performance across HP configurations for Adroit tasks.}
\label{tab:tradeoffs_overview2}
\begin{tabular}{lrrrrr}
\toprule
{} & Door expert & Door human & Hammer expert & Hammer human & Pen expert \\
\midrule
90\% &        0.71 &       0.25 &          1.03 &         0.45 &       0.70 \\
95\% &        0.89 &       0.71 &          1.25 &         1.19 &       0.86 \\
99\% &        1.04 &       2.12 &          1.36 &         2.95 &       1.07 \\
Max &        1.15 &       3.79 &          1.44 &         5.27 &       1.34 \\
\bottomrule
\end{tabular}

\end{center}
\end{table}\begin{table}[ht]
\begin{center}
\caption{Quantiles of the \emph{average} agent performance during training across HP configurations for OpenAI Gym tasks.}
\label{tab:tradeoffs_overview3}
\begin{tabular}{lrrrrr}
\toprule
{} &  Ant & HalfCheetah & Hopper & Humanoid & Walker2d \\
\midrule
90\% & 0.48 &        0.75 &   0.90 &     0.12 &     0.63 \\
95\% & 0.63 &        0.83 &   0.98 &     0.32 &     0.72 \\
99\% & 0.77 &        0.92 &   1.06 &     0.62 &     0.83 \\
Max & 0.89 &        1.00 &   1.10 &     0.85 &     0.92 \\
\bottomrule
\end{tabular}

\end{center}
\end{table}\begin{table}[ht]
\begin{center}
\caption{Quantiles of the \emph{average} agent performance during training across HP configurations for Adroit tasks.}
\label{tab:tradeoffs_overview4}
\begin{tabular}{lrrrrr}
\toprule
{} & Door expert & Door human & Hammer expert & Hammer human & Pen expert \\
\midrule
90\% &        0.38 &       0.26 &          0.54 &         0.39 &       0.50 \\
95\% &        0.53 &       0.49 &          0.71 &         0.65 &       0.63 \\
99\% &        0.74 &       1.02 &          0.91 &         1.21 &       0.82 \\
Max &        0.94 &       2.05 &          1.17 &         2.13 &       1.01 \\
\bottomrule
\end{tabular}

\end{center}
\end{table}

\clearpage

\begin{figure}[ht]
\begin{center}
\centerline{\includegraphics[width=\textwidth]{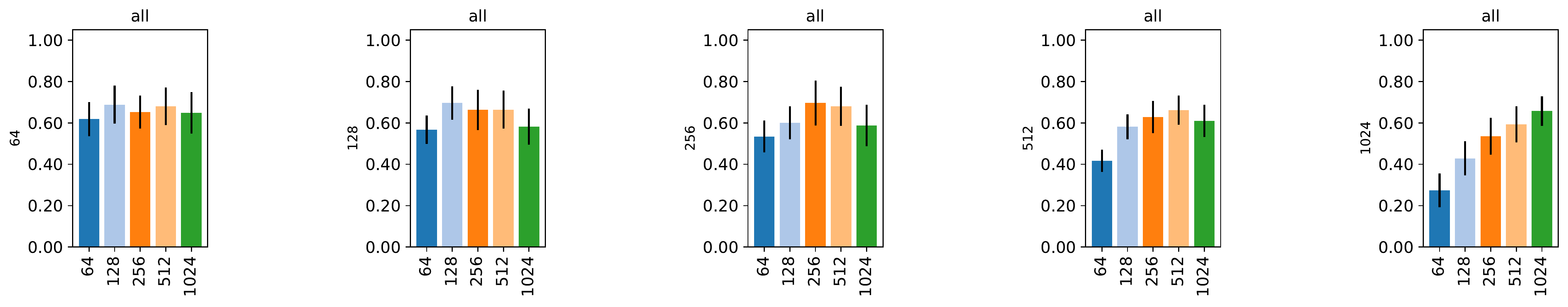}}
\caption{95th percentile of performance scores conditioned on \choicet{batchsize}(subplots) and \choicet{samplesperinsert}(bars).}
\label{fig:corr_batch_replay}
\end{center}
\end{figure}
\section{Additional experiments}\label{app:additional-plots}

\begin{figure}[ht]
\begin{center}
\centerline{\includegraphics[width=\textwidth]{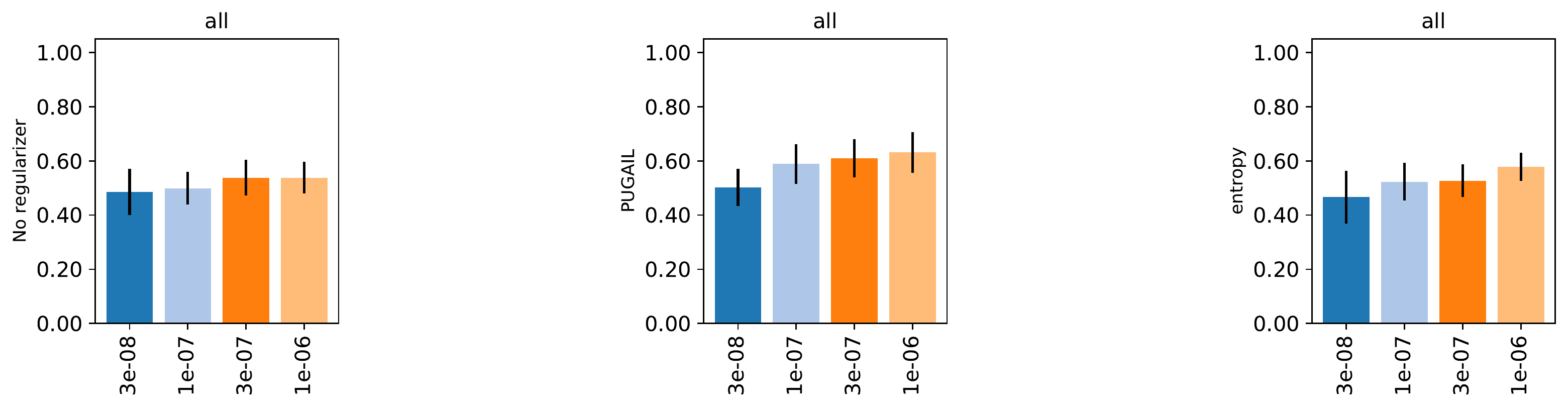}}
\caption{95th percentile of performance scores conditioned on \choicet{regularizer}(rows) and \choicet{gaildiscriminatorlearningrate}(bars).
The data comes from an experiment similar to the main one
but with smaller values of \choicet{gaildiscriminatorlearningrate}.}
\label{fig:corr_reg_lr_low}
\end{center}
\end{figure}

\begin{figure}[ht]
\begin{center}
\centerline{\includegraphics[width=\textwidth]{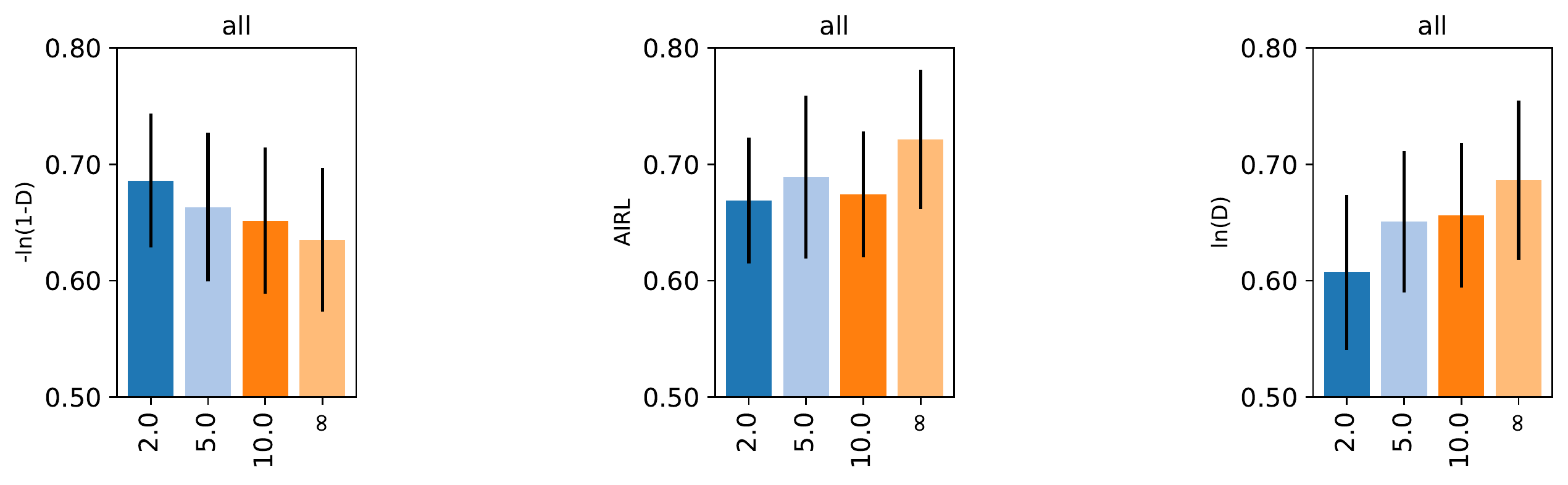}}
\caption{95th percentile of performance scores conditioned on \choicet{gailreward}(subplots) and \choicet{gailmaxrewardmagnitude}(bars). The data comes from an experiment similar to the main one
but with \choicet{gailmaxrewardmagnitude} swept.}
\label{fig:extra_reward_clipping}
\end{center}
\end{figure}

\begin{figure}[ht]
\begin{center}
\centerline{\includegraphics[height=4.5cm,width=1\textwidth]{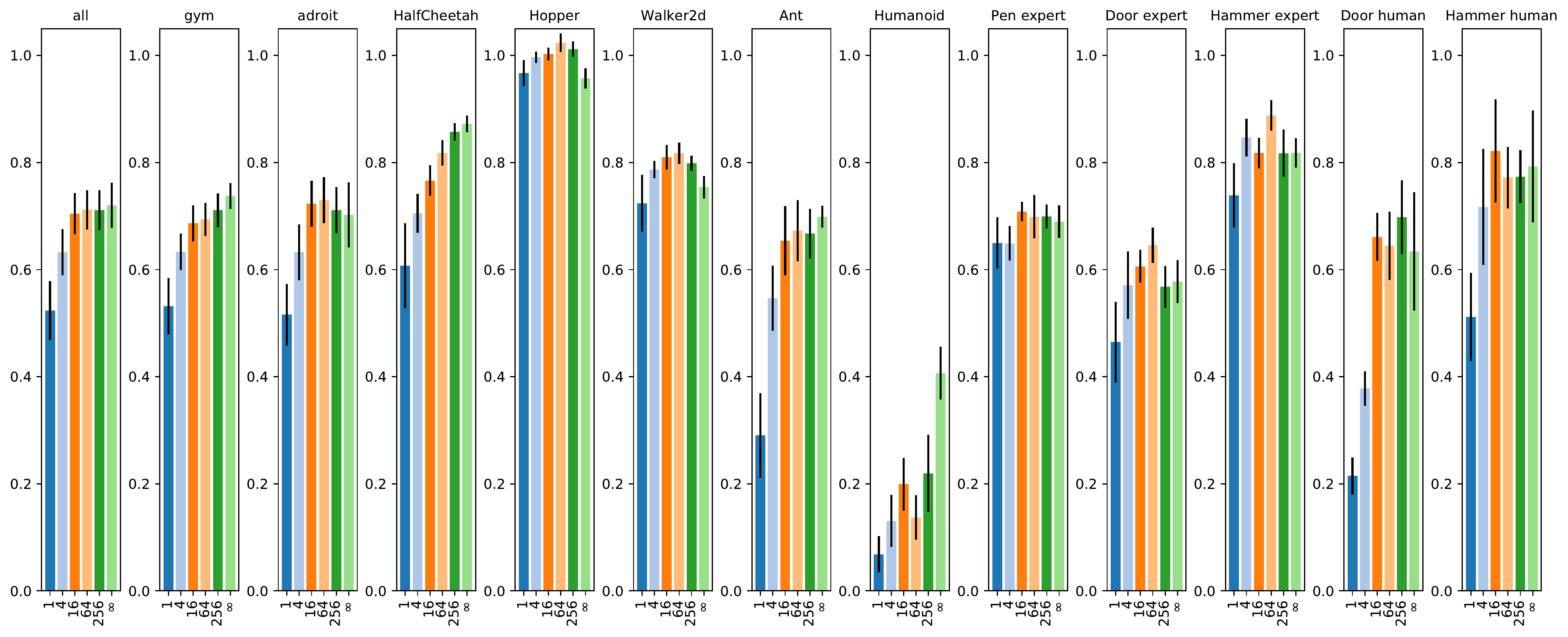}}
\centerline{\includegraphics[height=4.5cm,width=1\textwidth]{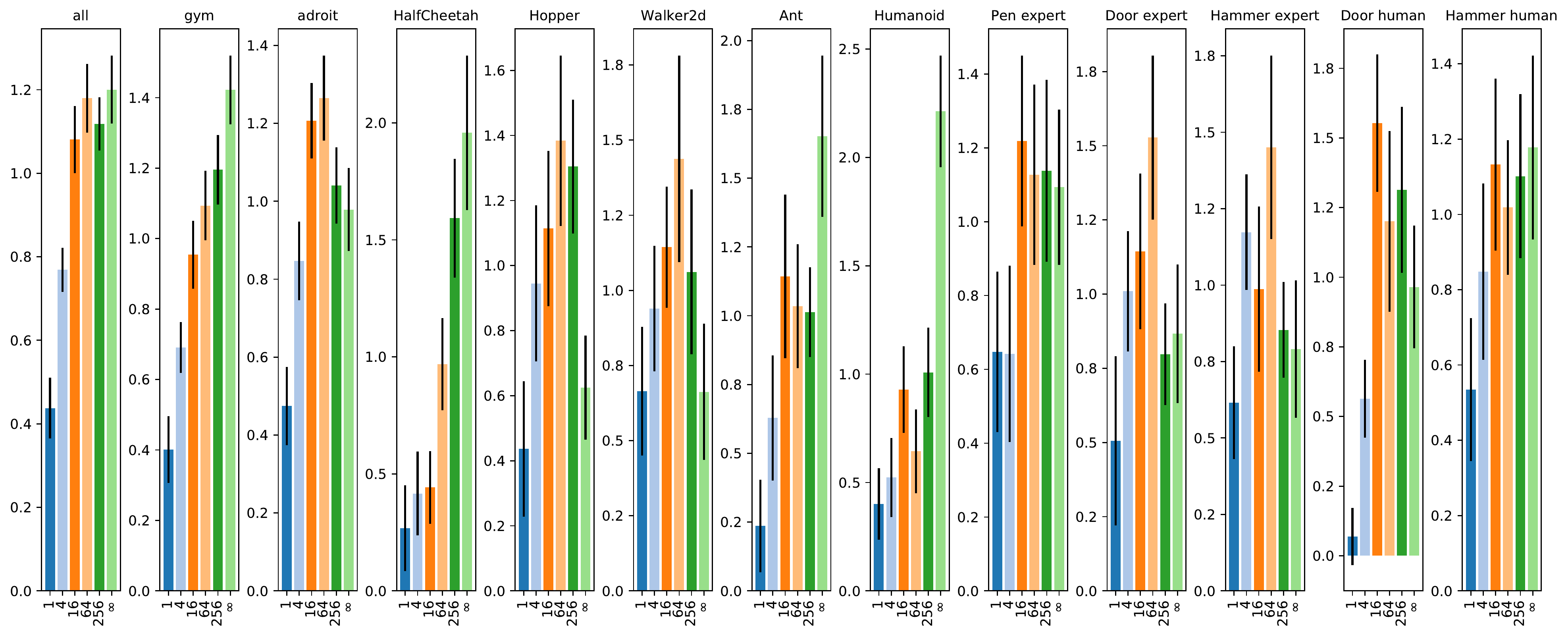}}
\caption{Analysis of choice \choicet{expertreplay}: 95th percentile of performance scores conditioned on choice (top) and distribution of choices in top 5\% of configurations (bottom).
\choicet{expertreplay}\texttt{=None} means that expert transitions are not replayed.
The data comes from an experiment similar to the main one but in which
we also sweep \choicet{expertreplay}.
All other experiments do not replay expert data.}
\label{fig:expert_replay}
\end{center}
\end{figure}
\end{document}